\documentclass{article}
\usepackage{graphicx}		
\DeclareGraphicsExtensions{.pdf,.jpeg,.png}
\usepackage{color}
\usepackage{subfig}
\usepackage{amssymb, amsmath}                            
\usepackage{amsfonts}
\usepackage{mathtools}
\usepackage{pdfpages}
\usepackage{booktabs}
\usepackage{colortbl} 
\usepackage{xcolor} 
\usepackage{xfrac}
\usepackage{multirow}
\usepackage{tabularx}
\usepackage[linesnumbered,ruled]{algorithm2e}
\usepackage[numbers,square,sort&compress]{natbib}
\usepackage{authblk}
\usepackage{rotating}
\usepackage{tabularx}
\usepackage{multirow}
\usepackage{booktabs}
\usepackage{geometry}
\newgeometry{vmargin={1in}, hmargin={1in}}   
\usepackage[utf8]{inputenc}
\usepackage{adjustbox}
\usepackage{url}
\usepackage{placeins}  

\makeatletter
\renewcommand\AB@affilsepx{, \protect\Affilfont}
\makeatother

\title{Intuitive Surgical SurgToolLoc and SurgVU Challenges Results: 2022-2025}

\author[1]{Aneeq Zia}
\author[1]{Max Berniker$^*$}
\author[1]{Rogerio Garcia Nespolo$^*$}
\author[1]{Xiaorui Zhang$^*$}
\author[1]{Conor Perreault$^*$}
\author[1]{Kiran Bhattacharyya}
\author[1]{Xi Liu}
\author[1]{Ziheng Wang}

\author[2]{Satoshi Kondo} 
\author[3,57]{Satoshi Kasai} 
% ...
\affil[1]{Intuitive Surgical, Inc.}
\affil[2]{Muroran Institute of Technology}
\affil[3]{Niigata University of Health and Welfare}
% ...
\affil[57]{Fujita Health University}
% ...

\footnotetext{These authors contributed equally.}

\begin{document}

\maketitle

\newpage
\begin{abstract}
Robotic assisted (RA) surgery promises to transform surgical intervention. Intuitive Surgical is committed to fostering these changes and the machine learning models and algorithms that will enable them. With these goals in mind we have invited the surgical data science community to participate in a yearly competition hosted through the Medical Imaging Computing and Computer Assisted Interventions (MICCAI) conference. With varying changes from year to year, we have challenged the community to solve difficult machine learning problems in the context of advanced RA applications. Here we document the results of these challenges, focusing on surgical tool localization (SurgToolLoc) and surgical visual understanding (SurgVU). The publicly released dataset that accompanies these challenges is detailed in a separate paper https://arxiv.org/abs/2501.09209
\cite{SurgVUpaper}.
\end{abstract}

\newpage

\tableofcontents

\newpage

\section{Introduction}
Robotic-assisted (RA) surgeries become more prevalent with each passing year \cite{mederos2022trends, grimsley2022exploring}. 
Importantly, RA surgeries also generate rich streams of data, e.g. video recordings, tool kinematics, and events gathered from the robotic system, that lend themselves to quantitative analyses. Pursuits such as the quantification of surgical performance, autonomous analysis of procedures, surgical skill evaluation, AI-guided surgical planning, and surgical data science, in general, can all exploit this rich source of data \cite{jiang2025surgisr4k, maier2022surgical, vedula2017surgical, guerin2022review, luongo2021deep, hung2019deep, brown2020bring, zia2018automated, zia2017temporal, zia2019novel, zia2018surgical}.
However, despite the ability to collect large amounts of it, annotating it to provide meaningful context remains a major bottleneck in the field.

Acquiring annotations for surgical data is particularly challenging as it requires personnel with in-depth domain knowledge. 
What's more, relying solely on human annotators is costly, time-consuming, and subject to inter-rater variability.
Yet RA procedures generate numerous, potentially instructive, labels that can be harvested automatically from the system.
For example, each instrument install (or uninstall) elicits events that record the tool type and the install (uninstall) time. Similarly, commands to these tools, such as the use of energy, elicit related events. While events such as this are informative, they are not always representative of the tools visible in the surgical field, let alone their spatial locations.
Machine learning techniques that could reliably exploit these events, by interpreting them as noisy, weak labels for training, would prove extremely valuable. While algorithms for this class of problem exist, there is a real need for advancements in their accuracy and robustness.

Machine learning (ML) challenges --also known as ML competitions-- have become an integral part of many top computer vision and machine learning conferences \cite{eisenmann2022biomedical}. These competitions aim to address specific problems within various fields where the organizers provide the required data set and/or labels. In the surgical domain specifically, EndoVis\footnote{https://opencas.dkfz.de/endovis/} is an annual challenge held at the international Conference on Medical Image Computing and Computer Assisted Intervention MICCAI\footnote{https://miccai.org/}. This challenge contains multiple sub-challenges targeting problems like tissue and instrument tracking and segmentation, 3D scene reconstruction, action recognition, and surgical skills assessment \cite{zia2021surgical, zhao2020learning, zia2022objective, allan2021stereo, AllanMICCAI2018}. Following suit, our team has organized a yearly challenge since 2020.

With varying differences, all our challenges have focused on problems related to RA surgical intelligence. Beginning in the year 2022, however, we have focused on the same problem of weakly supervised learning of surgical tool localization (SurgToolLoc). For these challenges we provided participating teams with an extensive set of endoscopic video data. Alongside these videos, the teams received noisy annotations for the surgical tools present in each video frame, in the form of tool presence information. The ultimate goal of these challenges was to utilize automatically extracted tool labels to build models that can classify and localize surgical instrumentation in a hidden test set. In addition, each year we have posed a unique secondary challenge, that also utilizes the same video data. However, the main objective of our repeated challenge is to facilitate year over year advances in the ML solutions to the weakly supervised problem of tool classification and localization.

Here we document the results of each year's challenge. We begin with an overview of the data set made available to the teams. Then we present each year's challenge as a separate chapter. We end with some general observations, outlining what strategies have proved most successful, and what we believe are the challenges moving forward. Finally, we note that in the interest of encouraging the surgical data science community to continue working on similar problems, the training data used in these challenges has be released publicly in a separate publication. To the best of our knowledge, this is the largest publicly available surgical data set to date. It is our firm hope that this ongoing challenge and the accompanying data sets become touchstones for the broader field of surgical data science.

\section{Data Overview}\label{chapter_dataoverview}

For this continuing challenge a large dataset of surgical videos and accompanying annotations and labels was generated. Video and system data were harvested from surgical training exercises performed on the da Vinci robotic system. During these exercises, surgical trainees (and some experts) performed a series of standardized activities on porcine tissue, such as dissecting, suturing, and cauterizing. The captured video displays the trainee's view from the surgeon's console, although it is only one of the two video streams from the stereoscopic imaging system. 

The training exercises that generated the data were performed on two distinct animal modalities, either an anesthetized living pig, or using organs harvested from a pig. This distinction leads to differences in the general appearance of the resulting video frames, and those of the former modality are generally redder in tone. The anatomy present in the videos was not relevant to this challenge. 

To obtain training labels system data was harvested while the exercises were performed, and a team of trained clinical experts annotated the videos. 
The system data provided the tools installed in the robot. The clinical experts annotated the videos, segmenting them into surgical steps. The challenges hosted in 2022 and 2023 drew from this dataset. A separate paper that accompanies the dataset provides more details along with download link \cite{SurgVUpaper}.

\section{Results and methods from the MICCAI 2022 SurgToolLoc challenge}

\subsection{SurgToolLoc 2022 Challenge Description}

\subsubsection{Overview}
\begin{figure}
 \centering
 \includegraphics[scale=1.2]{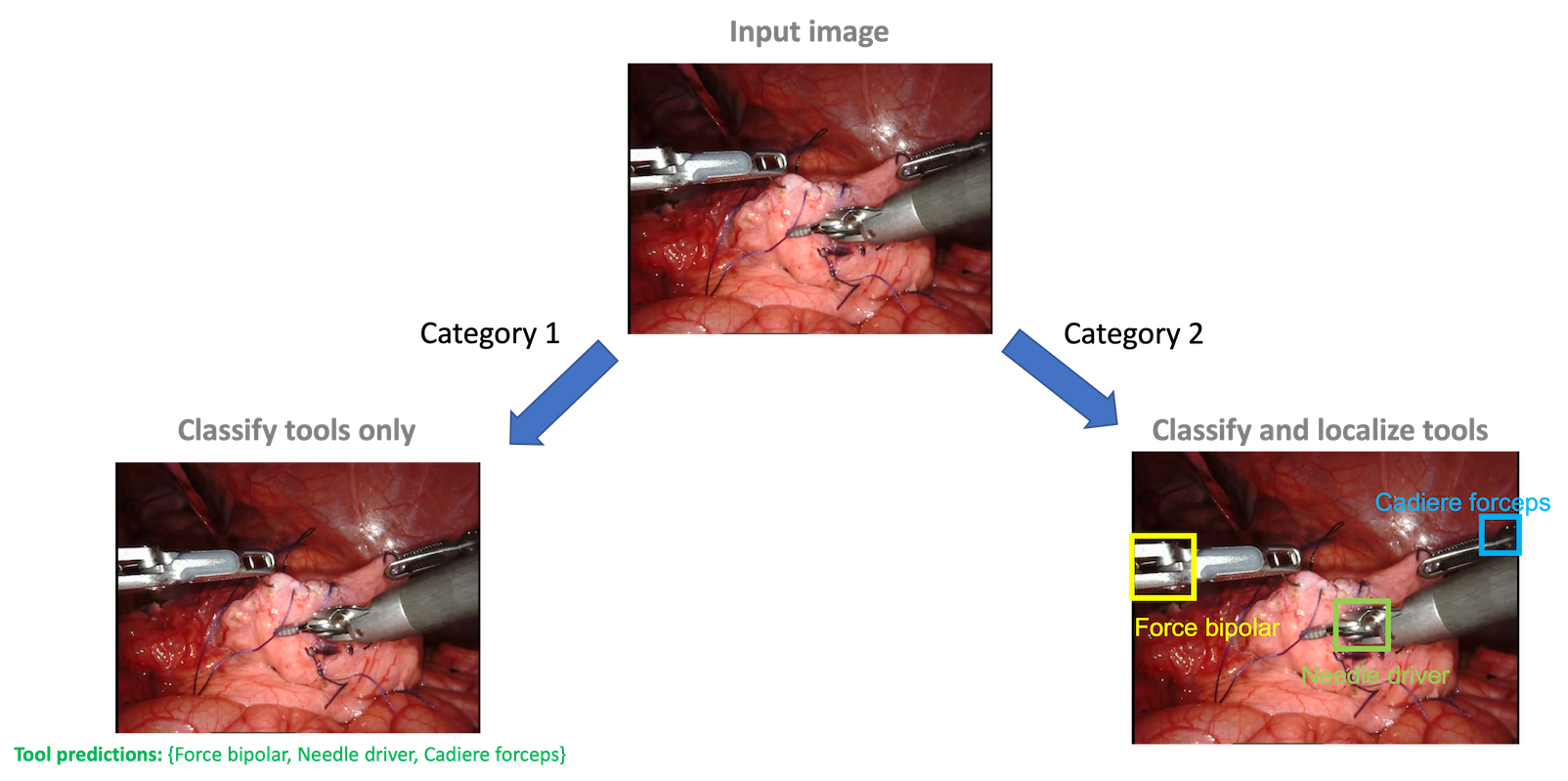}
 \caption{Overview of challenge categories 1 and 2}
  \label{fig:overview_categories}
\end{figure}

The sub-challenge was arranged as part of the Endoscopic Vision Challenge at MICCAI 2022. The challenge had two separate categories open to the participating teams (Figure \ref{fig:overview_categories}). Category 1 focused on identifying the presence of surgical tools in video frames, while Category 2 challenge was designed in order to identify both those tools present and their locations in the corresponding video frames.

Both categories required the development of weakly supervised algorithms. The first category relied on supervised learning for object detection with weak/noisy labels. The second category required weakly supervised object detection and localization with the same labels. Teams were allowed to participate in either challenge or both.

\subsubsection{Category 1: Surgical Tool Classification}

This category required the teams to submit a model that could classify all the tools present (up to 3 surgical instruments) within 30-second video clips.
This required the teams to train a multi-label model to classify the 14 possible tools.
A data set of 24,695 video clips was provided for training. The accompanying labels indicated which robotic tools were present in each clip. For technical reasons, a portion of the labels was missing or incorrectly assigned.

Submissions were evaluated on a hidden test data set of surgical videos through the Grand Challenge's automated docker submission and evaluation system. Teams were ranked by their model's average F1 score across all classes.

\subsubsection{Category 2: Surgical Tool Classification \& Detection}

This category required the models to perform the same classification problem as before while also localizing the identified tools with bounding boxes for each frame.
As in Category 1, the models localized the tools present within a set of 30-second video clips. 

\subsubsection{Training Data}

Video and system data were harvested from surgical training exercises performed on the da Vinci robotic system (see Section \ref{chapter_dataoverview}). From this dataset 30-second video clips were sampled. Across all the video clips, 14 tools were installed and visible. They were: needle driver, cadiere forceps, prograsp forceps, monopolar curved scissors, bipolar forceps, stapler, force bipolar, vessel sealer, permanent cautery hook/spatula, clip applier, tip-up fenestrated grasper, suction irrigator, bipolar dissector, and grasping retractor.

\begin{figure}
 \centering
 \includegraphics[scale=0.4]{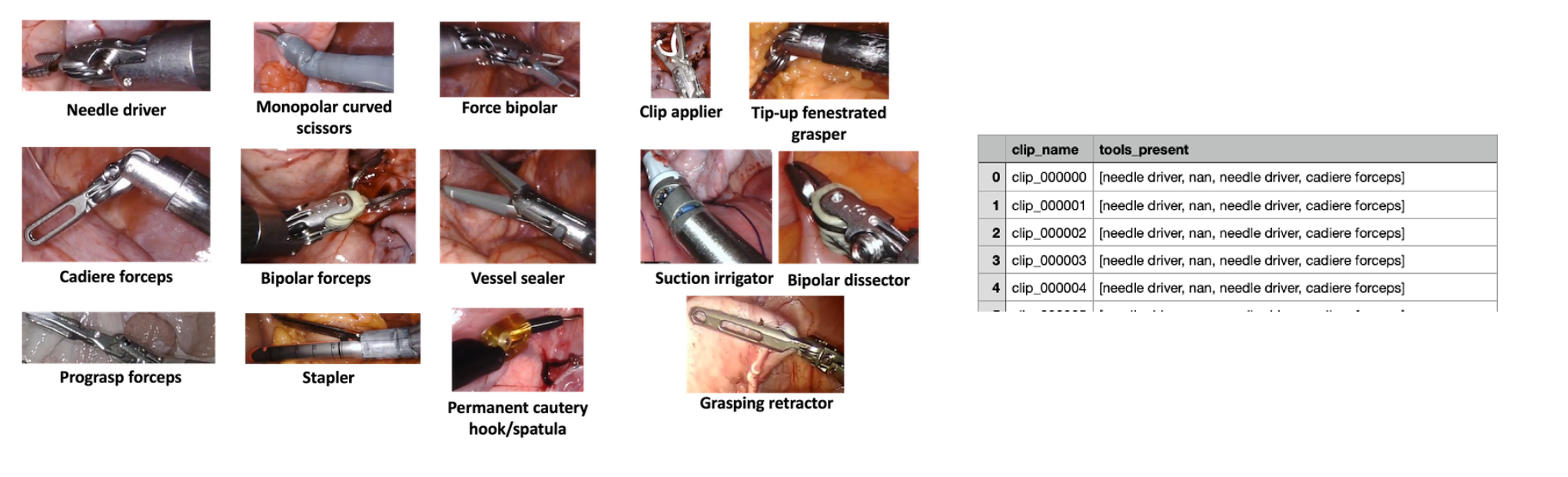}
 \caption{Sample frames with presence labels (left) and a snapshot of the labels CSV file (right)}
  \label{fig:test_labels_csv}
\end{figure}

From the data collected, 24,695 video clips from a set of 11 unique training exercises were made available for training purposes. Each clip was 30 seconds long and captured at 60 FPS with a resolution of 720p (1280 x 720) from one channel of the binocular endoscope (Figure \ref{fig:test_labels_csv}).  For the extent of each clip, there were up to three robotic surgical tools simultaneously installed within the surgical field. 
For each video clip within the training set, we provided the corresponding tool presence labels that indicated which robotic tools are installed. Since the tools are installed for the entire duration of each video clip, only one label per video clip was provided. Notably, for some clips, or portions of clips, tools were obscured or otherwise temporarily not visible. Moreover, the instrument distribution among the data was highly imbalanced. As such, the tool presence labels were noisy indicators of the presence of a surgical instrument in the video frames.

Presence labels were a list of four strings corresponding to the tools present in the robot's four arms [USM1, USM2, USM3, USM4], e.g, ['cadiere forceps', 'nan', 'force bipolar', 'grasping retractor], where the arm with the camera installed is labeled as "nan." The four arms are usually (though not necessarily) installed from left to right (i.e. USM1 is the leftmost arm, and USM4 is the rightmost arm). Figure \ref{fig:test_labels_csv} shows an example of the .csv file with the presence labels. This training data set has also been released publicly and can be accessed at https://console.cloud.google.com/storage/browser/isi-surgtoolloc-2022.

Finally, we note that the teams were also allowed to pre-train their models on publicly available surgical datasets such as Cholec80 \cite{Twinanda2016} and M2CAI \cite{jin2018tool}, and comprehensive datasets such as ImageNet \cite{russakovsky2015imagenet}.

\subsubsection{Testing Data}\label{Training_Data}

\begin{figure}
 \centering
 \includegraphics[scale=0.5]{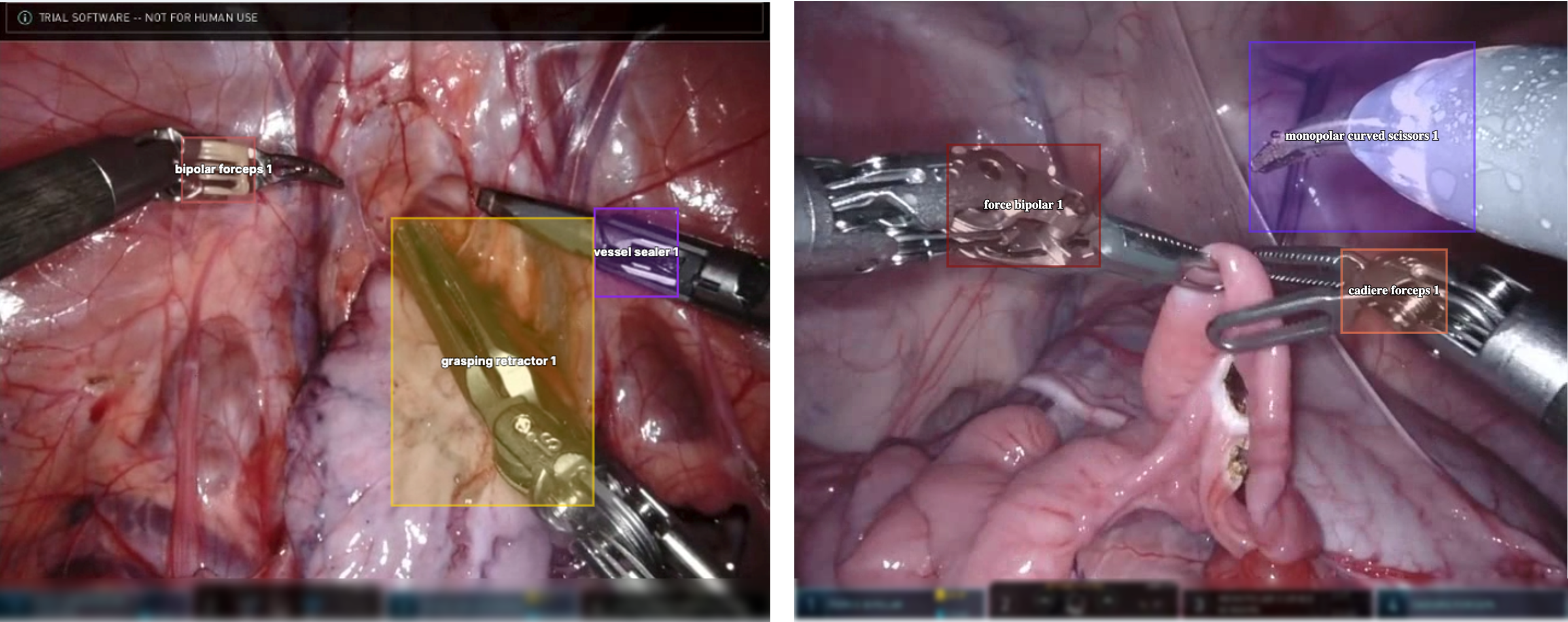}
 \caption{Sample frames with testing labels. The UI interface was blurred to avoid its embedded information (description of the tools) being used for predictions.}
  \label{fig:test_labels}
\end{figure}

The testing dataset consisted of 93 video clips taken from surgical training exercises---similar to the training set---using the da Vinci robotic system. The length of each video clip was variable-mean(SD), in seconds: 747.31(579.92)-, but had the same resolution and was sub-sampled to 1 frame-per-second (FPS) for inference purposes. 
The testing dataset had the tool presence labels in the same way as for the training set but also combined with  annotated bounding boxes around the instruments. These annotations were generated by a crowd (30) of experienced annotators. Two rounds of annotation reviews were conducted to ensure quality. It is important to note that the clevis of each robotic surgical tool was considered the ground truth for most tools as Figure \ref{fig:test_labels} shows. However, there are following exceptions to this rule:
If a tool's clevis is not well defined or if the tool is especially large and the clevis does not always show in the field of view, then the bounding box includes the surgical tip as well - e.g monocular curve scissor as shown in the purple-colored bounding box on the left of Figure \ref{fig:test_labels}.
Moreover, using the information available in the UI to make predictions was not allowed. To enforce this, the area comprising the UI interface in the test set was blurred, as seen in Figure \ref{fig:test_labels}.

\subsubsection{Submission Process}

The challenge was hosted as a ``Type 2 (T2) challenge" on the grand-challenge (GC) platform\footnote{grand-challenge.org}. T2 challenges on GC allow for the automated submission of containers, while the evaluation process for these submissions is executed via a standardized evaluation protocol to ensure the reproducibility of the algorithm. This modality also allows for each algorithm to be evaluated on a private test dataset that is hidden from the teams.

Each team developed its own algorithm following the guidelines provided by the GC platform\footnote{https://grand-challenge.org/documentation/algorithms/} and the organizing team of the challenge. Every submitted algorithm was configured to receive as input either a preliminary test dataset containing two videos for debugging purposes or the final test dataset described in \ref{Training_Data}. The algorithm's output had to follow a standardized file format for each category. For category 1 -surgical tool classification-, the output file had to generate a list of dictionaries containing information on the tools present in each frame of the input video in a boolean format. For category 2 - Surgical tool classification and localization-, the algorithm had to output a dictionary containing the set of tools detected in each frame with its corresponding bounding box cartesian coordinates (x, y).
Details on the creation of the algorithms, samples of input data, and the required format for output files were provided to the participating teams \footnote{https://surgtoolloc.grand-challenge.org/submission/}. 

Following each evaluation, the platform's embedded leaderboard displayed each team's submission scores. Then, the teams were asked to submit a final report with their methodology and results when evaluating the algorithm with the hidden test dataset. Each report is reproduced in this paper in section \ref{team_submissions}.

\subsection{Team Submissions} \label{team_submissions}

For this challenge, there were a total of 79 teams that had shown interest in participating and downloading the dataset. In the end, there were 17 teams that made full submissions in Category 1 whereas 11 made full submissions for Category 2. All teams that participated in Category 2 also participated in Category 1. A table showing the participating teams is presented below (Table \ref{table:TeamAffils}), while Table \ref{table:TeamMethodsSummary} summarizes the methodologies employed by each team. What follows in its section are the methodological details of the participating teams. Please note that these sub-sections were written by the participating teams so a difference in writing style and content will be observed.


\begin{table}[h!]
\centering
\caption{Team affiliations and challenge categories.}
\begin{tabular}{ccp{4cm}ccc}
\hline
Team \# & Team name & Institution  & Country & Category & Report \\ \hline 
1 & HRI\_MV & Hikvision Research Institute & China & 1 \& 2 & Y\\ 
2 & HKMV & South China University of China & China & 1 \& 2 & Y\\ 
3 & NVIDIA & NVIDIA & USA & 1 \& 2 & Y \\ 
4 & ANL-Surg & Argonne National Lab & USA & 1 \& 2 & Y \\ 
5 & HVRL & Keio University & Japan & 1 \& 2 & Y \\ 
6 & SK & Muroran Institute of Technology, Niigata University of Health and Welfare, Konica Minolta & Japan  & 1 \& 2   & Y  \\
7 & TeamZERO & University of West of England Bristol, University of Glasgow, University of Bristol & UK  & 1 \& 2 & N \\ 
8 & VANDY-VISE & Vanderbilt University & USA & 1 \& 2 & Y \\ 
9 & UKE & University Medical Center Hamburg & Germany & 1 only & Y \\ 
10 & Gatech & Georgia Institute of Technology & USA & 1 \& 2 & N \\ 
11 & ITeM & Furtwangen University, University of Lipzig & Germany & 1 \& 2 & Y\\ 
12 & MM & Chinese University of Hong Kong, National University of Singapore, University College London & Hong Kong, Singapore, UK & 1 only & Y \\ 
13 & BioMedIA & Muhmammad bin Zayed University of Artificial Intelligence & UAE & 1 only & Y\\ 
14 & WhiteBox & The University of Tokyo & Japan & 1 only & Y \\ 
15 & CAMMA & University of Strasbourg & France & 1 only & Y \\ 
16 & Vision\_HK & Tianjin University & China & 1 \& 2 & N \\ 
17 & lsgroup & Xiamen University & China & 1 only & N \\ 
\end{tabular}
\label{table:TeamAffils}
\end{table}


\begin{sidewaystable*}
\caption{Summary of methodologies}
\begin{adjustbox}{scale=0.6,center}
{\begin{tabular}{ l p{10em} l m{5em}l m{5em} l m{5em}l m{5em}l m{10em}l m{5em}l m{10em}l m{10em}l m{5em} l } 
Team Name & Architecture & Backbone & Data Preprocessing & Pretrain & Image Augmentation & Use Additional Data & Loss & Output  \\
\hline
HKMV & Faster-RCNN & efficient net-b3 &  & Endovis17, Endovis18 & N/A & Bounding box generated from segmentation mask &  & tool bounding box   \\
SK & Multiplicity Feature fusion & ResNet-50 & Resize & Segmentation data & Yes & N/A &  & tool class, tool bounding box   \\
NVIDIA & Hybrid (ConvNext-tiny, EfficientNet-B4, Ensemble), YoloV5 &  & Resize, crop &  & Yes & Pseudo label and manual bounding box &  & tool class, tool bounding box   \\
WhiteBox & CNN & ResNet-18 & Resize, crop &  &  &  & Asymmetric Focal Loss & tool class &  \\
HVRL & Hybrid (Swam Transformer, EfficientNet-V2, ResNeXt) &  & Resize, crop & ImageNet weights, Instagram weights & Yes &  & binary cross-entropy & tool class, heatmap, tool bounding box  \\
MM & Hybrid (ResNet50, ViT-Batch16) &  & Resize &  & Yes &  & Asymmetric Focal Loss & tool class   \\
CAMMA & Spatial Attention network (SANet) & ResNet-18 & Resize & ImageNet weights & Yes &  & weighted sigmoid cross entropy & tool class, attention map   \\
UKE & DINO-based ViT & ViT & Resize & ImageNet weights & Yes &  &  & tool class, attention map, tool bounding box  \\
BioMedIA & YOLOv5, ConvNext & ResNet 50 & Oversampling, UI removal & Cholec80, EndoVis’15, CholecSeg8k & Yes & Cholec80, EndoVis’15, CholecSeg8k datasets & focal loss, box regression loss, objectness loss & tool class, tool bounding box   \\
ANL-Surg & Ensemble (YOLOv5, ConvNext); Detectron2 &  & Resize, crop & MaskRCNN weights & - & Segmentation labels from  MICCAI17 and MICCAI18 &  & tool class, tool bounding box   \\
VANDY-VISE & Query2Label & ResNet, ViT &  &  & - &  & Asymmetric loss & tool class, tool bounding box   \\
ITeM & CNN + Squeeze and Excitation (SE) module & ResNet50 &  & Cholec80 & - &  & binary cross-entropy, softmax cross-entropy & tool class, tool bounding box  \\
\hline
\end{tabular}}
\end{adjustbox}
\label{table:TeamMethodsSummary}
\end{sidewaystable*}

\clearpage

\subsection{Hikvision Research Institute - Team HRI\_MV}

\noindent Team Member: Quan He\\


The HRI\_MV team made full use of the characteristics of the video data and adopt an object-tracking scheme with a semantic segmentation network to obtain the pseudo tags of the frames. In particular, an IOU scoring with the dual model cross-validation mechanism was employed to choose more reliable pseudo tags. Using the obtained pseudo tags, a detector was trained and integrated into a pseudo tag using a Kalman filter, iteratively updating the models.

In traditional instrument target detection, the whole object is considered as the ground truth (GT) and included in the detection frame, while in surgical instrument detection, only the clevis is considered as GT. The difficulty is that the appearance of the clevis of different devices is relatively similar. When human eyes distinguish different devices, they are actually mainly distinguished according to the clamps of the devices. To solve this problem, the ROI proposal box output by the RPN network was modified: for the regression head, the proposal box output by RPN was used, while for the classification head, the proposal box was expanded to obtain more information for device classification.

Finally, the integration algorithm was adapted to use the detection frame output from the previous frame for target tracking while combining the results of target detection and target tracking by weighting, which effectively solved the problem of target missing detection.


\subsubsection{Method Description}

\noindent{\textbf{Object Tracking with Cross Supervision:}}

The HRI\_MV team choose to label the frame sequence of the video by using the target tracking method: The first frame of the video was manually labeled, while automatically labeling the subsequent frames using the target tracking algorithm. However, the tracking process sometimes failed. The team found that this was caused by clevis’ small and inconspicuous features, reflection, and other factors. Therefore, a cross-validation method based on segmentation results was proposed.

Siamfc\texttt{++}\cite{xu2020siamfc++} was chosen for single target tracking (SOT). The advantage of this approach is that the trained SOT model can be used without retraining. Only the first frame of the sequence needed to be labeled so the SOT was able to track the subsequent picture frames effectively. Then, an unet-like segmentation network was trained with the open source data dVRK segmentation dataset, Event-based-2D-tracking-of-articulated-instruments-for-robotic-surgery dataset, EndoVis17 dataset, and EndoVis18 dataset. 


The detector was trained with part of pseudo labels and then used to generate new pseudo labels with the tracking output. 
For frames with an IoU less than the threshold value of 0.2, the generated pseudo label was considered untrustworthy. 
In this way, a dataset containing about 25000 frames with pseudo labels was obtained.






\noindent{\textbf{ROI Expansion}:}

Cascade RCNN \cite{cai2018cascade} was chosen as a target detection model. In the initial test results, although many instruments were successfully detected, they were misclassified. This is because, in traditional instrument target detection, the whole of the object is considered as GT and included in the detection frame, while in surgical instrument detection, only clevis is considered as GT. Although, the appearance of clevis components of different instruments is relatively similar, and the difference is mainly reflected in the clamp part.

\begin{figure}[tbh]
\centering
\includegraphics[width=0.3\textwidth]{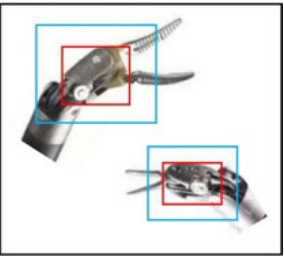}
\caption{HRI\_MV: The information required for detection and classification is different. Thus, the ROI proposal box output by RPN network was modified.}\label{fig1HRI}
\end{figure}

To solve this problem, the ROI proposal box output by the RPN network was modified. For the regression head, the proposal box output by RPN was used, while for the classification head, the proposal box was expanded twice in width and height, respectively, to obtain more information for device classification, as shown in Figure \ref{fig1HRI}.

\begin{figure}[tbh]
\centering
\includegraphics[width=0.7\textwidth]{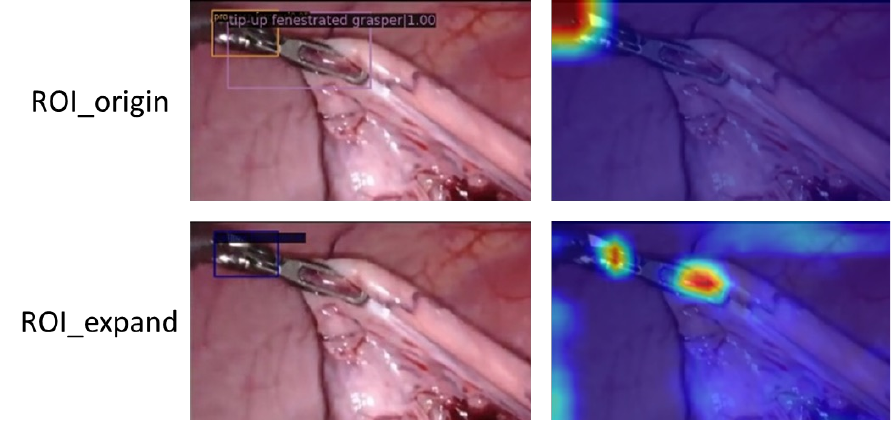}
\caption{HRI\_MV: ROI expand makes model identification device}
\label{fig2HRI}
\end{figure}

Figure \ref{fig2HRI} shows the feature map extracted from the backbone of the proposed model. In the original model, the instrument cadiere forceps was mistakenly considered prograsp Forceps and tip-up fenestrated Grasper, the thermodynamic values in its characteristic diagram are all concentrated in clevis. With the proposed expanded ROI, the cadiere forceps is correctly classified: in its feature map, there is not only a large thermal value distribution at the clevis but also a large thermal value at the clamp, which helps the model to better classify instruments.

\noindent{\textbf{Model ensemble:}}

\begin{figure}[tbh]
\centering
\includegraphics[width=0.4\textwidth]{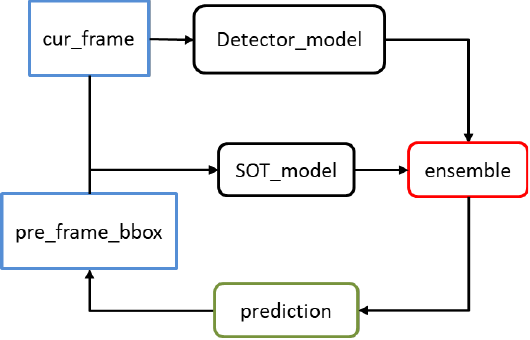}
\caption{HRI\_MV: model ensemble}
\label{fig_overall_HRI}
\end{figure}

The detection frame output from the previous frame algorithm was used for target tracking and integrated into the results of target detection and target tracking via weighting. In this way, as long as the device appears in a certain frame, it can be identified by target tracking in its subsequent frames, solving the problem of target missing detection.
The overall framework of our algorithm is shown in Figure \ref{fig_overall_HRI}.

\subsubsection{Preliminary Performance}

A test set was established to verify the performance of the detector to evaluate whether 
more data is needed. The test set was determined by manual review to include each sample category. \textit{Roundn} means nth times object tracking with the cross supervision method was used to get more data (see Table \ref{table:hri_mv1_method}).

\begin{table}[tbh]
\caption{HRI\_MV: Object tracking with the Cross Supervision method was used to get more data.}
\begin{center}
\begin{tabular}{lllll}
\hline
Method & Round1 & Round2 & Round3 & Round4  \\ \hline
mean mAP & 0.164 & 0.347 & 0.489 & 0.491 \\ \hline
\end{tabular}
\label{table:hri_mv1_method}
\end{center}
\end{table}

By the fourth round, performance has improved slightly. Then, these data were used for the next experiment. The entire dataset was then divided into a training set and test set (7:3) to verify what strategies could be taken to enhance the performance of the model (table \ref{table:hri_mv}).

\begin{table}[tbh]
\caption{HRI\_MV: Model Results}
\begin{center}
\begin{tabular}{ll}
\hline
Method                      & mean mAP  \\ \hline
Base              & +0.489      \\
+ swin pre-trained & +0.052      \\
+ roi expand 2            & +0.063      \\
+ ChannelShuffle-RGNShift  & +0.021      \\
+Fog-Blur-Noise         & +0.048      \\
+ focallosss           & +0.002      \\
+cbam            & +0.033      \\
+DetecoRS pre-trained  & +0.059      \\
swin pre-trained + roi expand 2 + Fog-Blur-Noise + ChannelShuffle-RGBShift &    +0.109      \\
detectors pre trained + roi expand 2 + Fog-Blur-Noise + ChannelShuffle-RGBShift               & +0.111      \\ \hline
\end{tabular}
\end{center}
\label{table:hri_mv}
\end{table}

The mentioned strategies can improve the performance of the model on the current test set; by adopting the ROI expand method, the overall performance was improved by 0.063. Then, the last two models in Table \ref{table:hri_mv} were selected as the models for the ensemble. On the Surgical tool localization leaderboard, the effectiveness of the ensemble scheme was confirmed: the single model reached a score of 0.3985 on the final test set, while the ensemble model reached a score of 0.4077 and ranked first. Since the detector can classify the target, it was also joined with the Surgical tool detection Tasks. However, on the Surgical tool detection leaderboard, The performance of the ensemble model deteriorated when compared to that of the single model, with the former score being 0.7485 and the latter being 0.7341.


%
\subsection{South China University of China - Team HKMV}

\noindent Team Member: Weiliang Chen


\subsubsection{Method Description}

In order to obtain an object detection model that can effectively locate surgical tools, the key is how to build a well-labeled training dataset at a low cost. In this method, two public datasets-endovis17 \cite{Allan2019-2017} and endovis18 \cite{AllanMICCAI2018}- were used as the primary training dataset. The clevis part in the mask label of the dataset was converted into a bounding box and then used to train an object detection model. This model was then used to infer the images of the competition dataset and then adjust the predicted bounding box slightly. After repeating several operations, a large-scale training dataset can be obtained.

In terms of the model, a two-stage object detection model was built based on Faster R-CNN. EfficientNet-b3 was chosen here, which takes into account both performance and speed, as the backbone to extract image features and add FPN structure to further optimize the feature map.\\

\noindent{\textbf{Dataset:}}

\begin{figure}[tbh]
\centering
\includegraphics[width=0.7\textwidth]{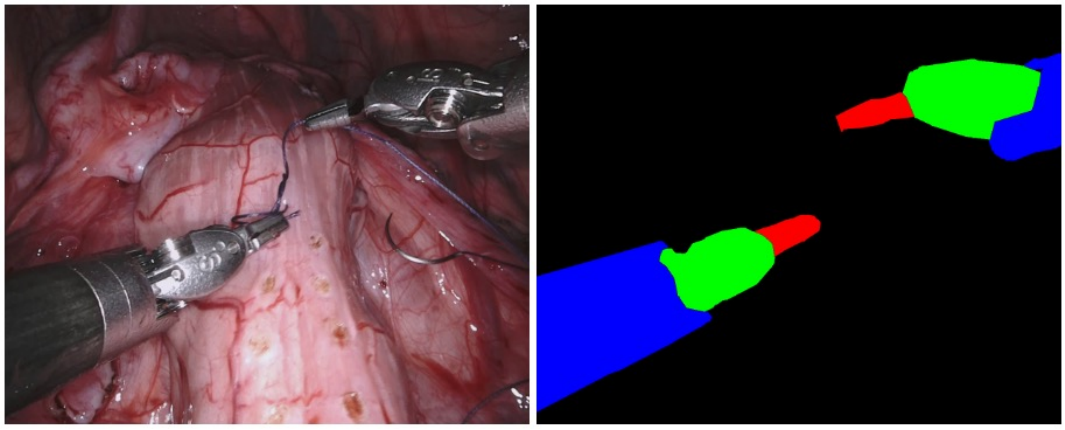}
\caption{HKMV: Primary training dataset}
\label{fig1hkmv}
\end{figure}

Two public datasets endovis17 and endovis18 were used to assist us to obtain a primary training dataset. These two data sets provide mask annotations, as shown in Figure \ref{fig1hkmv}.

The clevis part in the mask label of the dataset was converted into a bounding box label and used as the primary training dataset. Since endovis17 and endovis18 do not include all 14 surgical tools, part of the data was added to the training dataset to ensure that the dataset contains all 14 surgical tools. The final set contained 5212 images.\\

\noindent{\textbf{Surgical Tool Localization Algorithm:}}
\begin{itemize}
\item A Faster R-CNN object detection model was trained on the primary training dataset.
\item EfficientNet-b3, which takes into account both performance and speed, was chosen as the backbone to extract image features and add FPN structure to further optimize the feature map.
\item According to the size of the clevis of surgical tools in the training dataset, four layers of feature maps were included in the model, which were downsampled 8, 16, 32, and 64 times respectively.
\end{itemize}

\noindent{\textbf{Dataset Expansion and Adjustment:}}

\begin{figure}[tbh]
\centering
\includegraphics[width=0.7\textwidth]{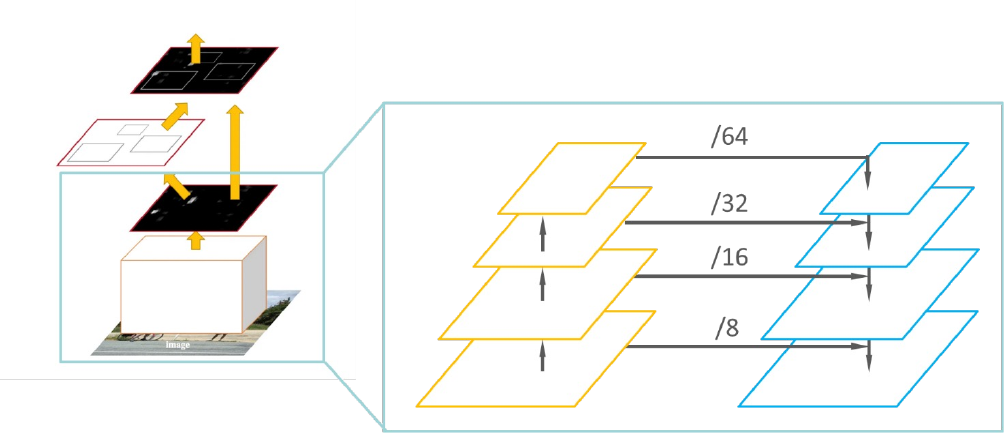}
\caption{HKMV: Model architecture}
\end{figure}

\begin{figure}[tbh]
\centering
\includegraphics[width=0.7\textwidth]{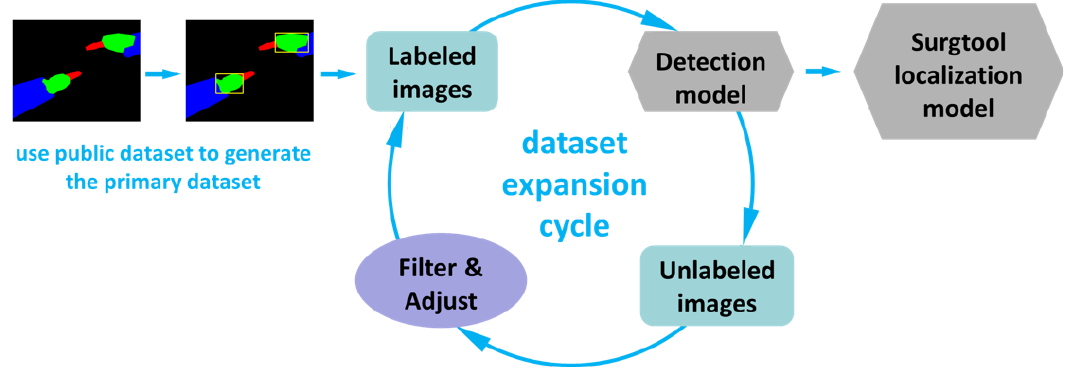}
\caption{HKMV: Data expansion scheme}
\label{fig2HKMV}
\end{figure}

In order to increase the size of the dataset, the trained model was used to infer the images of the competition dataset, adding these to the training dataset. A score threshold of 0.7 was employed to filter out the poor bounding box.
After several operations, the dataset was expanded to 11035 images. The dataset expansion flow diagram of the algorithm is described in Figure 
\ref{fig2HKMV}.

\subsubsection{Preliminary Performance}

For the preliminary experiments, 25\% of the dataset was used as the validation dataset, and only well-labeled images were used in order to accurately evaluate our algorithm. The model structure of EfficientNet-b3 was chosen as the backbone, downsampling 8, 16, 32, and 64 times the feature map. During preliminary experiments on this validation dataset, the  model was able to reach a mean mAP of 0.452. 






\subsection{NVIDIA - Team NVIDIA}

Team Members: Bo Liu, David Austin, Yiheng Wang, Michal Futrega, Jean-Francois Puget \\


This team consists of five NVIDIA employees. Three of them (Jean-Francois, David, and Bo) are members of the Kaggle Grandmasters team, with extensive experience in computer vision machine learning competitions. One member (Yiheng) works on MONAI, an open-source framework for deep learning in healthcare imaging. Another (Michal) works on building GPU-optimized models for NVIDIA's Deep Learning Examples repository\footnote{https://github.com/NVIDIA/DeepLearningExamples}. This team's motivation was to win prizes in both categories using the NVIDIA software stack. 

\subsubsection{Method Description}

Category 1 is a weakly supervised category in the sense that classification labels are given on video level instead of frame level, but test predictions need to be performed on frame level. Since many video-level tools don’t appear in all the frames of the video, the major challenge of Category 1 is to identify frames that contain clean labels. They came up with a unique way to tackle this: first use segmentation models to identify frames where they could identify three unique tools per frame. By identifying three tools, they could then apply the video-level labels to the given frame. In order to do this they used models \cite{Shvets2018} published from the MICCAI 2017 Robotic Instrument Segmentation Challenge\footnote{https://github.com/ternaus/robot-surgery-segmentation}. After applying
segmentation models they then used traditional computer vision techniques to count the contours of each unique instrument, and layered additional logic (i.e., instrument contour needs to touch
the frame edge) to filter out false positives. All frames where they could positively identify three unique instruments per frame then formed the basis of the training dataset. 

Another unique thing in this dataset is the fact that many videos are sequential, e.g., 30-second chunks of a continuous video that last for several minutes or even dozens of minutes. Grouping
videos from the same “scene” into the same fold split is critical for meaningful local validation. In order to do this, they compared the image hash of each video’s first frame and the last frame of the preceding video. If the similarity is lower than a threshold, they mark them as from the same scene and put all videos of the same scene into the same split when they split the data into 5 folds. This way, leakage between folds can be prevented. To simulate the test data, they downsampled training videos from fps=60 to fps=1. They also cropped the black empty space on the left and right sides of all the training videos. Then, they blurred out the bottom banner which may contain ground truth labels. Finally, they resized the frames into 640 W x 512 H, the same as the test videos.

After the above steps, the team trained ConvNext\cite{liu2022convnet} Tiny and EfficientNet\cite{tan2019efficientnet} B4 models, two model
architectures that consistently dominate Kaggle image competitions. Standard techniques were used such as cosine learning rate schedule, mixed precision training, mixup augmentations, and weighted loss functions. 

Category 2 required localization and classification of a given set of surgical instruments in each frame of a video. Only video-level labels were provided so the challenge can be tasked as a weakly supervised problem. In order to solve this problem they used a diversity of techniques taken from both supervised and unsupervised methods.

One of the key learnings from Category 1 was the importance of sampling rare classes and identifying high-confidence labels in the train set. They exploited these learnings by developing a sample methodology to get a balanced set of frames/labels from a diverse set of scenes using the models they developed for Category 1. After these frames were identified they attempted to get clean localization of the tools of interest through two methods, GradCam\cite{selvaraju2017grad} and explicit labeling of the tools of interest. GradCam was effective at identifying the clevis of the tools of interest, however, because there was a set of rules and detection logic that needed to be applied on top of the clevis identification (i.e., tip vs clevis detection for some tools and not others), they found that applying manual labels that incorporated these rules were important. GradCam was thus used for visualization only, and manual application of labels and competition logic was used for the start of training.

\begin{figure}[ht]
    \centering
    \includegraphics[width=0.7\textwidth]{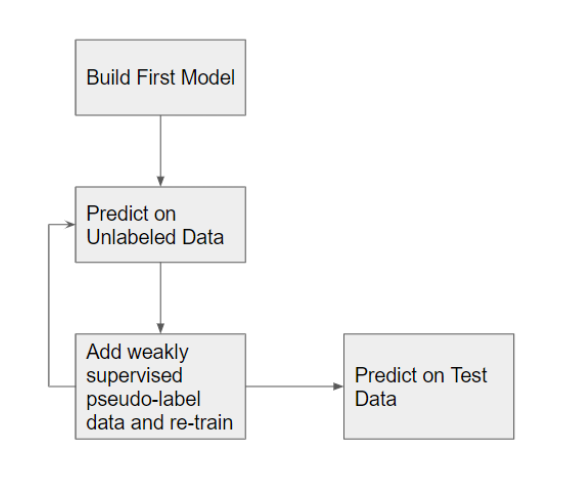}
    \caption{NVIDIA: Team NVIDIA Category 2 workflow}
    \label{fig:nvtask2}
\end{figure}

After identifying clean labels, the team trained YoloV5\cite{Jocher_YOLOv5_by_Ultralytics_2020} models first on a small set of manually created labels, according to the example labels on the challenge data page\footnote{https://surgtoolloc.grand-challenge.org/data/}, and then on weakly supervised pseudo-labels predicted from the first set of base models \cite{wang2021weakly}. The workflow is illustrated in Figure \ref{fig:nvtask2}.

\subsubsection{Model Training}

For Category 1, they trained 5 EfficientNet-B4 models (on 5 fold splits) and 5 ConvNext-tiny models and ensembled them using a trick called logit shift. The idea of the logit shift trick is that, when data is extremely imbalanced between classes as in this dataset, the minor classes' probabilities are extremely biased towards 0. The extent of the bias can be different across models, making ensembling difficult. Logit shifting is to shift the logit of each class by an optimal constant value (i.e. the same value for the same class, across all samples), so that 0.5 probability threshold is the optimal threshold for the F1 score. This optimal logit shift is tuned on local validation, and is different for each model. When ensembling, they shift logit first on individual models then simply take the average probabilities, since they are already adjusted to be centered around 0.5.

For Category 2, the team trained several versions of YoloV5 and found YoloV5m to be the best performing. Except for the default augmentations such as mixup, mosaic, and HSV transforms, they applied more augmentations like hue saturation value and random brightness contrast to make the model generalize better, especially for the minor tools.

All the experiments for the challenge were trained on NVIDIA DGX Station A100 (with A100 GPUs) and NVIDIA DGX Station (with V100 GPUs).

\subsubsection{Preliminary Performance}

For Category 1, The per-class F1 validation scores of both the ConvNext and EfficientNet models are provided in Table \ref{table:nv1}, as well as the ensemble score. The final ensemble has a validation score of 0.8526 and a leaderboard score of 0.7055.

For Category 2, the team achieved a mAP validation score of 0.3614 for the first model, and 0.4132 for the retrained model, which translated to a leaderboard score of 0.3058. The validation scores for each class are shown in Table \ref{table:nv2}.

\begin{table}[]
\caption{NVIDIA: Category 1 Results}
\hspace*{1.5cm} 
\begin{tabular}{lccc}
\centering
Class                          & ConvNext   Tiny & EfficientNet   B4 & Ensemble \\ \hline
needle driver                  & 0.9439          & 0.9565            & 0.9487   \\
cadiere forceps                & 0.6529          & 0.6826            & 0.6907   \\
bipolar forceps                & 0.9236          & 0.9305            & 0.9264   \\
monopolar curved scissors      & 0.9179          & 0.9330            & 0.9189   \\
grasping retractor             & 0.7082          & 0.6801            & 0.7188   \\
prograsp forceps               & 0.7549          & 0.7946            & 0.7742   \\
force bipolar                  & 0.9216          & 0.8665            & 0.9401   \\
vessel sealer                  & 0.9690          & 0.9728            & 0.9805   \\
permanent cautery hook/spatula & 0.8602          & 0.8417            & 0.8489   \\
clip applier                   & 0.8664          & 0.8700            & 0.8439   \\
tip-up fenestrated grasper     & 0.5333          & 0.5502            & 0.5222   \\
stapler                        & 0.9746          & 0.9748            & 0.9789   \\
bipolar dissector              & 0.9787          & 0.9744            & 0.9829   \\
suction irrigator              & 0.8619          & 0.8306            & 0.8619   \\ \hline
\textbf{Average   F1}                   & \textbf{0.8476}          & \textbf{0.8470}            & \textbf{0.8526}   \\ \hline
\end{tabular}
\label{table:nv1}
\end{table}

\begin{table}[]
\caption{NVIDIA: Category 2 Results}
\hspace*{2.5cm} 
\begin{tabular}{lcc}
\centering
Class                      & Frist model & Weakly supervised model \\ \hline
needle driver              & 0.5043      & 0.4933                  \\
cadiere forceps            & 0.5024      & 0.5111                  \\
bipolar forceps            & 0.3275      & 0.4441                  \\
monopolar curved scissors  & 0.3568      & 0.3898                  \\
grasping retractor         & 0.3101      & 0.4098                  \\
prograsp forceps           & 0.3619      & 0.4177                  \\
force bipolar              & 0.4341      & 0.4442                  \\
vessel sealer              & 0.4660      & 0.4874                  \\
permanent cautery hook     & 0.3866      & 0.4164                  \\
clip applier               & 0.2536      & 0.3520                  \\
tip-up fenestrated grasper & 0.4815      & 0.4472                  \\
stapler                    & 0.2262      & 0.2902                  \\
bipolar dissector          & 0.1046      & 0.3197                  \\
suction irrigator          & 0.3448      & 0.3617                  \\ \hline
\textbf{Average   mAP }             & \textbf{0.3614 }     & \textbf{0.4132}                  \\ \hline
\end{tabular}
\label{table:nv2}
\end{table}







\subsection{Argonne National Laboratory - Team ANL-Surg}
ANL-Surg team included Neil Getty, Fangfang Xia, Zixuan Zhao, Xiaotian Duan. This team was primarily interested in category 2, though given the available challenges, both categories were considered. ANL-Surg was motivated by the scale of the data, and the interesting need for a weakly-supervised approach. Their initial plan was to use traditional computer vision or pre-trained deep learning models to segment seen tools, and then make label assumptions about the unseen tools and retrain the model iteratively.

\subsubsection{Method Description}

Training data was sampled at 0.25 FPS and cropped to 880x520 to remove black borders and UI. ANL-Surg used segmentation labels from the two public competitions: MICCAI18: Robotic Scene Segmentation Sub-Challenge \cite{AllanMICCAI2018} and MICCAI17: Robotic Instrument Segmentation \cite{Allan2019-2017} for training initial Detectron2 segmentation models.

Their approach for tool presence classification was an ensemble of six ResNet50 models trained using the Fastai framework \cite{Howard_2020}. Five of the models were trained using cross-validation, and one was trained on all of the data. ANL-Surg averaged the softmax output and used a threshold to determine the final classification.

\begin{figure}[ht]
\centering
\includegraphics[width=0.9\textwidth]{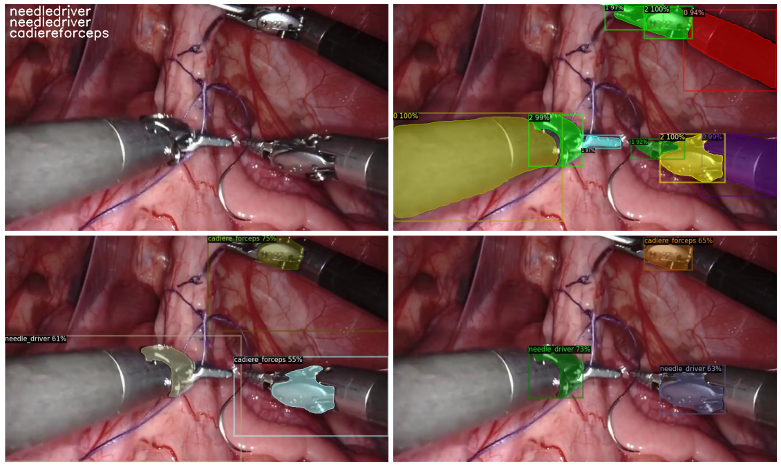}
\caption{ANL-Surg: Top right frame is the result of the parts detection model. Bottom left is the instrument tool detection model. Bottom right is the wrist final prediction with post-processing using the Category 1 model.}
\label{fig:anl_result}
\end{figure}

For tool localization, they trained a Detectron2 \cite{wu2019detectron2} model to segment parts of surgical tools. Then the parts were clustered to form bounding boxes for entire tools. Finally, individual tools are cropped out based on the predicted bounding box and query the Category 1 model about the class of the tool (Figure \ref{fig:anl_result}.


\subsubsection{Model Training}

ANL-Surg used 5-fold cross-validation, splitting the clips based on their computed similarity scores so that clips of the same video, or very close temporally were not included in both training and validation. However, validation was still a challenge, as a good result may actually be a bad result, due to the many frames with less than 3 visible instruments.

Detectron2 models were trained on a Tesla v100s with 32gGB VRAM for 2000 iterations, a batch size of 20, learning rate of 0.02, and 4 detections per image. MaskRCNN weights were used before finetuning on MICCAI 2017 and 2018 data.

\subsubsection{Preliminary Performance}
Internal validation results for tool presence classification are presented in Table \ref{tab:anl-table1}.

\begin{table}[ht!]
\centering
\caption{ANL-Surg: Preliminary F1 Scores for 5-fold cross-validation for tool presence classification.}
\begin{tabular}{cc}

Fold & F1      \\
\hline
0    & 0.87399 \\
1    & 0.91667 \\
2    & 0.87041 \\
3    & 0.84748 \\
4    & 0.86528 \\
\hline
\end{tabular}
\label{tab:anl-table1}
\end{table}


\subsection{Keio University - Team HVRL}

Team Members: Ryo Fujii, Ryo Hachiuma, Mana Masuda, and Hideo Saito\\


The main motivation of the HVRL team was to familiarize themselves with weakly supervised learning in a real surgery environment. The team has experience in building large-scale tool presence detection dataset in open surgery and deeply understand how expensive and tedious to annotate the surgery video \cite{app122010473}. Thus, applying weakly supervised learning reduces the annotation time to a real surgery video.

\subsubsection{Method Description}


For Category 1, multi-label tool presence classification models were trained in a fully supervised manner. Since only video clip-wise annotation is provided, all frames in a video clip were assumed to have the same annotation as the annotation attached to the video clip. The proposed solution was the ensemble of three different models, Swin Transformer V2 \cite{Liu_2022_CVPR}, EfficientNetV2 \cite{pmlr-v139-tan21a}, and ResNeXt \cite{DBLP:journals/corr/XieGDTH16}, which showed the high performance to the validation data. Models were implemented based on the PyTorch Image Models \footnote{https://github.com/rwightman/pytorch-image-models} \cite{rw2019timm}.

For Category 2, Grad-CAM++~\cite{8354201} was employed for the weakly-supervised tool localization task.

\begin{itemize}
    \item architecture design: 
    For category1, the team adopted the ensemble of three different models, Swin Transformer V2 \cite{Liu_2022_CVPR}, EfficientNetV2 \cite{pmlr-v139-tan21a}, and ResNeXt \cite{DBLP:journals/corr/XieGDTH16}. For Category 2, only EfficientNetV2 was employed.

    \begin{figure*}[tb]
\begin{center}
\begin{tabular}{ccc}
\begin{minipage}{0.3\hsize}
    \begin{center}
        \includegraphics[trim={0 1em 0 0}, clip, width=\hsize]{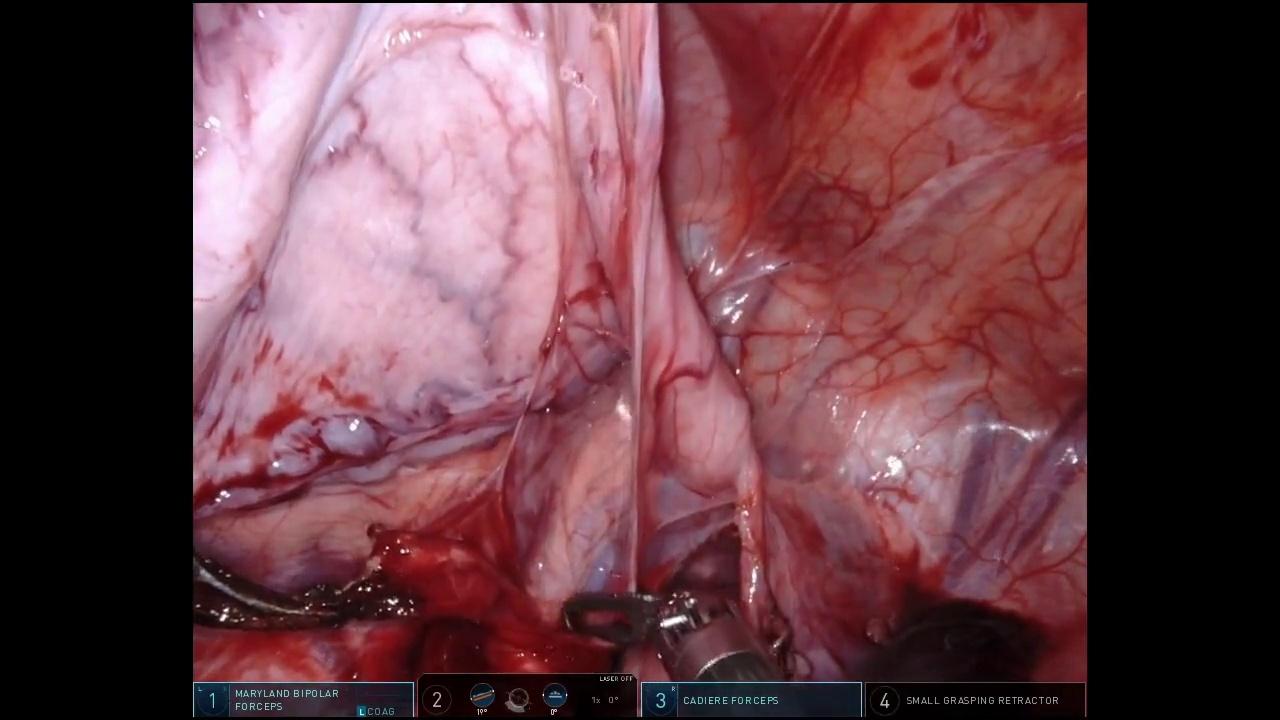} 
    \end{center}
\end{minipage}
&
\begin{minipage}{0.3\hsize}
    \begin{center}
        \includegraphics[trim={0 1em 0 0}, clip, width=\hsize]{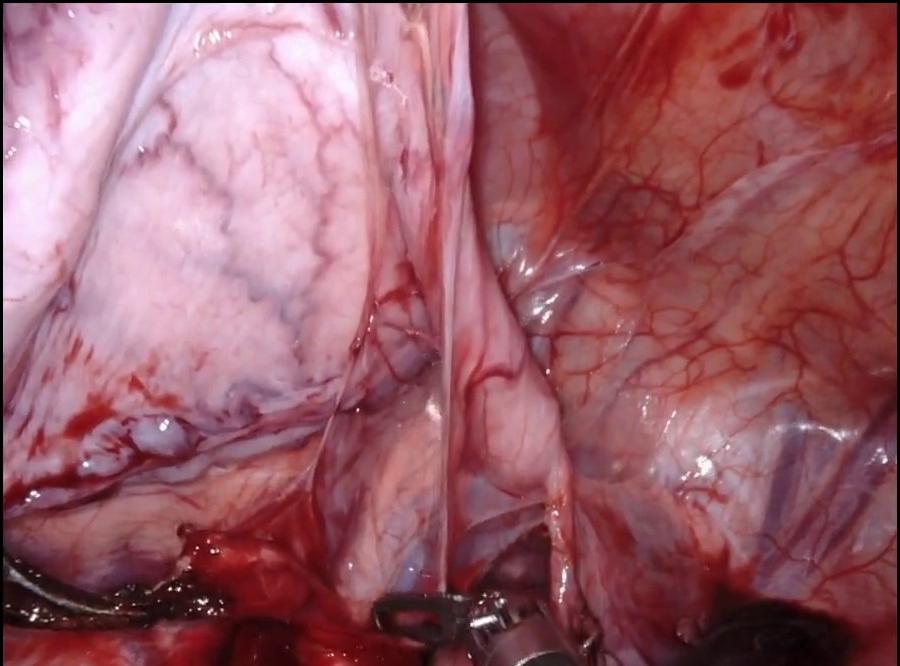} 
    \end{center}
\end{minipage}
&
\begin{minipage}{0.3\hsize}
    \begin{center}
        \includegraphics[trim={0 1em 0 0}, clip, width=\hsize]{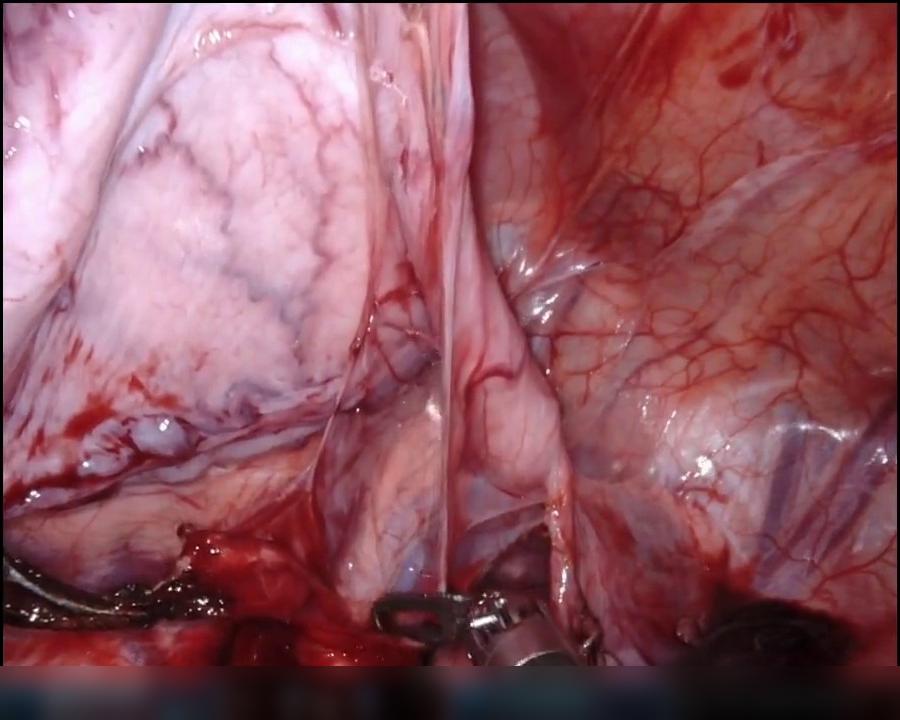}
    \end{center}
\end{minipage}
\\
{\footnotesize (a) Original image}
&
{\footnotesize (b) UI cropped image}
&
 {\footnotesize (c) UI blurred image}
\end{tabular}
\caption{HVRL: Examples of preprocessing. The original image included the black region on the left and right sides and the tool annotation UI on the bottom. First, the original image was cropped to remove the black region. Then, UI cropping and UI blurring with a probability of 0.5 was randomly conducted.}
\label{fig:1}
\end{center}

\end{figure*}
    \item data processing, reformatting, transformation, etc.:
    The image extracted from video clips includes a black region on the left and right sides and a tool annotation UI on the bottom. Firstly, these black regions were removed. Then, UI cropping and/or UI blurring with a probability of 0.5 were randomly conducted. These procedures prevented the model from getting the tool presence information from the UI directory. Examples are shown in Fig. \ref{fig:1}.

    \item data augmentation: several data augmentation techniques were applied for achieving robust training. Specifically, the team applied the data augmentation techniques shown in Table \ref{tab:aug}. These augmentations were applied to train all models.
    
    \item model inputs: RGB image.
    
    \item model outputs: For category 1, multilabel output. For category 2, pixel-wise heatmap.
    
    \item model loss function: For category 1, binary cross-entropy loss.
    For Category 2, the model trained by Category 1 was used.
    
    \item model pre-training:  For category 1, The weights of Swin Transformer V2 and EfficientNetV2 were initialized with the weights trained with ImageNet \cite{deng2009imagenet}, and that of ResNeXt was initialized with the weights trained with Instagram Datasets.
    For Category 2, the model trained in Category 1 was reused.
 
    \item post-processing: For category 1, test time augmentation (TTA) was employed (Table \ref{tab:aug}). For category 2, the gradient for the last convolutional layer and the penultimate convolutional layer was computed via back-propagation. After obtaining the pixel-wise heatmap for each layer, these two heatmaps are averaged to obtain the final heatmap. Then, the heatmap is binarized with threshold $\sigma=0.5$ to obtain the binary map. Finally, the connected component algorithm is applied in order to extract multiple bounding boxes from the binarized heatmap.

    \item Hyper-parameters (Table \ref{tab:hyper_hvrl}): For Category 1, Stochastic Gradient Descent was employed as the optimizer, with the momentum $0.9$, weight decay $0.0001$, and cosine annealing strategy. For the learning rate, batch size, and training epochs, different settings on three methods were used as shown in Table \ref{tab:aug}. For Category 2, the model trained in Category 1 was used, thus retraining was not conducted.
\end{itemize}

\begin{table}[tbh]
\centering
\caption{HVRL: Data augmentation}
\begin{tabular}{ccc}
\hline
Augmentation & Range & Probability \\ \hline
Horizontal flip & & 0.5\\
Color shift & 30 & 0.5\\
HSV Shift & 20 & 0.5\\
Brightness &[0.1, 0.3] & 0.2\\
\hline
\label{tab:aug}
\end{tabular}
\end{table}

\begin{table}[tbh]
\centering
\caption{HVRL: Hyperparameters}
\begin{tabular}{cccc}
\hline
methods & learning rate & batch size & epochs \\ \hline
Swin Transformer V2  & 0.6  & 24 & 50 \\
EfficientNetV2 & 0.57 & 48 & 30 \\
ResNeXt  & 0.36 & 192 & 50 \\
\hline
\label{tab:hyper_hvrl}
\end{tabular}
\end{table}


The trained model outputs a multi-class label. 
The Grad-CAM++~\cite{8354201} module generated the class activation maps. Then, the generated class activation maps with threshold $\sigma=0.5$ generated the binary map. Finally, a connected component algorithm was applied in order to extract multiple bounding boxes from the binarized heatmap.


\subsubsection{Preliminary Performance}

\begin{table}[tbh]
\centering
\caption{HVRL: Local validation results with different models.}
\begin{tabular}{c|c}
\hline 
Method & F1-score \\ \hline
Swin Transformer V2 &  0.925 \\
EfficientNetV2  & \textbf{0.930}\\ 
ResNeXt & 0.928 \\
Model ensemble & \textbf{0.930} \\
\hline
\end{tabular}
\label{tab:res}
\end{table}

Table \ref{tab:res} summarizes the results against the local validation data. Models which showed the best performance on the local validation data were chosen. Specifically on epochs 17, 25, and 23. The table shows that a single EfficientNetV2  and the ensemble of three models achieved the highest F1-score.









\subsection{Muroran Institute of Technology - Team SK} 

Team Members: Satoshi Kondo, Satoshi Kasai, and Kousuke Hirasawa

\subsubsection{Method Description}


Team SK proposed a weakly-supervised surgical tool localization method with a multiplicative feature fusion network. The network was trained to learn the presence of surgical tools in the input image with supervised learning. When detecting surgical tools (task in Category 1), this method predicts the presence of surgical tools by using the trained network. When localizing surgical tools (task in Category 2), this implementation first predicts the presence of surgical tools and then checks the class activation map corresponding to the predicted surgical tool, localizing the tool by thresholding the activation map, i.e., in a weakly-supervised manner.

\begin{figure}
\centering
\includegraphics[width=0.8\linewidth]{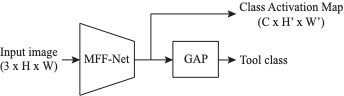}
\caption{SK: Overview of our proposed method.}
\label{fig:overview_sk}
\end{figure}

 Figure \ref{fig:overview_sk} shows the entire structure of the proposed network. Multiplicative Feature Fusion Networks (MFF-Net) \cite{SK} was used as the base network. MFF-Net is designed to aggregate features at different levels, which generates the class activation map (CAM) and categorical features for classification. Then, a CAM of the size $C \times H' \times W'$ is obtained. Here, $C$ is the number of classes in the tool classification, and $H'$ and $W'$ are the height and width of the CAM, respectively. ResNet-50 is used as the backbone of the MFF-Net. The training and prediction are performed frame-by-frame base although the input data is video.

The network was then trained by using the provided dataset, which has images and labels showing the presence of tools in the images. There were 14 surgical tools and 3 tools are shown in an image at most. Based on the team's observation that two ``needle driver'' tools are often shown in the same images, the model was trained not as a 14-class classification model but as a 15-class classification model, i.e., $C=15$. Since the dataset contains two ``needle driver'' classes in 15 classes. When only one needle driver is shown in the input image, only the first label in needle driver classes is active, i.e., give 1 as the ground truth label. When two needle drivers are shown in the input image, both labels for needle driver classes are considered active.

In inference, the probabilities of surgical tool presence were obtained from the CAMs through global averaging pooling and sigmoid activation. Tool classes whose probabilities are higher than the threshold (0.5) were selected. In this case, the selected tool classes were the results of the surgical tool detection task (Category 1). 
For the selected tools, this approach obtained CAM of the corresponding tool class in the order of the probability (from the highest probability to the lower ones). Then, the tool area was obtained by thresholding the CAM. The threshold was determined by using Otsu's binarization method. After identifying the bounding box surrounding the tool area, the IoU was calculated. If the IoU value is less than the threshold, the bounding box is employed as the result of the surgical tool localization (Category 2).

\subsubsection{Model Training}

For the training of the network, the team constructed a dataset by sub-sampling images from videos. The total number of images for training in this set was over 150k, being split into training and validation in an 8:2 ratio.

The augmentation techniques used were horizontal flip, shift, scale, rotation, color jitter, Gaussian blur, and Gaussian noise. The augmented images were resized to $640 \times 480$ pixels. The employed optimization method was Adam, and the initial learning rate was set to $1.0 \times 10^{-5}$ changing at every epoch with cosine annealing. The cross-entropy loss was used as the loss function. The batch size was set to 64 and the number of epochs to 30. The model taking the lowest loss value for the validation dataset was selected as the final model. Five models were trained by using 5-fold cross-validation, and CAMs obtained by five models were averaged in inference. An NVIDIA RTX3090 GPU for the training.

\subsubsection{Preliminary Performance}

When using the evaluation system provided by the challenge organizers, the F1 score was 0.816 for Category 1 (surgical tool detection). For Category 2 (surgical tool localization), mAP was 0.03.




\subsection{Team VANDY-VISE}
Team Members: Xing Yao, Ange Lou, Hao Yang, Jintong Han, Jack Noble, Jie Ying Wu\\


The goal was to take full advantage of the auto-correlation and cross-correlation between various frame-to-frame and video-to-video to solve this weakly supervised recognition problem. Therefore, an attention-based learning framework was selected to capture and detect causal features that were directly related to surgical tool detection. No public dataset apart from the challenge dataset was used in this work.

\subsubsection{Method Description}

\begin{figure}[htp]
    \centering
    \includegraphics[width=0.8\textwidth,height=0.5\textwidth]{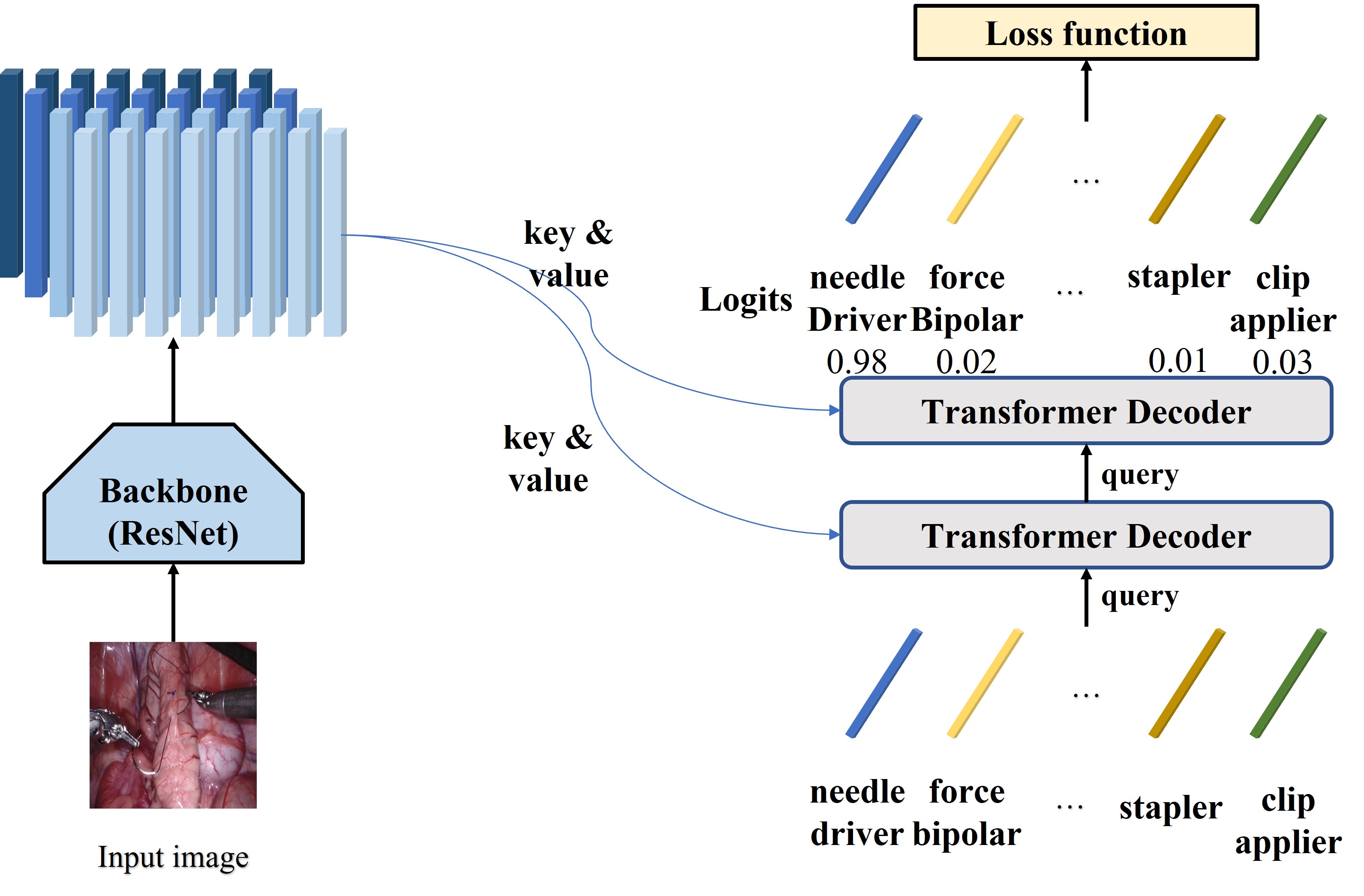}
    \caption{VANDY-VISE: Architecture of Query2Label \cite{liu2021query2label}}
    \label{fig:network}
\end{figure}

The VANDY-VISE team leveraged the same framework as Query2Label\cite{liu2021query2label} to do multi-label classification and localization (shown in Figure \ref{fig:network}). The input image $x$ was first sent through a ResNet backbone \cite{he2016deep} to extract the spatial features. The extracted spatial features were then sent into the transformer along with each of the label embeddings. The transformer decoders generated attention maps by comparing label embeddings with spatial features,  then pooling the spatial features by linearly combining these spatial features based on the attention map. After obtaining spatial features of the input image, each label embedding was sent to the transformer decoder to query and pool the desired feature. Finally, those features went through a linear projection layer to compute the logits for each class.

For a given image $x\in R^{H\times{W}\times{3}}$, a ResNet backbone was used to extract spatial features $f_0 \in R^{H_0\times{W_0}\times{d_0}}$ from the input image. Here, $H\times{W}$ and $H_0\times {W_0}$ represented the height and width of the input image and extracted features respectively, while $d_0$ denoted the dimension of features. The spatial features were then projected from $HW\times d_0$ to $HW\times d$ by using a linear projection layer, where $d$ was the desired feature dimension.

The class embedding was applied as a query that had dimension $Q^{K\times d}$. Here $K$ represented the number of classes. Then the cross-attention was calculated between the spatial features and queries to pool the class-related features by using multi-layer transformer decoders. 
In the implementation carried out by the Vanderbilt University team, the standard transformer decoder comprised cross-attention, self-attention, and a position-wise feedforward network.
The transformer decoder, $i$, updated the queries $Q_{i-1}$ as following:
\begin{equation*}
    \operatorname{self-atten}: Q_i^{(1)} = \operatorname{MultiHead}(\tilde{Q}_{i-1},\tilde{Q}_{i-1},Q_{i-1})
\end{equation*}
\begin{equation*}
    \operatorname{cross-atten}: Q_i^{(2)} = \operatorname{MultiHead}(\tilde{Q}_{i-1}^{(1)},\tilde{f},f)
\end{equation*}
\begin{equation*}
    \operatorname{FFN}: Q_i = \operatorname{FFN}(Q_i^{(2))}
\end{equation*}
where the hat represented variables with position encoding.

The query feature $Q_i$ was sent to a linear projection layer to a logit value for each class.
\begin{equation*}
    p_k = \operatorname{Sigmoid}(W_k^T Q_{i,k} + b_k)
\end{equation*}
in the right-hand-side of the function, $W_k$ and $b_k$ were parameters of the linear projection layer. The left-hand side, $p_k$, was the predicted probability of each class.

In the implementation, a simplified asymmetric loss \cite{Ridnik2020} was applied as shown below:
\begin{equation*}
    L = {1\over K} \sum_{k=1}^{K} \left\{\frac{(1-p_k)^{\gamma+}log(p_k),y_k=1}{(p_k)^{\gamma-}log(1-p_k),y_k=0}\right\}
\end{equation*}
here, $y_k$ was a binary label for $k$ classes, which corresponds to the $\gamma$ value to choose. By default, they are set to $\gamma+=0$, $\gamma-=1$. It was noteworthy that countless noisy labels existed in the training annotations, and it was very time-consuming to remove them manually. 
Most of the label noise came from pseudo-labels that were not present in the videos.
Based on this prior knowledge, during the training phase of the classification model, for videos with more than two labels, the team removed the largest loss among all existing labels. Compared with the original loss, this method improved the best validation mAP from 0.953 to 0.966.

In this work, the VANDY-VISE team post-processed the cross-attention weights created by the transformer decoder to generate bounding boxes for each predicted class. The post-processing included sparse matrix processing and bounding box generation based on OpenCV \cite{bradski2000opencv}. 
%
During the attention visualization process, the cross-attention weights matrix $M$, produced by the decoder, was initially reshaped from dimensions $B \times C \times HW$ to $B \times C \times H \times W$. This transformation enabled the representation of cross-attention between each pixel and its respective class.
Then, according to the classification results of the model, the classes in the normalized attention map were filtered, and only the classes with predicted True were retained. Because tools of different categories shared similar morphological features, the attention maps of different categories still had similarities. To further distinguish the critical vector of each class, for different positions of the same class, this method selectively retained one or several positions with the highest energy; for the exact position of different classes, it only retained the position corresponding to the class with the highest energy. Finally, the team binarized the sparse attention maps and used cv2.findContours and cv2.boundingRect \cite{bradski2000opencv} functions to generate corresponding bounding boxes.

\subsubsection{Model Training}

To make the problem more tractable, one frame from each video clip was randomly selected to build a relatively smaller dataset, which in total contained 24695 images. From this whole set, 3000 images were randomly partitioned as the validation set, and the rest 21695 frames were assigned to the training set. Blank areas in each frame were removed. Furthermore, AdamW was chosen as the optimizer with a learning rate of 1e-4. The training batch size was 32. All experiments were based on Pytorch and trained with a single RTX A5000 GPU.

\subsubsection{Preliminary Performance}

In multi-label classification tasks, the loss decreased steadily during the training phase, and 
the Vanderbilt University team achieved 0.966 mAP and 0.955 F1 scores on their validation set and a 0.6202 F1 score on the given test set.
For the localization task, their algorithm (multi-head attention weights map) could localize all surgical tools that appear in the given validation video.




\subsection{University Medical Center Hambrug-Eppendorf - Team UKE}

Team Members: Maximilian Nielsen, Samuel Schüttler, Thilo Sentker, Hümeyra Husseini, Ivo Baltruschat, Rüdiger Schmitz, Ren\'e Werner


\subsubsection{Method Description}

Recently, self-supervised deep learning gained attention by setting new benchmarks for different image classification tasks, showing potential for further investigation in related domains \cite{DINO,iBOT}. Different to supervised methods, self-supervision does not directly produce a meaningful output (e.g., a classification result or segmentation mask), but solely a representation, i.e. a feature-vector, of the input. To find an expressive representation of the input, such methods predominately rely on contrastive learning, which is characterized by forcing the model to output similar representations for different augmented versions of the same input. Advantages compared to supervised learning approaches are a relative strong performance with a fraction of labeled data and a high degree of generalizability.
Therefore, a self-supervised approach for both tasks of the SurgToolLoc challenge was employed. More specifically, deep image representations are extracted by self-supervised vision transformers (DINO). Subsequently, downstream classification is performed by a machine-learning classifier that is fitted on the model output of a smaller subset of labeled data. For tool localization, the individual outputs of the trained attention heads to generate bounding boxes were exploited. 

\begin{figure}[t!]
    \centering
    \includegraphics[width=1\textwidth]{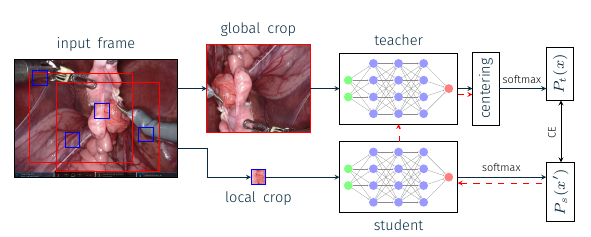}
    \caption{UKE: DINO-based self-distillation. Global and local crops are extracted from the input frame, individually augmented and fed into the teacher and student model, respectively. Note that a temporal global crop was used, i.e. the global crop is randomly extracted out of one frame that is in temporal correspondence to the input frame ($\pm$5\,s) to achieve temporal consistency. Further, the applied softmax function has a temperature parameter to control the sharpness of the output probabilities. After training, a machine-learning classifier is fitted on the DINO model output of a smaller subset of labeled data. For more details cf. the corresponding DINO-publication \cite{DINO}.}
    \label{fig:figure_1}
\end{figure}

For both tasks, the starting point was to extract feature vectors exploiting a self-supervised learning approach based on DINO (cf. figure \ref{fig:figure_1} for a schematic illustration). Prior to training, frames were resized to $308\times208$\,px and frame intensities normalized to ($-1,1$). A DINO model, pre-trained on the ImageNet data, was trained on the given pre-processed training data with slightly adapted default DINO-parameters. 

Several image augmentation techniques were utilized to improve the robustness of the model. The DINO augmentation techniques included random cropping to simulate different perspectives, solarization to introduce variations in brightness, flipping to incorporate left-right reversal, Gaussian blurring to mimic real-world blurs and color jittering to introduce variations in color.

A temporal global crop was applied to consider the temporal components of the input data. This approach involves randomly selecting a global crop from a frame that is in temporal correspondence with the input frame, with a tolerance of $\pm 5$\,s. This approach ensures temporal consistency between the input frame and the selected global crop, i.e. the temporal structure of the data is preserved which further diversifies the input data.

The loss function used was the DINO loss, which is a combination of temperature-weighted cross-entropy and feature normalization. The temperature-weighted cross-entropy term ensures a sharp feature distribution in the latent space, while the feature normalization term stabilized the training process and improved generalization by enforcing uniformity.

For task 1, a machine-learning classifier (logistic regression) was fitted on the DINO model output of a smaller subset of labeled data to learn the relationship between extracted feature vectors and corresponding training labels. Hyper-parameter optimization, feature selection and decision threshold optimization using a differential evolution algorithm were performed.

For task 2, a linear combination between multi-head attentions, RGB-values and minimum RGB-value to gray-value-distances was optimized to match the EndoVis 2017 instrument segmentations (shaft, tip, clevis) \cite{Allan2019-2017}. Subsequently, bounding boxes for the three largest connected components were extracted from the resulting segmentations. For each crop, a pseudo-label was created from possible ground-truth values and similar other crops (DINO latent space). The tool classifier was trained as in task 1.

\noindent 

\subsubsection{Model Training}

The model was trained for a total of 30,000 epochs using a batch size of 224 on a GPU server equipped with an A40 GPU, 1TB of RAM, and a CPU with 64 cores. Model training took about 35 hours. 

To ensure a deterministic output and a fair comparison between different runs, a fixed random seed, i.e. sum of int32 interpreted sha256-hash of string ``STL-Challenge'', was used throughout model training. Thus, consistent results for each run were obtained and it was possible to compare the performance of different models in a controlled environment. 

\subsubsection{Preliminary Performance}

Applying the proposed algorithm on the self-assessed validation split containing 30\% of the provided video data yielded a F1-score for task 1 of \textbf{0.82}. For task 2, validation was not possible as ground truth values were not available.  







\subsection{Furtwangen University - Team ITeM}

Team Members: Tamer Abdulbaki Alshirbaji, Nour Aldeen Jalal, Herag Arabian, Ning Ding, and Knut Moeller\\


The ITeM team is composed of four PhD students who are currently working under the guidance of Prof. Knut Moeller at the Institute of Technical Medicine, Furtwangen University, Germany. ITeM team members are involved with research projects in collaboration with industrial and clinical partners. These projects aim to develop and integrate smart technologies into the current setup inside the operating rooms (OR). Analyzing surgical videos using deep learning approaches, particularly surgical workflow recognition, represents a core topic that has been addressed by the researchers at ITeM. Spatio-temporal deep learning approaches for surgical tool classification and phase recognition have been developed. Additionally, data fusion approaches have also been investigated on data from surgery and anesthesiology. Therefore, participation in the \textbf{SurgToolLoc} challenge represents a great opportunity for the ITeM team to evaluate developed approaches on a new challenging dataset and to be in contact with other highly qualified teams from around the world.

\subsubsection{Method Description}

The SurgToolLoc challenge addresses supervised tool classification and weakly-supervised tool localization tasks in laparoscopic videos. In both tasks, only binary labels for tool presence can be used to train the proposed model. Similar to other public datasets (e.g., Cholec80 \cite{Twinanda2016}), surgical tool presence is imbalanced, which will cause a bias on the trained model to high-representative tools. Additionally, tool classification is a multi-object classification task, where up to 4 tools may appear in the laparoscopic image. Moreover, the model will be tested with images from video clips that are not included in the training data.

\begin{figure*}[ht]
\begin{center}
\includegraphics[width=\textwidth]{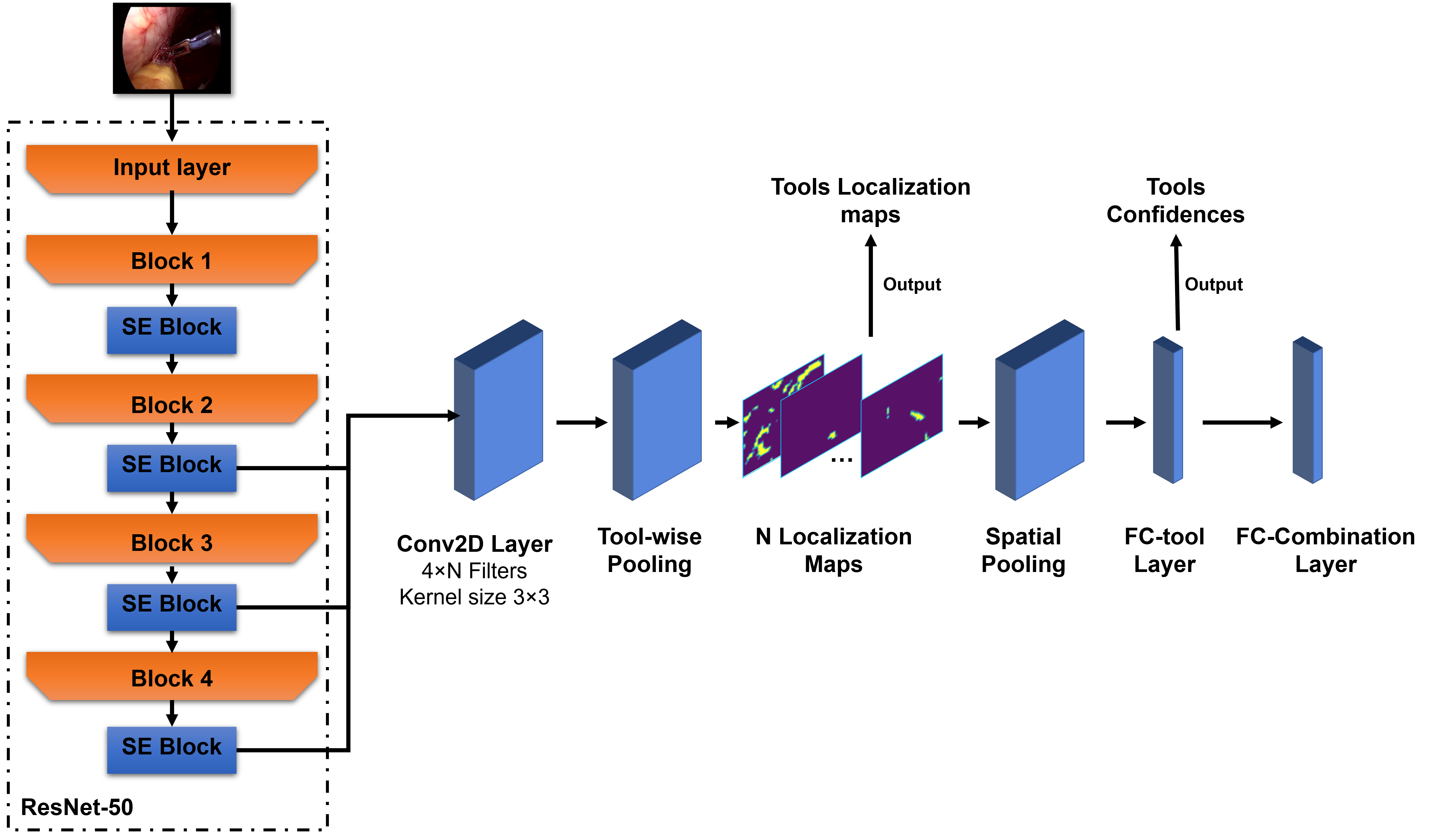}    
\caption{ITeM: The complete pipeline of the proposed model for tool presence detection and localization. SE blocks represent the squeeze-and-excitation attention modules; the Conv2D layer is the multi-map localization layer; FC-tool and FC-Combination layers are the fully-connected layers employed to obtain confidence values of the tool presence and tool-combination, respectively. N is the number of tool classes defined in the dataset (N=14).} 
\label{fig:Method}
\end{center}
\end{figure*}

A deep learning pipeline that addresses all the above-mentioned challenges was developed. The model consisted of the ResNet-50 architecture as a backbone model since it showed better performance over other CNN architecture for the tool classification task \cite{alshirbaji2021deep}. For weakly-supervised tool localization, the algorithm was built upon the work of Durand et al. \cite{durandWildcatWeaklySupervised2017a} and the follow-up approaches by Vardazarya et al. \cite{vardazary2018a} and Nwoye et al. \cite{nwoye2019a}, but introduced novel modifications. First, attention modules were adopted to the base CNN model to help generate better-refined features. Second, features from multiple stages (i.e., from lower and top layers) were fused for a better representation of laparoscopic images' contents. Third, a novel spatial pooling was applied to obtain tool presence confidence from the localization maps. Finally, a new loss function was proposed to model possible tool combinations of each video frame (Figure \ref{fig:Method} shows the architecture of the proposed model).

\begin{itemize}
    \item architecture design:
    \begin{itemize}
    \item{base model}\\
    The proposed architecture was built on the base CNN model ResNet50 \cite{he2016deep}. This model showed competitive performance for surgical tool classification in laparoscopic images \cite{alshirbaji2021deep}. ResNet50 consists of five convolutional blocks and on top a global average pooling and a fully-connected layer. The shortcut connections used in the model lead to more efficient optimization and higher accuracy \cite{he2016deep}. The model input was increased to 375×300 instead of 224×224. Additionally, the stride of convolutional layers was set to 1x1 in the last two blocks. Due to these modifications, higher spatial resolution is preserved at deeper layers.    
    
    \item{squeeze-and-excitation module}\\
    Squeeze-and-Excitation (SE) \cite{huSqueezeandexcitationNetworks2018} attention modules were integrated into the convolutional architecture to boost the model performance. Four SE modules were added after the fourth, seventh, thirteenth, and sixteenth convolutional blocks. The reduction ratio was set to 16 in all SE modules.
    
    \item{multi-stage feature fusion}\\
    In conventional methods, the CNN model performs the classification based on information learned at the last layer. In fact, features of deeper layers encode semantic information about classes. On the other hand, features of shallow and intermediate layers encode some generic information. Feeding such information for instance to a classification layer can support the classification decision \cite{alshirbajiImprovingGeneralisabilityDeep2022a}. Thus, in this approach, features at the output of the second, third, and fourth attention modules were combined forming a feature bundle. The combined features were passed to a batch-normalization layer.    
    
    \item{Conv2D layer}\\
    A convolutional layer was used to transfer the multi-stage features to a localization map for each class. The convolutional layers had a kernel size of 3x3. The number of filters was set to MxN, where M is the number of maps for each class and N is the number of tool classes (N=14). M was set to four, thus, four localization maps were generated for each surgical tool. In these convolutional layers, the stride was set to 1x1 to preserve the spatial resolution.    
    
    \item{pooling operations}\\
    To generate one localization map and presence confidence for each tool, tool-wise pooling, and spatial pooling operations were applied consecutively. The max operation was applied across the 4 localization maps of every tool. The input and output of this operation are $w{\times}h{\times}4N$ and $w{\times}h{\times}N$, respectively, where $w$ and $h$ represent the spatial dimension of the obtained feature maps, while $N$ is the number of tool classes in the dataset.

    Special spatial pooling (equations \ref{eq:PER}, \ref{eq:NER}) was then applied to transform the map of each tool into two features. These two features represent the positive evidence ratio (PER) and the negative evidence ratio (NER), and they were provided into a fully-connected layer with a Sigmoid activation to obtain tool-wise confidence values.

    \begin{equation*} \label{eq:PER}
        \begin{aligned}[c]
         PER = \frac{\sum_{i}^{h}\sum_{i}^{w}x_{i,j}}{h.w}
        \end{aligned}
        \qquad\Longleftrightarrow\qquad
        \begin{aligned}[c]
          x_{i,j} = 1 & \text{  for  } x>0.9\\
          x_{i,j} = 0, & \text{  Otherwise}
        \end{aligned}
    \end{equation*}

    \begin{equation*} \label{eq:NER}
        \begin{aligned}[c]
         NER = \frac{\sum_{i}^{h}\sum_{i}^{w}x_{i,j}}{h.w}
        \end{aligned}
        \qquad\Longleftrightarrow\qquad
        \begin{aligned}[c]
          x_{i,j} = 1 & \text{  for  } x<0.1\\
          x_{i,j} = 0, & \text{  Otherwise}
        \end{aligned}
    \end{equation*}

    where $h$ and $w$ are the spatital dimension of the localization maps, $x_{i,j}$ is the pixel value at $i$ and $j$ position.

    \item{fully-connected (FC) layers}\\
    Two fully-connected (FC) layers were added on top of the spatial pooling layer. The first FC layer (FC-tool) performs the tool presence detection and contains 14 nodes since 14 tools are defined in the dataset. The confidences obtained from the FC-tool were then provided into the second FC layer, termed FC-Combination), which was added to model the possible tool combinations in the dataset. Therefore, tool-combination classes were generated for all images prior to training the model. First, all possible tool combinations were extracted from the provided tool classes with their corresponding distributions. Then, 14 tool combination classes were generated. The first 13 classes represent the combinations of the highest thirteen distributions and the last class represents the remaining combinations.

    \end{itemize}
    
    \item data pre-processing:
    The video clips provided had a frame rate of 60 frames per second (FPS). Training images were extracted at 1 fps. Since some clips contained black frames, these frames were identified and excluded. The label provided for every clip was assigned to its corresponding frames.
    
    The image pre-processing included cropping out the black sides of the images, blurring the user interface (UI) information appearing at the bottom of the images, resizing the image into the model input (375×300×3), and image normalization.
    
    \item model input:
    The input of the model is an image of size 375×300×3.
    \item model outputs:
    The outputs of the model are confidences of the 13 tool categories defined in the dataset and localization maps of detected tools. 
    \item model loss function:
    Two loss functions were employed in this study. The binary cross-entropy function was used to compute the loss of each tool class, as in Eq. \ref{eq:tool_loss}. To reduce the imbalanced data effect, loss-sensitive learning approach \cite{abdulbaki2018surgical} was applied. Here, the loss of each tool was multiplied by a weighting factor that was estimated according to the tool-appearance distribution in the training data. Additionally, the Softmax cross-entropy loss was utilized to calculate the classification error for the tool combination (Eq. \ref{eq:loss_comb}).
    
    \begin{equation*} \label{eq:tool_loss}
    Tool_{loss} = \frac{-1}{B}\sum_{n=1}^{B}\sum_{t=1}^{T} w_t[l_t^n\log{(C_t^n)} + (1-l_t^n)\log{(1-C_t^n})]
    \end{equation*}
    
    where $Tool_{loss}$ is the total loss of all tools, $B$ is the batch size, $T$ is the number of tools in the dataset, $w_t$ represents the loss-weight calculated for every surgical tool, $l_t^n = [0,1]$ is the tool presence ground truth, and $C_t^n$ is the tool presence confidence obtained from the spatial pooling operation.

    \begin{equation*} \label{eq:loss_comb}
    loss_{comb.} = \frac{-1}{B}\sum_{n=1}^{B}\sum_{c=1}^{C}G_c^n\log{\sigma(C_c^n)}
    \end{equation*}
    
    where $loss_{comb.}$ is the total loss of tool-combination category, $B$ is the batch size, $C$ is the number of defined tool-combination in the dataset, $G_c^n$ is the ground truth of image $n$, $C_c^n$ is the output of the FC-Combination layer, and $\sigma$ represents the softmax activation function.

    \item model pre-training:
    The base ResNet-50 model was pre-trained with the Cholec80 \cite{Twinanda2016} dataset. The weights of the newly added layers were randomly initialized.

    \item post-processing:
    post-processing step was required to extract bounding boxes from the localization maps generated after applying the tool-wise pooling operation. Localization maps of the tools that were detected to be present in the image (confidence \textgreater 0.5) were extracted. A Gaussian filter with sigma = 1.0 was applied to smooth boundaries in the localization maps. Then, a binary mask was generated by thresholding the localization map with a threshold of 0.5. Bounding boxes were finally generated on top of generated masks. For the \emph{needle driver} tool, up to two bounding boxes were considered.
    
\end{itemize}

\subsubsection{Model Training}
The Adam optimizer with cyclical learning rate techniques was utilized. The cyclical learning rate boundaries were chosen as 0.005 and 0.001, and a step-size of $4\times iteration-per-epoch$ was chosen. The model was trained with a batch size of 50 images for 50 epochs. Images were shuffled for every epoch. All implementations and experiments were accomplished in Keras framework with Anaconda virtual environment platform and run on a desktop with an AMD Ryzen Threadripper PRO 3955WX 16-Cores 3.90 GHz, 512 GB RAM, and 64-bit Windows 10 Operating System with an NVIDIA RTX A6000 graphics processing unit (GPU).

\subsubsection{Preliminary Performance}
\begin{figure*}[ht]
\begin{center}
\includegraphics[width=\textwidth]{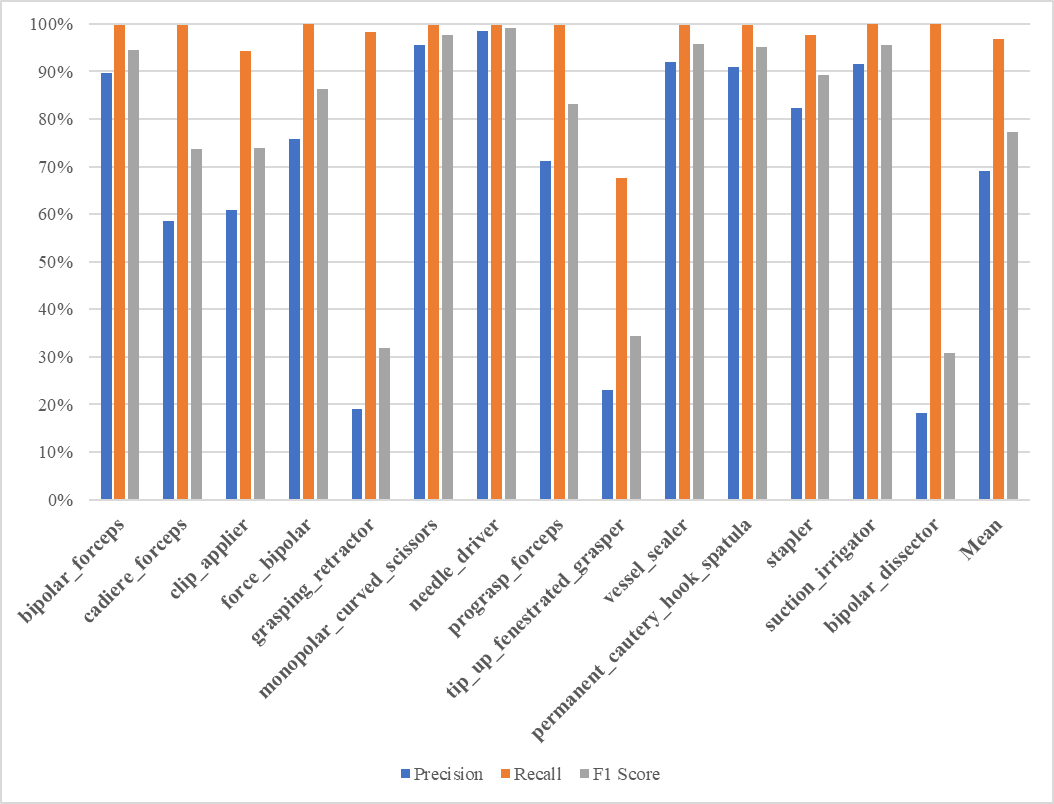}    
\caption{ITeM: Model performance on the validation data.} 
\label{fig:Results1}
\end{center}
\end{figure*}

To evaluate the trained models prior to submission, a small, denoised-label dataset was created by manually revising the provided labels. Figure \ref{fig:Results1}  shows the performance of the proposed model for surgical tool classification on the validation set. Additionally, the results of different models on the preliminary-test data are presented in Table \ref{tb:Results2}.

The experimental results of the proposed model showed high performance on both the validation and preliminary test data. However, an extensive ablation study is still required to show the advantage of every component of the model. Moreover, additional work should be done to denoise the tool labels provided.

\begin{table}[h!]
\begin{center}
\caption{ITeM: The results of the different models obtained from the preliminary test}\label{tb:Results2}
\begin{tabular}{|l|l|lll}
\cline{1-2}
\textbf{Model}                        & \textbf{F1-Score (\%)}  \\ \cline{1-2}
\textbf{Base model}                   & 51.76                   \\ \cline{1-2}
\textbf{Base model+Conv2d+Pooling}    & 76.23                   \\ \cline{1-2}
\textbf{Base model+Conv2d+Pooling+SE} & 77.51                   \\ \cline{1-2}
\textbf{Final Model}                  & 84.35                  \\ \cline{1-2}
\end{tabular}
\end{center}
\end{table}





\subsection{Shun Hing Institute of Advanced Engineering - Team Medical Mechatronics (MM)} 

Team Members: An Wang, Mengya Xu, Mobarakol Islam, Long Bai, Winnie Pang, Hongliang Ren



\subsubsection{Method Description}

Multi-label tool classification from surgical videos is challenging but demanding to increase the autonomy of computer-assisted interventions. The method from Medical Mechatronics (MM) utilized the hybrid ViT~\cite{b1} model (R50-ViT-B\_16)~\footnote{https://github.com/mobarakol/Hybrid\_ViT} to complete the classification task. Further, the team considered solving the noise and imbalance issues of the dataset by intentionally splitting the dataset into training and validation and with an unbalanced dataset sampler during training. They also applied augmentations like Cutout~\cite{b2}, and RandAugment~\cite{b3}, to increase generalization. 

There are always dataset issues like noisy labels, imbalanced datasets, and domain shifts in the practical dataset. It is important to consider proper data cleaning and manipulation before feeding the data to the deep learning models. The entire training dataset consists of over 20 thousand videos. There are 14 types of tools in these videos. The frequency of their appearance is quite different, making the dataset imbalanced. The first frame from each valid video (except for 9 corrupted videos) was extracted to represent each video clip to reduce redundancy and save memory. The corresponding label of each frame is a one-dimensional array with a length of 14. The element of the respective class was set to 1 if the tool exists and 0 otherwise. The dataset was not randomly split into training and validation sets. Instead, the team analyzed the dataset and ensured all tools existed in the training and validation periods. The number of training samples was about 80\% of the entire dataset, and the number of each class of samples for training was also around 80\%.

\begin{figure}[!hbpt]
\centering
\includegraphics[width=0.8\linewidth]{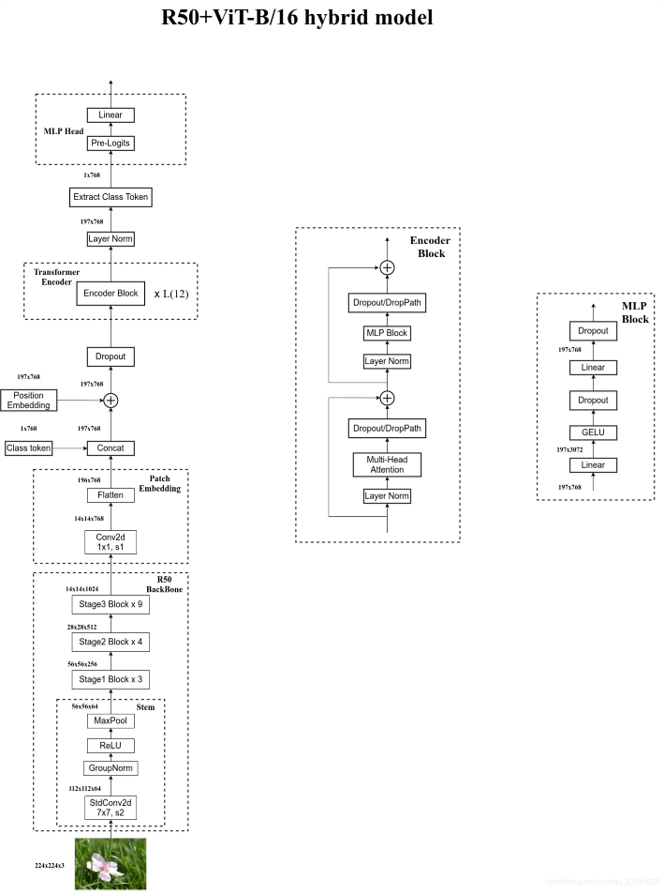}
\caption{MM: R50+ViT-B\_16 hybrid model}
\label{fig:model}
\end{figure}

The training images were resized to 224X224, and two advanced augmentation techniques, i.e., Cutout and RandAugment, were incorporated as data augmentation to avoid overfitting. The PyTorch-based multilabel balanced sampler~\footnote{https://github.com/issamemari/pytorch-multilabel-balanced-sampler} was adopted to sample the samples during training. Asymmetric Loss (ASL)~\cite{Ridnik2021} was used as the multi-label classification loss function, which performs differently on positive and negative samples by automatically reducing the weight and hard limit of easy negative samples, and eliminating samples that could be mislabeled. Figure~\ref{fig:model} describes their model architecture.

\subsubsection{Model Training}
During training, SGD was used as the optimizer, and a cosine scheduler was applied to the learning rate (max learning rate was 3e-2) for 1000 warm-up steps. The total number of training steps was 10000. The hybrid ViT model was trained on one RTX3090 GPU with a batch size of 128, taking around 3.5 hours to finish the training. The balanced multi-label sampler in the dataloader helped achieve training-time class balancing. The evaluation interval was set to 50 steps, and the final model was acquired from the step with the highest validation MAP.

\begin{figure}[!hbpt]
\centering
\includegraphics[width=0.8\linewidth]{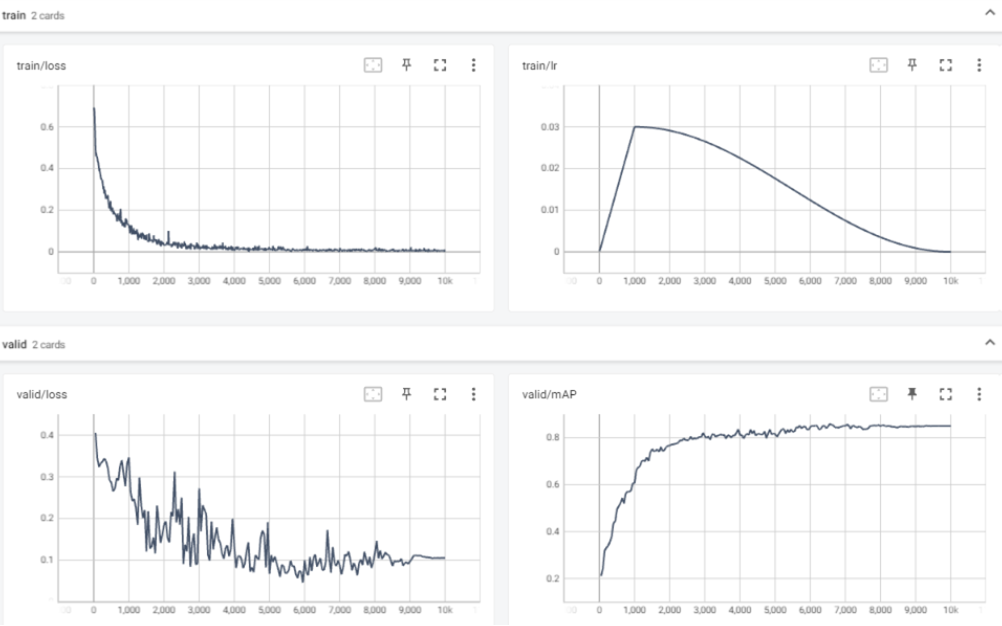}
\caption{MM: The training process.}
\label{fig:res}
\end{figure}

\subsubsection{Preliminary Performance}
An mAP of 0.86 was achieved on the validation dataset. Figure~\ref{fig:res} shows the training process.

\subsection{Bin Zayed University of Artificial Intelligence - Team BioMedIA}

Team Members: Aleksandr Matsun$\ast$, Mugariya Farooq$\ast$, Numan Saaed, Jose Renato Viera Restom and Mohammad Yaqub\\



The team was focused on the task of tool classification but they targeted the localization task as well considering the extent to which the task of localization could help the classification problem. The authors experimented with simple architectures like ResNet 50 first and planned to build on these simple architectures to increase the performance of the model. In the end, the team used a combination of two architectures: ConvNext and Yolov5 which helped them improve the performance metrics.

\subsubsection{Method Description}

The authors developed a method that localized different surgical instruments in surgical videos with a two-stage model, a detection foundation model based on YOLOv5 \cite{bochkovskiy2020yolov4} to derive the bounding boxes and a classifier based on the ConvNext \cite{liu2022convnet} architecture. 

\begin{itemize}
    \item \textbf{Architecture Design:} The authors proposed a two-stage framework by combining two standard architectures YOLOv5 and ConvNext. Yolov5 was used for the task of object detection. ConvNext, on the other hand, was used for image classification. The training was organized in a way that would reduce the effect of noisy labels.
    
    \item \textbf{Data Preprocessing:} According to the authors, the dataset posed many challenges due to the following reasons:
    \begin{itemize}
        \item Dataset Imbalance: As per the data exploration done by the authors, the imbalance in the dataset was huge. The videos contained some instruments which appeared thousands of times as compared to the others which appeared only tens of times as shown in Figure~\ref{fig:dimb}. The authors balanced the dataset by sampling videos with different frame rates taking into account the types of tools present in each video. They also optimized (minimized) the standard deviation of the frequency of each tool in the resulting dataset.
        \item Presence of banners: The banners showed the tools used in the surgery, and as proposed by the authors, the removal of the banners was crucial as it could have caused the model to learn the labels from the banners and not actually identify the instruments for their proper classification.
        \item Noisy Labels: The authors, upon careful inspection of the videos in the dataset, discovered that some tools might never appear in the video even though the label suggested otherwise, which made the labels noisy and the training challenging.
    \end{itemize} 
\begin{figure}
 \centering
 \includegraphics[scale=0.23]{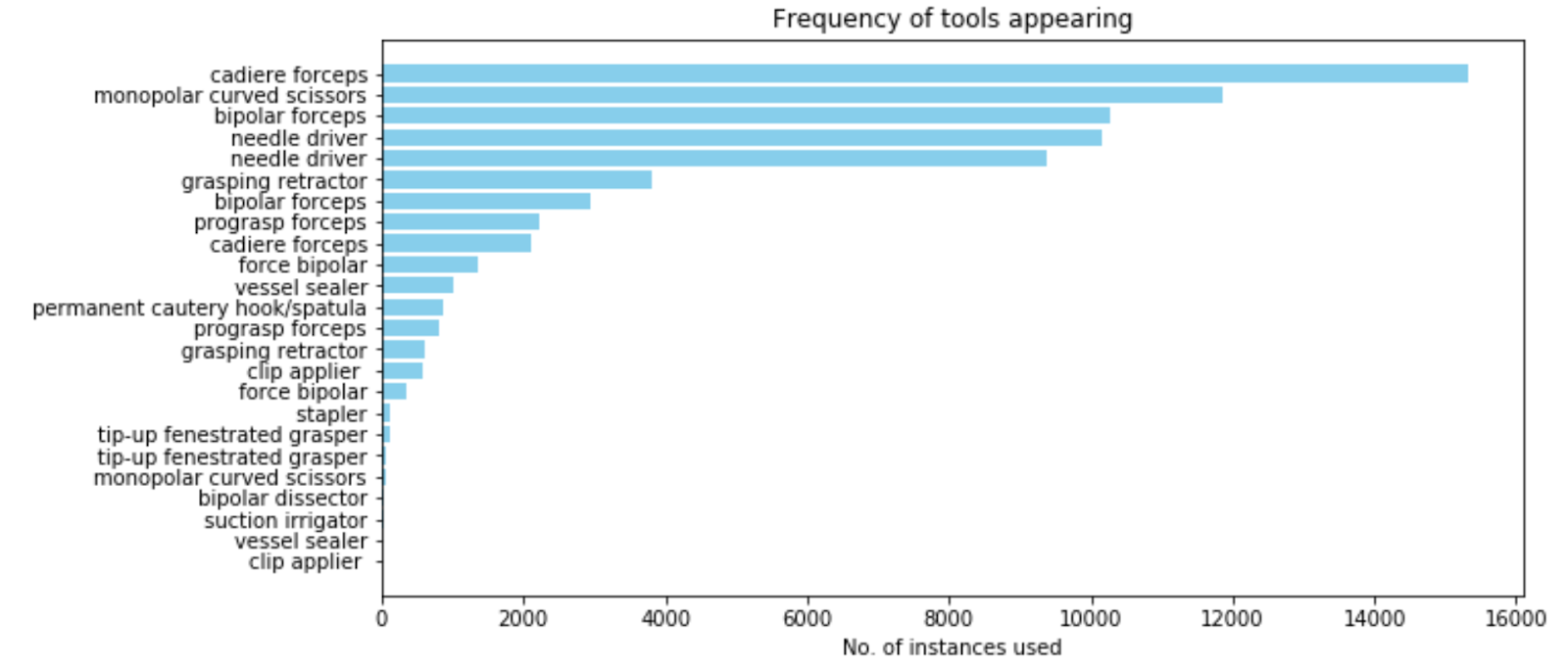}
 \caption{BioMedIA: Frequency of each tool across the whole dataset. The graph illustrates the frequent appearance of some tools compared to the other tools. This creates a huge imbalance in the dataset.}
  \label{fig:dimb}
\end{figure}
    
    \item \textbf{Data Augmentation:} Due to the nature of the dataset, where numerous frames were similar, the authors proposed different transformations that could help in training. Different types of augmentations were used for the detection model and the classifier model. For detection, the authors used mosaic augmentation, mixup augmentation, and random perspective transformations. On the other hand, for the classifier, the authors used random image flipping, rotation, and color jittering.
    \item Model Input(s): As an end-to-end model, the proposed framework was designed to have a video frame (RGB image) as an input. The frames were obtained by processing the video data provided.
    \item Model Output(s): For each of the input frames, the framework independently outputs a set of labels for surgical tools present in the respective frame.
    \item Model Loss Function: The authors used a combination of box regression loss and objectness loss in order to train the localization model and a focal loss as the criterion for the classifier in order to partially mitigate the impact of the huge class imbalance in the dataset.
    \item Model Pretraining: The authors maintained that pretraining the model on different datasets that had similar characteristics as the provided data would, to some extent, boost the performance of the models. They trained ResNet50 and ConvNext on Cholec 80 \cite{Twinanda2016} and M2CAI \cite{jin2018tool}. Cholec 80 dataset consists of 80 cholecystectomy surgeries. The labels present are primarily phase and tool presence annotations, where the ground truth for phase annotations is shown in Figure \ref{fig:cholec} and has been confirmed by a surgeon. On the other hand, the m2cai16-tool-locations dataset contains 15 videos with tool annotations for seven surgical instruments, as shown in Figure \ref{fig:m2cai}. The labels are given per frame for 7 classes for each video.

\begin{figure}
 \centering
 \includegraphics[scale=0.35]{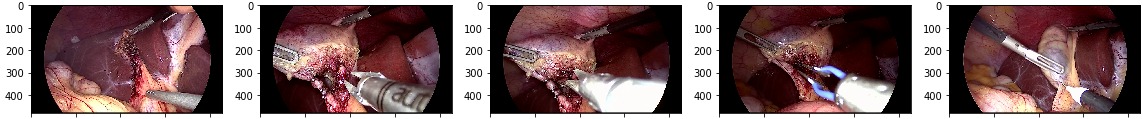}
 \caption{BioMedIA: Surgical instruments present in Cholec80 Dataset. The tools are similar to the tools present in the challenge dataset. }
  \label{fig:cholec}
\end{figure}

\begin{figure}
 \centering
 \includegraphics[scale=0.32]{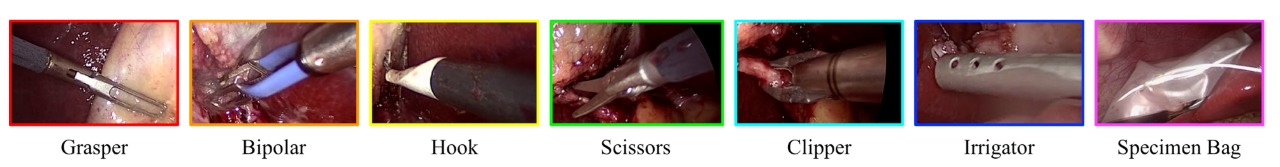}
 \caption{BioMedIA: Surgical instruments present in M2CAI Dataset. Some of the tools present are similar to the ones present in the given dataset.  }
  \label{fig:m2cai}
\end{figure}
\end{itemize}
For the task of classification, out of various CNN architectures, ConvNext-Small provided the best results. However, the authors made the following assumptions about the data:
\begin{enumerate}
    \item They considered and filtered frames where all three instruments were present. They assumed that these frames were sufficient for the task of classification. They also assume that the distribution of surgical tools in these frames would be approximately the same as that in the actual dataset.
    
    \item In the vast majority of frames, the order in which the tools are listed on the label of a clip corresponds to the counter-clockwise order of their positions, starting from the upper left corner and ending with the upper right corner.
\end{enumerate}
\textbf{Training algorithm:} The authors used the localization model Yolov5 pre-trained on a combination of three datasets:  EndoVis'15 Instrument sub-challenge Dataset, CholecSeg8k \cite{hong2020cholecseg8k} and Endoscopic instrument segmentation with crowdsourced data challenge dataset \cite{maier2014can} to detect the surgical tools in a frame with bounding boxes. The bounding boxes were saved as separate images alongside their respective image-level ID number derived from their position relative to the center of the image in an anti-clockwise fashion. The bounding boxes are then assigned the labels from the corresponding video clip (from left to right, depending on the id number of the bounding box). The outputs obtained were a set of predictions for the tools in the frame (Figure \ref{fig:archi}. 

\begin{figure}[h]
    \centering
    \includegraphics[scale=0.5]{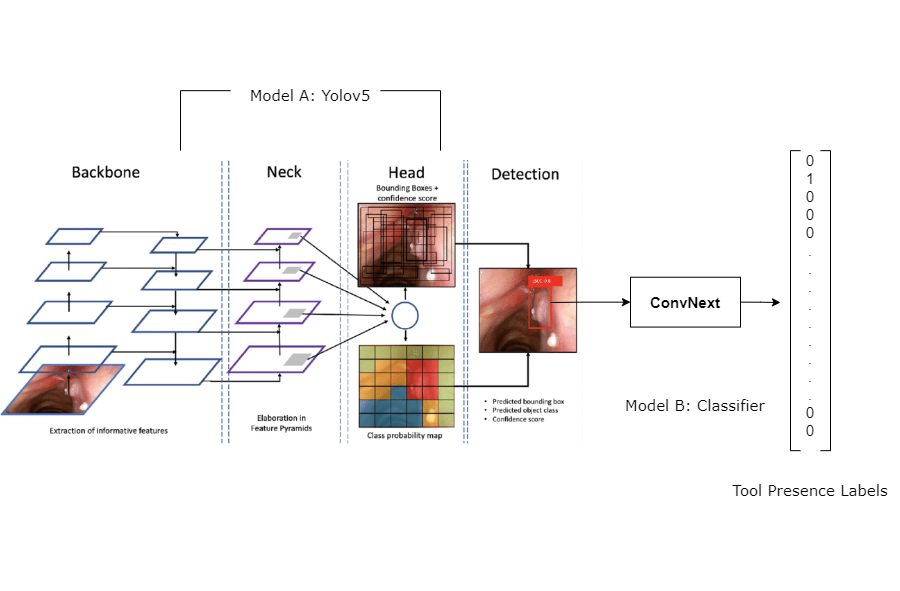}
    \caption{BioMedIA: Architecture of the two-tier model used for classification. Images are passed through Yolov5 for generating bounding boxes around the tools. Then they are passed to ConvNext for classification.}
    \label{fig:archi}
\end{figure}

\subsubsection{Model Training}
To train the detection model, the authors used a batch size of 128, a learning rate of 0.001, and a total number of epochs of 300. The classifier was using a batch size of 128, a learning rate of 0.0001, and the number of epochs was set to 5 due to the fast convergence of the network.

Given the size of the dataset, the authors used a cluster to run their experiments on a GPU with 2x AMD EPYC 7742 CPUs (128 cores total with 256 threads), 256GB RAM, 4x Nvidia A100 SXM 40G GPUs.

\subsubsection{Preliminary Performance}

The performance of the model during training was quite high with models like ResNet-50 and ConvNext. When pre-trained on the Cholec-80 dataset, ResNet produced a training F1 score of 0.989, while for ConvNext it was 0.985. Moreover, both of them produced around the same F1 scores when trained on the M2CAI dataset. The authors believe that the unrealistically high-performance metrics were due to the similarity across the frames in different surgeries and the usage of the same set of tools for different surgeries with the same background. The validation performance for ConvNext-Small and the proposed two-tier model is shown in Table~\ref{table:loss_exps}. However, the testing performance for the proposed architecture was around 0.552.

\begin{table}[h]
\caption{BioMedIA: Comparison of different models on the performance metrics. Models were pre-trained on Cholec 80 and M2CAI.}
\vskip 0.15in
\begin{center}
\begin{small}
\begin{sc}
\begin{tabular}{lcc}
\toprule
Model & F1-Score & Accuracy \\
\midrule
ResNet-50 & $0.20$ & $0.10$ \\
ConvNext-S  & $\textbf{0.83}$ & $0.39$ \\
Yolov5+ConvNext-S & $0.75$ & $\textbf{0.50}$ \\
\bottomrule
\end{tabular}
\end{sc}
\end{small}
\end{center}
\vskip -0.1in
\label{table:loss_exps}
\end{table}











\subsection{University of Tokyo - Team WhiteBox}

Team Members: Zhenqiang Li, Yoichi Sato\\


The authors introduced a network architecture for multi-label surgical tool classification from endoscopic videos. To address the challenge of subtle variance in tool appearance, the network utilizes feature maps from an intermediate layer of ResNet, which retain a higher level of spatial detail. Additionally, the long-tailed class distribution of tools is addressed by employing the asymmetric focal loss in place of the commonly used binary cross-entropy loss.

\subsubsection{Method Description}
The proposed network architecture comprises a feature map extraction module based on the ResNet-18 architecture and a feature post-processing module for multi-label tool classification. A significant challenge in this task is the subtle variance in the appearance of various tools. To address this issue, the network utilizes the feature maps generated by an intermediate layer of the ResNet architecture, as opposed to the final feature obtained after the 2D Average Pooling layer of ResNet. This practice is motivated by the observation that the feature maps obtained at this intermediate stage retain a higher level of spatial detail, thereby enabling the detection of tools with fine variance in appearance. Another critical challenge in this task is the long-tailed class distribution of tools, where the number of samples for the most frequent tool class in the training set is 1149 times that of the least frequent one. To address this imbalance in the training data, the network utilizes the Asymmetric Focal Loss (ASL)~\cite{Ridnik2021} in place of the commonly used binary cross-entropy loss as the training loss. 

\begin{figure}[ht]
\centering
\includegraphics[width=\textwidth]{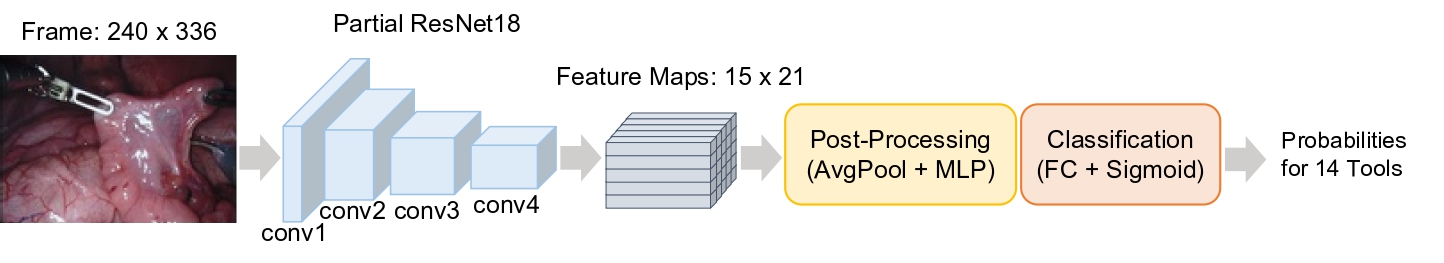}
\caption{WhiteBox: The tool presence classification framework used by WhiteBox team.}
\label{fig:framework}
\end{figure}

The overall framework for classifying tools' presence in each frame is illustrated in Figure \ref{fig:framework}. The network utilizes the cascaded first four convolutional blocks of ResNet to extract feature maps for each frame, which have a higher spatial resolution and retain more details in the frame compared to those extracted from the fifth convolutional block or the final average pooling of ResNet. A post-processing module, consisting of an average pool layer (AvgPool) and a Multi-Layer Perceptron (MLP), is utilized to aggregate features in spatial dimensions and conduct feature transformation. The presence probabilities for 14 tools are then predicted on each frame by feeding the post-processed feature into a fully-connected layer (FC) and a Sigmoid activation layer.

The network is trained via asymmetric focal loss (ASL) to overcome the long-tailed class distributions in the training dataset. ASL adapts focal loss for the multi-label classification task by decoupling the focusing levels of the positive and negative samples. Given $K$ labels, the network predicts one probability per label, $p_k$, whose ground truth is denoted by $y_k$. The total classification loss $L_{total}$ is obtained by aggregating the binary loss from $K$ labels as $L_{total}=\sum^{K}_{k=1}-y_{k}L_{+}(p_k)-(1-y_{k})L_{-}(p_k)$. The positive and negative loss parts can be formed as follows by following the focal loss:
\begin{equation*}
    \begin{cases}L_+=(1-p)^{\gamma}\log{(p)}, \\L_-=p^{\gamma}\log{(1-p)},\end{cases}
\end{equation*}
where p is one predicted probability. ASL incorporates two key modifications to the loss in order to improve the multi-label training effect. First, ASL assigns distinct weights, denoted as $\gamma_+$ and $\gamma_-$, to the positive and negative loss parts, respectively. By emphasizing the contribution of positive samples through setting $\gamma_- >\gamma_+$, ASL ensures that the network is able to learn meaningful features from positive but infrequent samples. Additionally, in order to reduce the ratio of very easy negative samples on the overall loss, ASL fully discards negative samples when their probability is very low by performing the hard thresholding as $p_m = \max (p-m, 0)$. Here $m\geq0$ is a tunable hyperparameter for the probability margin. In general, ASL is defined as follows:
\begin{equation*}
    ASL=\begin{cases}L_+=(1-p)^{\gamma_+}\log{(p)}\\
    L_-=(p_m)^{\gamma_-}\log{(1-p_m)}.\end{cases}
\end{equation*}

\subsubsection{Implementation Details}

During the training phase of the network, the video frames were resized to a height of 270 and a width of 480. Subsequently, the upper-middle regions (shaped in 240$\times$336) of the resized frames were cropped as input to the network. This cropping process removed the black regions present on both sides and the text labels located at the bottom of the frames. Each training sample for one video was formed by splitting it into 32 segments and randomly sampling one frame from each segment. At the network test phase, the continuous regions with pixel values less than 1, which represented the black background, were first identified and removed from the frames. The remained foreground was then resized to 270$\times$336, from which the upper 240$\times$336 regions were cropped as input to the network. All frames of a video were taken as input for making predictions at the test phase. The input RGB frames were normalized using the means of (0.432, 0.395, 0.376) and the standard deviations of (0.228, 0.221, 0.217). The network outputted the existence probabilities of all 14 tools, from which the first 3 classes having probabilities larger than 0.5 were selected as the predicted tool labels.

The framework employed the first four convolutional blocks of a ResNet-18 model pre-trained on ImageNet as the feature extractor. To preserve the high spatial resolution of the output feature maps, the max pooling layer following the first convolutional block is eliminated. The resulting feature maps have a spatial shape of 15$\times$21 and 256 channels for a 336$\times$240 frame. The MLP for post-processing preserves the feature channel number at 256. These features are finally fed through a linear transformation network and a Sigmoid activation layer to predict frame-wise tool existence probabilities.

The parameters of all components were trained in an end-to-end manner using a Stochastic Gradient Descent (SGD) optimization algorithm for a total of 60 epochs. The learning rate was initially established at 0.01 and decreased by a factor of 10 after the 40th epoch. The training process was conducted using an Nvidia A100-SXM4-40GB GPU, with a batch size of 18. Additionally, the weights for the positive and negative samples in the Asymmetric Focal Loss (ASL) were set as $\gamma_+=1.0$ and $\gamma_-=4.0$, respectively. Furthermore, a probability margin of $m=0.05$ was employed to constrain the easy negative samples.






\subsection{University of Strasbourg - Team CAMMA}
Team Members: Chinedu Nwoye, Luca Sestini, and Nicolas Padoy.

\subsubsection{Method Description}
To detect surgical tools from videos acquired from Da Vinci robotic system, a spatial attention network (SANet) was proposed. This was motivated by the characteristics and constraints of the provided dataset: (1) the videos were annotated with binary presence labels from the Da Vinci user interface (UI) - in this case, a set of tool labels was fixed for a video with a high level of inconsistency when considered at a frame level, (2) utilizing the UI information was not allowed for the SurgToolLoc challenge and would also be blurred for the test videos, and (3) the test labels would be frame-based, more accurate, and thus, not a fair representation of the provided training videos and labels.
Taking these limitations into consideration, the SANet model was designed to discover more useful features that could represent the tools' presence and yet with some level of tolerance to the noisy ground truths.

\begin{figure}[th]
    \centering    
    \includegraphics[width=.9\linewidth]{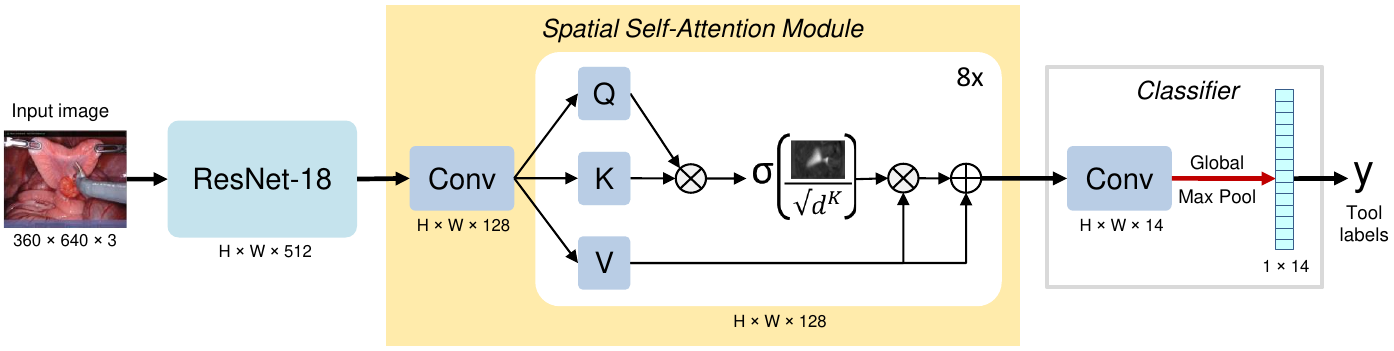}
    \caption{CAMMA: Architecture of the spatial attention network (SANet) for surgical tool presence detection}
    \label{fig:architecture-camma}
\end{figure}

The model and its implementation details are summarized as follows:

\begin{itemize}
    \item Architecture design: The SANet consists of 3 modules as shown in Figure \ref{fig:architecture-camma}: (1) a ResNet-18 \cite{he2016deep} backbone to extract spatial features from the input images, (2) a spatial attention module, consisting of 8 layers of a self-attention unit \cite{vaswani2017attention}, to highlight the discriminating regions of the input features for the tools with each feature channel constrained by designed to focus on a distinct tool class, and (3) a global pooling-based classifier \cite{nwoye2019weakly} to transform the attention features to a vector of class-wise probabilities for the 14 tool classes considered in the dataset.
    
    \item Data processing: The images were cropped to eliminate UI sections before being resized to a $360\times640$ spatial dimension because the video frames contain a sort of tool information from the Da Vinci UI that would not be available during testing.  
    
    \item Data augmentation: The input images were augmented using color jittering, flipping, random rotation, and random adjustment of brightness/contrast.
    
    \item Model input: During training, the model received pre-processed and augmented inputs, while during inference, it received solely pre-processed inputs.
    
    \item Model outputs: The output was a vector representing the log probability of each tool category appearing in the image. The output attention maps also included location-based activations for tool instances, which helped localize the tools in a weakly-supervised scenario.   
    
    \item Model loss function: The model was trained using weighted sigmoid cross entropy loss. The class weights were computed by median-weighted inverse frequency given in \cite{nwoye2019weakly}.
    
    \item Model pre-training: The ResNet-18 backbone was pretrained on ImageNet \cite{russakovsky2015imagenet}.
    \item Post-processing: The output vector of the model was converted to binary labels at a threshold of 0.5 and the positive class identities were mapped to the tool labels.
\end{itemize}

Using its ResNet-18 backbone, the SANet model extracts spatial features from an input image. 
The spatial self-attention module highlights the feature regions that belong to the tools while suppressing those that belong to the background.
The classifier pools the maximal activations of the feature channels into a vector of logits for each tool type, using the same methodology as in \cite{nwoye2022rendezvous} and then transforms the logits to presence probability scores by a sigmoid operation. Casting the values to binary at a threshold of 0.5 yields the desired result.

\subsubsection{Model Training}

The provided dataset was split into train, validation, and test sets for the experiment.
Training elapsed after 30 epochs using a batch size of 30 frames randomly composed from 3 video clips extracted at 1 fps.
Model weight optimization was performed using a Stochastic Gradient Descent (SGD) optimizer (Momentum $\rho=0.9$) with an initial learning rate ($\lambda=7e^{-3}$), decayed by a factor ($\eta=1e^{-4}$) and regularized by an $L_2$ norm. Optimal hyper-parameters were selected using the validation set.

The presented SANet model was trained on 4x Quadro RTX 6000 24GiB GPU provided by HPC Unistra Mesocenter managed by the University of Strasbourg, France.




\subsection{TeamZero}




\noindent Members: Muhammad Bilal, Taofeek Akinosho, Adnan Qayyum, Massimo Caputo, Hunaid Vohra, Michael Loizou, Anuoluwapo Ajayi, Ilhem Berrou, and Faatihah Niyi-Odumosu\\

\noindent TeamZero has a multi-disciplinary focus that involves experts from academia as well as from surgical healthcare. In this challenge, TeamZero participated in Category 1, where the task was to classify different surgical tools appearing in the field of view of endoscopic surgical videos. The objective of the team was to develop a systematic methodology to solve multi-label (multi-class) classification problems by harnessing the capabilities of ensemble models. The methodological approach that was followed by TeamZero for their entry in the SurgToolLoc challenge along with the discussion of findings from their experimentation is presented below. 

\subsubsection{Method Description}

\noindent Methodology proposed by TeamZero for robust detection of surgical tools consists of three key components: (1) intelligent data preprocessing; (2) model training; and (3) model finetuning (as depicted in Figure \ref{fig:method}). Details about the various processes of each component are presented below.  

\begin{figure*}
    \centering
    \includegraphics[width=\textwidth]{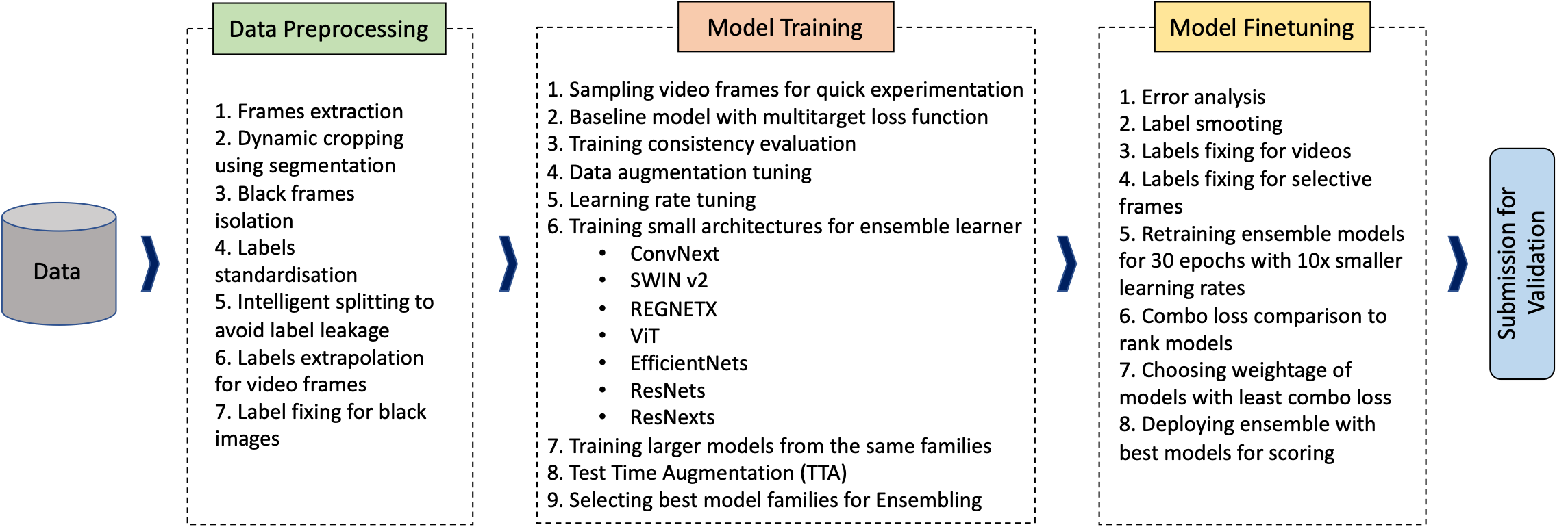}
    \caption{TeamZero: Proposed methodology for robust detection of surgical tools with noisy labels.}
    \label{fig:method}
\end{figure*}

\begin{figure*}
    \centering
    \includegraphics[width=\textwidth]{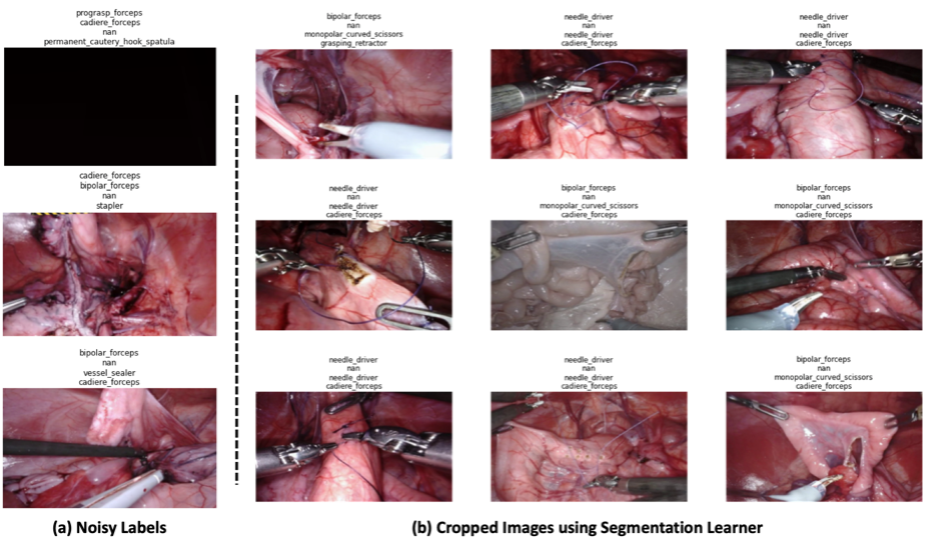}
    \caption{TeamZero: Examples of noisy labels (a) and cropped Images using Segmentation Learner (b). }
    \label{fig:dynamic}
\end{figure*}

\paragraph{Data Description and Pre-processing} TeamZero performed intuitive data preprocessing to remove different types of noises present in the data (as shown in Fig. \ref{fig:dynamic} (a)). This ensures good quality data and ultimately helps develop efficient models. The following are the key steps involved in preprocessing the surgical videos.

\begin{enumerate}
    \item \textit{Frames Extraction:}  Firstly, they extracted frames from surgical videos. Each video clip contains an average of 1,800 frames, resulting in a collection of over 4 million images with dimensions of 1280x780 pixels. The sheer volume of data, exceeding 1TB, presents a challenge when attempting to conduct meaningful experimentation using the entire dataset. To address this issue, the team initially extracted 10 random frames from each video for preliminary experimentation and hyperparameter tuning purposes. Subsequently, they extracted frames at a rate of 1fps (30 frames per video) for training the final models. 
    \item \textit{Dynamic Cropping using Segmentation:} TeamZero performed dynamic cropping of frames to extract the region of interest (ROI) in the camera field of view (FoV). Specifically, they trained a segmentation network (UNet model) using a small dataset that was randomly selected from the data. They used the Prodigy tool and a custom algorithm for the annotation of randomly selected frames from half of the surgical videos. They selected one frame per video from the annotated dataset for creating data for the training segmentation model that was trained for binary image segmentation into foreground and background. Their learned segmentation network efficiently removes (1) the user interface control panel that appears at the bottom; (2) the disclaimer notice at the top; and (3) the black borders appearing on the left and right sides of the videos, as these extra pixels can only contribute towards prolonged network training and wasted computation. A visual example of their dynamic segmentation-based cropping is shown in Figure \ref{fig:dynamic} (b). 
    \item \textit{Black Frames Isolation:} In the dataset, a few videos contain entirely black frames without any visual information, while they have also received labels due to extrapolating their parent ones. They have isolated such frames for fixing their labels.  
    \item \textit{Labels Standardisation:} TeamZero used lambda expressions to clean label strings, i.e., removing different symbols that include brackets ([]), quotes (“), hyphens (-), and slashes (\slash). Also, the underscore (\_) symbol was replaced with white space.  
    \item \textit{Intelligent Splitting:} TeamZero realised that randomly splitting frames into the training and validation set will allow frames of one video to end up in both datasets which will also lead to data leakage. Therefore, they devised a logic to intelligently split the dataset into training and validation sets, where the frames from videos are allocated to either training or validation sets unless the labelled tool combination has only one video in the dataset. This ensures that each tool combination is present in both training and validation sets, as there are eight tool combinations with just one video. 
    \item \textit{Label Extrapolation:} The dataset provides one label for each video, while the expected algorithm is required to predict labels for instruments' presence for all the frames in the video. Therefore, video labels are required to be extrapolated for frames. This extrapolation introduced a lot of noise in the training labels. There are many examples where the label might indicate the presence of three instruments but only two or fewer tools can be seen in the given frame. This is often the case when the surgeon moved the tool out of view even though it is installed on the robotic system. Learning from noisy labels is one of the unique challenges this competition presented.
    \item \textit{Label Fixing:} There are many cases in which video frames are entirely black. This happened when the surgeon uninstalled the camera or wanted to move it to another arm. These are totally or partially black videos. Instead of assigning actual labels to such frames, TeamZero decided to introduce a new 'blank' category and assigned it to all four arms for black images ([blank, blank, blank, blank]). They used the PIL library's {\it getextrema} method to identify such void images. 

\end{enumerate}

\paragraph{Model Training Strategy}

\noindent As illustrated in Fig. \ref{fig:method}, nine different steps are involved in the development of the model(s) for the detection of tools.  Here TeamZero provides the description and intuition of using each method. Their modelling approach involved experimentation on data augmentation, model stability checking, customising loss function, grid search optimisation for architectural search to design ensemble learners, progressive resizing, label smoothing \cite{szegedy2016rethinking}, fine-tuning with weighted data loaders, and fine-tuning with the cleaned dataset. 

\noindent The team carried out exhaustive experimentation to understand three learning strategies for modelling similar multi-classification problems from data with noisy labels. The meta-strategies under consideration included binary classification, multi-label classification, and multi-target classification. The team started with the most intuitive approach of binary classifiers to detect the tool’s presence or absence from surgical images. Binary classifiers train models that can learn to perform one task with great precision such as the presence or absence of bipolar forceps. An obvious deployment challenge in this meta-strategy phase is the huge computational time. The system needs to run and collate predictions from several models (in this case 14 models; one for each tool) to make the final prediction. Most importantly, this meta-strategy cannot learn unique intrinsic relationships among objects which is key in tools usage patterns in the surgery. For example, certain tools are often used in tandem with each other. Since binary classifiers learn and predict in isolation, the resulting model fails to learn similar intrinsic usage pattern features. They then tried multi-label classification in which a model learns to predict multiple labels at once. This strategy compromises the learning abilities of binary classifiers in lieu of learning intrinsic tool usage features, but it is susceptible to learning noisy and spurious correlations. To deal with the limitation of both meta-learning strategies, they then employed multi-target classification which is predominantly used to fuse diverse data types in modern deep learning (DL) applications such as images and text. This strategy allowed the learner to learn some interesting features informing robotic arm-specific tools, which are used often or seldom or never with each other. 

\noindent TeamZero started the experimentation with minimalistic data to fast-track iterations and see what will work for training the multi-class learner. An exhaustive hyperparameter tuning was performed to choose the appropriate data augmentation types, learning rates, and architecture types. Major algorithm families including ResNet, ConvNext, Swin V2, EfficientNets, ResNexts, REGNETX, and ViT were explored. They selected the final model to be an ensemble of pre-trained models from various families through bagging to predict the presence of tools in surgical videos. To allow the model to learn some of the difficult classes, they also employed weighted data loaders and curriculum learning for extrapolating certain scarce classes. They finetuned the ensemble models for several epochs using a weighted data loader to improve learner performance. All experimentations were conducted using the FastAI library in Python language \cite{howard2020fastai}.

\subsubsection{Preliminary Performance}
\paragraph{Results Description} The following are the key observations from their experimentations for surgical tool detection. 

\noindent \textbf{\textit{Effect of Dynamic Cropping:}} Their binary segmentation strategy to dynamically crop images for the borders and UI control which were leaking the tool names phenomenally. It increased the model accuracy and generalisation ability significantly as images without leaky labels forced the models to learn the actual representation of instruments rather than some spurious correlations.

\noindent \textbf{\textit{Data Augmentations:}} TeamZero tested three data augmentation approaches including Squish, Crop and Pad. They found that squishing images works better than cropping and padding. In addition, they also checked for the rectangular vs squared resizing. Their baseline algorithm (ConvNext) performed better when used on a dataset which is scaled with a similar aspect ratio to the original dataset which has a rectangular dimension.  

\noindent \textbf{\textit{Multi Target Classification:}} TeamZero found that multi-target classification supersedes the simple multi-label classification on this problem. In multitarget classification, the ML model tends to learn to model each robotic arm. This allows the ML model to learn the intrinsic relationship between tools usage patterns. The multi-target version exceeded the detection accuracy over a single multi-label classifier by 5\%.

\noindent \textbf{\textit{Model Training using Progressive Resizing:}} TeamZero employed progressive resizing for the training baseline model. The training starts with small images (4x smaller in this case) and completes the training using large images (mostly original image size if that is not too large). They retrained the aspect ratio while resizing as discussed above. The final model reduces the images to $3/4$ of the original input image by keeping the original aspect ratio. They performed 12 epochs for the smaller image experimentation and then used 6 epochs each for two successive resized datasets. The model was able to improve the accuracy by 1.35\%.

\noindent \textbf{\textit{Using Test Time Augmentations:}} To improve the generalisability of their model, TeamZero employed test time augmentation (TTA) which reduced the overall error rate slightly in the 3rd digit place after the decimal. In TTA, they created various augmentations of the original image and used them to make a prediction in addition to the original image and then averaged the probabilities. A key reason for TTA to not work in this case is that they squished the input images rather than using a random resize crop where TTA has shown huge improvements. It did not affect the training time but significantly increased the validation and inference time as the model has to go over images n times, which is the number of augmented versions they specified for the learn.tta() method.

\noindent \textbf{\textit{Hyperparamter Selection:}} They used a higher learning rate for first-time training of the models with 12 epochs and then lowered the learning rate in all fine-tuning afterwards. The learning rate of 1e-2 seemed to have worked very well for all the architectures. They used FastAI’s built-in function to inform this decision. A typical learning rate chart can be seen in Figure \ref{fig:chart}:

\noindent \textbf{\textit{Effect of Ensemble Model:}} The approach that dramatically improved TeamZero's rank on the preliminary testing leaderboard was the strategy they used for ensembling. Rather than ensembling using typical approaches such as cross-validation, they trained models across different families of architectures. They used the timm library for downloading these models which have been nicely integrated into the FastAI library. They performed exhaustive experimentation to see which model families can learn the given classification tasks with greater accuracy. They found that ConvNext and SWIN outperform all other architectures. Besides, ViTs and REGNETX were also equally performing well. So, they chose to use four models of different families across these four families. They used smaller architectures to choose the right families and then tested their larger versions for ensemble learning. They found that larger architectures were not improving the overall performance but instead were massively overfitting on the given task. So, they decided to use the smaller architectures instead. Tyhe also doubled the weightage of the predictions from the top-performing models. These included Convexts and SWINs.

\begin{figure*}[t]
    \centering
    \includegraphics[width=0.55\textwidth]{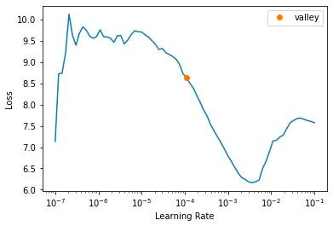}
    \caption{TeamZero: Learning rate chart showing 5e-3 where ViT model can learn the most on the given dataset. }
    \label{fig:chart}
\end{figure*}

\noindent \textbf{\textit{Error Analysis:}} Once TeamZero had finalised their ensemble model, they then performed error analysis. They held out at least one video across different label combination types to understand the learning capability of the ensemble model. The first insight they got was that model is learning the wrong labels with high confidence and accuracy as the right ones. The arm where they observed the most noise was USM4 and USM1. The instrument which is predicted largely wrong by the model is Cadiere forceps. This led us to perform label cleaning at the video level. There were many videos where just two or fewer tools were present, but the label said otherwise. They fixed those videos and fine-tuned their models on newer less noisy data which seemed to improve their performance slightly. But the difference was not dramatic as in most of the videos labels were very wrong and this small fix would still not guide the algorithm to learn that difference.

\noindent \textbf{\textit{Manual Label Fixing:}} They then chose 5 videos per label combination to fix labels at the image level. It was an exhaustive exercise. They used Prodigy for label fixing. They created a custom recipe for this purpose to perform this correction. They fine-tuned their model on this dataset but there was no considerable increase in the accuracy of models even after performing label fixing.

\paragraph{Conclusions and Future Work} Classification tends to model the analytical problem using the one-hot encoding of target labels. The model is forced to learn classes with high confidence since they are eventually rounded to 0 or 1. They felt that this is very harmful as the provided data has various types of noise. Many images are mislabelled. In real life, no data is noise-free. Even human-curated labels end up with wrong labels. Label smoothing allows models to relax label boundaries slightly. So instead of 1s and 0s, the loss function is engineered to use a number less than 1 for 1s and a number a bit more than 0 for all 0s in the encoding vector. Retraining this should have worked. They tried fine-tuning their model using label smoothing but they didn’t have enough time due to wasting a lot of time figuring out docker submission issues. They trained their models for 10 epochs but there was no significant improvement, so they discarded the idea. For label smoothing, models shall be trained for longer to pick up confusing cases. In their future work, they plan to use different noise modelling approaches for fixing noisy labels \cite{karimi2020deep} such as iterative label clearing during model training, automatically identifying, and fixing incorrect labels, using Bayesian models, etc. A major limitation of their proposed methodology is that they did not start off with those prevalent strategies. 

\begin{figure*}[t]
    \centering
    \includegraphics[width=0.55\textwidth]{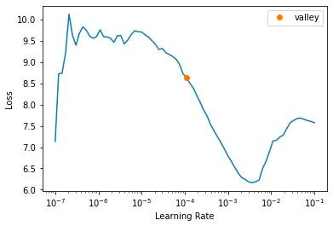}
    \caption{TeamZero: Learning rate chart showing 5e-3 where ViT model can learn the most on the given dataset. }
    \label{fig:chart}
\end{figure*}


\clearpage
\subsection{Results}
The team evaluations for the challenge were performed on a private, hidden test set containing individual files of surgical videos at 1 FPS. 
The results were generated through the Grand Challenge automated algorithm submission and evaluation system.
Below we present a summary of the challenge results.

\subsubsection{Category 1: Surgical tool presence detection}

For Category 1, the performance of surgical tool presence detection was evaluated with the mean F1-score \cite{scikit-learn} across all tools. Overall, the teams performed well in this category with some of them obtaining F1-scores above 0.7 (Table \ref{tbl:c1_results}). 
It is important to notice that the presented F1-scores take into consideration data imbalance: the scores are weighted based on how often each tool is present in the test data. This was an important consideration because some tools did appear much more often than others in this data set.



\begin{table}[h!]
\centering
\caption{Results for Category 1}
\begin{tabular}{|c|c|}
\hline
\textbf{Team} & \textbf{Avg. F1-score} \\ \hline
HRI\_MV       & 0.7485                 \\ \hline
HKMV          & 0.72                   \\ \hline
NVIDIA        & 0.7055                 \\ \hline
ANL-Surg      & 0.6691                 \\ \hline
HVRL          & 0.6605                 \\ \hline
SK            & 0.643                  \\ \hline
TeamZERO      & 0.6299                 \\ \hline
VANDY-VISE    & 0.6206                 \\ \hline
UKE           & 0.6203                 \\ \hline
ITeM          & 0.5978                 \\ \hline
CAMMA         & 0.5744                 \\ \hline
MM            & 0.5637                 \\ \hline
BioMedIA      & 0.5521                 \\ \hline
\end{tabular}
\label{tbl:c1_results}
\end{table}

\subsubsection{Category 2: Surgical tool localization and classification}

\begin{table}[h!]
\centering
\caption{Results for Category 2}
\begin{tabular}{|c|c|}
\hline
\textbf{Team} & \textbf{mAP} \\ \hline
HRI\_MV       & 0.4077       \\ \hline
HKMV          & 0.3726       \\ \hline
NVIDIA        & 0.3058       \\ \hline
ANL-Surg      & 0.0961       \\ \hline
SK            & 0.0247       \\ \hline
VANDY-VISE    & 0.0003       \\ \hline
HVRL          & 0            \\ \hline
ITeM          & 0            \\ \hline
\end{tabular}
\label{tbl:c2_results}
\end{table}

For this category of the performance of surgical tool localization and classification was evaluated with the standard COCO dataset evaluation metric \cite{lin2014microsoft}, i.e. the mean average precision (mAP) of intersection over union (IoU), discretized with values $0.5:0.05:0.95$ at 1 FPS. 
This challenge proved to be very demanding for most teams. Only the top 3 teams were able to achieve a reasonable performance of above 0.3 mAP on the test set. The remainder of the teams struggled with detecting bounding boxes accurately and could only achieve mAP values between 0 and 0.1 (Table \ref{tbl:c2_results}).

Unsurprisingly, teams that performed well in Category 2 also performed well in Category 1. Interestingly, lower performance in Category 2 could still translate to reasonable performance in Category 1. This indicates that detection models may be learning image features that do not help with localization. Critically, the method that distinguishes higher from lower-performing results in Category 2 is pre-training on some data sets with ground truth tool segmentation labels. This may suggest that ``nudging'' the model weights to some portions of the weight-space that are informed by tool segmentation, helping learn features that are more relevant to localization.

\subsection{Discussion}
%

Alternatives to fully supervised learning, such as self-supervision and weak supervision, continue to advance in the areas of object classification, detection, and localization. This is particularly impactful in surgical data science since large volumes of data can be exploited if noisy or weak labels can be utilized. Our SurgToolLoc 2022 EndoVis challenge provided the community with just such a dataset. With all the trappings of a large, real-world dataset (e.g. variability, imbalanced classes, noisy and weak labels), teams could explore how different model architectures and training strategies could solve a relevant surgical task.

In total, there were 13 complete submissions in Category 1 and nine complete submissions in Category 2. Team reports were included to provide a comprehensive view of the approaches taken. Teams performed relatively well in Category 1 but struggled in Category 2. Good performance in Category 1 generally translated to relatively good performance in Category 2. This was perhaps not surprising since discovering an accurate and robust representation of the tool presence, at least conceptually, seems like a prerequisite for localizing these same tools. 

For both categories 1 and 2, supplementing training data with publicly available surgical data (e.g. Cholec80), or building off of previously trained backbones (e.g. resnet) proved to be key in getting high-performing models. Specifically, five out of the 13 teams used a clinical dataset to pre-train the model. For Category 1, three out of the top five teams used the EndoVis instrument segmentation dataset to pre-train the model, including HRI-MV, HKMV, and ANL-Surg. Similarly for category 2, two out of the top three teams used the same dataset, including HRI-MV and HKMV.
For Category 2, the use of class activation maps was crucial but only proved moderately successful. On the whole, the challenge was difficult and the proposed tasks remain largely unsolved.

Performance in Category 2 was clearly worse than that of Category 1, but also poor by conventional standards. This was not surprising as only noisy, weak labels were available for training. All other things being equal, we assume supervised training would improve performance on this task. By how much is not clear. It could be that performance of these weakly supervised models would be viewed more favorably in light of the limits of similar models trained with strong labels. This would have important implications given the difficulties and resources required in obtaining strong labels. We plan to pursue this question in future work.


There are a host of topics that can be explored with the dataset from this competition. For example, surgical gesture and activity recognition, and tool-tissue interaction are areas of study that this data set can speak to, albeit with various annotation efforts. Similarly, automated, objective skill assessment is an important area that could potentially benefit from this and similar data sets. We hope that this data set can contribute to many future advances.


In this challenge we encouraged the surgical data science community to tackle the problem of automatically detecting and tracking surgical instruments in endoscopic videos. This had to be accomplished by leveraging only tool presence data, as this data is readily available in robotic-assisted surgery. For category 1, designed to classify tools present in each surgical frame, the participating teams developed algorithms that performed well. For category 2, classifying and detecting the spatial locations of surgical instruments, the algorithms performed relatively poorly.
The models used by the teams represent many state-of-the-art classifier algorithms. While these models perform well on natural and conventional image data sets, they failed to exhibit similar performance on these surgical videos. We can only speculate as to the reasons for these shortcomings. For instance, it could be due to the noisy labels; it could be that these surgical images represent an entirely different distribution of data that is more difficult to categorize; it could be that this problem has exposed the limitations of these algorithms; or, it could be that these algorithms simply require more training data. While the lack of sufficient training data is the usual culprit for poor performance, it is worth noting once more that we believe this is the largest publicly available surgical data set to date. Regardless of where the models fell short, it is clear that this problem has not been adequately solved.

\subsection{Acknowledgements}

Work done by the CAMMA team was supported by French state funds managed within the Plan Investissements d’Avenir by the ANR under references: National AI Chair AI4ORSafety [ANR-20-CHIA-0029-01], Labex CAMI [ANR-11-LABX-0004], DeepSurg [ANR-16-CE33-0009], IHU Strasbourg [ANR-10-IAHU-02] and by BPI France under references: project CONDOR, project 5G-OR. 

Work done by the ITeM team was supported by the German Federal Ministry of Research and Education (BMBF) under grants CoHMed/DigiMedOP grant no. 13FH5I05IA, CoHMed/PersonaMed-A grant no. 13FH5I06IA, and CoHMed/PersonaMed-B grant no. 13FH5I09IA.

\newpage
\section{Results and methods from the MICCAI 2023 SurgToolLoc challenge}

\subsection{SurgToolLoc 2023 Challenge Description}

\subsubsection{Overview}
This subchallenge was arranged as part of the Endoscopic Vision Challenge at MICCAI 2023 and was a continuation of the SurgToolLoc 2022 challenge. However, this time there was only one category of surgical tool detection (the same as category 2 of 2022 challenge). Please see the description provided in Section $3.1$ for more details.

\subsubsection{Training Data}

The training data used for this challenge was the same as 2022 SurgToolLoc challenge - please see Section $3.1$ for specific details on the dataset.

\subsubsection{Testing Data} 

The format of the testing dataset was exactly the same as that of 2022 Challenge, however, a new set of 108 video clips collected from surgical training exercises were used as the test set. This meant that the algorithm performance numbers (mAP values) for the two years cannot be directly compared. However, we did evaluate the winning algorithms from 2022 in the 2023 test set to be able to compare and see if the algorithm performances improved over the year.

The testing data was a set variable mean (SD) clips in seconds: 687.45 (424.68), downsampled to 1 FPS for inference purposes. The test data was annotated with bounding boxes around the instruments, using a crowd of experienced annotators. The process of creating the 2022 testing set labels was reproduced here; see section $3.1$ for more details.


\subsubsection{Submission Process}

The submission process for this challenge was the same as 2022 challenge; please review Section $3.1$ for more details.


\subsection{Team Submissions}\label{team_submissions_23}

For this year's challenge, there were a total of 18 teams that had shown interest in participating and downloading the dataset. In the end, only 7 teams submitted full submissions in the single category available. Table \ref{table:TeamAffils2} shows the participating teams, while Table \ref{table:TeamMethodsSummary2} summarizes the methodologies employed by each team. Team methodological details follow below. Note that these sub-sections were written by the participating teams.

\begin{table}[h!]
\centering
\small
\caption{Team affiliations and challenge categories - 2023}
\begin{tabular}{ccccc}
\toprule
\textbf{Team \#} & \textbf{Team name} & \textbf{Institution} & \textbf{Country} & \textbf{Report} \\ \midrule
1 & AIT & King's College London & UK & Y\\
2 & ANL-Surg & Argonne Nat. Lab. & USA & Y\\
3 & CAIR-HK & City U. of Hong Kong & Hong Kong & Y\\
4 & HVRL & Keio U. & Japan & Y\\
5 & Jmees & Jmees Inc. & Japan & N\\
6 & LabRen-CUHK & Chinese U. of Hong Kong & Hong Kong & N\\
7 & LozaGera & U. of Leeds & UK & N\\
8 & MapleLab & Vanderbilt U. & USA & Y\\
9 & PUMCH & Peking Union Med. Coll. Hosp. & China & Y\\
10 & SDS-HD & German Cancer Res. Ctr. (DKFZ) & Germany & Y\\
11 & seventeen & Beijing U. of Posts \& Telecom. & China & N\\
12 & TUE-VCA & Eindhoven U. of Tech. & Netherlands & Y\\
13 & ZJURealdoctor & Zhejiang U. & China & Y\\ \bottomrule
\end{tabular}
\label{table:TeamAffils2}
\end{table}
\begin{sidewaystable*}
\caption{Summary of methodologies - 2023}
\begin{adjustbox}{scale=0.7,center}
{\begin{tabular}{ p{8em} p{10em} l m{5em}l m{5em} l m{5em}l m{5em}l m{10em}l m{5em}l m{10em}l m{10em}l m{5em} l }
Team Name & Architecture & Backbone & Data Preprocessing & Pretrain & Image Augmentation & Use Additional Data & Loss & Output \\
\hline
AIT & Faster-RCNN + U-Net & ResNet & OCR for UI detection & Y &  & EndoVis2017 & Binary cross-entropy & Tool bounding box \\
ANL-Surg & Detectron2 Faster RCNN + YOLOv8 & ResNet &  & & Manually annotated frames & Team Nvidia’s dataset from 2022 challenge & & Tool bounding box \\
CAIR-HK & WS-YOLO & CSPDarknet53 & Pseudo-labels & Y & & & & Tool bounding box \\
MapleLab & TernausNet + ResNet-101 & ResNet & Attention masks for tools + pseudo labels & Y & & & Binary cross-entropy & Tool bounding box \\
PUMCH & OsTrack & Transformer tracking model & & & Auto-generated pseudo-labels & & Custom (see report) & Tool bounding box\\
SDS & YOLOv5 + SwinTransformer & CSPDarknet53 &  & Y & Semi-automatically generated bounding boxes &  & Binary Cross-Entropy & Tool bounding box \\
TUE-VCA & YOLOv8 & CSPDarknet53 & Filtering & Y & Pseudo-labeling & & Symmetrical
cross-entropy & Tool bounding box\\
\hline
\end{tabular}}
\end{adjustbox}
\label{table:TeamMethodsSummary2}
\end{sidewaystable*}

\clearpage

\subsection{AIT}

Aiming to keep within the spirit of weakly-supervised object detection, our approach tried to make use of only a small amount of human annotated data. We also note that the presence labels provided are often incorrect and the UI is a better source of information as well as providing additional information such as activation and rough positioning which we make use of.

We start with EndoVis2017 tool parts segmentation dataset to train models for binary tool segmentation and binary tool tip detection. We then use this in conjunction with the data parsed from the UI of the SurgToolLoc data to generate a pseudo labeled dataset. We then train our final model on this data, using any SurgToolLoc data which could not be assigned pseudo labels as weak supervision.

\subsubsection{Method Description}
%
%
%
%

The EndoVis2017 dataset was used to train models for both binary tool-tip object detection and binary tool segmentation.
For the detection model we used a basic FasterRCNN~\cite{Ren2016}. The detection labels were generated using the tool parts classification maps.
The bounding box was drawn around the clevis, if no clevis is visible, it is instead drawn around the tip.
As tool part segmentations do not exist for the testing samples, we make our own training and validation splits from the training samples with 6 and 2 clips respectively.
For the segmentation model we train a standard UNet~\cite{Ronneberger2015arxiv} on the segmentations provided with the training/validation split following the training/testing splits provided in the original dataset.

The UI present in many of the clips provides richer and more accurate information than the ground truth tool presence labels.
We implement a parser using EasyOCR\footnote{https://github.com/JaidedAI/EasyOCR} for optical character recognition (OCR) to read the tool names and note the order these tools appear from left to right.
The ordering of the UI elements representing the tools and the camera can be used to determine if the robot is configured with two tools on the left and one on the right, or visa-versa.
We also use a simple color model to determine which tools are currently active, as these are displayed with a pale-blue UI instead of gray.

\begin{figure}[tbh!]
\begin{center}
\resizebox{5in}{!}{\includegraphics{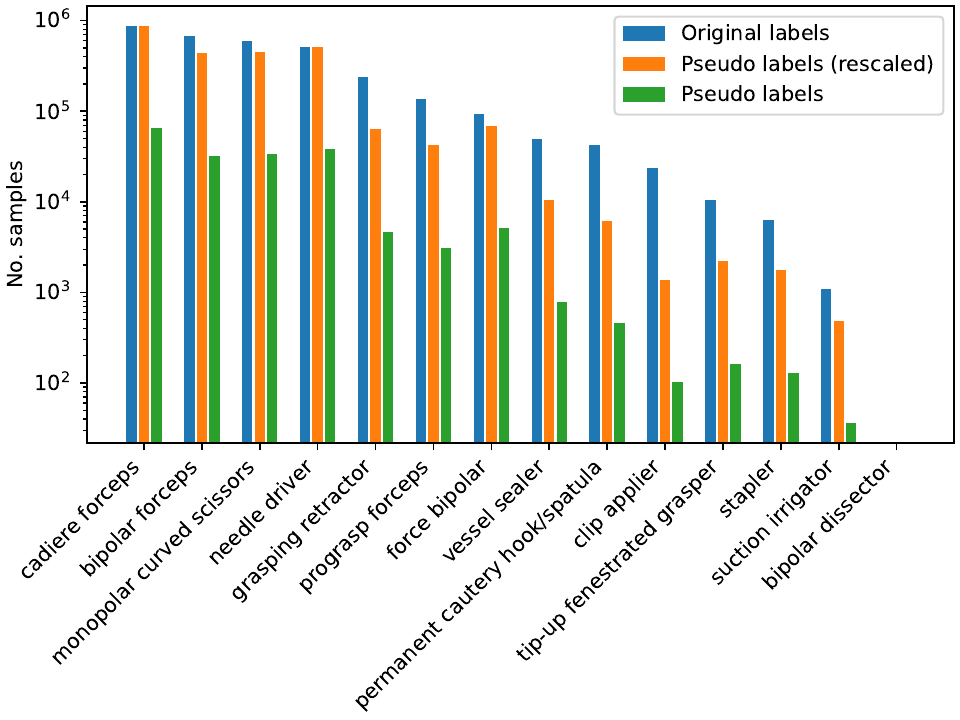}}
\caption{Class wise distribution for the original SurgToolLoc labels and the generated pseudo labels. The pseudo labels are also presented re-scaled to better compare the distributions with the SurgToolLoc labels}
\label{class_dist}
\end{center}
\end{figure}

To generate our pseudo labeled dataset we combine inferences from our two trained models with the information parsed from the UI.
We sample the provided SurgToolLoc video clips every 30 frames, to reduce residency in the data, and run our UI parse on these frames
If the OCR finds no tool names the clip is assumed to have no UI and is discarded from the pipeline.
We then run our binary segmentation and detection models on the frames.
The segmentations are processed by keeping connected components above 400 pixels in area.
Any frames that do not have exactly three connected components remaining are discarded.
The detected bounding boxes are then assigned to one of the three connected components of the binary segmentation based on intersection over union (IoU).
We then keep the most confident detection for each connected component.
The tools are determined to be either on the left or the right based on where the binary mask meets the edge of the image.
We also run RAFT~\cite{teed2020raft} optical flow to both the previous and next frame, averaging the forwards and backwards magnitudes of the flow vectors.
We average the flow magnitudes over each connected component of the segmentation map to arrive on an average motion for each detected tool, with tools with average motion above a threshold are determined as active.
We now use our detect labels (left, right, active) for each bounding box and compare these to the labels parsed form the UI.
The correct class label is assigned to the bounding box, if there is a discrepancy, the sample is discarded.
This whole process has a success rate of around 6
The class distributions for the data can be seen in Figure \ref{class_dist}.

To incorporate the discarded data, we use the classification logits of the detections to produce a multi-label prediction for the image. Specifically, the classification logits of the detections are summed together, weighted by the confidences.
A loss can then be calculated between the prediction and the presence label provided in the original SurgToolLoc data.

\subsubsection{Model Training}
We again use a FasterRCNN, initialized with the weights from the model pretrained on EndoVis2017.
We batch together images from the weak and the strongly labeled datasets, employing class balanced sampling to ensure classes are more evenly represented.
We apply random photo-metric and spatial augmentations to all the images.
The strong supervision loss is calculated as the standard FasterRCNN loss summed and the weak supervision loss calculated using binary cross entropy.
The model parameters were optimised using stochastic gradient descent with learning rate of $1$$\times$$10^{-3}$ , moment $0.9$ and weight decay of $1$$\times$$10^{-4}$.
We reserved 10\% of our pseudo labeled data to use as a validation set for early stopping.

\subsubsection{Preliminary Performance}
Training on our pseudo labeled data with additional weak supervision we were able to consistently achieve around 0.33 mAP on our hold-out validation data. However, submission to the testing server yielded a much lower score of 0.11 mAP suggesting our model was badly over-fitting to some aspect(s) of the pseudo labeled data. An ablation study showed our weak supervision, provided only a modest improvement of just 0.03 mAP on our validation data.

While our pseudo labels generation provided accurate detection labels, too much of the data was discarded to ensure correctness of the retained data, leading to a small overall dataset. Furthermore, it is likely that the classes which were not present during the training of our binary object detection model were underrepresented in the final dataset. Ideally, weak supervision would provide a means to overcome these deficiencies, however we found our method was insufficient to effectively incorporate the information in the weakly labeled data. Perhaps, given more time, a more effective weak supervision approach would have been settled on.

\subsection{Argonne National Laboratory - Team ANL-Surg}

Team ANL-Surg includes Assistant Computational Scientist Neil Getty from Argonne National Laboratory (ANL), and Ruoxi Zhao, a recent graduate in Applied Mathematics from UC Merced. The team was primarily interested in Category 2. Motivated by the potential of weakly-supervised learning methods, ANL-Surg trained Detectron2 Faster RCNN \cite{Ren2016} and YOLOv8 \cite{Jocher_Ultralytics_YOLO_2023} on custom dataset for localizing, detecting and tracking each surgical instrument.

\subsubsection{Method Description}

%
%
%
%
Team ANL-Surg's goal was to train the Detectron2 Faster RCNN model on approximately 2500 manually labeled frames, and utilized the Mixformer \cite{cui2022mixformer} model for tracking the clevis of each tool starting from the manually selected frame in the clips, then performed backward and forward tracking at 30 frames per second (FPS). The team also trained the YOLOv8 huge model on the extended dataset, with a mean Average Precision (mAP) of 59.7\% on their validation holdout. The motivation of choosing the described models was to get accurate results of tracking and recognizing the surgical tools in the frames, without excessive human intervention and manually labeling all the frames in the training dataset. \par
Their approach began with the careful annotation of data. They manually annotated approximately 1500 frames from the competition dataset. These frames were selected halfway through the video clips, with a focus on the underrepresented surgical tools in the dataset, such as clip applier, permanent cautery hook/spatula, stapler, suction irrigator, tip-up fenestrated grasper, and vessel sealer. The reason for building this dataset was to increase the visibility of the underrepresented tools to the model, which the model tended to perform poorly in the previous training sessions due to the larger size of such tools. In addition, they edited team Nvidia’s dataset from MICCAI 2022 SurgToolLoc challenge that had around 1000 labeled frames, to be more accurate in capturing the clevis of the surgical tools, adapting to the competition guidelines for this year’s challenge. The foundation of their methodology lay in model training. The team initiated their efforts by training a Detectron2 Faster RCNN model using those 2500 manually annotated frames. This model served as their baseline for subsequent improvements. \par

\subsubsection{Model Training}

To enhance the temporal understanding of tool localization, the team employed the Mixformer object tracking model. This tracking model was used to follow labeled clevis of the tools in both forward and backward directions within the video clips, operating at a frame rate of 30 FPS. Tracking was initiated from the manually annotated frames. After using the trained model to infer tool detections of all the 30 FPS frames, the team merged tracking boxes generated by the object tracking model with prediction boxes produced by the Detectron2 model. This integration allowed them to rename predictions based on tracking data, effectively propagating manual labels throughout the entirety of the video clips. To reduce redundancy in the dataset, image hashing techniques were implemented. This step aimed to identify the frames that exhibited high similarity, remove frames where tracking differed from prediction by threshold, mitigating duplication and streamlining the dataset for training and evaluation. \par

\begin{figure}[h!]
\caption{Label frequency of manually annotated tools.}
\centering
\includegraphics[width=0.75\textwidth]{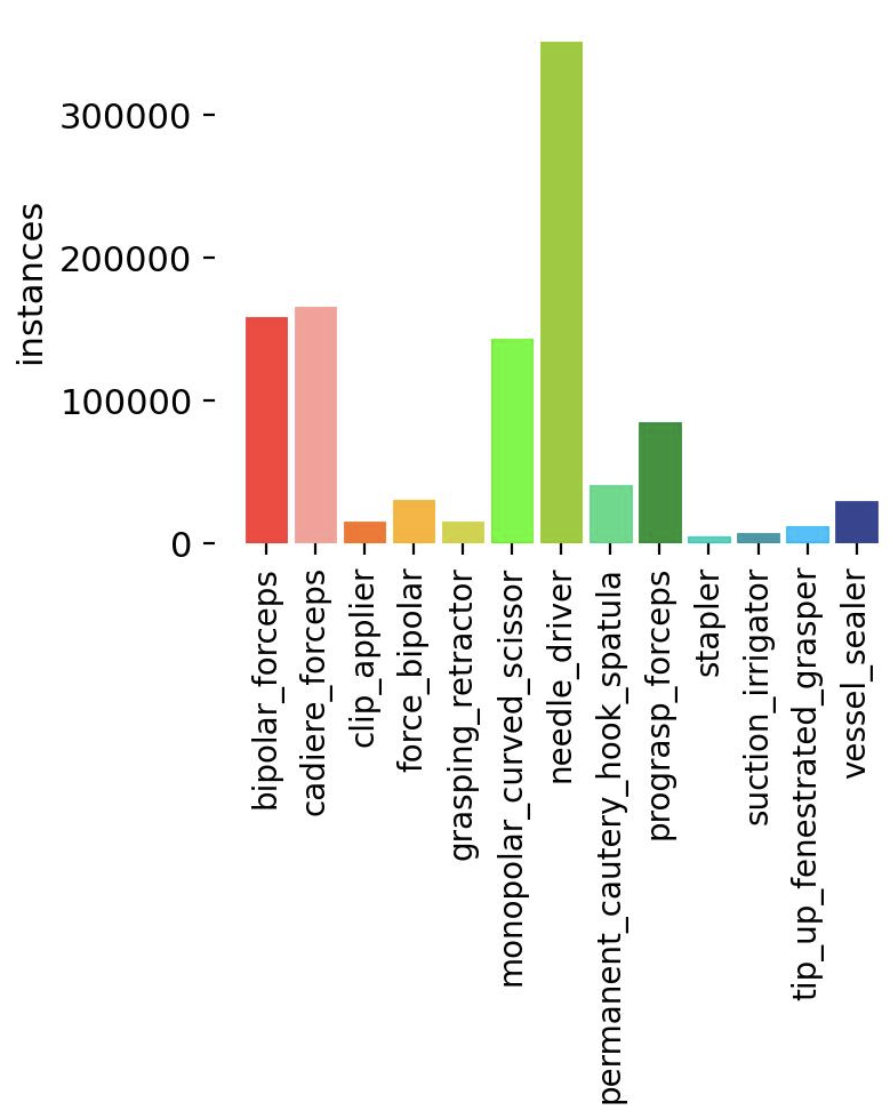}
\end{figure}

\begin{figure}[h!]
\caption{Example image showing tracking propagated labels (red) and overlapped predicted class.}
\centering
\includegraphics[width=0.75\textwidth]{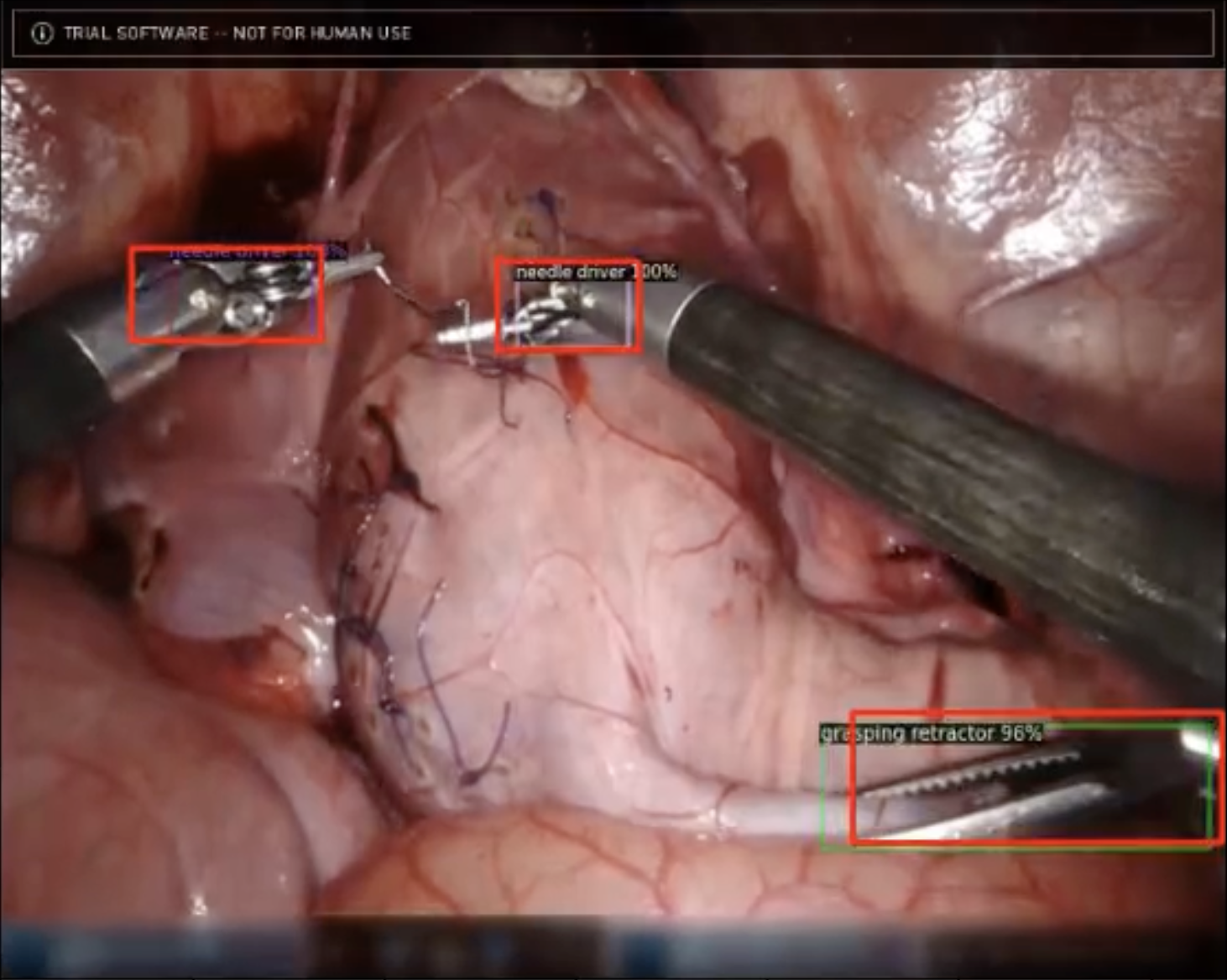}
\end{figure}

The model refinement process continued with the training of a YOLOv8 model. This model was trained on the expanded dataset, including a validation holdout. Its primary focus was on detecting frames containing three surgical tools, which the frames would have more potential of matching the three global labels in model predictions. For enhanced reliability and accuracy, the team enforced strict label matching criteria. Predicted labels were required to match global labels with high model confidence. Frames that did not meet this criterion were removed from consideration. They implemented a confidence threshold of 70\% and max detection to 4 to classify detections as trustworthy, acknowledging the model's certainty in its predictions. \par

For data augmentation, random 30\% rotation and 10\% shearing was used. The model was trained up to 300 epochs with a learning rate of 5e-4 and a batch size of 102 images. 6 V100 GPUs with 32gb memory were utilized for training.

\subsubsection{Preliminary Performance}

The comprehensive methodology yielded promising results. Initially, the YOLOv8 model trained solely on manual labels achieved a validation mAP of 42.7\% on the holdout set. As they progressed, the model trained on tracked images, consisting of 136,237 frames, demonstrated an improvement, achieving a validation mAP of 49.1\%. Notably, further advancements were observed as the model was trained on tracked and predicted frames, specifically those containing three tools. This iteration achieved a validation mAP of 55.7\%. However, the most significant leap in performance occurred when the team trained the model on the entire dataset, incorporating both high and low threshold detections, resulting in the highest validation mAP of 59.7\%. Finally, the ultimate model, trained on the complete dataset without a holdout, comprised 432,918 images and exhibited substantial enhancements in surgical tool localization accuracy. These results highlight the effectiveness of the methodology in progressively improving model performance. \par

\begin{figure}[h!]
\caption{Confusion matrix for all classes.}
\centering
\includegraphics[width=0.75\textwidth]{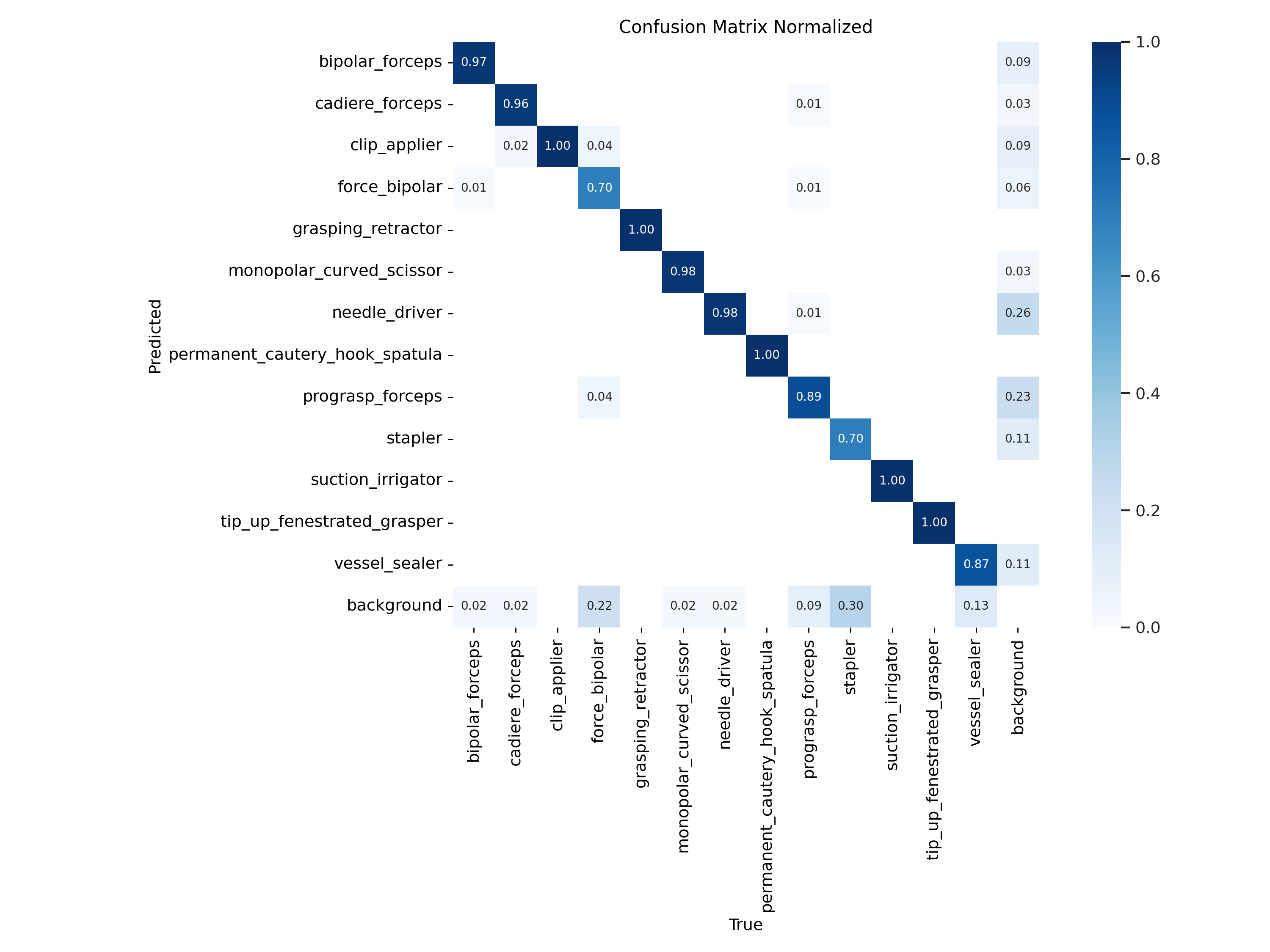}
\end{figure}

In conclusion, after testing out different methods and models, the team achieved a validation mAP of 59.7\% with the YOLOv8 model. The use of the Mixformer object tracking model helped the team to gain more training data by tracking the tools from one labeled frame, significantly reducing the time and labor of manual labeling medical dataset. Additionally, the use of confidence thresholding helps identify untrustworthy detections, enhancing the reliability of the localization model. In future research, the team is interested in focusing on the frames where one or two tools were present, which usually contains larger tools such as clip applier, stapler, and tip-up fenestrated grasper. The team is also interested in the unsupervised methods for this challenge, such as training DINOv2 \cite{oquab2024dinov2} on the training data to learn the features of the surgical tools in a self-supervised way. \par 
\subsection{MapleLab}
Team Members: John Han, Ayberk Acar, Jumanh Atoum, Yinhong Qin, Jie Ying Wu




The general intuition of team MapleLab's model was that by seeing many examples of different tools and their combinations, a model would be able to associate which tools are which while learning the image labels during training. Therefore, the model could correctly classify tools by recognizing the overlap of tools across different video samples and distinguishing each tool from the rest. 

\subsubsection{Method Description}
The MapleLab team developed a pipeline that did not require any additional manual labeling and learned the tool classification and localization from the complete surgical scene. The inputs of the model were the color image, segmentation mask generated from TernausNet, and their element-wise multiplication (referred as the attention mask). 

The application consisted of three stages: data preparation and bounding box initialization, training, and reiterated inference for classification. 

In the first stage, the team employed TernausNet \cite{Shvets2018}, a binary segmentation model trained on the EndoVis 2017 dataset \cite{Allan2019-2017}, to generate tool masks from the input RGB image. These two images were multiplied and Gaussian blurred to create an "attention mask." These three images (the original color image, the binary segmentation mask, and the attention mask) were cascaded to create a 7-channel tensor, which is used for the next two stages. In addition, the method obtained bounding boxes for each tool by identifying the rectangle around each connected component of the segmentation mask. Because the evaluation metric requires bounding boxes around the clevis of each tool, each bounding box was scaled down with a manually-selected scale factor of 0.6.

The goal of the training stage was to produce a network that could correctly identify which tools were presented in a frame, using only the tool presence labels. Their model, a fine-tuned ResNet-101 model \cite{he2015deep} pretrained on the ImageNet-1K dataset, took the input of the 7-channel tensor created in the first stage and predicted the tools that existed in that frame in one-hot encoding format. Figure~\ref{modelFig} shows an example of the input 7-channel tensor and model behavior.


\begin{figure}[tbh!]
    \begin{center}
        \resizebox{\textwidth}{!}{\includegraphics{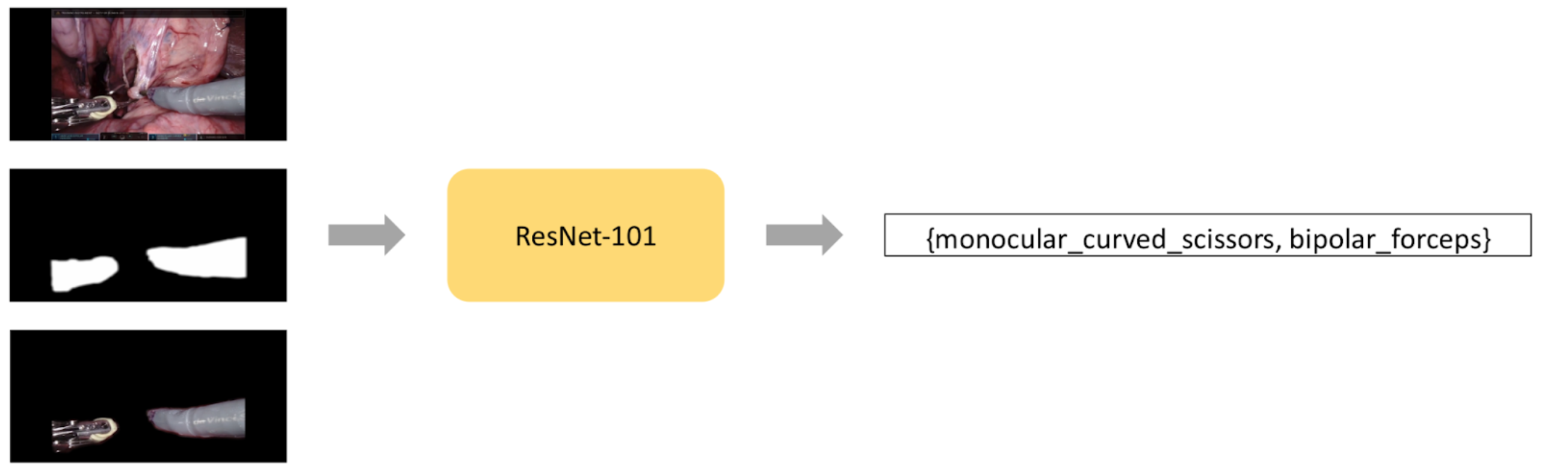}}
        \caption{A fine-tuned ResNet-101 model takes the concatenated 7-channel image as input and outputs the tool presence vector.}
        \label{modelFig}
    \end{center}
\end{figure}

\begin{figure}[tbh!]
    \begin{center}
        \resizebox{\textwidth}{!}{\includegraphics{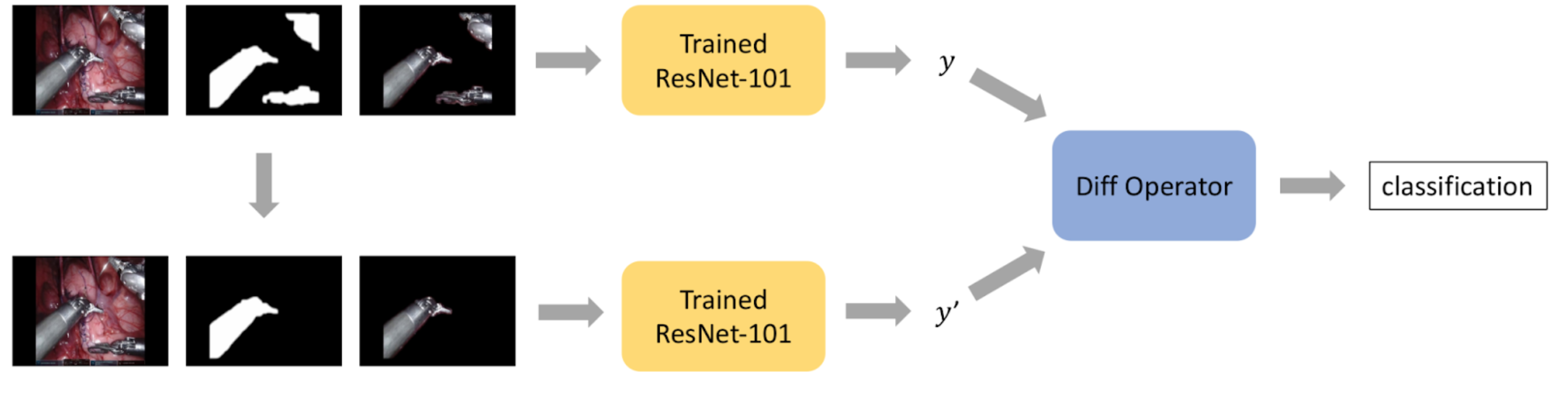}}
        \caption{In the final reiterated inference stage, predictions are made on each bounding box via masking out the other tools.}
        \label{inference}
    \end{center}
\end{figure}

\begin{figure}[tbh!]
    \begin{center}
        \resizebox{\textwidth}{!}{\includegraphics{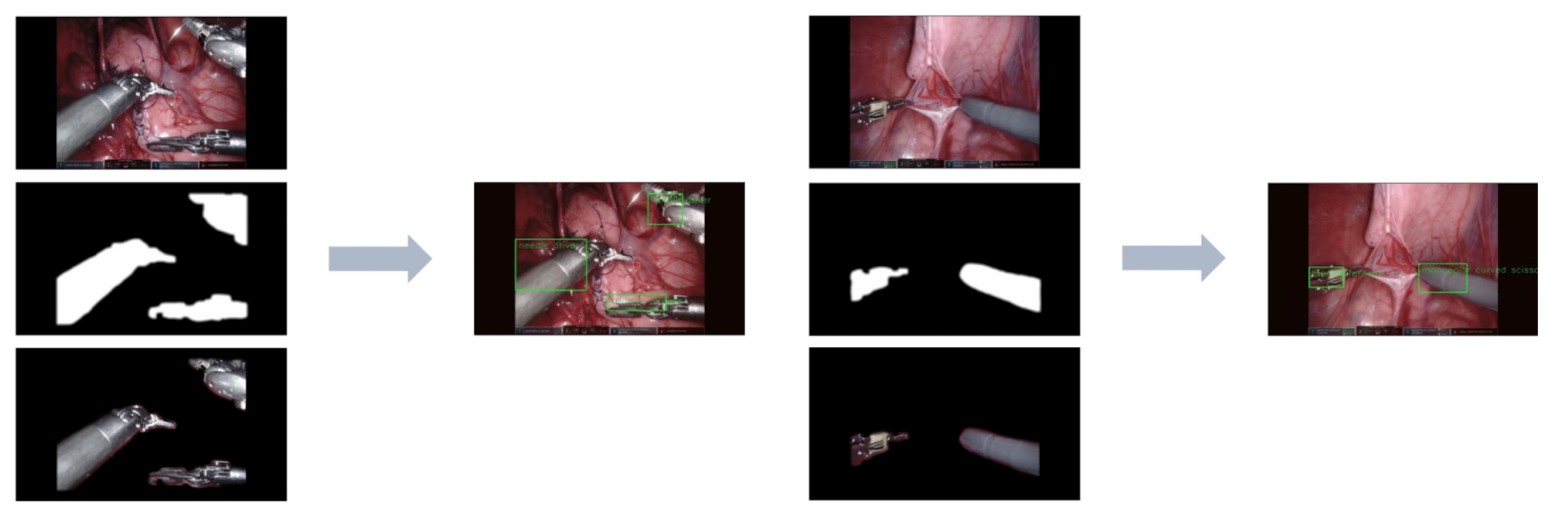}}
        \caption{Qualitative Results}
        \label{outputs}
    \end{center}
\end{figure}

In the final stage, every bounding box was labeled by reiterated inferences with the trained ResNet model. A score vector $y$, which represented the model’s confidence values of what tools exist in the frame, was first extracted. Next, all tools in the binary segmentation mask and attention mask were masked out except for the targeted tool to be classified. The model received this new set of images and generated another score vector $y'$. These two score vectors, $y$ and $y$', represented the model’s confidence values of what tools were present in the original set of images and the altered set of images. The classification label of a tool was determined to be the maximum of the values of $y'$ where there were positive confidence values from $y$. In other words, the most confident label was selected after the other tools had been masked out from the input data. This procedure was repeated for all tools to assign class labels to each bounding box. Figure~\ref{inference} displays the workflow of the classification process, and Figure~\ref{outputs} shows some example output images.

\subsubsection{Model Training}
To improve the quality of training data, a morphological opening and closing was applied to the segmentation mask to remove noise and close holes after running TernausNet. The images with poor segmentation masks were removed, via comparing the size of the segmentation blob with certain thresholds. Segmentation blobs with areas $> 0.02 \times \textrm{total image area}$ were kept and segmentation masks with more than 40\% predicted foreground pixels in the image were removed. Furthermore, the team also down-sampled given images to the size of $512 \times 512$ with 0 mean and unit standard deviation. The train-test split was 70-30.  The initial learning rate was set to 1 changing at every epoch with cosine annealing and the optimizer was stochastic gradient descent (SGD). The loss function was binary cross entropy. An NVIDIA RTX4090 GPU was used for the model training.





\subsubsection{Preliminary Performance}
During the training stage, team MapleLab's model had an accuracy of 98\% in correctly identifying the present tools. The model got a final test score of 0.0007 mAP (mean Average Precision). The low accuracy of the model may be due to the model’s inability to correctly identify which tools are which from the team's approach, as well as the fact that the bounding boxes were not refined specifically to be ROIs of the clevis of each tool. 

While the team's method acquired high accuracy in the initial training phase, results could be improved by the selection of frames with tool presence matching the provided labels, since the data did not always have all the tools in the field of view. Finally, overlapping tools in the camera view created a challenge for our pipeline, since the model relies on accurately separated tools from the binary segmentation.




%
%

 %

\subsection{PUMCH}
\begin{itemize}
  \item Team name:  PUMCH
  \item Members:  Surong Hua, Lu Ping, Wenming Wu
  \item Research field: Computer Vision
  \item Institution: Peking Union Medical College Hospital
  \item City: Beijing, China
  \item motivation and plan: At first glance, we were thinking of weakly supervised learning. But then we found that the video tags provided for training only contain information on whether a certain instrument exists in this video, it was not certain that whether the instrument appears in every frame, since it might be partially or completely obscured, which reduced the performance. 
  We also noticed that almost all the winning teams last year were based on semi-supervised method. Therefore, we proposed a framework by combining semi-supervised learning, object detection, and object tracking together.
\end{itemize}
 
\subsubsection{Method Description}

\begin{figure}
    \centering
    \includegraphics[width=0.75\textwidth]{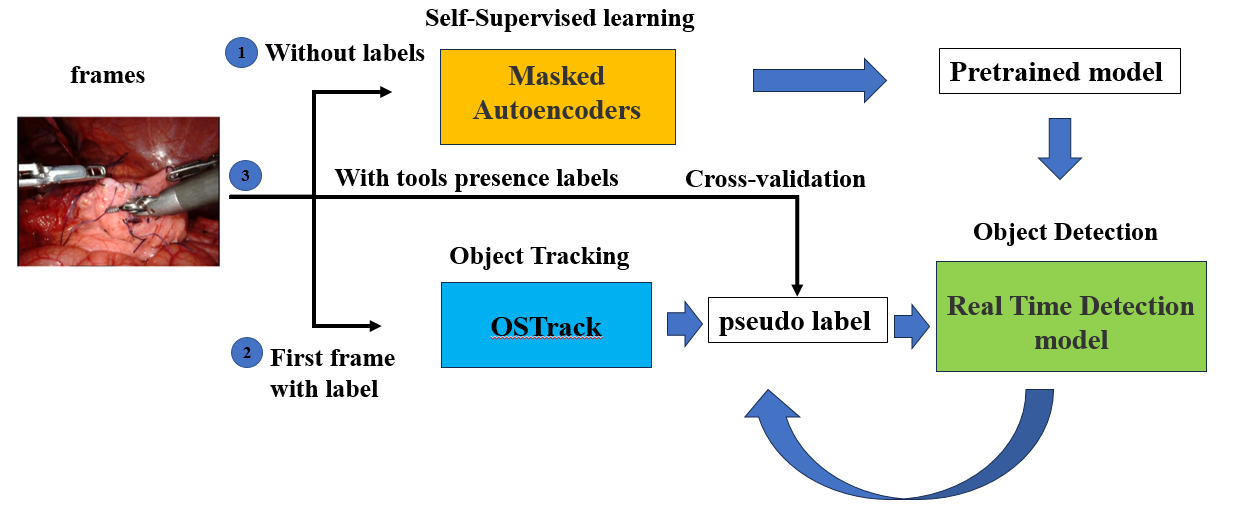}
    \caption{Data Processing}
    \label{fig:fig1}
\end{figure}

\paragraph{Data Processing}
See Figure \ref{fig:fig1} ,in this framework, we adopt the following three methods to process data. Firstly, for frames without labels, we use the MAE\cite{He_Chen_Xie_Li_Dollar_Girshick_2022} method to train the feature extraction ability of the backbone network. Secondly, we referred to last year's champion method to merge and process consecutive video segments to get sequence information, and then manually annotate the target in multiple stages of the video. Then, we used a transformer-based target tracking model (ostrack)\cite{Ye_Chang_Ma_Shan} to generate pseudo labels. Thirdly, we use tools presence labels to conduct cross validation and manual correction. Then, the model is continuously looped and trained to update pseudo labels. Ultimately, we obtained 30,000 pseudo labels of data.

\begin{figure}
    \centering
    \includegraphics[width=0.75\textwidth]{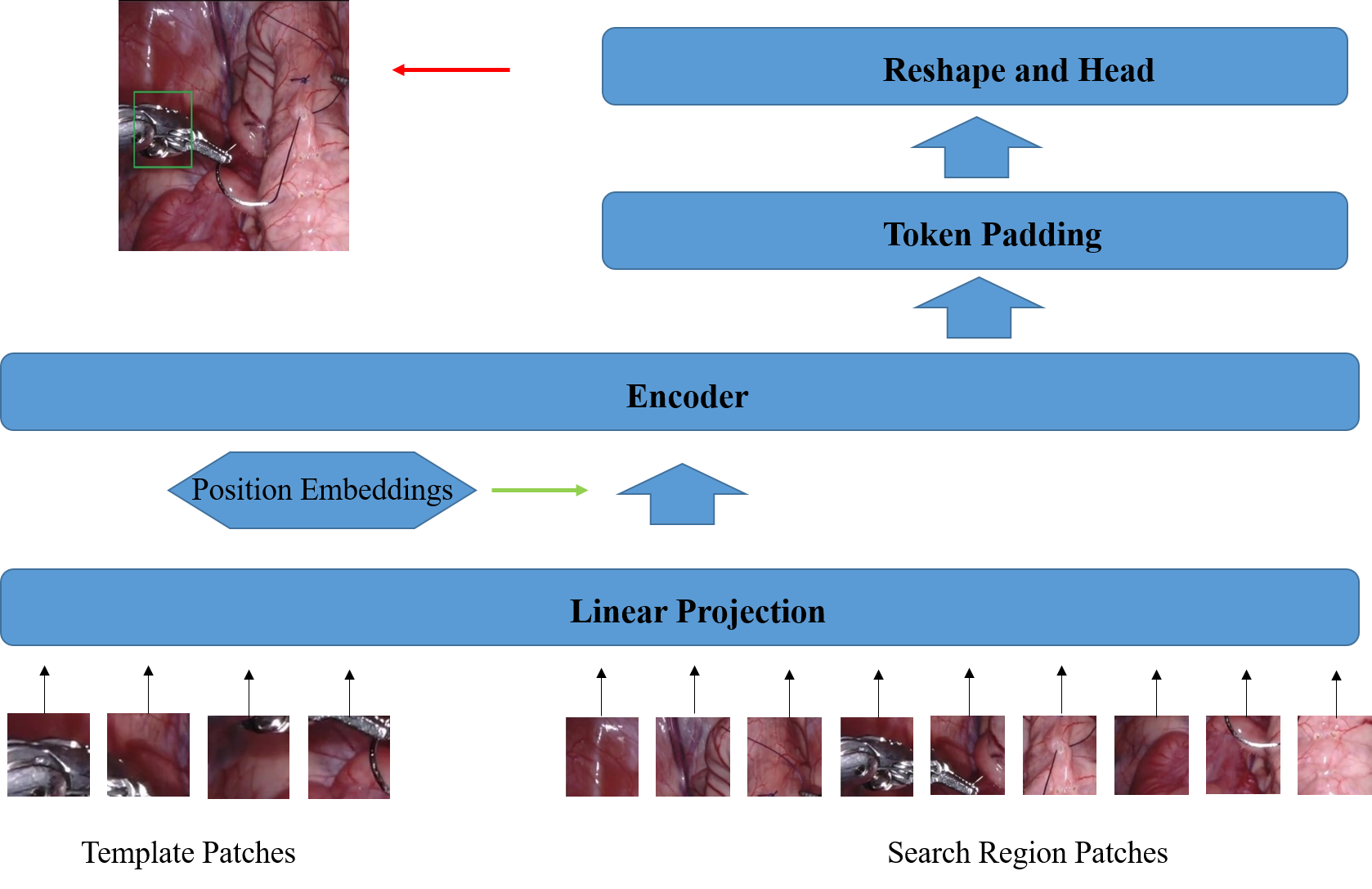}
    \caption{OsTrack Pipeline}
    \label{fig:fig2}
\end{figure}

\paragraph{Architecture Design}
Specifically, As Figure \ref{fig:fig2} shows, OStrack\cite{Ye_Chang_Ma_Shan} in our strategy consists of the following four steps: (1) The template and search region undergo patch embedding and spreading to recover the token, which is added with the positional encoding and spliced to form the overall token. (2) The overall token is fed into the encoder, in which the similarity between each token and the template token is calculated, and the foreground token is retained while the background token is rounded off in descending order. 
(3) Token padding is then performed to complement the rounded tokens with zeros and output to the head. (4) The head section outputs the classification score, the predicted offset value to compensate for down-sampling quantization error, and the normalized bounding box size, respectively.

\paragraph{Model Loss Function}
To train the one-stage object detector, we adpoted a dynamic soft label assignment strategy based on SimOTA \cite{Ge_Liu_Wang_Li_Sun_Technology}. As RTM-DET \cite{Lyu_Zhang_Huang_Zhou_Wang_Liu_Zhang_Chen_Conv_Concat}does, we used the IoU between the predictions and ground truth boxes as the soft label to train the classification branch, used the logarithm of the IoU as the regression cost instead of GIoU used in the loss function, used a soft center region cost instead of a fixed center prior to stabilize the matching of the dynamic cost.

\paragraph{Post Processing}
We adopted Kalman filtering approach based on weighted boxes fusion (WBF) \cite{Solovyev_Wang_Gabruseva_2021} to achieve a higher detection rate.To be specific, the implementation of WBF has the following steps: (1) The prediction frames of both the detection and the tracking model are normalized by coordinates and confidence, and clustered into a single set B and sorted in descending order using the confidence of the detection frames. (2) two empty lists are initialized, list one is used to hold clusters of detection frames for each target and list two is used to hold fusion frames for each target. (3) The algorithm would iterate through the predictions in B and calculate Intersection over Union (IoU) with each fused frame in list two. (3) When IOU is greater than a manually set threshold “T”, the clusters are considered to be in the same class, and coordinates and confidence of the fused frames are calculated. On the other hand, when IOU is smaller than T, the new box is defined as a new cluster center and is added to the tails of list one and list two. Then the iterative traversal for all the boxes in B is repeated. Eventually the box in List two would be the result.

\subsubsection{Model Training}
\begin{figure}
    \centering
    \includegraphics[width=0.75\textwidth]{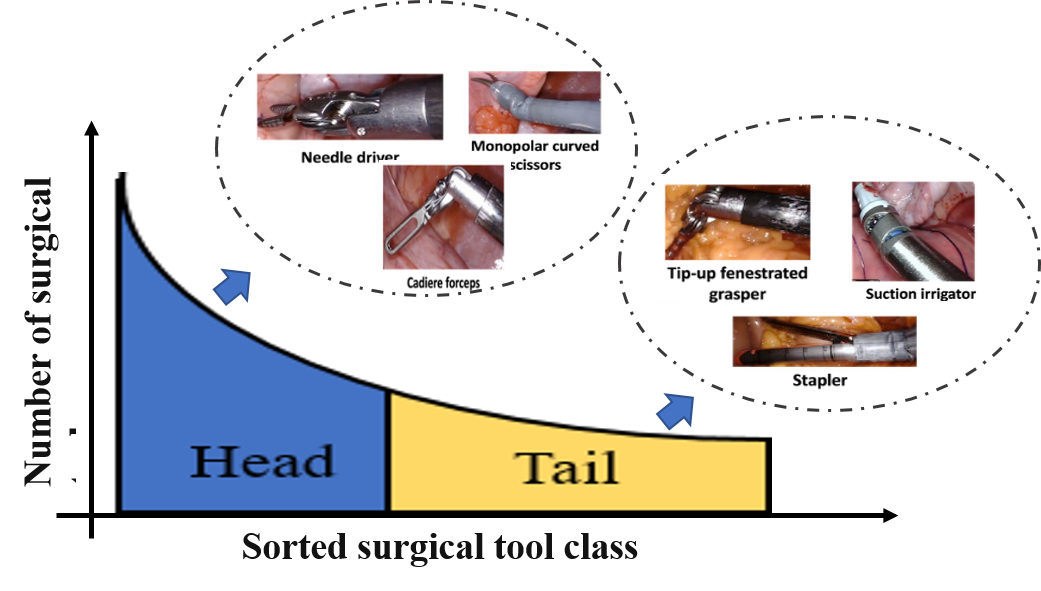}
    \caption{Long Tail}
    \label{fig:fig3}
\end{figure}

\paragraph{Long Tail}
As we noticed that our test metric is MAP, and we realized that the 14 low frequency instruments have long tail distributions, which will lead to the model overfitting,see Figure \ref{fig:fig3} .To avoid this, we utilized the tools presence labels to generate pseudo labels in similar size by instrument categories.

\paragraph{Implementation Details}
We divided all the data into a training set and a validation set by 7:3, and then conducted model experiments, data augmentation, category balance, and ensemble experiments, respectively. For every experiment, we trained our models using AdamW optimizer with an initial learning rate of 0.004, momentum 0.9, and a batch size of 64 for 100 epochs on the training set and validated them on the validation set. We resize the input images to 640 x 640 and the learning rate is annealed following the poly learning rate policy. All experiment were trained on 8 NVIDIA A100 GPUs. We evaluate the model performance on object detection and instance segmentation by bbox AP.

\subsubsection{Preliminary Performance}
\paragraph{Results}
Tab1-Tab3  show which modules perform best on the current validation set.Then, we choose RTM-DET \cite{Lyu_Zhang_Huang_Zhou_Wang_Liu_Zhang_Chen_Conv_Concat} as our detection model, cbnet-swim \cite{Liu_Wang_Wang_Liang_Zhao_Tang_Ling_2020} as our backbone, auto-augment as data augmentation, and adopt category balancing strategy. Finally, we adopted the integration of RTM-DET and the fine-tuned tracking model as our final submission.

\begin{table}[]
\centering
\begin{tabular}{|l|l|l|}
\hline
\textbf{Model}          & \textbf{Size} & \textbf{Bbox\_map} \\ \hline
Cascade R-CNN-Detectors & (894,682)     & 63.6               \\ \hline
RTMDet-l                & (864,864)     & 70.6               \\ \hline
YOLOv8-l                & (864,864)     & 69.1               \\ \hline
Co-DETR-swin-s          & Multi   scale & 64.6               \\ \hline
\end{tabular}
\caption{ Model Experiments}
\label{table:table1}
\end{table}

\begin{table}[]
\centering
\begin{tabular}{|l|l|}
\hline
\textbf{Backbone}    & \textbf{Bbox\_map} \\ \hline
cspnext              & 70.6               \\ \hline
Convnext             & 70.6               \\ \hline
Convnext-v2          & 70.9               \\ \hline
swintransformer\_7   & 70.6               \\ \hline
swintransformer\_12  & 71.1               \\ \hline
Swintransformer\_MAE & 66.0               \\ \hline
resnet               & 63.6               \\ \hline
convnext             & 70                 \\ \hline
Cbnet\_swin          & 71.5               \\ \hline
Cbnet\_res2net101    & 70.1               \\ \hline
\textbf{Data Augmentation}                             & \textbf{Bbox\_map} \\ \hline
Mosaic\_mixup                                          & 70.6               \\ \hline
Mosaic\_mixup9                                         & 70.8               \\ \hline
Fog-Blur-Noise and   channelshiift                     & 71.2               \\ \hline
RandomBrightnessContrast   MotionBlur ShiftScaleRotate & 71.1               \\ \hline
Autoaugment                                            & 71.2               \\ \hline
convnext                                               & 70                 \\ \hline
Cbnet\_swin                                            & 71.5               \\ \hline
Cbnet\_res2net101                                      & 70.1               \\ \hline
\textbf{Class Balance} & \textbf{Bbox\_map} \\ \hline
w/o                    & 70.6               \\ \hline
w                      & 71.4               \\ \hline
\end{tabular}
\caption{ Main Experiments}
\label{table:table2}
\end{table}

\begin{table}[]
\centering
\begin{tabular}{|l|l|l|}
\hline
Model              & Fusion    & \textbf{Bbox\_map} \\ \hline
rtmdet-siamfcpp    & wbf\_mean & 65                 \\ \hline
rtmdet-OStrack     & wbf\_mean & 66.7               \\ \hline
rtmdet-OStrack\_FT & wbf\_mean & 67.9               \\ \hline
rtmdet-OStrack\_FT & wbf\_max  & 64.2               \\ \hline
\end{tabular}
\caption{ Ensemble experiments}
\label{table:table3}
\end{table}

\paragraph{Conclusion and Discussion}
We have adopted the method of integrating detection models with tracking models. Although it has shown effectiveness in offline evaluations, it has not shown greater advantages in online evaluations. We believe that the online test set may be more complex, and our tracking model does not perform well. Compared to the previous semi supervised schemes, we have introduced self supervised learning, optimized the detection and tracking models, and considered class imbalance, which are still directions for future optimization.

\subsection{Team Name: CAIR-HK}


\subsubsection{Method Description}
Due to varying instrument visibility across frames, relying solely on video-level labels can induce substantial noise. Developing models capable of effectively detecting surgical instruments necessitates designing pseudo-label generation algorithms to maximize the utilization of available labels. Many existing approaches employ manual annotation to supervise models, but the vast number of videos in the dataset renders manual labeling extremely laborious and non-autonomous. Our approach substantially reduces the requisite human annotation effort while achieving an excellent compromise between manually labeled data volume and model performance. 

In this work, we proposed a Weakly Supervised Yolo Network (WS-YOLO)\cite{wei2023ws} for Surgical Tool Localization in Endoscopic Videos. Specifically, we preliminary trained an object detection model to distinguish the 3 parts of surgical tools (shaft, clevis, and tip) using the annotated public dataset (Endovis \cite{endovis}). Then, we sampled the 24695 video clips provided by the challenge \cite{zia2023surgical} at a frequency of 1/30 and performed inference on the sampled dataset using the trained model, obtaining an initial pseudo-label dataset via the Initial Label Filter strategies. Subsequently, we trained a new model with surgical instrument detection capabilities based on this initial pseudo-label dataset. We then selected new pseudo-labels by screening the inference results from this and previous models, leveraging Multi-round Label Filter strategy. Generating a new pseudo-label was iterated multiple times to elevate the performance. Our approach significantly diminishes the necessary human annotation labor while striking an optimal balance between the quantity of manually annotated data and detection performance. 
\subsubsection{Method}
\begin{figure}[t]
\includegraphics[width=0.75\textwidth]{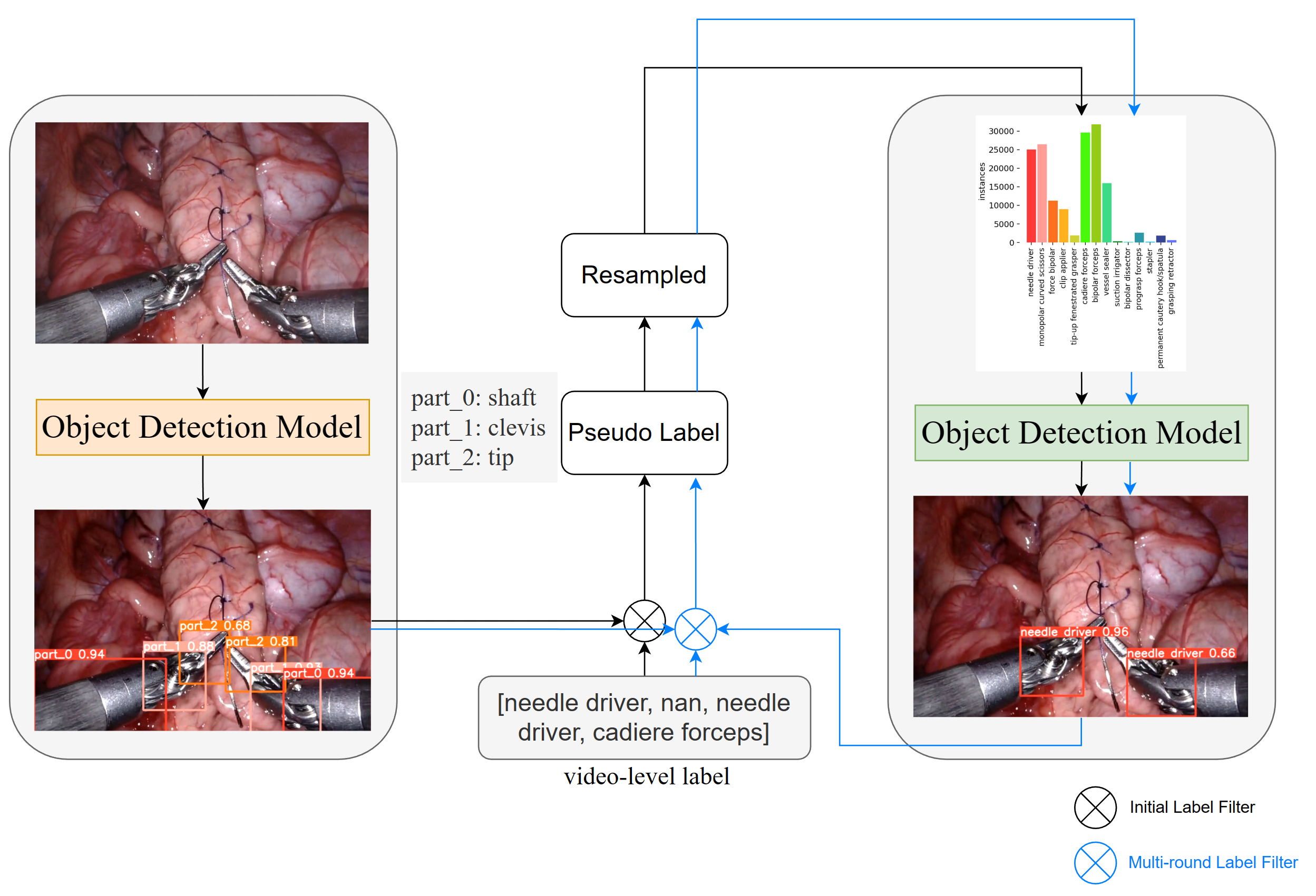}
\caption{The overall pipeline of our proposed method.} \label{fig1}
\end{figure}
Fig. \ref{fig1} presents the overall pipeline of our proposed method. Initially, we transformed the semantic masks in the Endovis \cite{endovis} into labels for object detection. We trained a model $Det_{parts}$ based on YOLOv8 \cite{yolo2023v}, capable of localizing the shaft, clevis, and tip, as depicted on the left of Fig. \ref{fig1}. We sampled every 30th frame from each video clip for the dataset. We performed inference with the trained model $Det_{parts}$, on the sampled dataset, obtaining corresponding bounding boxes (localizing the shaft, clevis, and tip in each sampled frame).
\begin{figure}[t]
\includegraphics[width=\textwidth]{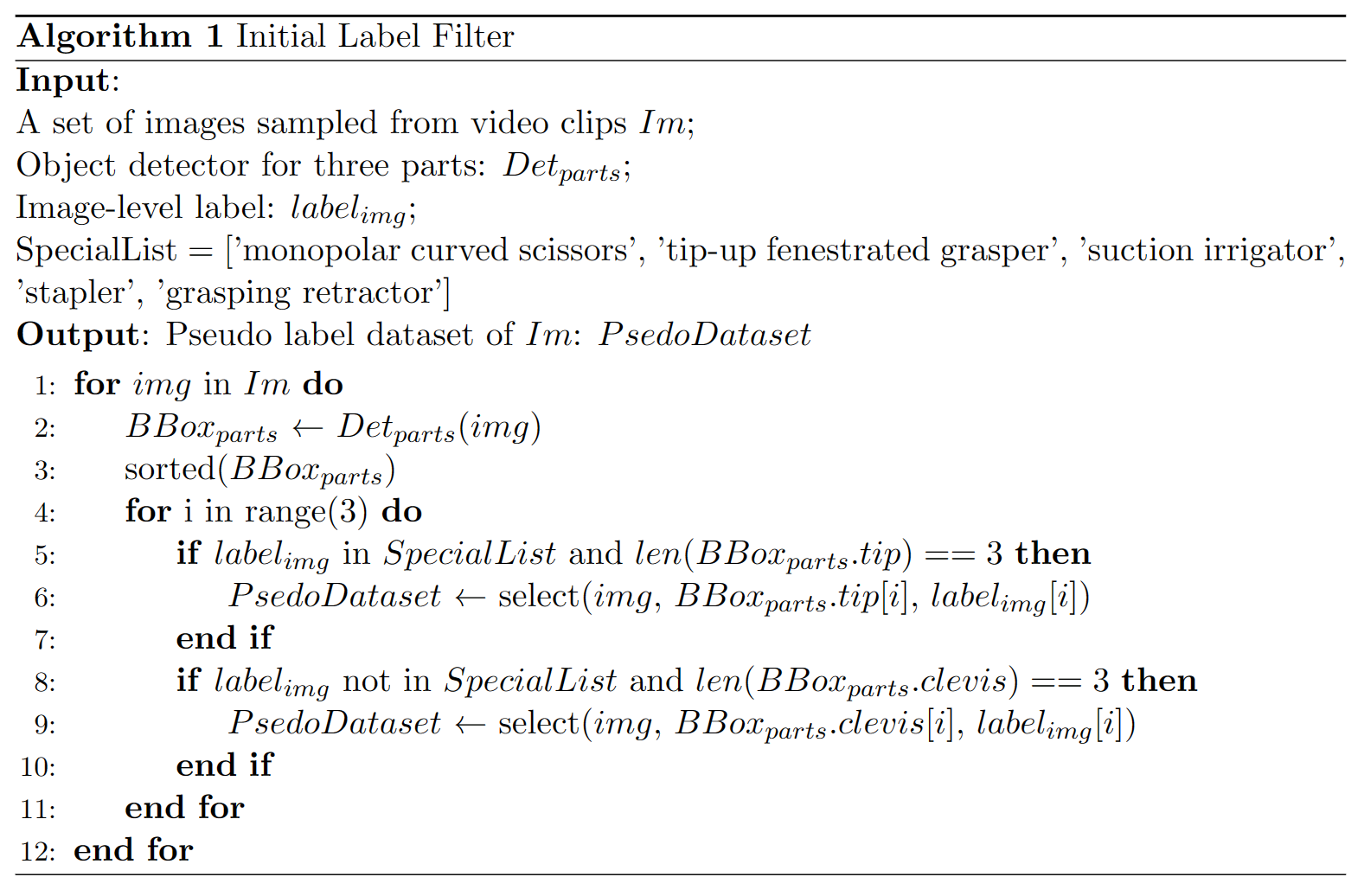}
\caption{Pseudo-code of Initial Label Filter Algorithm.} \label{fig2}
\end{figure}
Due to varying instrument visibility across frames, relying solely on video-level labels can induce substantial noise. Developing models capable of effectively detecting surgical instruments necessitates designing pseudo-label generation algorithms to maximize the utilization of available labels. Many existing approaches employ manual annotation to supervise models, but the vast number of videos in the dataset renders manual labeling extremely laborious and non-autonomous. Our approach substantially reduces the requisite human annotation effort while achieving an excellent compromise between manually labeled data volume and model performance. 

\begin{figure}[t]
\includegraphics[width=\textwidth]{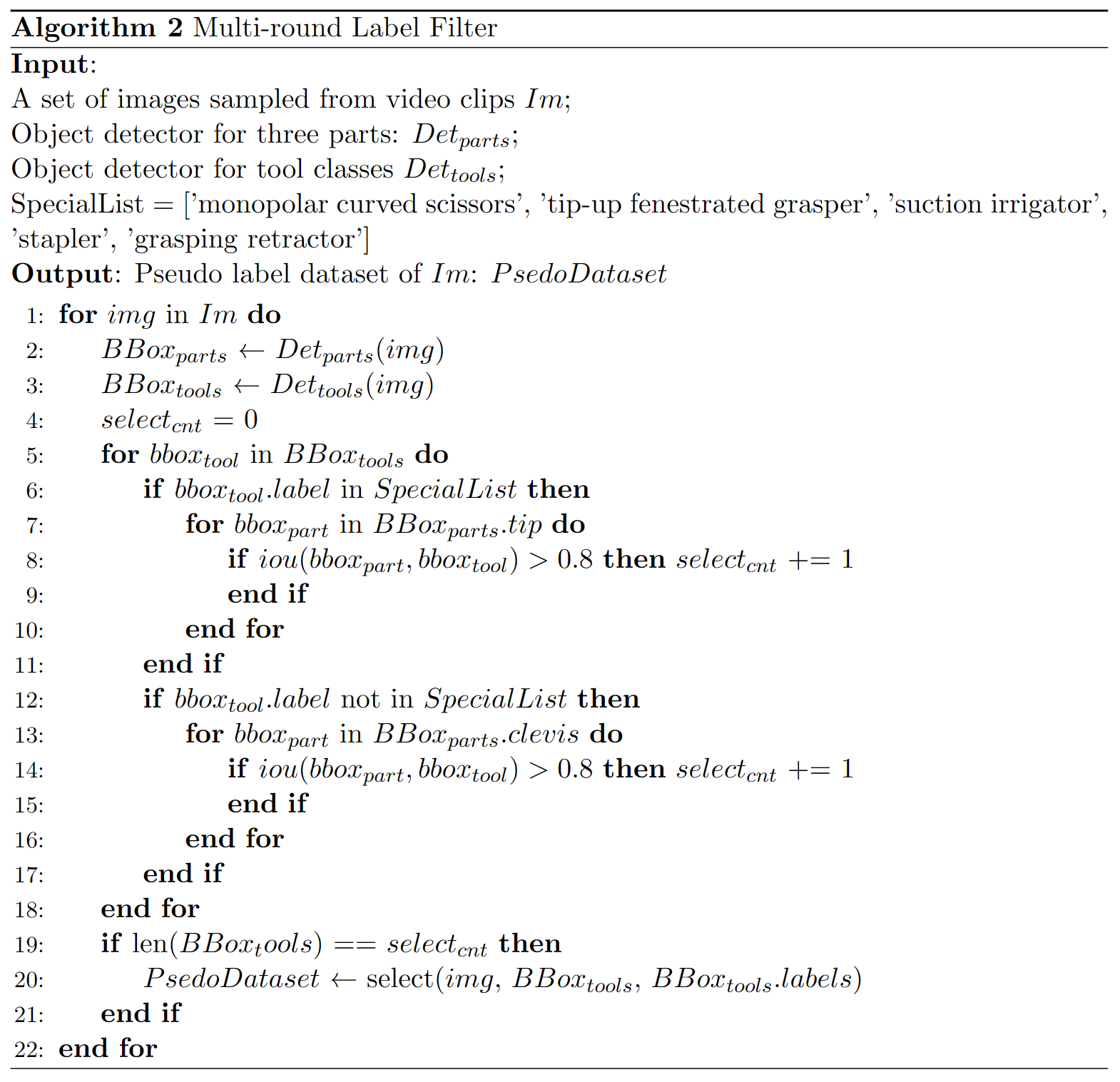}
\caption{Pseudo-code of Multi-round Label Filter Algorithm.} \label{fig3}
\end{figure}

Since the bounding boxes output by $Det_{parts}$  did not contain category information of surgical tools, we adopted an Initial Label Filter approach to match the outputs of $Det_{parts}$ with video-level labels, as illustrated in Fig. \ref{fig2}. 

Given that each video-level label contains at least three tool categories arranged in the order of the robot's three arms (from left to right), based on this prior knowledge, we posited that in most cases the tools from left to right in the surgical scene should correspond one-to-one with the video-level label order. Thus, in Algorithm 1, we first sorted the bounding boxes output by $Det_{parts}$  from left to right. In the initial label filter, we only selected frames that concurrently detected three clevises or three tips to generate pseudo-labels. We designated 'monopolar curved scissors', 'tip-up fenestrated grasper', 'suction irrigator', 'stapler', and 'grasping retractor' as special classes requiring tip detection, such that frames were only included in the pseudo-label dataset if $Det_{parts}$  concurrently detected three tips when the label belonged to these categories. 

After generating the preliminary round pseudo-labeled datasets, we trained a detection model called $Det_{tools}$ using YOLOv8 \cite{yolo2023v}. Similarly, we performed inference on the sampling dataset using the trained $Det_{tools}$  model to obtain detection boxes $BBox_{tools}$ corresponding to instruments in each video frame (localizing all instruments in every frame). Since both $Det_{parts}$ and $Det_{tools}$ contain positional information of instruments in the video frames, but the first round of pseudo-labels had considerable noise, $Det_{tools}$ had lower accuracy. Therefore, we adopted a multi-round label filtering approach to select high-quality pseudo-labels, as shown in Algorithm 2 (Fig. \ref{fig3}). For each detection box $bbox_{tool}$ from $BBox_{tools}$ that corresponds to a particular instrument category, we checked if there exist any $bbox_{part}$ with category tip that has IoU $>$ 0.8 with $bbox_{tool}$, and any $bbox_{part}$ with category clevis that has IoU $>$ 0.8 with $bbox_{tool}$. Only when all $bbox_{tool}$ in a frame satisfy these conditions, the pseudo-label is considered trustworthy.
%
%
%
%

\subsubsection{Model Training}
The training strategy of basic detection model is following the default setting of YOLOv8, and the iteration strategy of training and data pre-processing are illustrated above. We take 1 Tesla A100 with batch size set to 8 to train 1000 epochs. 

\subsubsection{Preliminary Performance}

In the preliminary testing, we obtained a mean average precision (mAP) of 0.043 using the initial round of pseudo-labeled data generated by Algorithm 1. In addition, we manually annotate a few video clips to figure out the problem of data unbalanced ($clip_{5670}$, $clip_{5672}$, $clip_{5675}$, $clip_{5985}$, $clip_{5986}$, $clip_{5987}$, $clip_{16647}$, $clip_{16650}$, $clip_{16652}$, totally 549 images are manually annotated, where $clip_i$ denotes the $i_{th}$ video clip in dataset). 

\begin{table}
\centering
\caption{The performance of WS-YOLO in terms of iterations.}
\label{tab1}
\begin{tabular}{c|ccccc}
\hline
Iteration & $round_0$ & $round_1$ & $round_2$ & $round_3$ & $round_4$\\

\hline
mAP & 0.043 & 0.109 & 0.135 & 0.147 & 0.157\\
\hline
\end{tabular}
\end{table}
In the process of filtering high-quality tags over and over again, our algorithm creates a positive feedback mechanism that makes performance improve with each iteration. As shown in the Table. \ref{tab1}, through iterative refinement of the pseudo-labels using Algorithm 2, we achieved mAP scores of 0.109, 0.135, 0.147, and 0.157 after successive rounds, demonstrating the effectiveness of the multi-round pseudo-label filtering approach for improving model performance. 

\clearpage

\subsection{DKFZ - Team SDS-HD}


SDS-HD team members: Amine Yamlahi, Piotr Kalinowski, Dominik Michael, Tim Rädsch, Marco Hübner and Lena Maier-Hein.

SDS-HD was motivated by developing an innovative semi-automatic annotation pipeline leveraging the recently introduced keypoints tracking algorithm.

\subsubsection{Method Description}
Our approach focused on the development of an innovative annotation pipeline, leveraging the recently introduced keypoints tracking algorithm TAPIR ~\cite{doersch2023tapir}. More specifically, we addressed the absence of bounding box annotations in the training data by a pipeline based on reliable keypoints tracking. To this end, we designed a semi-automatic annotation pipeline, comprising the steps of manual keypoint annotation, automatic keypoint tracking,automatic conversion of keypoints into bounding boxes, and filtering of bounding boxes. The semi-automatically generated bounding boxes were used to train a multi-label instrument detection model. Multi-class classification was achieved with a Swin Transformer model ~\cite{liu2021swin} which received the (cropped) bounding boxes, generated by the detection model, as input.

We propose a two-stage pipeline consisting of a YOLOv5 detection model ~\cite{jocher2020yolov5} applied to individual frames and a Swin Transformer classifier, individually applied to the preprocessed bounding boxes of the detector. The core innovation of our approach is related to how the training data was assembled:
Training of the detection model: The semi-automatic annotation of the training data with bounding boxes was achieved with TAPIR for keypoint tracking. Based on a manual initialization of keypoints, the track any point (TAPIR) ~\cite{doersch2023tapir} model was used to transfer the keypoints to the subsequent frames of the same video. Subsequently, the keypoints were converted to bounding boxes, labeled with the corresponding instrument. After filtering of implausible results, the subset of frames with plausible bounding boxes was used to train an ensemble of multi-label YOLOv5 detection models. The trained ensemble was then applied to an additional set of clips to enlarge the final training set for the classification model. 
Training of the classification model: For training of the classifier, the data with which the detection model was trained was converted to a set of cropped bounding boxes resized to 224x224 pixels with corresponding instrument label. The same workflow (detection, filtering, cropping) was applied to a set of external videos from five sources~\cite{Allan2019-2017, AllanMICCAI2018,garcia2021image,yoon2022surgical, cuzzolin2021saras} disregarding any existing labels from these datasets. 

%
%
%
%
\begin{figure}[h!]
\begin{center}
\resizebox{6in}{!}{\includegraphics{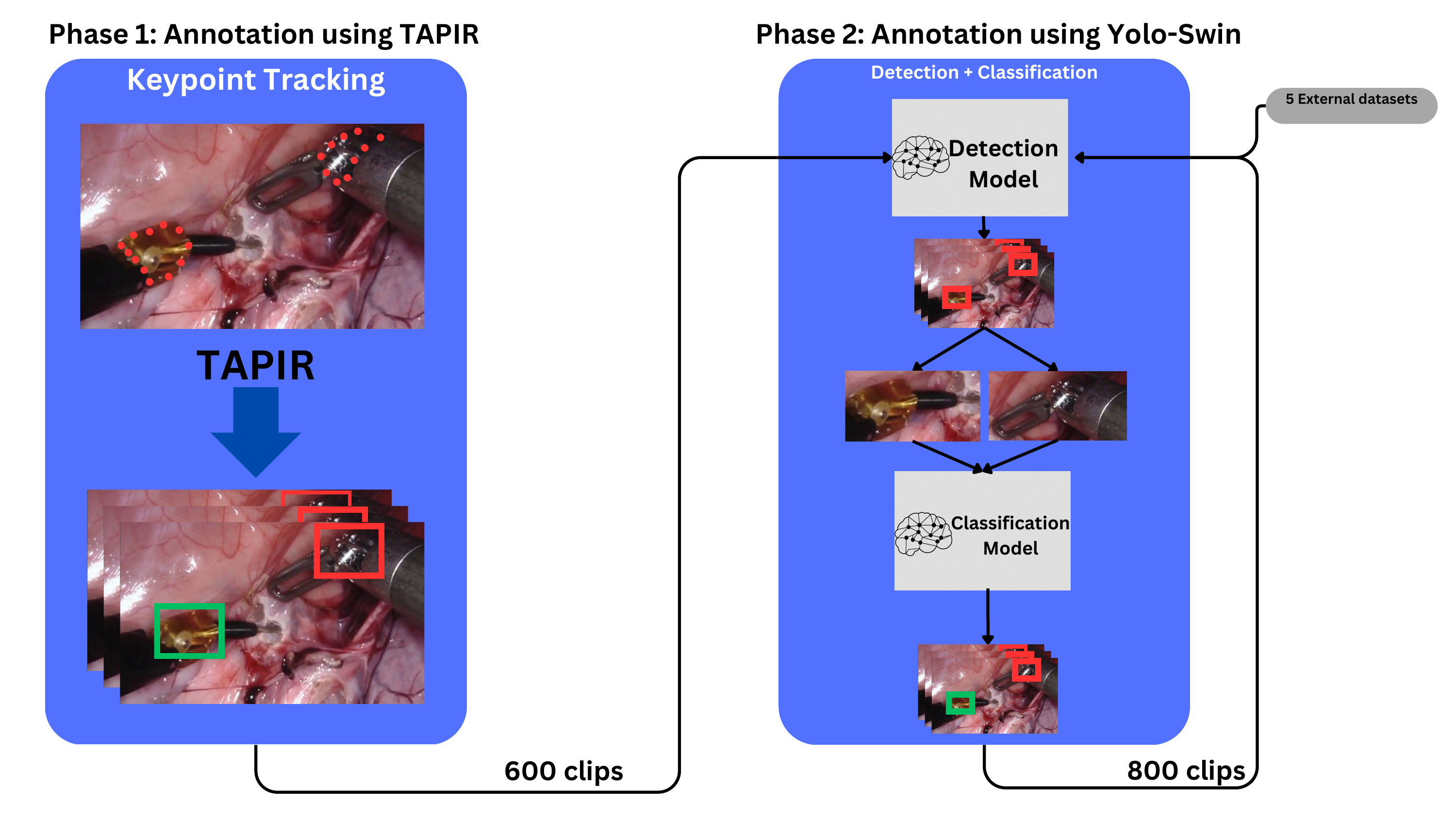}}
\caption{Proposed training workflow: (a) Keypoints tracking-based bounding box generation. (b) Classification based on cropped bounding boxes. }
\label{modelFig}
\end{center}
\end{figure}

\subsubsection{Model Training}
\paragraph{Data}

Semi-automatic annotation of training data: We selected 1400 clips (6 \%) of the provided training data for semi-automatic annotation. Criteria for selection were class prevalence as well as diversity in motion, number of instruments and surgical scene diversity.
Incorporation of External Data: To enhance the provided training set with five external detection and segmentation datasets ~\cite{Allan2019-2017, AllanMICCAI2018,garcia2021image,yoon2022surgical, cuzzolin2021saras} representing three distinct surgical procedures: nephrectomy, prostatectomy, and gastrectomy. These were used to enhance the variety within our training dataset. 

\paragraph{Bounding box generation using TAPIR}
For 600 out of the 1,400 clips selected for semi-automatic annotation, we implemented the following procedure: 

\textbf{(1) Initialization}: We first identified the first frame in which the relevant parts of at least two of the tools were visible. The clevises of the instruments were then annotated with keypoints and instrument label. We generated our annotation protocol based on the shared information of the challenge organizers. Following best practices ~\cite{radsch2023labelling}, we added additional images for the annotation classes and discussed ambiguous scenarios, in order to apply a consistent annotation style regardless of the annotator. Our annotation protocol is presented in the shared code repository.

\textbf{(2) TAPIR-based tracking}: All keypoints set manually were subsequently tracked in the entire clip using TAPIR. We displayed example images for the relevant classes directly in our annotation tooling to enable an easy access for the annotator to the annotation protocol information. 

\textbf{(3) Bounding box conversion:} Each set of keypoints corresponding to one instrument was converted to a bounding box. 

\textbf{(4) Filtering:} To overcome annotation uncertainties of TAPIR, resulting in keypoints “jumping” from one tool to another, or in moving to the background, we removed frames that met one of the following criteria:

\begin{itemize}
  \item Area Rule: At least one of the bounding boxes in the frame is substantially smaller (20\%) or bigger (80\%) than the corresponding initial, manually annotated box
  \item Narrowness Rule: The frame contains at least one box with a height-to-width or width-to-height ratio less than 0.2.
  \item Jumping Points: It features bounding boxes based on keypoints that transition from the instrument to the background. This is achieved by implementing a minimum distance margin of 20 pixels between two consecutive frames. 
\end{itemize}

\textbf{(5) Manual refinement:} As a final step of quality control, any noisy annotations that managed to pass through the filters were manually removed. Optionally, the original keypoints could be refined, leading to a repetition of steps (2)-(4). 
This process applied to the 600 clips yielded 114134 frames, which were used to train a multi-label YOLOv5 detection model. The trained model was applied to further 800 clips from the training data as well as 6500 frames from the five external datasets. After manual cleaning, the process resulted in a total of 268020 frames annotated with bounding boxes and corresponding instruments.
\paragraph{Instrument detection}
Instrument detection was accomplished by training three YOLOv5 ~\cite{jocher2020yolov5} models: two L6 (with and without external data respectively) and one X6, utilizing annotations generated by the keypoints tracking model. The training process employed an initial learning rate of 0.003 and a final learning rate of 0.004, stochastic gradient descent (SGD) optimizer and Binary Cross-Entropy (BCE) loss function. Augmentations applied during training encompassed rotation, flipping, adjustments to hue, saturation, and value, in addition to mosaic augmentation techniques. An ensemble of the three models, created using weighted box fusion ~\cite{Solovyev_Wang_Gabruseva_2021}, yielded the final model.

\paragraph{Instrument classification:}
The performance of the multi-label YOLOv5 detection model in terms of localization was adequate; however, classification of the instruments occasionally posed challenges, particularly when dealing with instruments that shared similar clevises, such as the cadiere forceps and the prograsp forceps. To address this limitation in our pipeline, we introduced a Swin Transformer ~\cite{liu2021swin} second-stage classifier to improve the classification results generated by the integrated classifier of the detection model (Fig. 1). This classifier operates on the cropped bounding boxes generated by the detection model, with the added step of enlarging the bounding boxes to encompass the instrument's tip. This enlargement provides additional context and aids the classifier in distinguishing between similar instruments more effectively. The model underwent training for 20 epochs, employing a cosine annealing scheduler that initiated with a learning rate of 2e-4 and concluded at 2e-5. The training process employed a binary cross-entropy loss function and incorporated augmentations such as flips, rotations, and color jitter.

All the models were trained with Nvidia RTX 3090 GPUs.





\clearpage
\subsection{Team Name: TUE-VCA}
Team member: T.J.M. Jaspers

\subsubsection{Method Description}

\begin{figure}[h!t]
    \centering
    \includegraphics[width=\textwidth]{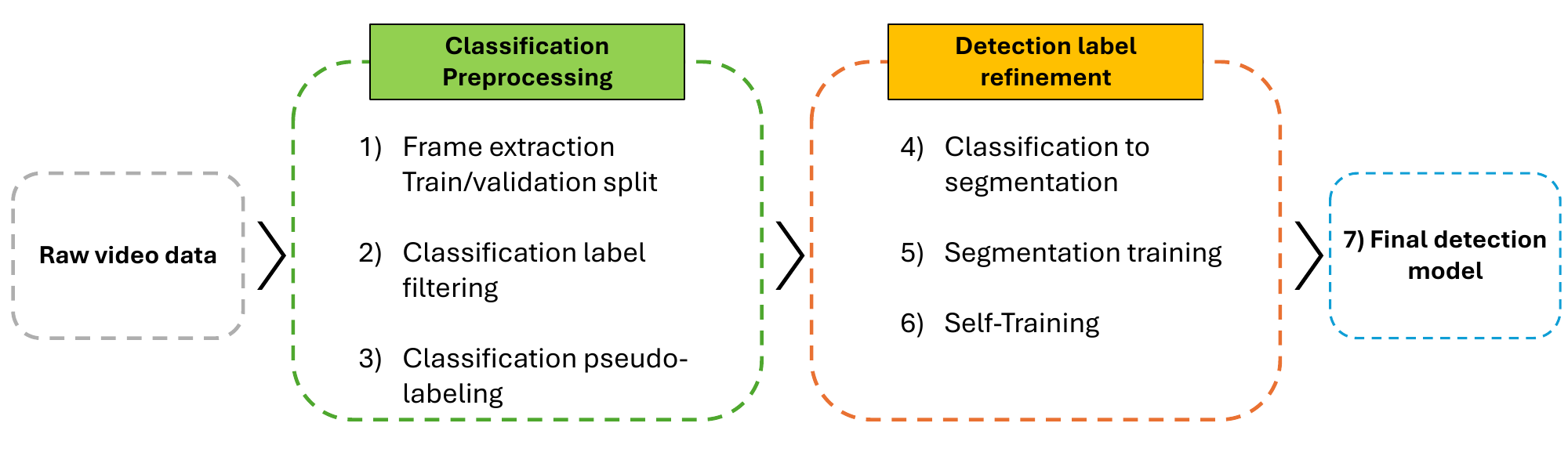}
    \caption{Team TUE-VCA Schematic representation of the proposed workflow.}
    \label{fig:method}
\end{figure}

Figure \ref{fig:method} presents a schematic representation of our proposed methodology used for the SurgLoc2023 challenge. Our method comprises several steps that culminate in the development of our final detection model. Initially, we work on preprocessing the data at a classification level, minimizing the amount of label noise present in the data. Following this, we transition from classification to detection. The refined detection labels serve as the final data for training our YoloV8 detection model.
\\
\\
\textbf{Classification preprocessing}
\\

  1) To initiate the process, we adopted the data-processing pipeline recommended by Team NVIDIA in the SurgLoc 2022 challenge, which employs a strategy of dividing the video chunks based on the differences in their hash values \cite{zia2023surgical}. This preliminary step of preprocessing facilitated the efficient organization of the video data. We decided to construct a small validation set, deliberately including at least one instance of each instrument type. The remainder of the video data was allocated for training purposes. Given the constraints of computational resources, we opted to extract 2 frames from each video chunk. This resulted in a total of 47,227 frames for training and 1,073 frames for validation.
\\

2) Initially, our dataset encountered a challenge due to the occasional invisibility of the tool's clevis within the frame. This resulted in a discrepancy between the automatically generated labels and the tools visible in the frames. To address this, we employ a segmentation model, leveraging the dataset of the EndoVis17 challenge, which is specifically tailored for robotic instrument segmentation \cite{Allan2019-2017}. The segmentation model adopts the Unet architecture with a Mit-b4 encoder \cite{Ronneberger2015arxiv, xie2021segformer}, and is trained using the cross-entropy loss function with the Adam optimizer, employing a learning rate of 1e-3. This model identifies the surgical tools' presence within the frames extracted from our video data. We select the frames in which the count of detected tools matches the number of provided labels. This filtering process yielded a refined dataset comprising 29,562 frames for training and 904 frames for validation.
\\

3) To increase the dataset, we use a form of pseudo-labeling. Initially, we train a classification model using the filtered dataset. This model uses the same encoder architecture described in the previous section, with the inclusion of a classification head. Additionally, we maintain the same loss function and optimizer configurations. After the training phase, we used the classification model to refine the labels for frames where the number of detected tools did not match the provided labels. We accomplish this by removing labels with the lowest probability, which contributes to the refinement of our final dataset. As a result, our final dataset includes all 47,227 training images and 1,073 validation images.
\\
\\
\textbf{Detection label refinement}
\\

4) To transition from classification to detection we first create a segmentation network similar to the one described in step (2), however, at the end of the encoder we add a classification head.  Initially, we train the encoder with a classification head on the entire classification dataset. Subsequently, we generate gradient-based saliency maps from the encoder and filter them using the output of the segmentation network described in step (2), solely using the prediction mask of the tool-clevis, or tools-tip for the specified tools in the data description. 
\\

5) While we acknowledge the potential presence of noise in our pseudo-segmentation masks, we have taken steps to mitigate its impact. Specifically, we adopt a symmetrical cross-entropy loss function proposed by Wang et al. \cite{wang2019symmetric}, which enhances the robustness of our segmentation model against label noise. The total loss function during this phase is an equally weighted combination of the classification loss (cross-entropy) and segmentation loss (symmetrical cross-entropy). We retain the Adam optimizer with a learning rate of 1e-3 for consistency.
\\

6) Additionally, we assume that, with several iterations, the output of our segmentation network would surpass the quality of the pseudo masks. Consequently, we apply a simple form of self-training, often employed in semi-supervised learning methods \cite{Yang_2023}. After 5, we computed a segmentation loss (symmetrical cross-entropy) for both the pseudo-segmentation masks generated in step (4) and the output of the segmentation network. The final loss function is as follows:
\\
\begin{equation}
\mathcal{L}_{\text{total}} = \alpha \mathcal{L}_{\text{cls}} + (\beta \mathcal{L}_{\text{seg}_{\text{masks}}} + (1-\beta) \mathcal{L}_{\text{seg}_{\text{output}}})
\label{eq:1}
\end{equation}
\\
Here the initial value of $\alpha$ and $\beta$ is 1 and decreases every epoch with 0.05. Causing the weight of the loss calculated based on the segmentation output to increase.
\\
\\
\textbf{Final detection model}
\\

7) After obtaining predictions from the final segmentation model, a filtering process is applied to retain only the largest areas corresponding to the image classification labels, while eliminating smaller regions. Subsequently, the processed masks are transformed into a Yolo format, which serves as the input for training our YoloV8 detection model. 

\subsubsection{Model Training}
The segmentation model, as described in (2), undergoes training for 25 epochs with a batch size of 4, utilizing images of resolution 512x512 pixels. Following this, the classification model outlined in (3) is initially trained for 10 epochs on the primary datasets, employing the same batch size and image resolution.

The encoder component of the segmentation model, detailed in (4), (5), and (6), is first trained on the complete classification dataset, using identical hyperparameters as the previous step. Subsequently, saliency masks are generated, and the entire encoder-decoder structure undergoes training for 5 epochs. For the subsequent 20 epochs, the model is trained utilizing the loss function described in Equation \ref{eq:1}.

The final YOLOv8 detection model is trained for 25 epochs, with a batch size of 4. All models utilize ImageNet-initialized weights, and experiments are conducted on an RTX 3090Ti GPU (NVIDIA Corp., CA, USA).

\subsubsection{Preliminary Performance}
This method eventually resulted in a mean average precision of 0.0934 on the final test set. While our current method has shown promise in localizing surgical tools within individual frames, it does not yet harness the valuable temporal information present in video sequences. An immediate enhancement could involve integrating a tracking algorithm in conjunction with our approach, enabling more robust tool tracking over time.

Nevertheless, one of the standout advantages of our methodology lies in its scalability. This scalability paves the way for future developments with even larger datasets. Currently, we have limited the data input to just 2 frames per video. However, with additional resources and computational power, we have the potential to significantly increase this number without relying on additional human inference.

\clearpage


\subsection{Results} 

Challenge evaluations were conducted on a private, concealed test set comprising individual surgical video files at 1 FPS. The Grand Challenge automated algorithm submission and evaluation system generated the results. Participants did not have access to the test data and labels. The evaluation of submissions employed the standard COCO dataset bounding box detection metric to determine the winners. The COCO (Common Objects in Context) dataset bounding box detection metric is a widely used evaluation measure in computer vision, particularly in the field of object detection. The bounding box detection metric evaluates the accuracy of object localization and classification by comparing the predicted bounding boxes of objects in each frame to the ground truth bounding boxes provided by the dataset. The metric is expressed here as mAP (mean Average Precision) at various Intersection over Union (IoU) thresholds. The IoU measures the overlap between the predicted bounding box and the ground truth bounding box, and IoU thresholds range from 0.5 to 0.95, with a step size of 0.05.
Table \ref{tab:performance_results} is a summary of the challenge results.

\begin{table}[h!]
\centering
\begin{tabular}{clr}
\toprule
\textbf{Rank} & \textbf{Team} & \textbf{mAP} \\
\midrule
1st  & PUMCH & 0.4669 \\
2nd  & ZJURealdoctor & 0.3899 \\
3rd  & Jmees & 0.3478 \\
4th  & seventeen & 0.3404 \\
5th  & ANL-Surg & 0.3377 \\
6th  & SDS-HD & 0.3074 \\
7th  & HVRL & 0.1774 \\
8th  & LozaGera & 0.1688 \\
9th  & CAIR-HK & 0.1489 \\
10th & AIT & 0.1139 \\
11th & TUE-VCA & 0.0934 \\
12th & MapleLab & 0.0007 \\
13th & LabRen-CUHK & 0 \\
\bottomrule
\end{tabular}
\caption{Highest Performance Results by Team}
\label{tab:performance_results}
\end{table}

There is a significant variation in performance among the teams, as indicated by the range of mAP scores. A common characteristic among these top-performing teams is the generation of pseudo-labels to aid the training process. For example, PUMCH employed OSTrack for object tracking based on the provided annotated frames, ANL-Surg performed backward and forward tracking using Mixformer, and SDS-HD developed a semi-automatic annotation pipeline. All of these teams utilized attention-based models at some point.

\subsection{Discussion} 



Considering the team's results as a whole, some general observations can be made. As with the previous year, training the models with the noisy labels has proved to be challenging. Most successful strategies employed pseudo-labels to aid the training process. Many teams also made use of small manually annotated data sets, either labeled themselves, or by a previous year's team. These labeled data sets where then bootstrapped to created larger labeled data sets. And, importantly, many teams followed approaches laid out by teams from the prior year.

\begin{figure}
 \centering
 \includegraphics[scale=0.25]{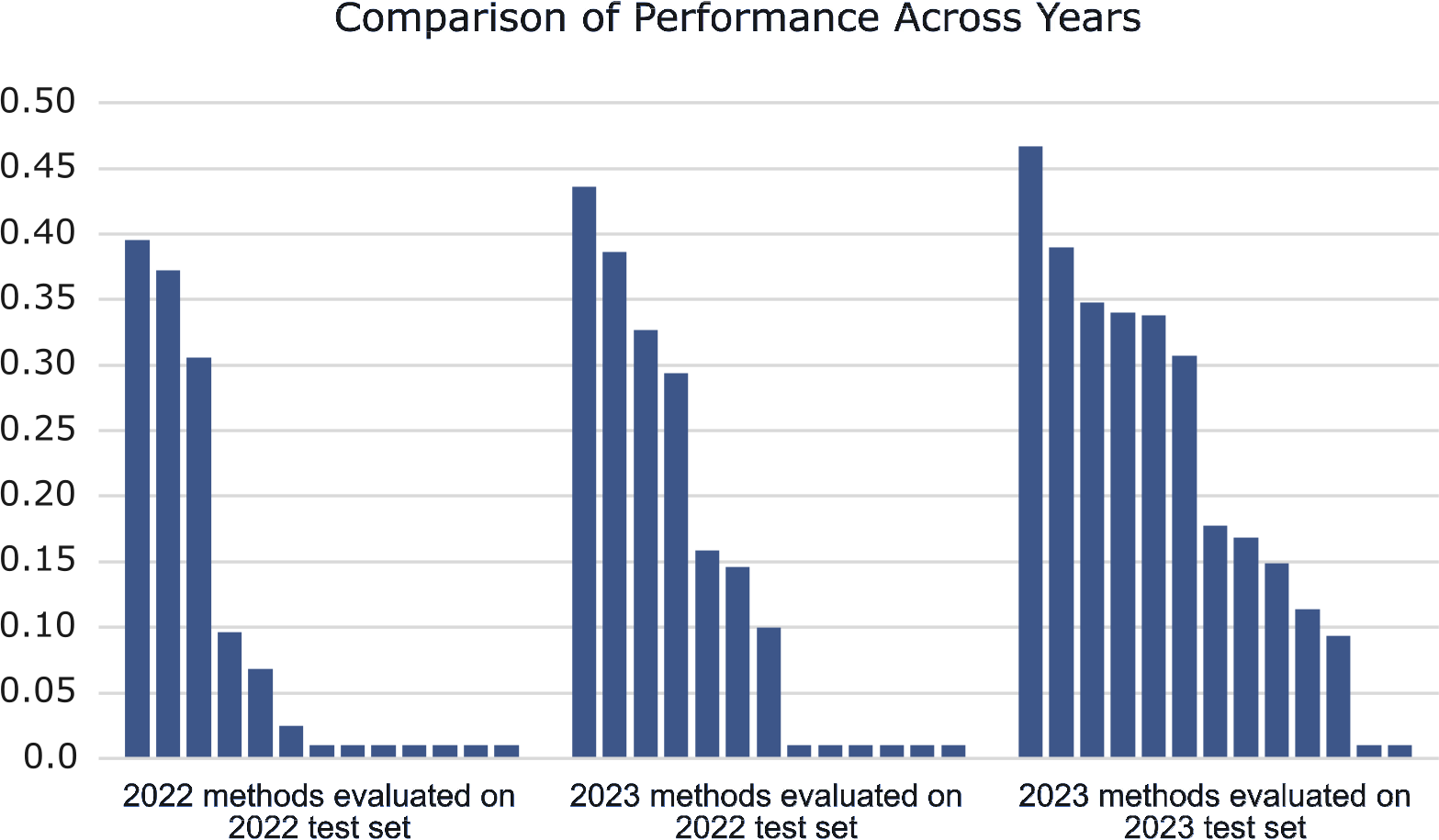}
 \caption{Overview of challenge results from 2022 to 2023 demonstrating a robust increase in model performance. Results from the 2022 and 2023 challenge are plotted alongside the 2023 models evaluated on the 2022 test set.}
 \label{fig:2023_vs_2023_results_fig}
\end{figure}


To provide more context for this year's results we compared them alongside those from the identical challenge of 2022 (Category 2). On the whole the models clearly performed better (compare the left and right bar plots, Fig \ref{fig:2023_vs_2023_results_fig}). However, the 2023 test set was different from that of 2022. To control for this we evaluated this year's models on the 2022 test set (middle bar plots, Fig \ref{fig:2023_vs_2023_results_fig}). As can be seen from the overall decreases in performance, the 2022 test set was more challenging. However, despite this the 2023 models still outperformed those from the previous year. 
This is general trend of year-over-year improvements is precisely what the challenge was designed to foster, and why we intend to continue with the challenge in the future.

The current state-of-the-art models for object detection, such as CO-DETR \cite{zong2023detrs}, have achieved remarkable mAP values of up to 0.66 when trained and evaluated on general datasets such as COCO. Comparing these high-performance models with the those of the teams participating in this challenge reveals an apparent performance gap. 
This disparity can likely be attributed to many factors, but chief among them may be the specialized nature of surgical tool detection, which presents unique challenges not encountered in general object detection tasks.
While our dataset contains thousands of video clips, the individual frames are often very similar and highly correlated, due to the nature of the surgical field (e.g. a background of red colored organs and metallic tools in the foreground). When contrasting this dataset with images from the COCO dataset, which may contain passenger jets or images of flowers, the differences are readily apparent. As such this dataset demands a relatively high degree of exactitude when distinguishing categorical labels.
Closing the performance gap between specialized tasks like surgical tool detection and general object detection remains an ongoing challenge, with implications for improving medical imaging and healthcare.

These complications notwithstanding, the efforts of the participating teams demonstrate valuable progress in this domain and lay the groundwork for future advancements. 
The diverse range of teams participating in the challenge, drawn from institutions around the world, showcases the competitive landscape of surgical tool detection research, and a cause for an optimistic outlook.

\newpage
\section{Results and methods from the MICCAI 2024 SurgVU challenge}

\subsection{SurgVU 2024 Challenge Description}

\subsubsection{Overview}

\begin{figure}[tbh]
 \begin{center}
\resizebox{5in}{!}{\includegraphics{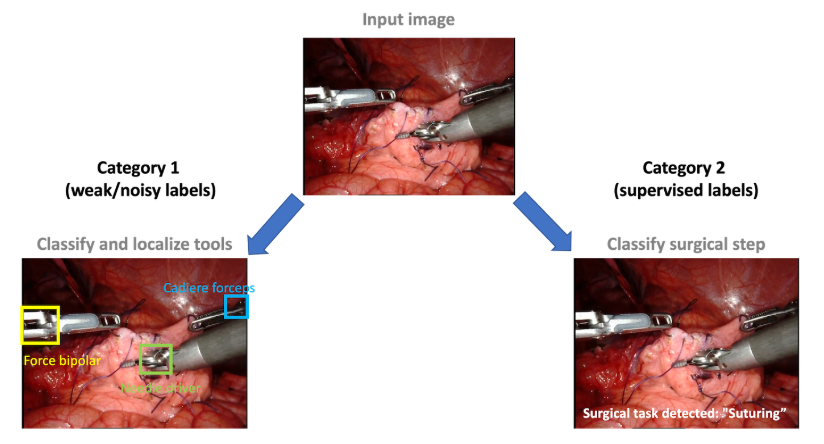}}
 \caption{Overview of challenge categories 1 and 2 (2024)}
  \label{fig:overview_categories_2024}
\end{center}
\end{figure}

The SurgVU 2024 sub-challenge was arranged as part of the Endoscopic Vision Challenge at MICCAI 2024. It consisted of two categories designed to address complementary tasks in surgical video understanding (Figure~\ref{fig:overview_categories_2024}). 

Category 1 focused on \textbf{surgical tool classification and localization} under a weakly supervised setting. Participants were required to develop models capable of detecting and localizing surgical instruments using only noisy tool presence labels provided at the video or clip level, without access to ground truth bounding box annotations.

Category 2 addressed \textbf{surgical task recognition}, where teams were required to classify the surgical step being performed at the frame level. In contrast to Category 1, this task was fully supervised, with frame-level annotations provided for training.

\subsubsection{Training Data}

The 2024 challenge used an expanded version of the training data from the 2022 and 2023 SurgToolLoc challenges. The dataset consisted of 280 long videos from 155 training sessions, captured at 60 fps with a resolution of 720p (1280 $\times$ 720) from one channel of the endoscope. This translates to over 840 hours of surgical video, resulting in over 18 million frames. Each clip contained up to three of 12 possible surgical tools from the same instrument set as in prior years (see Section~\ref{chapter_dataoverview}).

For each training session, two types of labels were provided: (1) \textit{tool presence labels}, indicating which robotic tools were installed in each frame, and (2) \textit{task labels}, indicating the start and stop times of each surgical task. Tool presence labels were derived from robot system data and remained noisy, as in prior years. Task labels segmented each video into one of eight mutually exclusive surgical tasks, as defined in Table~\ref{tab:task_definitions_2024}. These task labels were used exclusively for Category~2. Any unannotated segments of the videos were assigned the ``Other'' label.

\begin{table}[h!]
\centering
\small
\caption{Definitions of the eight surgical task labels used in Category~2 (2024).}
\begin{tabular}{p{4.5cm}p{9.5cm}}
\toprule
\textbf{Task Label} & \textbf{Description} \\ \midrule
Suturing & Suturing with the robotic system, encompassing basic suturing skills and techniques as well as their application to clinical scenarios. \\
Uterine Horn & Retraction and dissection along the uterine horn connective tissue of the porcine model, practising proper tension application paired with monopolar curved scissors dissection using both cut and coagulation energy modes. \\
Suspensory Ligaments & Blunt dissection of soft tissue at varying abdominal depths, focusing on skill development and proper camera movement and zooming during dissection. \\
Rectal Artery/Vein & Careful isolation of vessels from larger organ structures, demonstrating proper use of advanced tools and imaging functions. \\
Skills Application & Large-scale steps focused on combining previously taught skills in a clinically applicable scenario; these actions are relatively unstructured and variable. \\
Range of Motion & Navigation of the endoscope and tools around the body cavity to avoid collisions, demonstrating the full range of the workspace with proper port placement. \\
Retraction and Collision Avoidance & Navigation of tools through a stationary field of view while avoiding tool collisions; quadrants are established and cycles of movement through these quadrants are performed, with or without bladder retraction. \\
Other & Activity outside of the structured training curriculum. \\
\bottomrule
\end{tabular}
\label{tab:task_definitions_2024}
\end{table}

\subsubsection{Testing Data}

The testing dataset consisted of surgical training videos captured under the same conditions as the training set. Videos were sub-sampled to 1 FPS for inference purposes. The test set was annotated with bounding box labels for Category~1 and frame-level surgical task labels for Category~2. Bounding box annotations were generated by experienced annotators following the same protocol described in Section~\ref{Training_Data}. As in prior years, the UI region was blurred to prevent participants from exploiting embedded text information.

\subsubsection{Submission Process}

The submission process for the 2024 challenge followed the same Type~2 (T2) challenge format on the Grand Challenge platform as in 2022 and 2023; see Section~3.1 for details. Each algorithm was configured to receive a preliminary test dataset for debugging or the final hidden test dataset. For Category~1, the output format required bounding box predictions in JSON following the COCO detection format. For Category~2, a per-frame surgical step prediction in JSON was required. All teams were asked to submit a final report alongside their algorithm submission. Only teams that submitted both results and a report were considered complete submissions.

\subsection{Team Submissions}\label{team_submissions_24}

For this year's challenge, a total of 90 teams registered and showed interest in participating. In the preliminary testing phase, 94 submissions were received for Category~1 and 160 submissions for Category~2. In the final testing phase, 32 submissions were received for Category~1 and 24 submissions for Category~2. After filtering for teams that submitted both final results and a report, 6 complete submissions remained for Category~1 and 6 for Category~2. Two teams (PDMYR and InspireLab) participated in both categories. Table~\ref{table:TeamAffils2024} shows the participating teams with complete submissions, while Table~\ref{table:TeamMethodsSummary2024} summarizes the methodologies employed by each team. Team methodological details follow below. Note that these sub-sections were written by the participating teams.

\begin{table}[h!]
\centering
\small
\caption{Team affiliations and challenge categories -- 2024}
\begin{tabular}{ccp{4.5cm}ccc}
\toprule
\textbf{Team \#} & \textbf{Team name} & \textbf{Institution} & \textbf{Country} & \textbf{Category} & \textbf{Report} \\ \midrule
1 & PDMYR & Southern Medical University & China & 1 \& 2 & Y \\
2 & InspireLab & InspireLab Technology & Vietnam & 1 \& 2 & Y \\
3 & MULTIS & Chongqing U. of Posts \& Telecom. & China & 1 only & Y \\
4 & SJTUB & Shanghai Jiao Tong U. & China & 1 only & Y \\
5 & SKJP & Muroran Inst. of Tech. & Japan & 1 only & Y \\
6 & SEU-MIA & Southeast U. & China & 1 only & Y \\
7 & SurgOp & IRT Saint Exup\'ery & France & 2 only & Y \\
8 & TeamBCU & Birmingham City U. & UK & 2 only & Y \\
9 & SmartLab\_HKUST & Hong Kong U. of Sci. \& Tech. & Hong Kong & 2 only & Y \\
10 & MIDAS & Samsung Medical Center & South Korea & 2 only & Y \\
\bottomrule
\end{tabular}
\label{table:TeamAffils2024}
\end{table}

\begin{sidewaystable*}
\caption{Summary of methodologies -- 2024}
\begin{adjustbox}{scale=0.6,center}
{\begin{tabular}{ p{8em} p{12em} p{8em} p{6em} p{6em} p{6em} p{8em} p{8em} p{8em} }
Team Name & Architecture & Backbone & Data Preprocessing & Pretrain & Image Augmentation & Use Additional Data & Loss & Output \\
\hline
PDMYR (C1) & YOLOv5 ensemble (WBF) & CSPDarknet & HITL active learning, OCR-based sampling & ImageNet, CLIP & Mosaic, Mixup, MotionBlur, OpticalDistortion & 220k manually labeled frames & Standard YOLO loss & Tool bounding box \\
PDMYR (C2) & Video Swin Transformer & Swin-B & Residual-based frame removal, temporal resampling & Kinetics-400 & 3D-CutMix, ShiftScaleRotate, CoarseDropout & N/A & Cross-entropy & Surgical step \\
InspireLab (C1) & YOLOv10 & CSPDarknet & Semi-supervised labeling, copy-paste augmentation & COCO & Copy-paste & 19k manually labeled frames & Standard YOLO loss & Tool bounding box \\
InspireLab (C2) & SlowFast & ResNet-101 & Outlier removal, 1 fps downsampling & Kinetics-400 & N/A & N/A & Cross-entropy & Surgical step \\
MULTIS & YOLOv8-seg & CSPDarknet & BoT-SORT tracking, iterative refinement & EndoVis2017 & Rotation, clipping & 3.8k labeled frames & Standard YOLO loss & Tool bounding box \\
SJTUB & YOLOv10 (SAM2V10) & CSPDarknet & SAM2-assisted annotation, UI blurring & N/A & Gaussian blur & 300k+ SAM2-labeled frames & Standard YOLO loss & Tool bounding box \\
SKJP & YOLOv10 + ConvNeXt & CSPDarknet, ConvNeXt-v1-small & Synthesized images from EndoVis data & EndoVis2017, EndoVis2018 & Random resize, translate & 55k synthesized images & Standard YOLO loss & Tool bounding box \\
SEU-MIA & CLIP + DualCoOp & CLIP ViT & 0.5 fps extraction & CLIP & Spatial augmentation & N/A & Generalized cross-entropy & Tool class, CAM \\
SurgOp & MViT & MViT & Central crop, UI removal, resize & Kinetics-400 & RandAugment, RandomResizedCrop & N/A & Cross-entropy & Surgical step \\
TeamBCU & Ensemble (ConvNeXt, RegNetY, ViT) & ConvNeXtV2, RegNetY, ViT & ViT-based clustering for sampling & ImageNet & N/A & N/A & Cross-entropy & Surgical step \\
SmartLab\_HKUST & Surgformer & TimeSFormer & 1 fps, resize & Kinetics-400 & RandAugment, random erasing & N/A & Cross-entropy & Surgical step \\
MIDAS & MS-TCN + KAN & Feature extractor & N/A & N/A & N/A & N/A & Focal loss & Surgical step \\
\hline
\end{tabular}}
\end{adjustbox}
\label{table:TeamMethodsSummary2024}
\end{sidewaystable*}

\clearpage


\subsection{PDMYR}

Our team participated in both tracks of this year's competition. For Category 1, we adopted a Human-in-the-Loop (HITL) strategy to minimize manual effort and enhance model performance. We initially actively selected a small batch of representative video frames for manual labeling and trained a first-stage model on this data. We then generated predictions on another batch of unlabeled data, reviewed the results, and manually labeled a minimal portion where the model performed poorly. This iterative process continued until we were satisfied with the model's predictive performance. For Category 2, we focused on frame-level surgical task recognition in videos. With just one submission in the final phase, we significantly outperformed other teams, securing first place. Post-competition analysis revealed that our success stemmed from a robust and diverse regularization system.

\subsubsection{Method Description}

\textbf{Category 1: Data-Driven Surgical Instrument Detection Pipeline}

The competition provided only information about the installation of surgical tools on the da Vinci robot, without object detection box labels. While weakly supervised object detection methods can predict object locations using image-level labels, they are not yet practical for real-world applications due to significant precision gaps compared to fully supervised methods. Therefore, we needed to manually label the objects using the tool installation information as guidance.

Lacking a professional data labeling team, we aimed to achieve high model performance with minimal labeling effort. We designed an active learning-based sample mining and interactive labeling strategy within the HITL framework, conducting multiple rounds of cyclic labeling on the dataset.

\begin{figure}[tbh]
\begin{center}
\resizebox{3.5in}{!}{\includegraphics{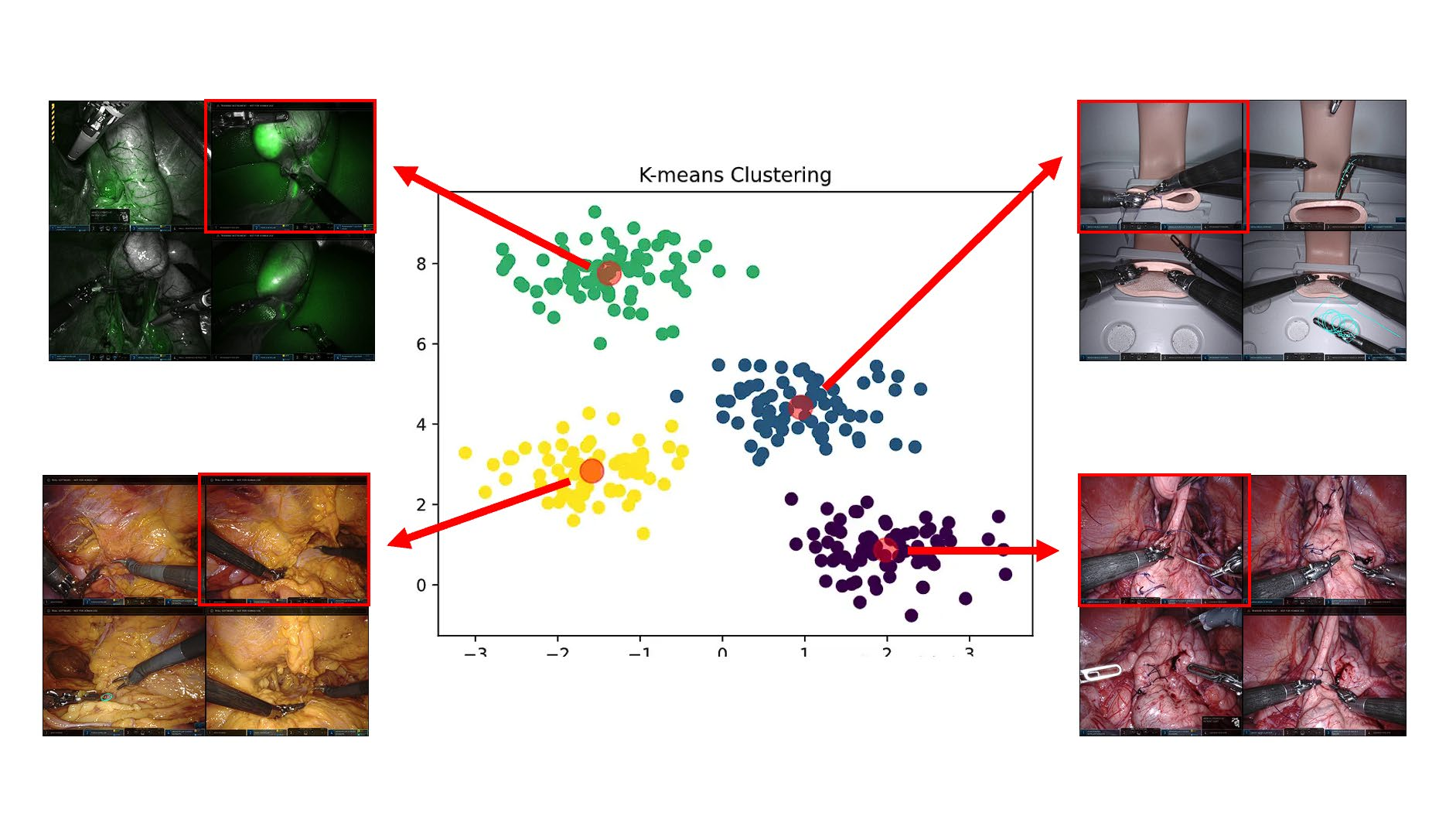}}
\caption{Feature-based clustering for initial sample selection in Category 1.}
\label{fig:C1_C2_PDMYR_c1_clustering}
\end{center}
\end{figure}

\textit{Feature-Based First-Round Sample Selection:} To obtain initial training data, we sampled the entire dataset at 1 frame per second (fps) and extracted features from all images using pre-trained backbone networks like ImageNet classifiers or CLIP. We performed K-Means clustering into 1,000 clusters and selected the images closest to each cluster center as representative samples for labeling (Figure \ref{fig:C1_C2_PDMYR_c1_clustering}).

\begin{figure}[tbh]
\begin{center}
\resizebox{3.5in}{!}{\includegraphics{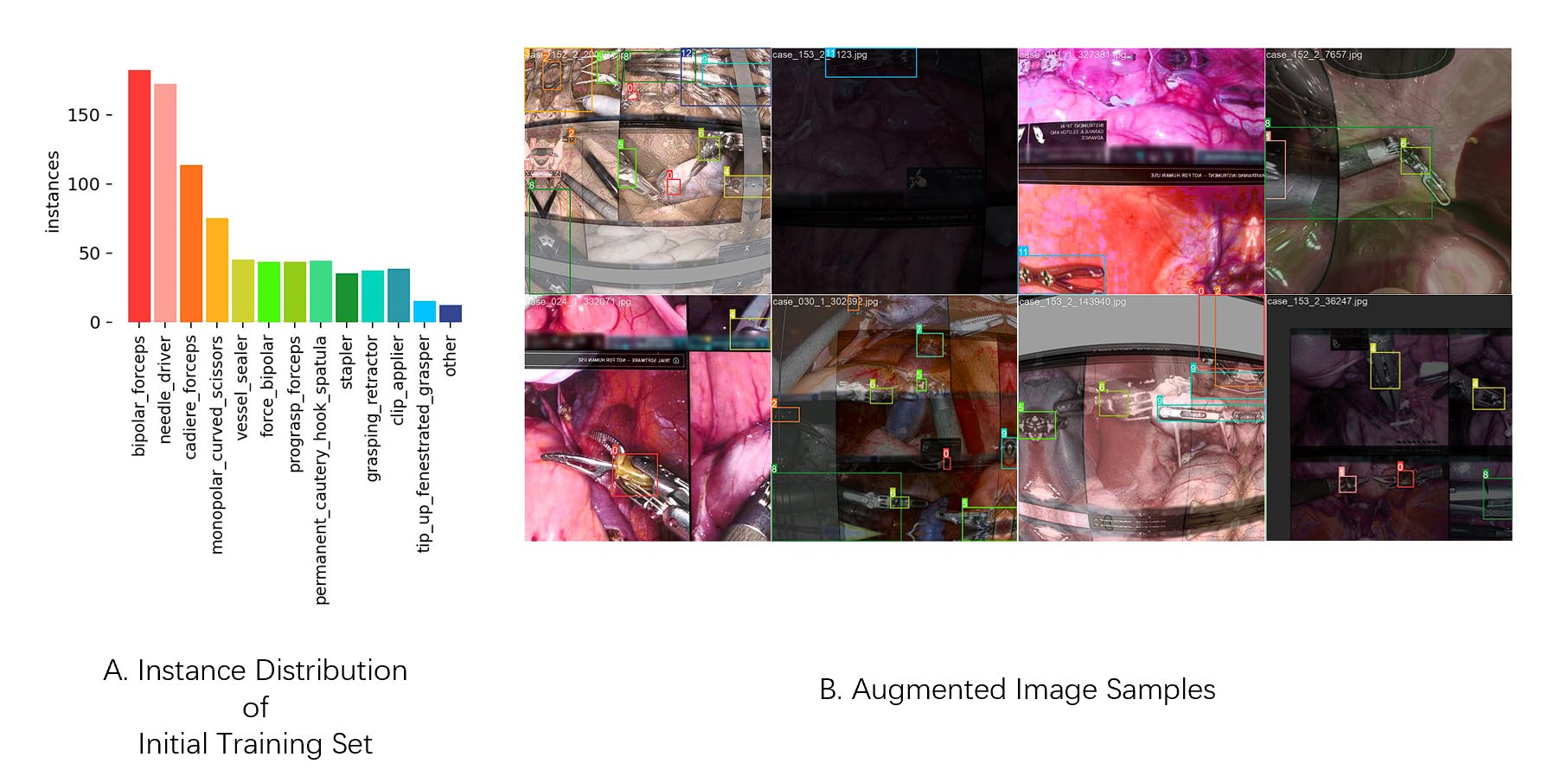}}
\caption{Data distribution analysis showing instrument frequency imbalance (A) and data augmentation examples (B) for Category 1.}
\label{fig:C1_C2_PDMYR_c1_distribution}
\end{center}
\end{figure}

\textit{Active Mining of Difficult Samples:} After labeling these 1,000 images, we analyzed the data distribution and identified two major issues: (1) a significant imbalance between high-frequency and low-frequency instruments (Figure \ref{fig:C1_C2_PDMYR_c1_distribution}), and (2) numerous challenging scenarios where instruments were difficult to distinguish due to factors like fog, bleeding, dirty lenses, and exposure errors.

\begin{figure}[tbh]
\begin{center}
\resizebox{3.5in}{!}{\includegraphics{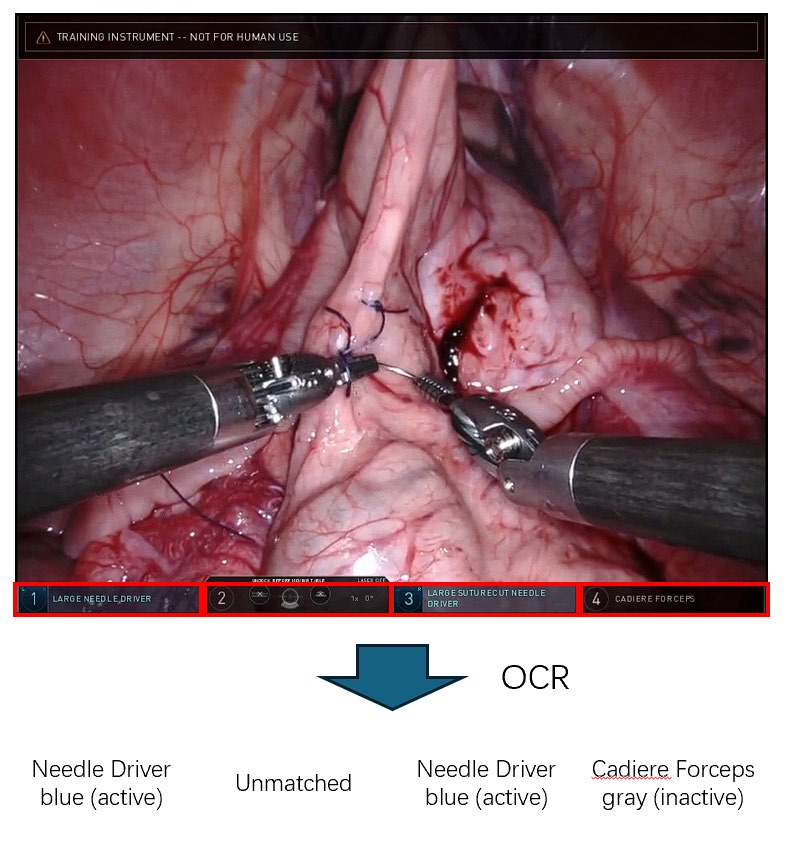}}
\caption{OCR-based extraction of instrument information from the da Vinci GUI for Category 1.}
\label{fig:C1_C2_PDMYR_c1_ocr}
\end{center}
\end{figure}

To address the low-frequency instrument issue, we focused on video segments where these instruments appeared. Meanwhile, high-frequency instruments also appeared in these segments, this approach increased the diversity of our labeled data. We attempted to use the official labels indicating instrument presence but found them unreliable due to duplicates and errors. Instead, we extracted instrument information from the graphical user interface (GUI) of the da Vinci system using optical character recognition (OCR) with the EasyOCR package, accurately identifying frames with low-frequency instruments (Figure \ref{fig:C1_C2_PDMYR_c1_ocr}).

\begin{figure}[tbh]
\begin{center}
\resizebox{3.5in}{!}{\includegraphics{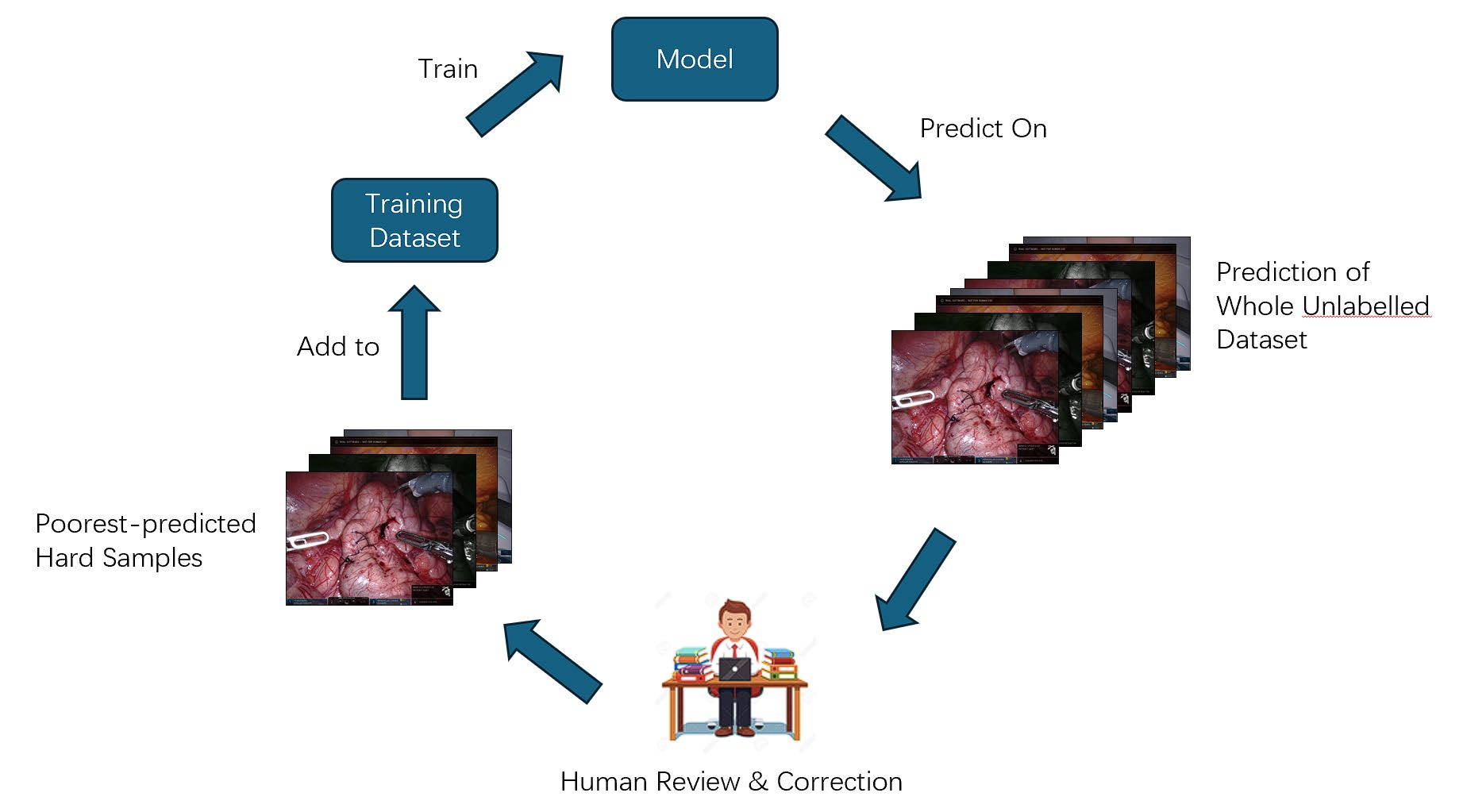}}
\caption{Examples of difficult scenes selected for labeling to enhance model robustness in Category 1.}
\label{fig:C1_C2_PDMYR_c1_difficult}
\end{center}
\end{figure}

For challenging scenes, we implemented two strategies: (1) Within the HITL process, we predicted the entire unlabeled dataset and selected sequences with the poorest predictions for labeling. These sequences often contained defects, and labeling them enhanced the model's robustness (Figure \ref{fig:C1_C2_PDMYR_c1_difficult}). (2) We designed data augmentation techniques reflecting the defects in the data, such as MotionBlur, OpticalDistortion, GridDistortion, along with built-in augmentations like Mosaic and Mixup in YOLOv5.

\textit{Tracking-Assisted Video Annotation:} Previous competition winners utilized target tracking to assist in instrument identification across consecutive frames. However, existing pre-trained single-object tracking (SOT) and multi-object tracking (MOT) models were ineffective at 1 fps due to significant variability between frames. Effective tracking requires high frame rates and manual corrections to ensure accuracy.

\begin{figure}[tbh]
\begin{center}
\resizebox{3.5in}{!}{\includegraphics{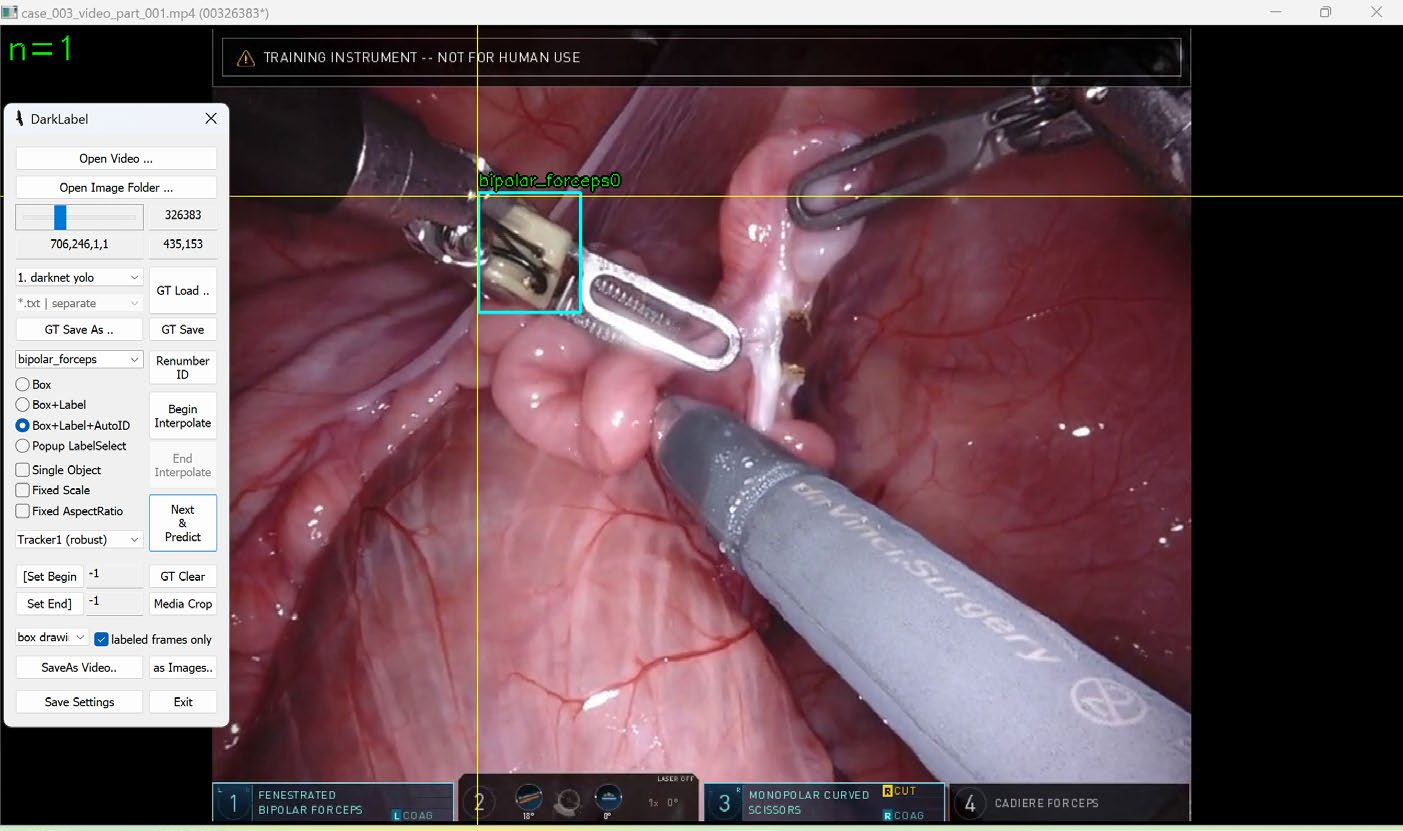}}
\caption{DarkLabel tool interface for tracking-assisted video annotation in Category 1.}
\label{fig:C1_C2_PDMYR_c1_darklabel}
\end{center}
\end{figure}

We used the DarkLabel tool, which incorporates a traditional tracking algorithm similar to Kernelized Correlation Filters (KCF). Although less advanced than deep learning-based trackers, DarkLabel offers an excellent user interface that facilitates immediate manual correction of tracking errors (Figure \ref{fig:C1_C2_PDMYR_c1_darklabel}). This approach enabled us to annotate 220,000 frames in just three days, significantly reducing annotation costs.

\textbf{Category 2: Robust and Diverse Regularization for Video Classification}

\begin{figure}[tbh]
\begin{center}
\resizebox{3.5in}{!}{\includegraphics{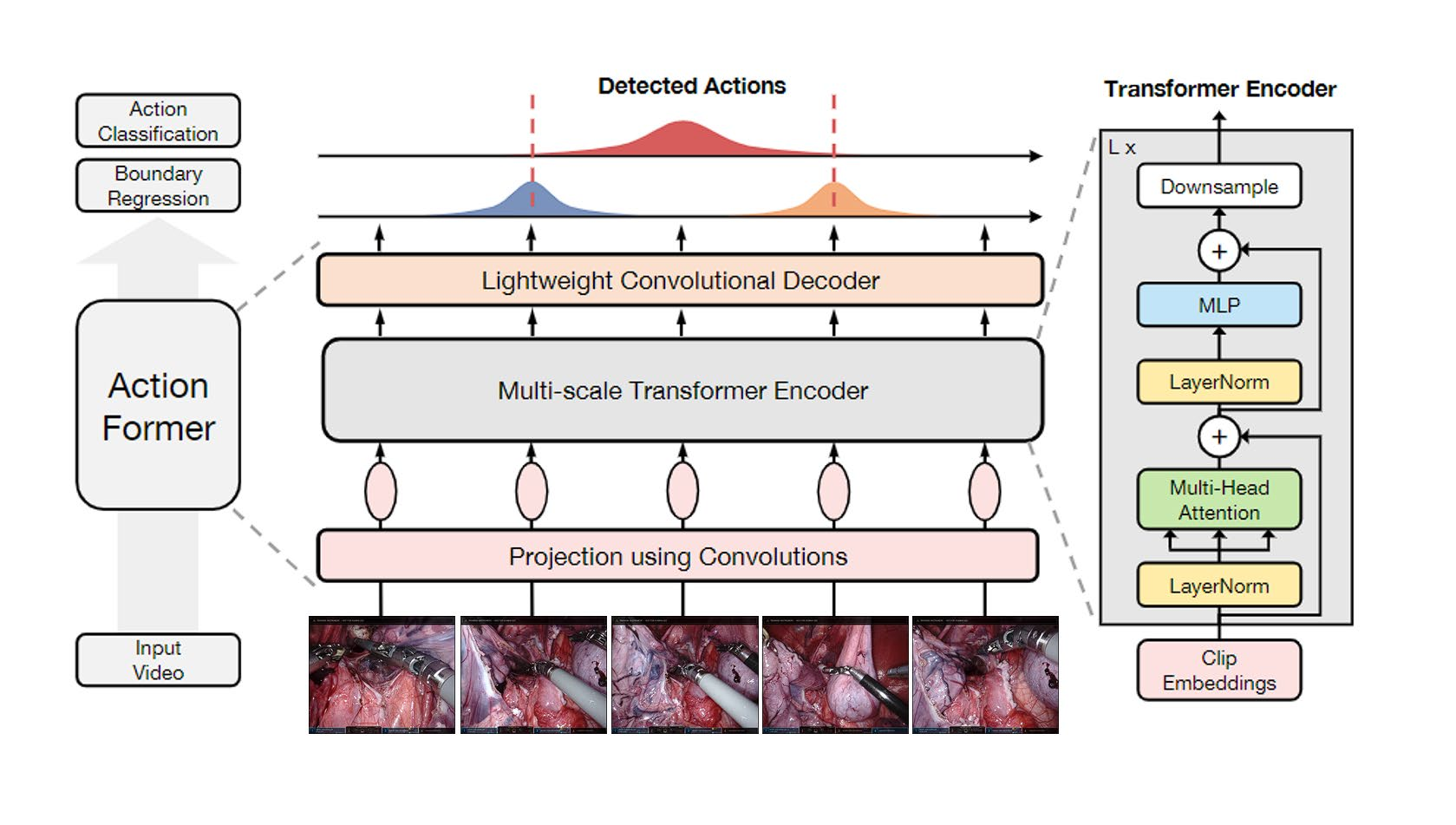}}
\caption{Task simplification from temporal action localization to segment-level classification for Category 2.}
\label{fig:C1_C2_PDMYR_c2_task}
\end{center}
\end{figure}

While the task aligns with spatio-temporal action detection (TAL)—requiring per-frame action predictions with annotations for specific "positive" actions and treating others as background (Figure \ref{fig:C1_C2_PDMYR_c2_task})—we simplified it to segment-level video classification due to: lack of suitable pre-trained feature extractors, insufficient annotations, and variable event lengths.

\begin{figure}[tbh]
\begin{center}
\resizebox{3.5in}{!}{\includegraphics{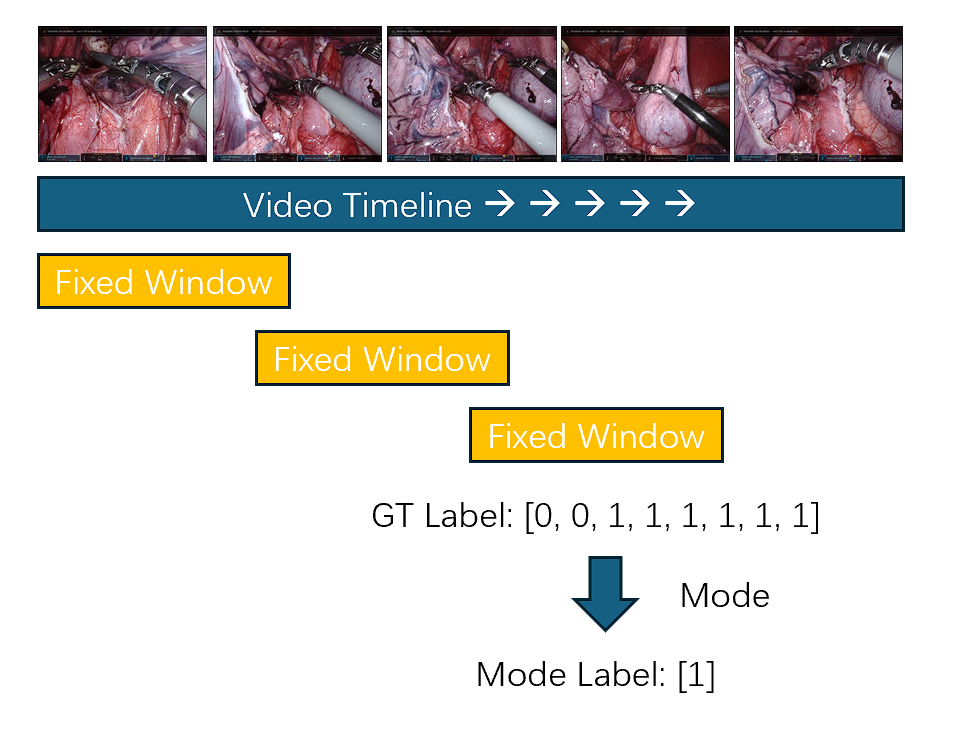}}
\caption{Fixed-length window approach with mode label assignment for Category 2.}
\label{fig:C1_C2_PDMYR_c2_windows}
\end{center}
\end{figure}

We used fixed-length video windows as input for a segment-level classification task (seven classes plus one "other"). For windows containing boundary cases with different labels, we assigned the mode label to the entire window (Figure \ref{fig:C1_C2_PDMYR_c2_windows}).

\begin{figure}[tbh]
\begin{center}
\resizebox{3.5in}{!}{\includegraphics{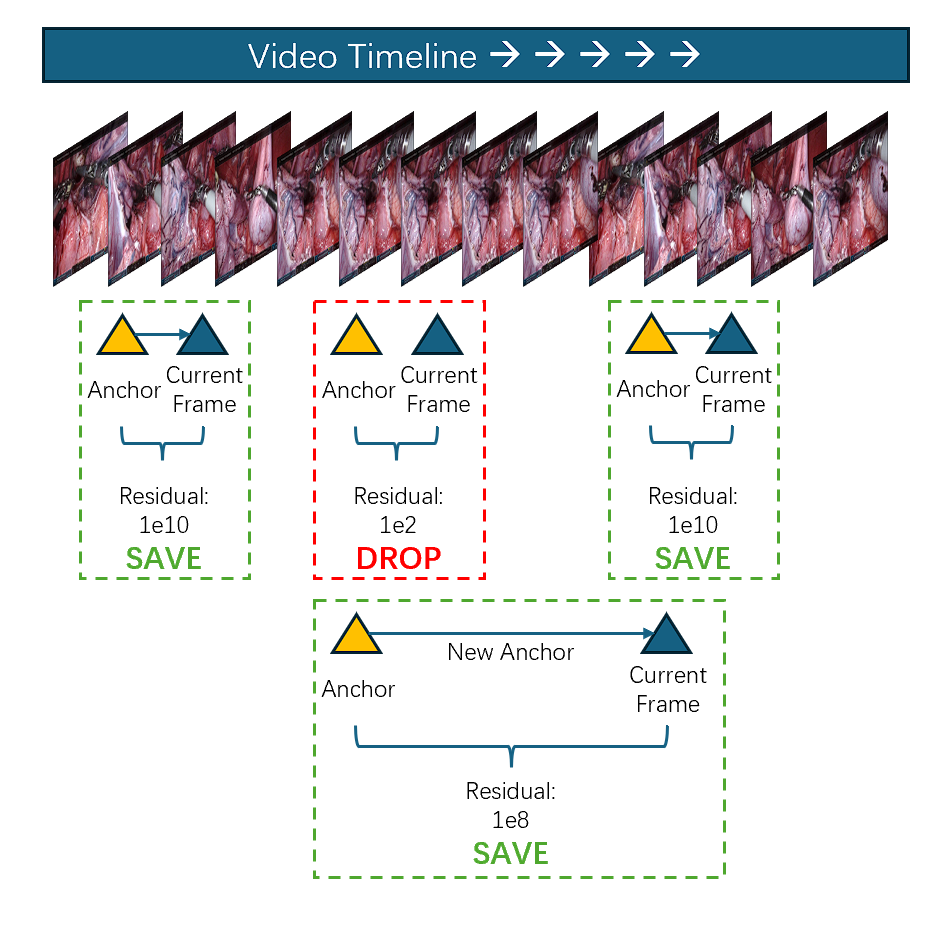}}
\caption{Residual-based invalid frame removal strategy for Category 2.}
\label{fig:C1_C2_PDMYR_c2_residual}
\end{center}
\end{figure}

\textit{Residual-Based Invalid Frame Removal:} To eliminate noise from static and black frames—often due to pauses or the endoscope leaving the surgical field—we designed a residual-based frame removal strategy. Using OpenCV, we set the first frame as an anchor. Subsequent frames are compared to the anchor; if the pixel-level residual sum exceeds a threshold T (set to 1e6), the frame is retained and becomes the new anchor (Figure \ref{fig:C1_C2_PDMYR_c2_residual}). This method discards long sequences of uninformative frames, retaining only meaningful data for training.

\begin{figure}[tbh]
\begin{center}
\resizebox{3.5in}{!}{\includegraphics{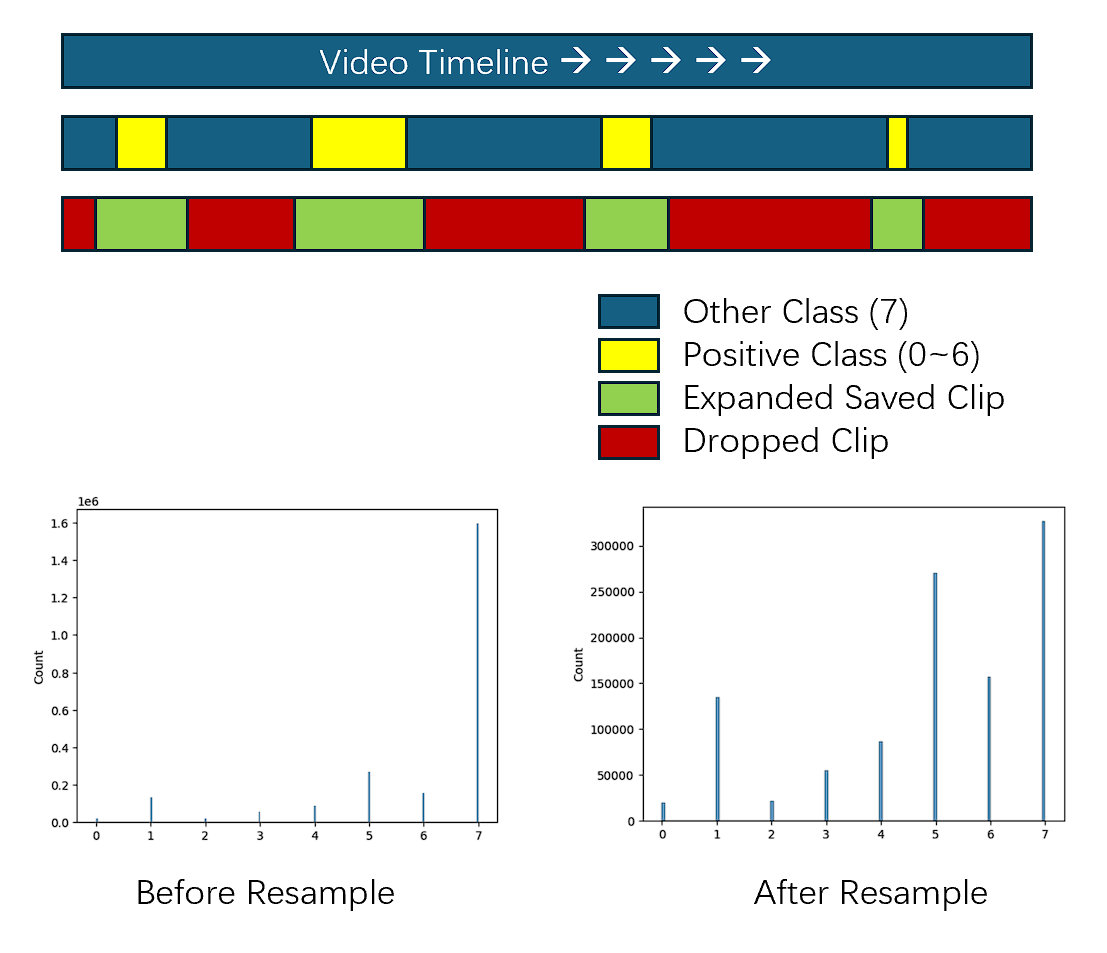}}
\caption{Temporal resampling strategy centered on positive samples for Category 2.}
\label{fig:C1_C2_PDMYR_c2_sampling}
\end{center}
\end{figure}

\textit{Temporal Resampling Centered on Positive Samples:} Noticing that the "other" class dominated the dataset, we focused sampling around positive samples. "Other" class segments near positive ones are harder to distinguish and serve as valuable hard samples. We retained 150 frames around positive segments for random sliding window sampling during training. Discarded samples were included at a lower proportion to ensure a balanced dataset (Figure \ref{fig:C1_C2_PDMYR_c2_sampling}).

\FloatBarrier

\subsubsection{Model Training}

\textbf{Category 1:} In this task, the quantity and quality of annotations are crucial, so the model architecture contributes less to the final performance. We chose YOLOv5 for its speed and ease of iteration. As a single-stage detector, YOLOv5 effectively relies on global information for instrument classification without requiring additional Region of Interest (ROI) processing.

To prevent label leakage, we segregated different cases in the training and validation sets based on the provided case information. We conducted extensive tuning and parameter searches within YOLOv5, including 300 rounds of hyperparameter evolution to identify the optimal training parameters (Table \ref{tab:C1_C2_PDMYR_c1_hyperparams}). We used a batch size of 12 across four NVIDIA RTX 4090 GPUs and maximized the input image resolution.

\begin{table}[htbp]
\centering
\caption{Hyperparameter evolution results for Category 1.}
\label{tab:C1_C2_PDMYR_c1_hyperparams}
\begin{tabular}{ll}
\hline
Config & mAP \\
\hline
hyp.scratch-low & 0.4515 \\
hyp.scratch-high & 0.5022 \\
custom evolution & 0.5548 \\
\hline
\end{tabular}
\end{table}

Our best individual model was YOLOv5x with a resolution of 1280. We combined predictions from models of various sizes and resolutions using Weighted Box Fusion with weights [0.75, 1.0, 1.0, 0.5, 0.5] to produce our final submission (Table \ref{tab:C1_C2_PDMYR_c1_results}).

\begin{table}[htbp]
\centering
\caption{Model ensemble results for Category 1.}
\label{tab:C1_C2_PDMYR_c1_results}
\begin{tabular}{llll}
\hline
Model Size & Input Resolution & Training Image Num. & mAP \\
\hline
YOLOv5-X & 1280 & 220k & 0.5548 \\
YOLOv5-X & 1280 & 170k & 0.5462 \\
YOLOv5-X & 640 & 170k & 0.5470 \\
YOLOv5-L & 1600 & 220k & 0.5453 \\
YOLOv5-M & 1920 & 220k & 0.5336 \\
WBF-ALL & - & - & 0.5661 \\
\hline
\end{tabular}
\end{table}

\textbf{Category 2:} To prevent overfitting and ensure generalization across different cases, we implemented GroupKFold cross-validation and diverse data augmentation strategies.

\textit{GroupKFold Cross-Validation:} We used four-fold GroupKFold cross-validation, isolating different cases in separate folds. Adjustments were based on out-of-fold (OOF) F1 scores, ensuring the model did not overfit intra-case features and could generalize across cases.

\begin{figure}[tbh]
\begin{center}
\resizebox{3.5in}{!}{\includegraphics{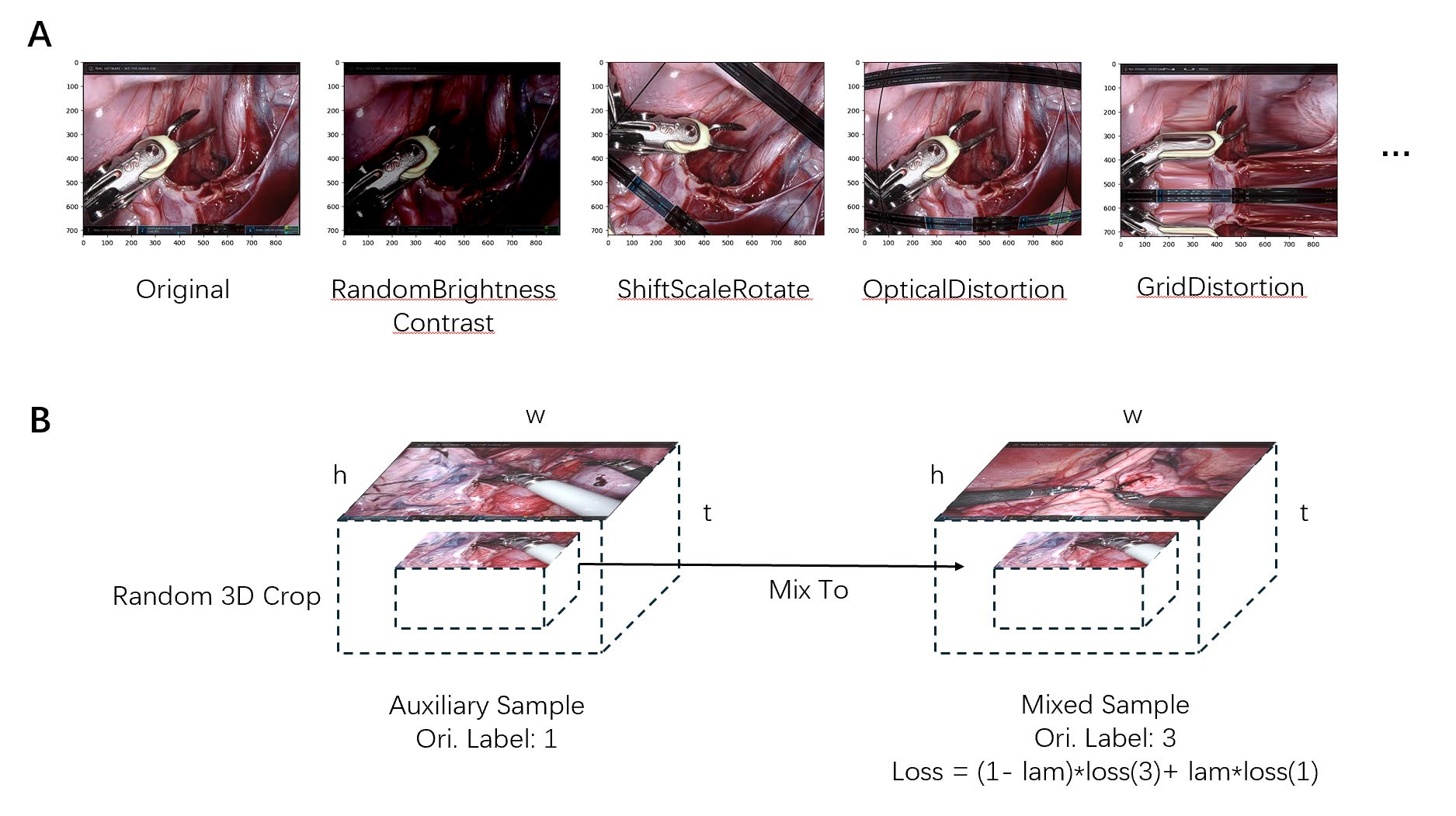}}
\caption{Data augmentation strategies including spatial augmentations (A) and 3D-CutMix (B) for Category 2.}
\label{fig:C1_C2_PDMYR_c2_augmentation}
\end{center}
\end{figure}

\textit{Diverse and Strong Data Augmentation:} Common spatial augmentations were synchronously applied to all frames of a video segment, including: ShiftScaleRotate, RandomCrop, HorizontalFlip, VerticalFlip, RandomBrightnessContrast, OpticalDistortion, GridDistortion, ElasticTransform, CoarseDropout (Figure \ref{fig:C1_C2_PDMYR_c2_augmentation}).

We also introduced 3D-CutMix, extending the traditional 2D CutMix to the temporal dimension. This technique performs random window slicing and mixing in time dimension, enhancing robustness by encouraging the model to learn from shorter durations and varied perspectives.

\textit{Model Selection and Data Preprocessing:} We chose the convolution-based R3D model as a proof of concept and scaled up to a Video Swin Transformer for the final solution. These choices were relatively arbitrary; we simply chose models with simple and easy-to-adapt open-source code and provided KINETICS400 pre-trained weights. Videos were cropped to remove black borders, resized to 320×256, and randomly cropped to 224×224 for input. Constrained by Swin Transformer's architecture, we maintained a 16-frame input to match the pre-trained weights. To capture various temporal dynamics, we trained models at frame intervals of 1, 2, 4, 8, and 16 at 1 fps (input lengths from 16 to 256 seconds), considering all predictions during inference.

\textit{Test-Time Augmentation with Temporal-Probability Fusion:} For better frame-level predictions and to fuse models with different input windows, we designed a test-time augmentation using temporal-probability fusion. During inference, we made predictions every 4 frames, creating overlapping predictions. Each softmax probability prediction was applied to all frames in the window, and overlapping regions were averaged. This strategy produces smooth probability transitions between clips, allowing for accurate segmentation.

\begin{figure}[tbh]
\begin{center}
\resizebox{3.5in}{!}{\includegraphics{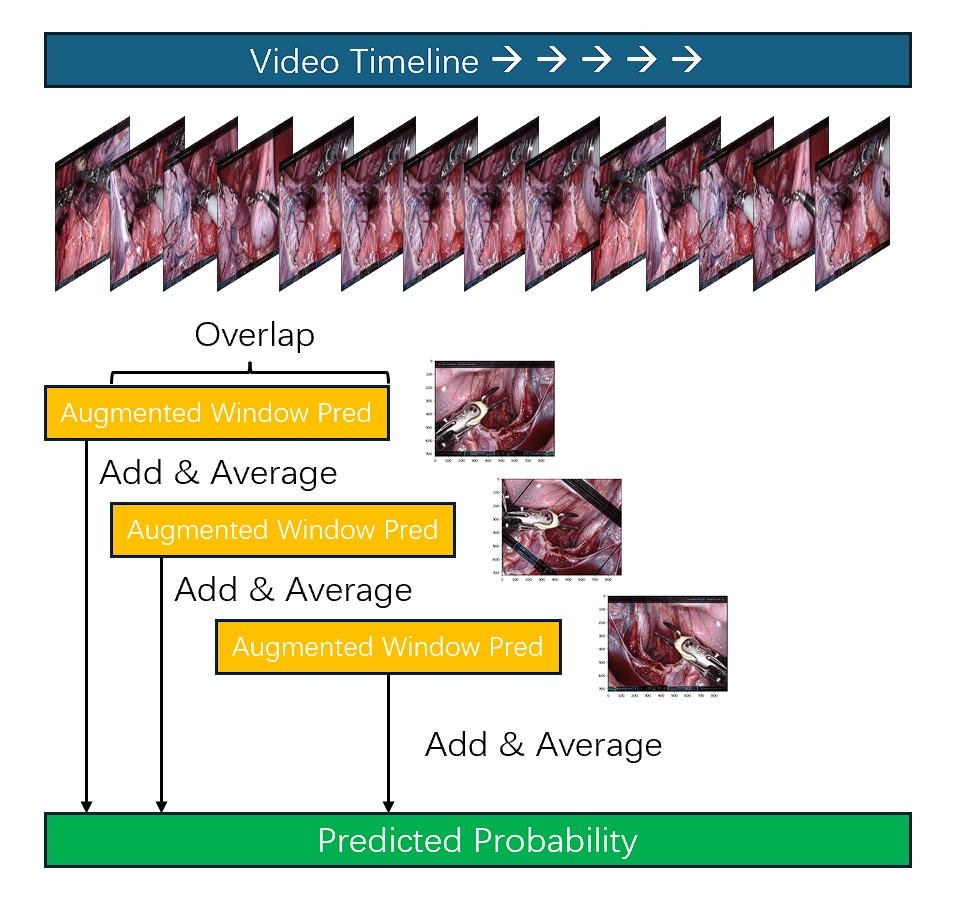}}
\caption{Test-time augmentation strategy for Category 2.}
\label{fig:C1_C2_PDMYR_c2_tta}
\end{center}
\end{figure}

Since the input window (224×224) is slightly smaller than the video size (256×320), we used test-time data augmentation to improve robustness. Destructive augmentations were removed, keeping only mild transformations: ShiftScaleRotate, RandomCrop, HorizontalFlip, VerticalFlip (Figure \ref{fig:C1_C2_PDMYR_c2_tta}).

\begin{figure}[tbh]
\begin{center}
\resizebox{3.5in}{!}{\includegraphics{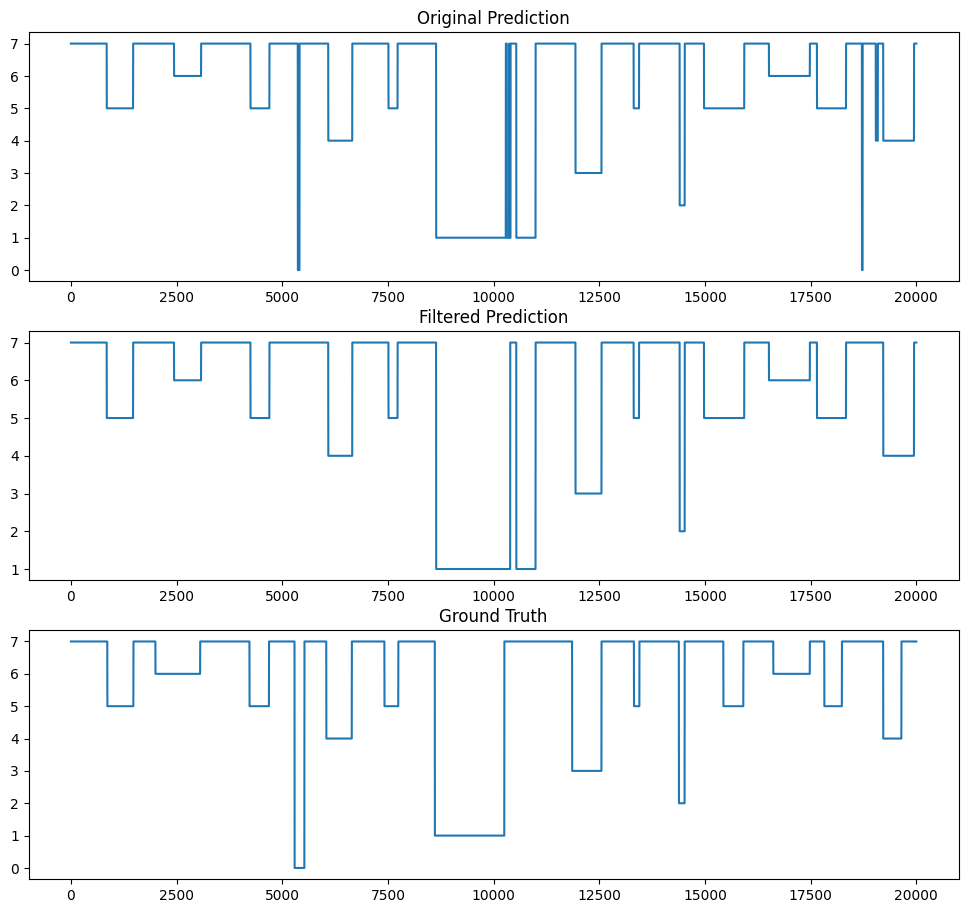}}
\caption{Mode filtering for temporal noise reduction in Category 2.}
\label{fig:C1_C2_PDMYR_c2_mode}
\end{center}
\end{figure}

\textit{Mode Filtering:} To reduce temporal noise from local misclassifications—common with small input windows—we applied a mode filter. A window of 140 seconds slides over the prediction sequence, assigning the mode of the predictions within this window as the new prediction. This method smooths the final output, enhancing the reliability of the frame-level predictions (Figure \ref{fig:C1_C2_PDMYR_c2_mode}).

\FloatBarrier

\subsubsection{Preliminary Performance}

\textbf{Category 1:} Despite employing various active learning and assistive techniques to enhance labeling efficiency, our effectively labeled data represented less than 3\% of the entire 840-hour video dataset. While pseudo-labeling and semi-supervised training are effective for object detection data, time constraints prevented us from adopting these methods before the competition deadline. We anticipate that incorporating these approaches in future competitions will further improve performance.

\textbf{Category 2:} By implementing these strategies, we effectively addressed the challenges posed by the competition and achieved superior performance (Table \ref{tab:C1_C2_PDMYR_c2_ablation}). Our approach demonstrates the importance of task simplification, data preprocessing, balanced sampling, robust validation, and innovative data augmentation in building effective models for complex video analysis tasks. It is hoped that subsequent competitions will further enrich the amount of data for segment labeling, and our powerful feature extractor with a fully trained TAL model will further improve the performance of this task significantly.

\begin{table}[htbp]
\centering
\caption{Ablation study showing progressive improvements for Category 2.}
\label{tab:C1_C2_PDMYR_c2_ablation}
\begin{tabular}{ll}
\hline
Model Design & OOF F1 \\
\hline
S3D Baseline & 0.6829 \\
Swin-B Scale Up & 0.7091 \\
Swin-B +Augmentations & 0.7189 \\
Swin-B +Cutmix 3D & 0.7322 \\
Swin-B +Window Ensemble & 0.7406 \\
Swin-B +Mode Filter & 0.7458 \\
\hline
\end{tabular}
\end{table}

\FloatBarrier
\clearpage
\subsection{InspireLab}

Our team participated in both Category 1 (surgical tool classification and localization) and Category 2 (surgical task recognition) of the SurgVU 2024 challenge. For Category 1, we employed a data-centric approach focusing on meticulous data labeling and augmentation methods to enhance model performance. For Category 2, we leveraged state-of-the-art video understanding models with thorough data preprocessing and post-processing techniques to achieve robust surgical task recognition.

\subsubsection{Category 1: Method Description}

We first explored the challenge dataset, which consisted of over 280 endoscopic surgery videos. The primary goal of this phase was to understand the data structure and identify any inconsistencies or issues that might affect model performance. Videos were divided into two parts: Part 1, significantly longer in duration (around 300 minutes), and Part 2, much shorter (under 100 minutes).

After exploration, we moved on to preprocessing. Key steps included extracting frames from the videos at a rate of 1 frame per second, followed by the creation of ground truth files using tool presence labels. We utilized one-hot encoding for each tool's presence to match the corresponding time intervals. This preprocessing allowed for efficient handling of large video data and facilitated later annotation processes.

\begin{figure}[tbh]
\begin{center}
\resizebox{5in}{!}{\includegraphics{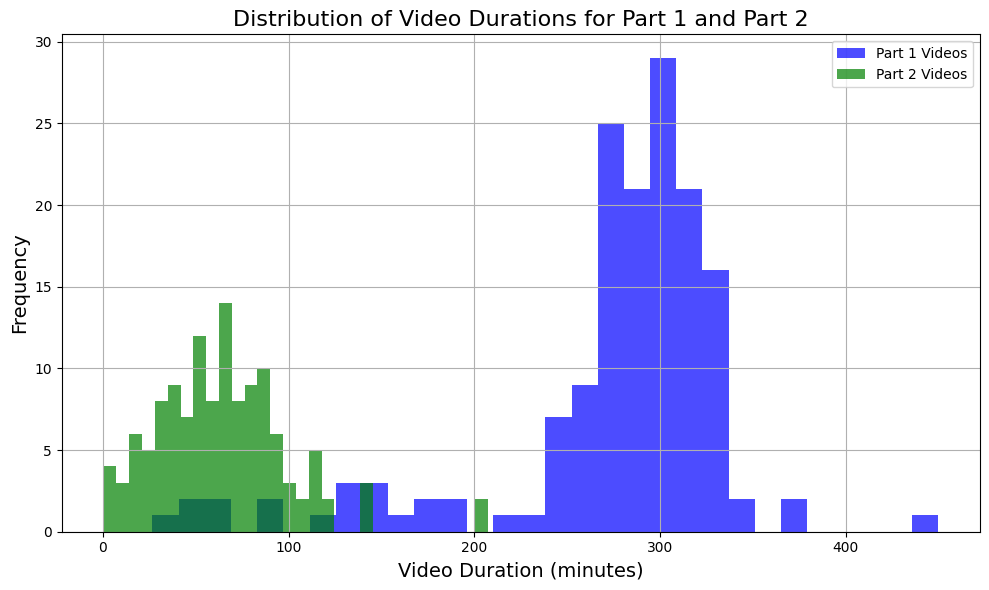}}
\caption{The chart illustrates the significant difference in length between Part 1 and Part 2 videos, with Part 1 typically much longer.}
\label{fig:C1_C2_InspireLab_c1_video_length}
\end{center}
\end{figure}

After preprocessing, manual annotation of frames was undertaken. Approximately 8,000 images were labeled to identify surgical tools. Recognizing the imbalance in labeled instances across different tools, we addressed this using Copy-Paste augmentation. Tools with fewer than 300 labeled instances were manually labeled in additional images, cropped, and pasted onto new frames. This helped increase representation for underrepresented tools, providing a more balanced dataset for model training.

\begin{figure}[tbh]
\begin{center}
\resizebox{5in}{!}{\includegraphics{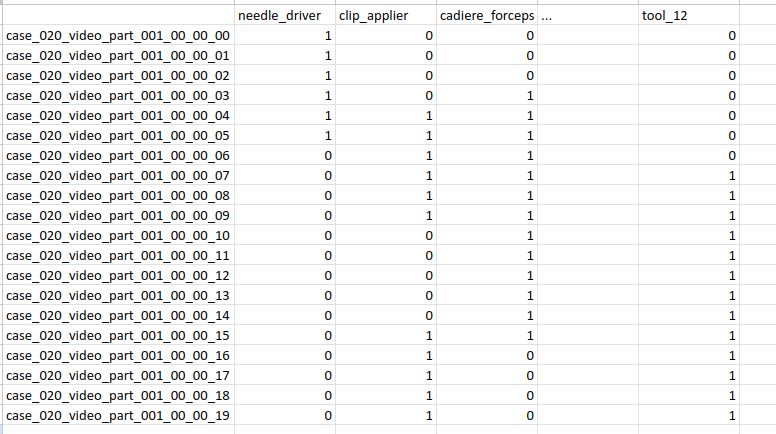}}
\caption{Tool encoding in ground truth files showing how each tool's presence is encoded over time, with binary indicators showing which tools are present in each frame.}
\label{fig:C1_C2_InspireLab_c1_tool_encoding}
\end{center}
\end{figure}

\begin{figure}[tbh]
\begin{center}
\resizebox{5in}{!}{\includegraphics{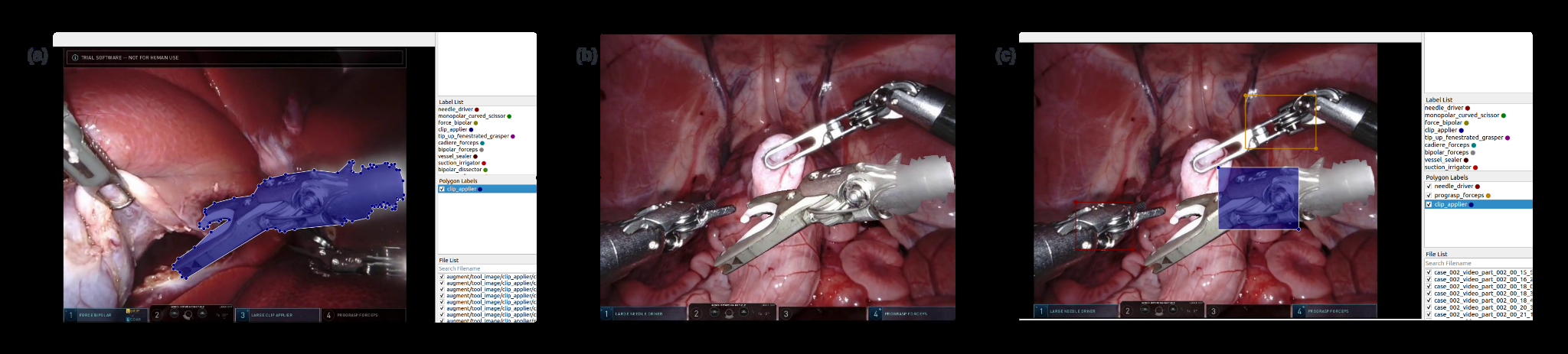}}
\caption{(a) Label polygon for a tool using LabelMe. (b) Copy-Paste Augmentation applied to new images. (c) Labeled with bounding boxes.}
\label{fig:C1_C2_InspireLab_c1_annotation}
\end{center}
\end{figure}

We employed a semi-supervised learning approach using the YOLOv10 model to address dataset size limitations. Predictions were generated on additional frames, and these were manually corrected, expanding the dataset to 19,016 images. This iterative process included testing different versions of YOLOv10 (YOLOv10-X and YOLOv10-M) to balance accuracy and computational efficiency.

The model performance was evaluated using the mAP50-95 metric. If the model did not meet the optimized threshold, additional data annotation and refinement were performed. This iterative loop continued until the mAP50-95 score was satisfactory.

\begin{figure}[tbh]
\begin{center}
\resizebox{5in}{!}{\includegraphics{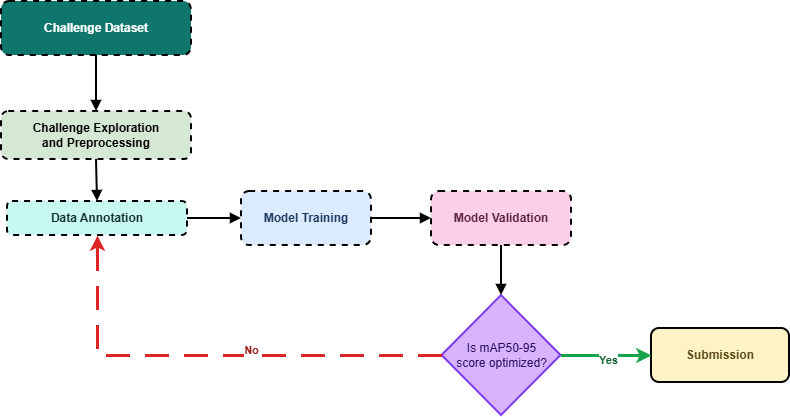}}
\caption{The complete pipeline of the dataset processing, model training, and evaluation phases.}
\label{fig:C1_C2_InspireLab_c1_pipeline}
\end{center}
\end{figure}

\subsubsection{Category 1: Model Training}

We tested different versions of YOLOv10 (YOLOv10-X and YOLOv10-M) to balance accuracy and computational efficiency. The semi-supervised learning approach involved generating predictions on additional frames and manually correcting them, which expanded our dataset to 19,016 images. We applied a non-max suppression filter to eliminate overlapping bounding boxes and further refine predictions.

\subsubsection{Category 1: Preliminary Performance}

Through experimentation, we achieved a mAP50-95 score of 0.4289 in the preliminary phase and 0.4217 in the final phase, thanks to the YOLOv10-M model's balance of performance and efficiency. Our model ranked 2nd in the competition.

\subsubsection{Category 2: Method Description}

We leveraged the unique characteristics of the video dataset and its associated labels to develop a robust analytical approach. Our methodology incorporated several preprocessing techniques and cross-validation methods to identify the most reliable model for our analysis.

In our initial exploration, we evaluated a range of popular models, including SlowFast \cite{InspireLab_Feichtenhofer2019}, UniFormerv2 \cite{InspireLab_Li2022uniformerv2}, MViTv2 \cite{InspireLab_Li2022mvitv2}, and SwinVideo \cite{InspireLab_Liu2022swin}. After thorough assessment, we narrowed our focus to two models that demonstrated superior performance and consistency with the provided dataset: SlowFast and UniFormer.

Upon testing these models on the dataset, we observed that while they exhibited high precision, their recall was comparatively low. Notably, the models struggled to accurately recognize certain tasks, with particular difficulty in identifying the "Other/Unannotated" class. In response to these findings, we implemented multiple post-processing techniques to enhance both the recall and precision of our results.

\begin{figure}[tbh]
\begin{center}
\resizebox{5in}{!}{\includegraphics{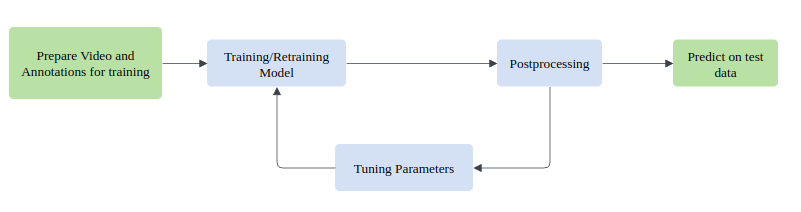}}
\caption{Surgical Task Recognition Workflow.}
\label{fig:C1_C2_InspireLab_c2_workflow}
\end{center}
\end{figure}

While the latest version of the label data showed significant improvements in cleanliness compared to previous iterations, several issues persisted that required addressing. We identified and rectified temporal inconsistencies including negative stop-time values, overlapping tasks, and instances where the stop time preceded the start time. We also noted that the provided video part 1 of case 71 was unavailable.

Our preprocessing strategy involved removal of outliers (with the exception of "Other/Unannotated" samples) to ensure data integrity. While we retained "Other/Unannotated" entries in the dataset, we excluded them from the training phase to prevent potential noise in the model's learning process. After removing outliers and "Other/Unannotated" samples, our final dataset comprised 1,151 videos. Finally, all videos were subsequently downsampled to 1 frame per second (fps) to align with the challenge's testing sample specifications.

\begin{figure}[tbh]
\begin{center}
\resizebox{5in}{!}{\includegraphics{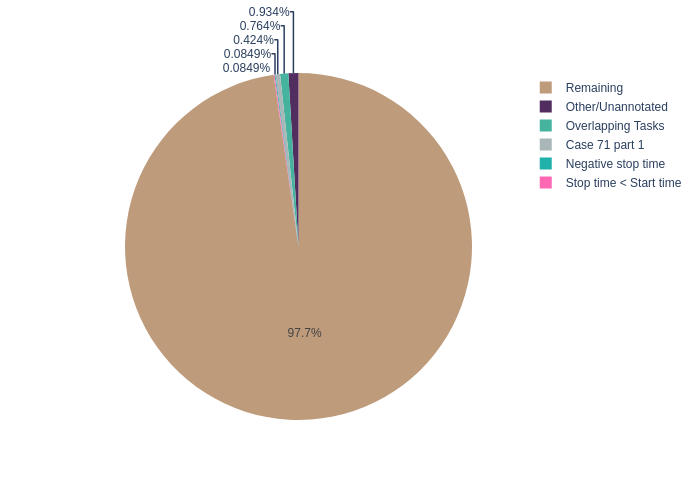}}
\caption{Distribution of Data Outliers and Valid Entries in Video Annotation Dataset.}
\label{fig:C1_C2_InspireLab_c2_outliers}
\end{center}
\end{figure}

The original dataset comprised a total of 155 video cases. We employed a 4-fold cross-validation technique, initially setting aside 27 cases as a fixed test set. The remaining 128 cases were randomly partitioned four times to create four distinct cross-validation sets. It is important to note that our cross-validation split was based on video cases rather than individual video tasks.

\begin{table}[htbp]
\centering
\caption{Data Distribution for Cross-Validation Sets}
\label{tab:C1_C2_InspireLab_c2_cv_split}
\begin{tabular}{ll}
\hline
Training Set & 96 cases \\
Validation Set & 32 cases \\
Test Set & 27 cases (fixed) \\
\hline
\end{tabular}
\end{table}

\subsubsection{Category 2: Model Training}

We trained both models on MMAction2 \cite{InspireLab_MMAction2} using the hyperparameters and settings detailed in Table \ref{tab:C1_C2_InspireLab_c2_hyperparams}. The training process was conducted using our cross-validation dataset to ensure robust performance evaluation. Additionally, we implemented a confidence threshold of 0.9 for the "Other/Unannotated" label assignment during the preliminary stage.

\begin{table}[htbp]
\centering
\caption{Hyperparameters and Training Configurations for SlowFast and UniFormerv2 Models}
\label{tab:C1_C2_InspireLab_c2_hyperparams}
\begin{tabular}{lllllp{3cm}}
\hline
Model & Pretrained & Learning rate & Epoch & Batch size & Scheduler \\
\hline
SlowFast-R101 & Kinetics-400 & 5e-04 & 50 & 4 & MultiStepLR \\
UniFormerv2-B/16 & Clip, Kinetics710, Kinetics-400 & 2e-05 & 50 & 4 & LinearLR, CosineAnnealingLR \\
\hline
\end{tabular}
\end{table}

One of the main challenges in SurgVu2024 was managing predictions for the "other" class and fluctuations in "suturing" scores, which often resulted in redundant predictions and disrupted the sequence flow of surgical tasks. To mitigate this, we fine-tuned the thresholds, reducing false positives and improving the accuracy of "suturing" task boundaries. Additionally, we applied smoothing techniques to ensure smoother transitions, better capturing the continuous nature of surgical procedures.

\begin{figure}[tbh]
\begin{center}
\resizebox{5in}{!}{\includegraphics{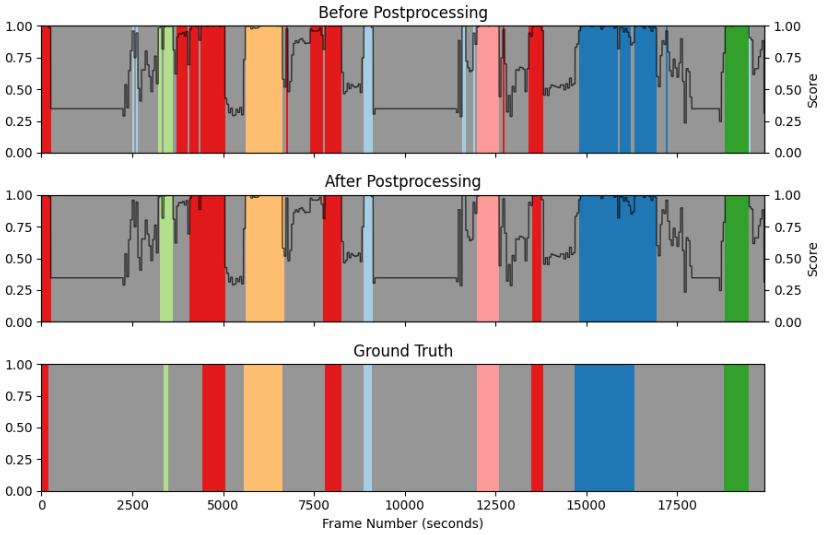}}
\caption{Visualize Post-processing result.}
\label{fig:C1_C2_InspireLab_c2_postprocessing}
\end{center}
\end{figure}

\subsubsection{Category 2: Preliminary Performance}

Based on our preliminary results, the SlowFast model demonstrated superior performance. Consequently, we selected the SlowFast model for further refinement and optimization in preparation for the Final Stage of the challenge.

\begin{table}[htbp]
\centering
\caption{Results at the Preliminary Phase}
\label{tab:C1_C2_InspireLab_c2_preliminary}
\begin{tabular}{llll}
\hline
Model & F1 Score & Recall & Precision \\
\hline
SlowFast-R101 & 0.7873 & 0.7779 & 0.8311 \\
UniFormerv2-B/16 & 0.6636 & 0.6814 & 0.7435 \\
\hline
\end{tabular}
\end{table}

\begin{table}[htbp]
\centering
\caption{Results at Post-processing Phase}
\label{tab:C1_C2_InspireLab_c2_final}
\begin{tabular}{llll}
\hline
SlowFast-R101 & & Preliminary & Final \\
\hline
F1 Score & & 0.7873 & 0.8103 \\
Recall & & 0.7779 & 0.8053 \\
Precision & & 0.8311 & 0.8699 \\
\hline
\end{tabular}
\end{table}

\FloatBarrier
\clearpage

\subsection{MULTIS}

Inspired by the previous methods of NVIDIA (SurgToolLoc 2022)~\cite{MULTIS_Zia2023} and ANL-Surg (SurgToolLoc 2023) and ZJURealDoctor (SurgToolLoc 2023), we believe that to deal with such a detection problem, the use of segmentation model to realize the detection can have a better recognition effect, based on the yolov5-detect model previously used by NVIDIA and the yolov8-detect model used by ANL-Surg. We believe that using the more advanced YOLOv8-segment model~\cite{MULTIS_Varghese2024} to solve the detection problem can achieve better results.

\subsubsection{Method Description}

\begin{figure}[tbh]
\begin{center}
\resizebox{3.5in}{!}{\includegraphics{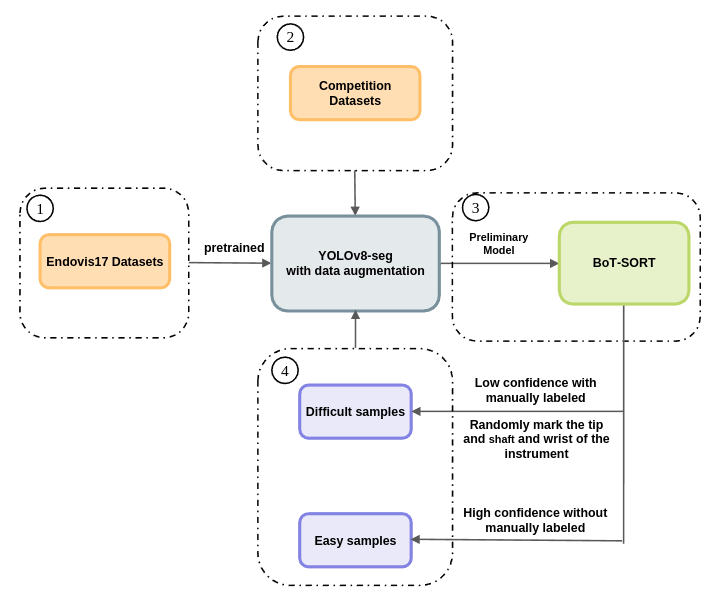}}
\caption{Pipeline overview of the MULTIS approach showing the four-phase training methodology.}
\label{fig:C1_MULTIS_pipeline}
\end{center}
\end{figure}

As illustrated in Figure~\ref{fig:C1_MULTIS_pipeline}, our framework employs a systematic approach composed of four distinct phases to process the data and enhance the accuracy of the model.

\textbf{Phase 1: Pre-training Phase.} Inspired by the methodology established in the ZJURealDoctor (SurgToolLoc 2023) algorithm, our approach begins with the pre-training of a model using the EndoVis17 dataset. This foundational step is critical for initializing the model's parameters and providing it with a broad understanding of surgical instrument appearances within laparoscopic environments.

\textbf{Phase 2: Constructing an Initial Manually Annotated Dataset.} Given that the EndoVis17 dataset~\cite{MULTIS_Allan2019} does not encompass all the surgical instrument categories pertinent to our specific application, we undertook the task of augmenting the dataset with additional categories. Initially, we segmented the dataset based on the provided tool presence labels to isolate frames featuring each instrument category. Subsequently, we manually annotated a total of 1,063 frames, ensuring comprehensive coverage across all instrument types required for the competition. In addition, occlusion is common in surgical scenes, so we use data augmentation methods such as rotation and clipping.

\textbf{Phase 3: Segmentation Model Training and Sample Tracking.} Using the augmented dataset from Phase 2, we proceeded to train an initial segmentation model. Recognizing the importance of both challenging and straightforward samples in improving the model's performance, we employed the BoT-SORT algorithm~\cite{MULTIS_Aharon2022} to track and categorize these samples. High-confidence and low-confidence frames were segregated, with the former representing easier instances and the latter indicating more challenging ones. These segmented frames were then stored separately for each instrument type. The rationale behind this separation is to facilitate the identification of difficult samples, which are then manually reviewed and labeled, while easy samples are directly utilized for training.

\begin{figure}[tbh]
\begin{center}
\resizebox{3.5in}{!}{\includegraphics{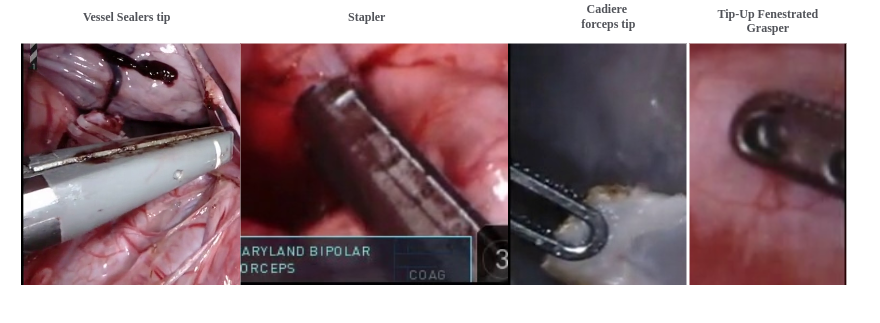}}
\caption{Error-prone results showing common misclassification patterns in surgical instrument tip detection.}
\label{fig:C1_MULTIS_errors}
\end{center}
\end{figure}

It is noteworthy, as depicted in Figure~\ref{fig:C1_MULTIS_errors}, that the model occasionally encounters difficulties in correctly identifying the tips of certain surgical instruments. Specifically, the tips of Prograsp forceps and Cadiere forceps are sometimes misclassified as those of a Tip-Up Fenestrated Grasper, while the tips of Vessel Sealers are incorrectly identified as those of a Stapler. To address these misclassification issues, we implemented a targeted annotation strategy. During the annotation process, we randomly selected a subset of images to explicitly mark each component of the devices, including their tips. This approach was adopted to ensure that the model receives a diverse set of examples, thereby enhancing its capability to accurately recognize the distinctive features of the instrument tips and reducing identification errors.

\textbf{Phase 4: Iterative Model Refinement.} In the final phase, the difficult samples identified and labeled during Phase 3, along with the easy samples, were reintroduced into the manually annotated dataset established at the outset. This expanded dataset, now comprising 3,842 images, served as the basis for retraining the model.

\FloatBarrier

\subsubsection{Model Training}

We initially trained a preliminary YOLOv8-seg model using manually labeled data. We then iteratively refined the model through a process involving BoT-SORT tracking and filtering to distinguish between difficult and easy samples. The final training dataset comprised 3,842 images. We utilized the EndoVis17 dataset for pre-training and applied data augmentation methods such as rotation and clipping to address occlusion scenarios common in surgical scenes.

\subsubsection{Preliminary Performance}

On the Final Testing Phase - Category 1 Leaderboard, our mAP reached 0.3631. This performance is comparable to that of the ANL-Surg (2023) model, which was trained on a significantly larger dataset of 432,918 images. Despite utilizing only 3,842 images, our model achieved competitive results, demonstrating the potential for achieving strong performance with smaller, carefully curated datasets.

However, there are several avenues for improvement. One such area involves refining the confidence thresholds used to delineate difficult from easy samples, which could lead to the identification of more informative key samples. Additionally, increasing the dataset size to tens of thousands of images is anticipated to further elevate the mAP to a higher level.

\FloatBarrier
\subsection{SJTUB}

We participated in Category 1 of the SurgVU challenge. We believe that the key of the challenge lies in generating a trainable high-quality dataset with weak supervision. After carefully observing the officially provided weakly supervised labels, we found many problems where the labels did not correspond to the data, and at the same time, in order to make it easier for subsequent training, we decided to use the weakly supervised approach to generate a dataset by ourselves that could be directly utilized. Then, in July, Segment Anything V2 (SAM2) \cite{SJTUB_Ravi2024} came. We developed a weakly supervised annotation tool for the challenge via SAM2 and generated 300K+ frames from different videos as our dataset, finally training the state of the art method, YOLOv10 \cite{SJTUB_Wang2024}, as our detector. Therefore, we named our method SAM2V10.

\subsubsection{Method Description}

\begin{figure}[tbh]
\begin{center}
\resizebox{3.5in}{!}{\includegraphics{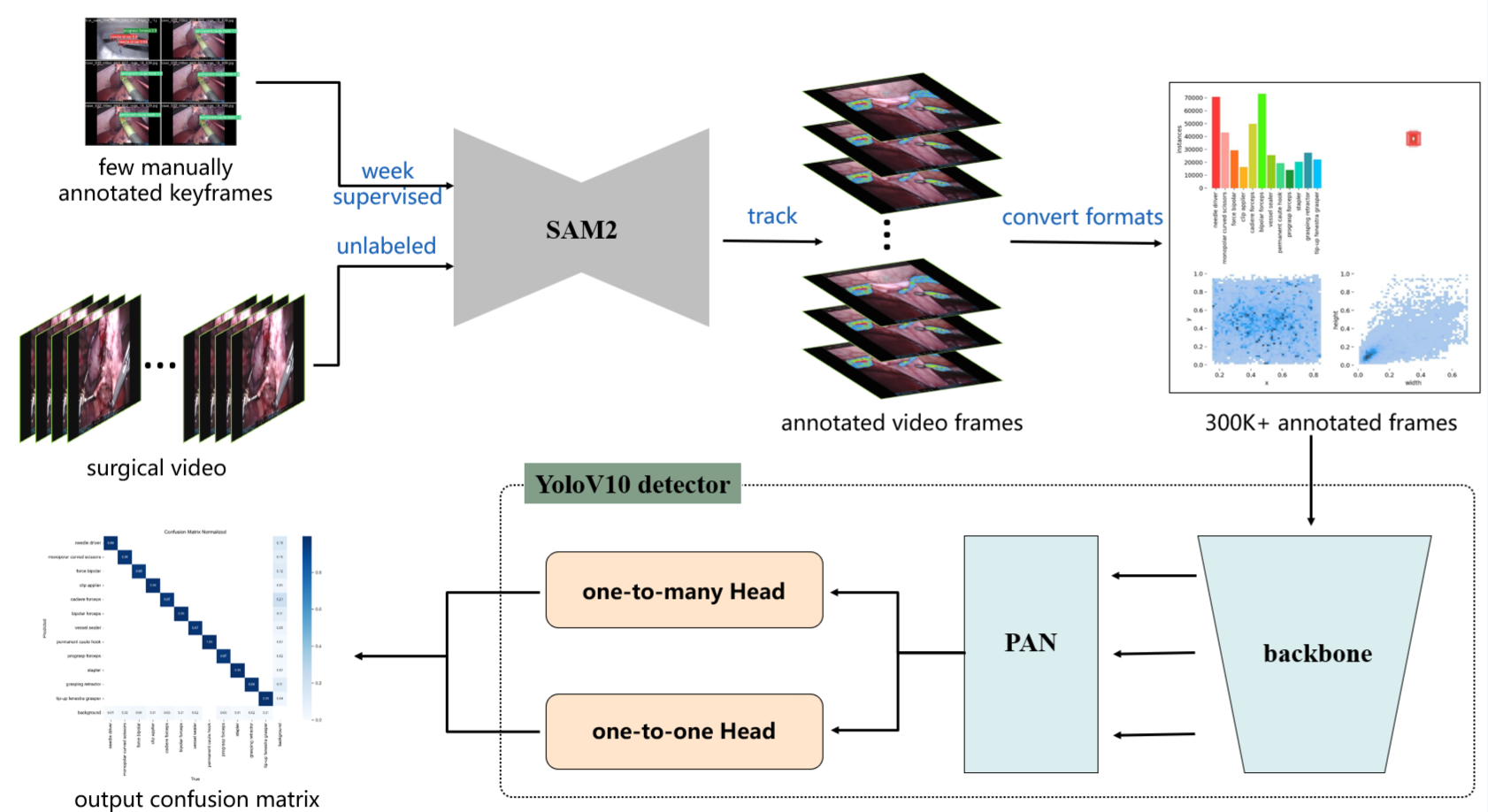}}
\caption{The workflow of SAM2V10.}
\label{fig:C1_SJTUB_workflow}
\end{center}
\end{figure}

Figure \ref{fig:C1_SJTUB_workflow} illustrates the workflow of SAM2V10. We sample several hundred frames from a specific video featuring the same surgical tools, then semi-automatically segment the target location using SAM2 for initial localization. Following this, we allow the system to automatically track the tools, resulting in several hundred automatically labeled images at a time. Notably, SAM2 does not always achieve perfect tracking directly; therefore, we select a number of keyframes from the video for cueing. Our tests indicate that, to ensure dataset quality, we typically maintain a keyframe-to-automatically labeled frame ratio of 1:40. By manually labeling around 20 keyframes, SAM2 can effectively track up to 800 frames. For convenience, these keyframes are generally sampled at equal intervals. Based on this approach, a dataset with over 300K+ frames of the 12 tools from 21 selected videos was generated. The data distribution is shown in Figure \ref{fig:C1_SJTUB_workflow}. We maintained the proportions of various instruments in the dataset to closely reflect their frequency of appearance in the videos.

\begin{figure}[tbh]
\begin{center}
\resizebox{3.5in}{!}{\includegraphics{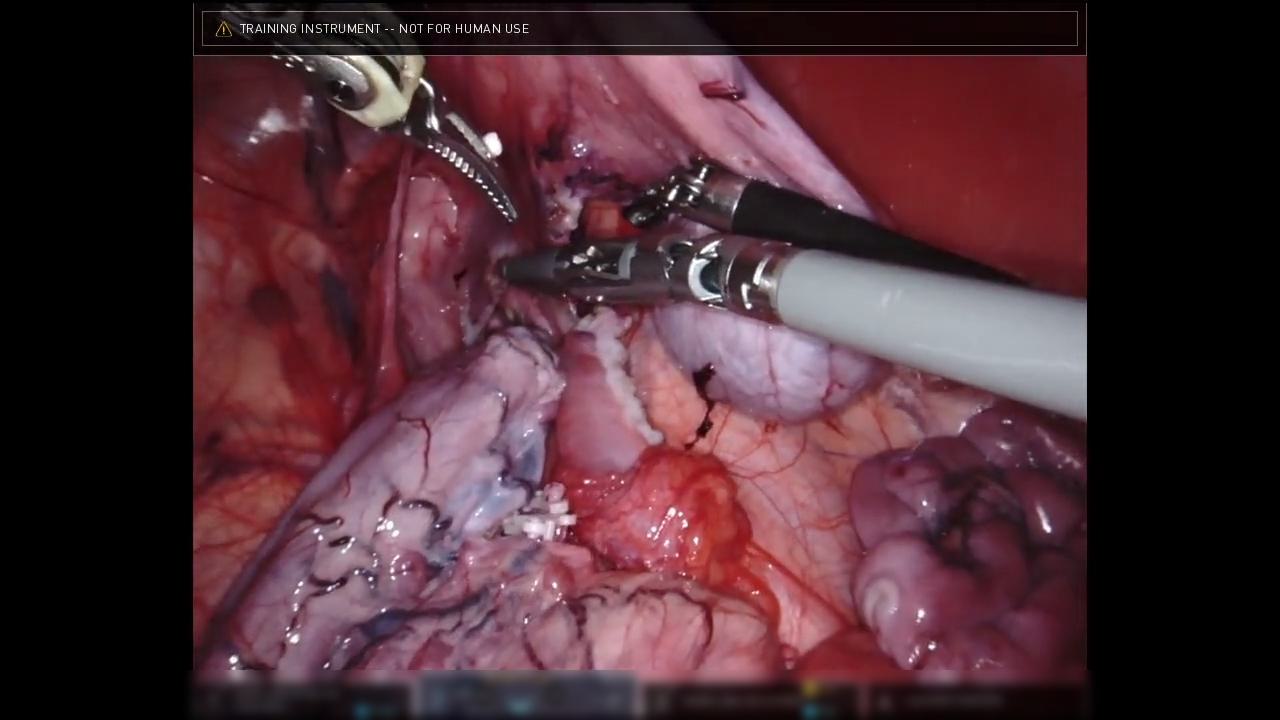}}
\caption{We applied Gaussian blurring to the bottom of the frames.}
\label{fig:C1_SJTUB_blur}
\end{center}
\end{figure}

Simultaneously, we conducted purposeful data augmentation to more realistically simulate test conditions. Specifically, we applied Gaussian blurring to the bottom 50 rows of pixels in the dataset to minimize the impact of the text UI on model training, as illustrated in Figure \ref{fig:C1_SJTUB_blur}.

\subsubsection{Model Training}

\begin{figure}[tbh]
\begin{center}
\resizebox{3.5in}{!}{\includegraphics{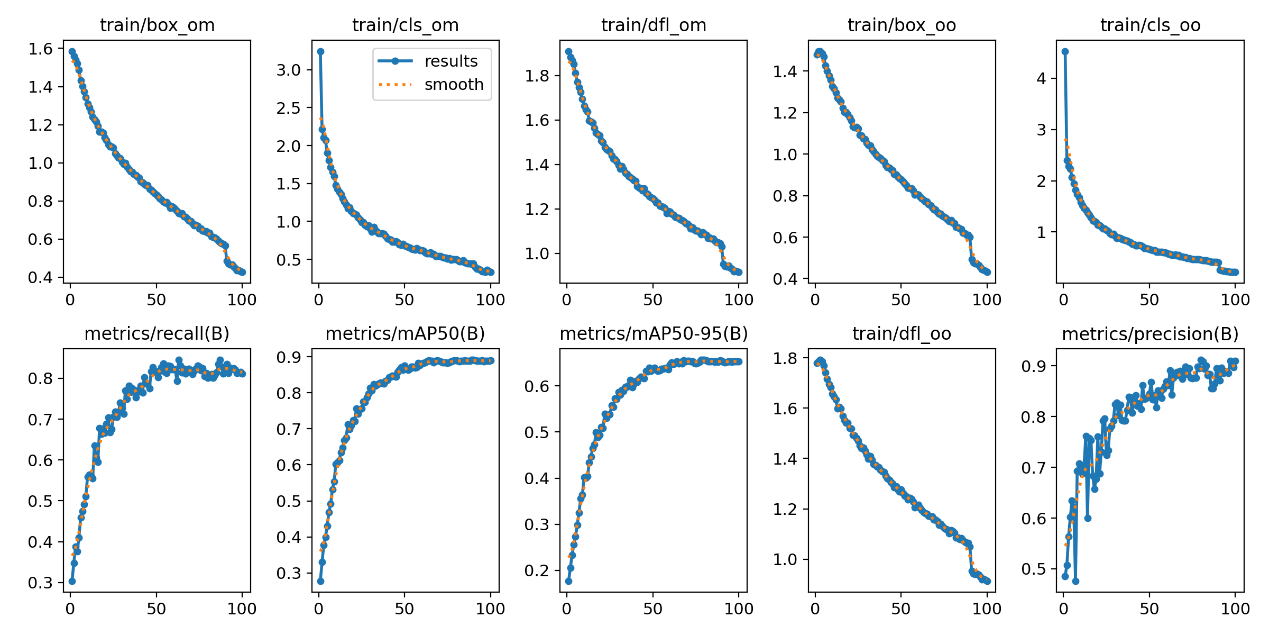}}
\caption{The training process.}
\label{fig:C1_SJTUB_training}
\end{center}
\end{figure}

Finally, we trained a YOLOv10X model using the dataset, with an approximate split of 80\% for training and 20\% for validation. The training process is shown in Figure \ref{fig:C1_SJTUB_training}.

\subsubsection{Preliminary Performance}

After multiple rounds of optimization, the mAP@50-95 of our model reached 0.90. However, the result from the Docker submission was only 0.2754. This discrepancy suggests there may be issues in our dataset or post-processing code, such as overfitting or data bias. We hope this discussion inspires further exploration and improvement.

\FloatBarrier
\subsection{SKJP}
Training object detection models by using weakly annotated dataset is a challenging task. To tackle this problem in surgical tool localization, we propose to utilize synthesized dataset constructed from existing datasets, then train an object detection model by using the synthesized dataset.

\subsubsection{Method Description}
In category 1, the provided dataset consists of videos taken from surgical training exercises using the da Vinci robotic system and tool presence labels for each frame in the videos. The tool presence labels are "weakly" labels for training surgical tool localization (object detection) models. Note that the tool presence labels are noisy because there are cases where the label might indicate 3 tools present, but only 2 or fewer tools can be seen in the video.

Training object detection models by using weak annotated datasets is a challenging task. To tackle this problem in surgical tool localization, we propose to utilize synthesized dataset constructed from existing datasets, then train an object detection model by using the synthesized dataset. We should detect clevis of each robotic surgical tool, so we need datasets which localize parts of surgical tools, i.e., clasper, clevis, and shaft. We use the 2017 Robotic Instrument Segmentation Challenge dataset \cite{SKJP_Allan2019} (hereinafter, we call this dataset as "Endovis2017" dataset) and the 2018 robotic scene segmentation challenge dataset \cite{SKJP_Allan2020} (hereinafter, we call this dataset as "Endovis2018" dataset). In these datasets, segmentation masks for clasper, wrist (clevis) and shaft parts are provided for each surgical tool.

First, we extract rectangle regions including tools from the Endovis2017 and Endovis2018 datasets and save the extracted images and the corresponding masks to files. We exclude cases where the image sizes are smaller than a threshold or two or more tools are included in a rectangle region. The tool images are categorized by both tool classes and tool positions. The tool positions are selected from "left", "right" or "bottom", which are decided by the original positions of the tools. The number of tool images are 1,052 and 1,447 for the Endovis2017 and Endovis2018 datasets, respectively. Note that the tools included in Endovis2017, Endovis2018 and this challenge (we call it "SurgVU2024") datasets are different as shown in Table \ref{tab:C1_SKJP_tools}. Several tools in the SurgVU2024 dataset are not included in the Endovis2017 and Endovis2018 datasets. These tools are force bipolar, cadiere forceps, permanent cautery hook / spatula, stapler and tip-up fenestrated grasper.

\begin{table}[htbp]
\centering
\caption{Tools included in three datasets.}
\label{tab:C1_SKJP_tools}
\begin{tabular}{lccc}
\hline
Tool name & Endovis2017 & Endovis2018 & SurgVU2024 \\
\hline
Need driver & \checkmark & \checkmark & \checkmark \\
Monopolar curved scissors & \checkmark & \checkmark & \checkmark \\
Force bipolar &  &  & \checkmark \\
Clip applier &  & \checkmark & \checkmark \\
Cadiere forceps &  &  & \checkmark \\
Bipolar forceps & \checkmark & \checkmark & \checkmark \\
Vessel sealer & \checkmark &  & \checkmark \\
Permanent cautery hook / spatula &  &  & \checkmark \\
Prograsp forceps & \checkmark & \checkmark & \checkmark \\
Stapler &  &  & \checkmark \\
Grasping retractor & \checkmark &  & \checkmark \\
Tip-up fenestrated grasper &  &  & \checkmark \\
Ultrasound probe &  & \checkmark &  \\
Suction instrument &  & \checkmark &  \\
\hline
\end{tabular}
\end{table}

\begin{figure}[htbp]
\begin{center}
\resizebox{5in}{!}{\includegraphics{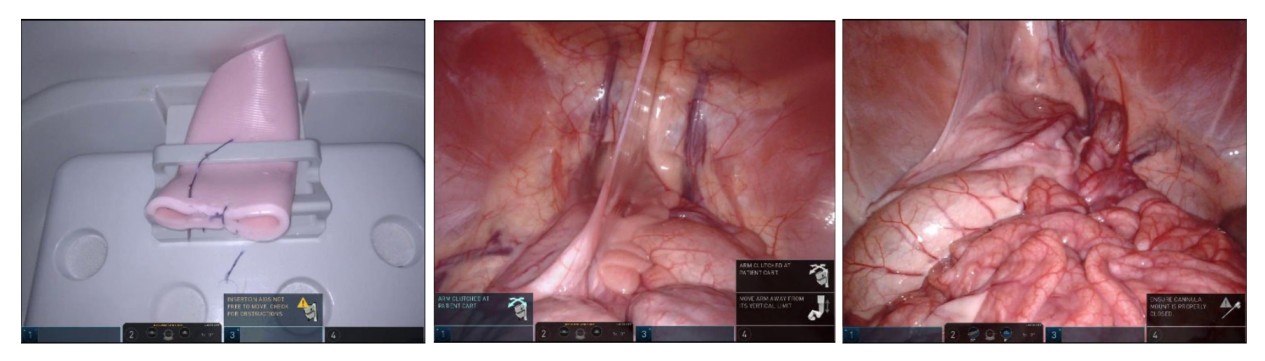}}
\caption{Examples of background images.}
\label{fig:C1_SKJP_background}
\end{center}
\end{figure}

Next, we extract background images from the SurgVU2024 dataset manually. The background images are selected from images where no tools are presented. The number of background images are 2,744, which are selected from about 30 videos and categorized to 55 scenes. Some examples of the background images are shown in Figure \ref{fig:C1_SKJP_background}.

\begin{figure}[htbp]
\begin{center}
\resizebox{5in}{!}{\includegraphics{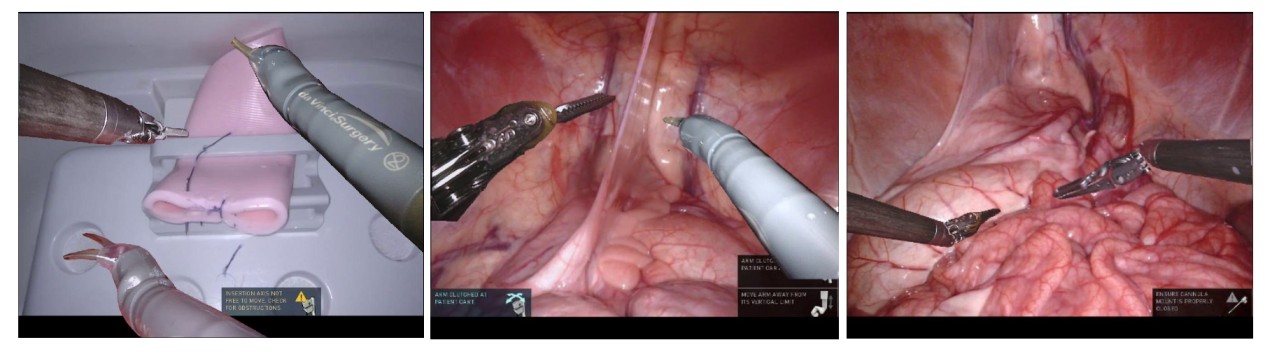}}
\caption{Examples of synthesized images.}
\label{fig:C1_SKJP_synthesized}
\end{center}
\end{figure}

Then, we synthesize surgical scene images by overlaying the tool images onto the background images. In this stage, a background image is selected randomly at first and then tools images are selected. When the tool images are selected, the number of tools is selected from 2 or 3. The possibilities selected 2 tools and 3 tools are 80\% and 20\%, respectively. When the number of tools is 2, one tool is selected from the left category and another tool is selected from the right category. When the number of tools is 3, the first, second and third tools are selected from the left, right and bottom categories, respectively. A selected tool image is resized. The ratio of the resize is randomly selected from 0.5 to 1.5. Then the resized tool image is translated in horizontal and vertical randomly. We have some restriction for the translation, e.g., a left tool does not translate to right side since the left side of the tool is apart from the edge of the image. Similar rules are applied to right and bottom tools. Finally, the resized and translated images are overlayed on to the background image and the overlayed image is saved. Also, bounding box information about each part (clasper, clevis and shaft) is stored. The total number of synthesized images is 54,879. Examples of the synthesize images are shown in Figure \ref{fig:C1_SKJP_synthesized}.

We train an object detection model by using the synthesized images and the corresponding bounding box information. We use YOLOv10 \cite{SKJP_Wang2024} as the object detector.

We train a second model to identify tool presence. The second model is trained by using the images and the tool presence labels in the SurgVU2024 dataset and as a multi-label classifier. We use ConvNeXt-v1-small \cite{SKJP_Liu2022} as the classifier.

\begin{figure}[htbp]
\begin{center}
\resizebox{5in}{!}{\includegraphics{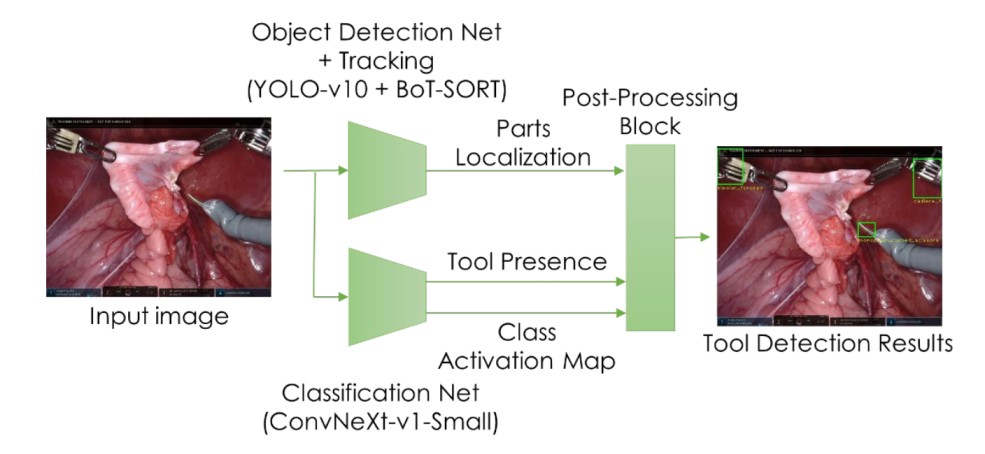}}
\caption{Block diagram in inference stage.}
\label{fig:C1_SKJP_blockdiagram}
\end{center}
\end{figure}

Figure \ref{fig:C1_SKJP_blockdiagram} shows the block diagram in inference stage. In inference stage, an input image is processed with two networks, i.e., object detection and classification. We obtain parts localization results from the object detection networks. Note that we use "object detection + tracking" mode after the second frame in the input video. BoT-SORT \cite{SKJP_Aharon2022} is used for the tracking. From the classification networks, we obtain tool presence results. We also obtain class activation maps (CAMs) from the classification networks. Grad-CAM \cite{SKJP_Selvaraju2017} is used for obtaining the CAMs.

\begin{figure}[htbp]
\begin{center}
\resizebox{3.5in}{!}{\includegraphics{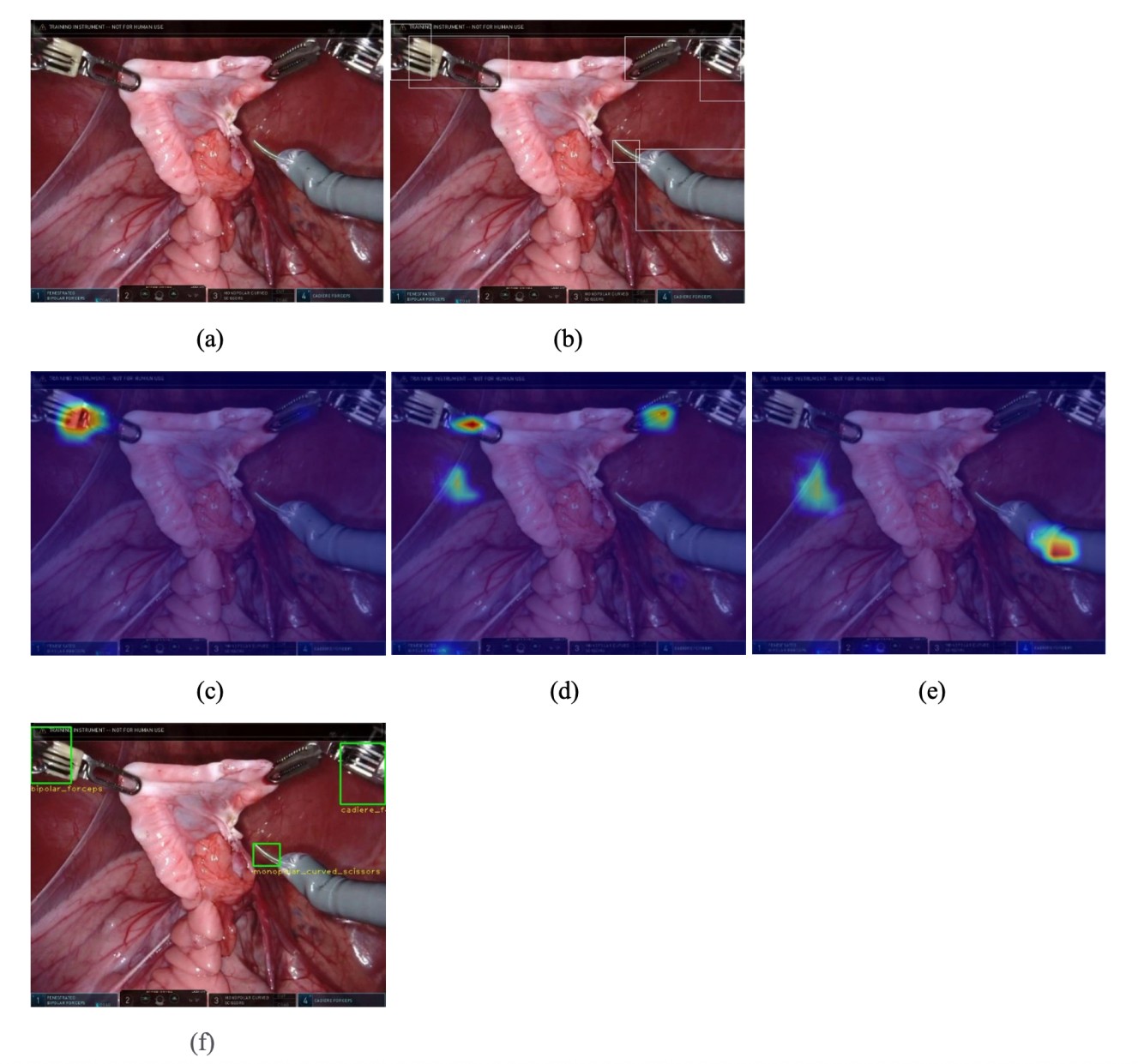}}
\caption{Examples of input image, intermediate results and final result.}
\label{fig:C1_SKJP_results}
\end{center}
\end{figure}

The parts localization results, the tool presence results, and the CAMs are processed in a post-processing block. We explain the procedure in the post-processing block by using Figure \ref{fig:C1_SKJP_results}. Figure \ref{fig:C1_SKJP_results}(a) shows the input image. We can see bipolar forceps at top-left, cadiere forceps at top-right, and monopolar curved scissors at bottom-right. Figure \ref{fig:C1_SKJP_results}(b) shows the parts localization results. While it is not shown in Figure \ref{fig:C1_SKJP_blockdiagram}, presence of "bipolar forceps", "cadiere forceps", and "monopolar curved scissors" is detected by the classifier in this case. And Figure \ref{fig:C1_SKJP_results}(c)-(e) shows CAMs for "bipolar forceps", "cadiere forceps", and "monopolar curved scissors", respectively. In the post-processing block, correspondence between parts localization results and CAMs is checked and a tool class is assigned for each group of parts. Figure \ref{fig:C1_SKJP_results}(f) shows the result.

\subsubsection{Model Training}
We train an object detection model using the synthesized images with YOLOv10 as the object detector. A second model for tool presence identification is trained using the images and tool presence labels in the SurgVU2024 dataset as a multi-label classifier, using ConvNeXt-v1-small as the classifier. For tracking in the inference stage, BoT-SORT is used after the second frame in the input video.

\subsubsection{Preliminary Performance}
As for the experimental results, mAP at the final testing phase was 0.0308.

We proposed a surgical tool localization method by using synthesized image dataset for category 1. In our synthesized images, tools not included in Endovis2017 and Endovis2018 are not included. It is a limitation of our method.

\FloatBarrier
\subsection{SEU-MIA}

We participated in Category 1 of the SurgVU challenge, focusing on surgical tool classification and localization. These tasks are critical in robotic-assisted surgery, where understanding the presence and position of tools can significantly enhance automation, safety, and precision. Given the complexity of surgical environments and the challenge posed by noisy labeled tools, we propose a multi-label recognition method based on prompt learning following DualCoOp \cite{SEUMIA_Sun2022}. Our hypothesis is that the visual-language model with learnable prompt have high robustness to noisy labels. The novelty of our approach stems from its use of learnable prompts that provide contextual understanding of the tools. By learning dual prompts designed for multi-label datasets, the model can better predict tool presence and classify multiple tools per frame.

\subsubsection{Method Description}

\begin{figure}[tbh]
\begin{center}
\resizebox{5in}{!}{\includegraphics{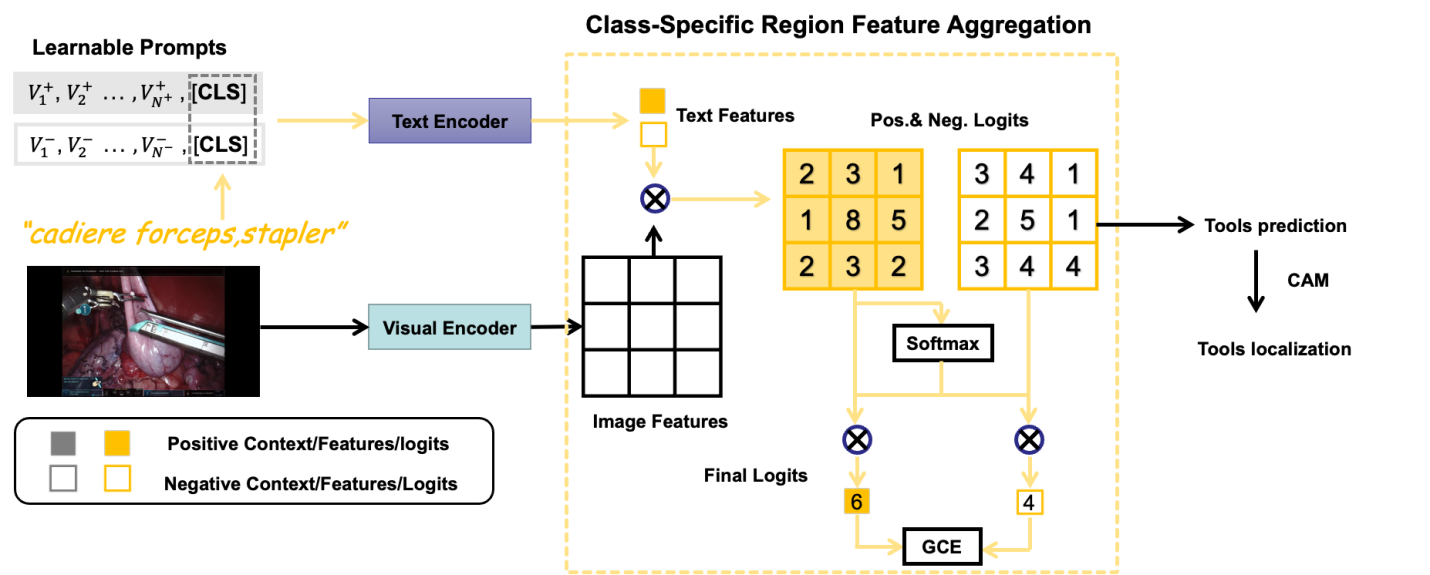}}
\caption{The framework of our proposed multi-label recognition and localization method.}
\label{fig:C1_SEU-MIA_framework}
\end{center}
\end{figure}

Our model is a prompt-based vision-language model that leverages the strong alignment between textual and visual representations. For this challenge, we adapted the model to predict both the class of the surgical tool and its spatial localization within each video frame. The workflow of our method is shown in Figure \ref{fig:C1_SEU-MIA_framework}.

The proposed method is based on the Contrastive Language-Image Pre-Training (CLIP) framework with context optimization. To make it suitable for multi-label context, we adopted the dual prompts from \cite{SEUMIA_Sun2022} by introducing dual prompts for both positive and negative contexts. A class specific region feature aggregation is proposed to adaptively aggregate region features in the multi-label setting.

We adopted the generalized cross entropy (GCE) loss \cite{SEUMIA_Zhang2018} to effectively address label noise and class imbalance in the weakly labeled dataset. By tuning the parameter q, GCE smoothly transitions between cross entropy and mean absolute error (MAE), offering better noise tolerance. This loss function is robust to noisy labels and helps the model mitigate the impact of label noise, improving generalization to both common and rare tools. We initialized the model using the pretrained CLIP backbone. The dual prompts were trained using a weakly supervised approach, where only tool presence served as the supervision signal. For localization, we utilized Class Activation Mapping (CAM) to highlight the spatial regions associated with each tool.

\subsubsection{Model Training}

The dataset was preprocessed to extract individual frames from the surgical videos with 0.5fps. We performed spatial augmentation and normalization to ensure the model could generalize across different sessions and tools.

\subsubsection{Preliminary Performance}

The model achieved an average precision (AP) of 62\% across all tool categories, with higher precision on frequently occurring tools.

In addressing the SurgVU challenge, we employed the DualCoOp network with a primary focus on managing the multi-noisy label issue, without tackling the detection problem directly. Our primary objective was to mitigate the impact of label noise, which is prevalent in surgical video data. During our investigations, we explored the use of class activation maps (CAM) for detection purposes but found that this approach was not suitable for our specific challenge. CAM helps visualize which areas of the image contribute the most to the model's classification decision, but it is not inherently designed for detecting and precisely identifying objects or events, especially when there are multiple or overlapping objects, as in detection tasks. This outcome suggests that alternative approaches, which are more resilient to noisy labels and better suited for precise detection in surgical video analysis, need to be considered for future work.

\FloatBarrier
\subsection{SurgOp}

We participated in Category 2 of the SurgVU2024 challenge, focusing on phase recognition in surgical videos. Our approach emphasizes simplicity, employing a single-stage model with minimal pre- and post-processing steps. This design allows us to concentrate on the primary issue of class imbalance, a key factor affecting model performance in this challenge. By using pre-trained weights from the Kinetics-400 dataset\cite{SurgOp_Kay2017kinetics}, we mitigate the need for large-scale surgical datasets, leveraging a widely used action recognition dataset to fine-tune our model for surgical phases. This method simplifies training while maintaining high accuracy, distinguishing our approach from more complex multi-stage pipelines typically seen in medical video analysis.

\subsubsection{Method Description}

\textbf{Data Preparation:} The videos were converted into sequences of frames, retaining 1 frame per second. Each frame was cropped and resized according to the following steps, illustrated in Figure~\ref{fig:C2_SurgOp_preprocessing}: a central crop with a 4:3 aspect ratio was performed to remove black borders; 8\% of the bottom of each frame was removed to hide the information band, while the top band was retained; the resulting image was resized to 256x256, allowing random crop and resize augmentation.

\begin{figure}[tbh]
\begin{center}
\resizebox{5in}{!}{\includegraphics{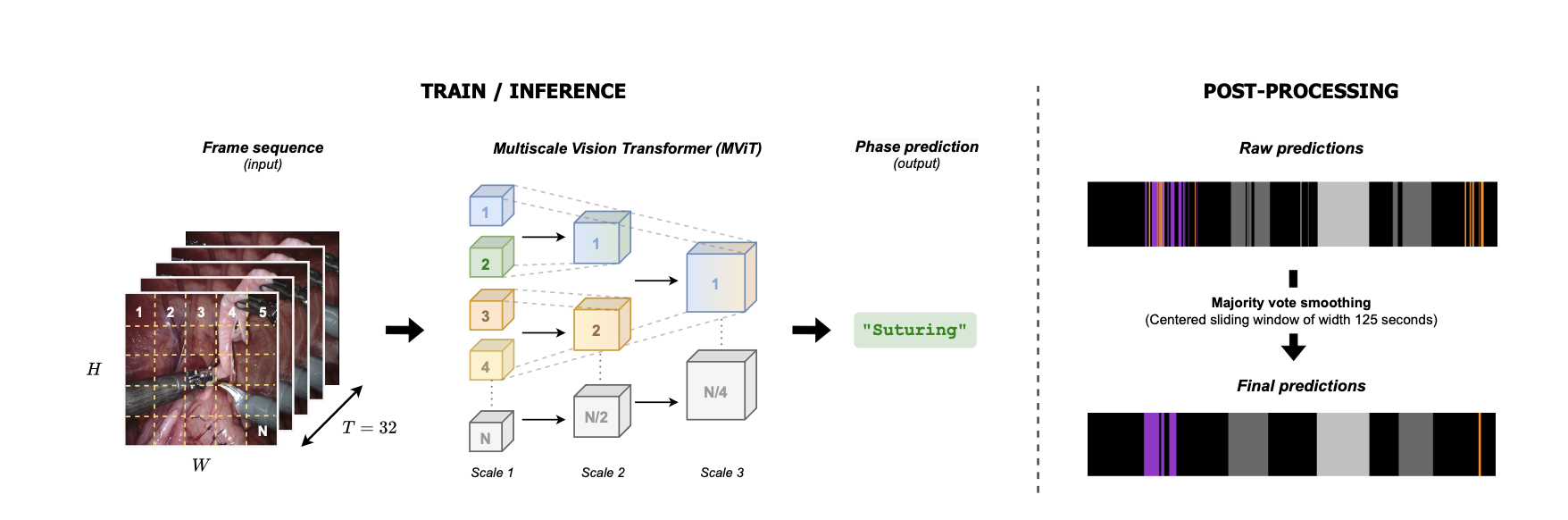}}
\caption{Overview of the pipeline. During inference, phase predictions are generated from sequences of 32 successive frames passed through an MViT model. Final predictions are refined using majority vote smoothing on the predicted phases.}
\label{fig:C2_SurgOp_pipeline}
\end{center}
\end{figure}

\begin{figure}[tbh]
\begin{center}
\resizebox{3in}{!}{\includegraphics{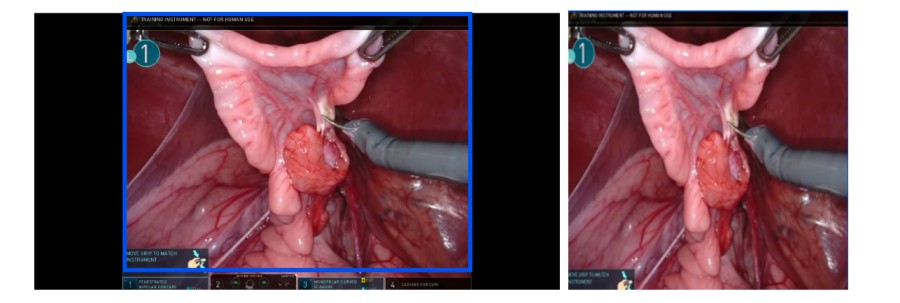}}
\caption{Pre-processing illustration: (right) the blue rectangle indicates the retained portion of the image; (left) the retained portion is reshaped to 256x256.}
\label{fig:C2_SurgOp_preprocessing}
\end{center}
\end{figure}

We used the PyTorchVideo\cite{SurgOp_Fan2021pytorchvideo} library to build the network and perform data augmentation. The training loop was executed using the PyTorch Lightning\cite{SurgOp_Falcon2019lightning} library, which provides clean and structured code.

\textbf{Model Architecture:} We employed an MViT architecture\cite{SurgOp_Fan2021mvit}, which accepts sequences of 32 frames as input, with an online sliding window for predictions. We utilized pre-trained weights from PyTorchVideo model zoo. Training employed the standard cross-entropy loss function with the AdamW optimizer, using a learning rate of $1 \times 10^{-4}$ and a minimal weight decay of $1 \times 10^{-8}$. The batch size was set to 5 samples per GPU on a dual-GPU machine, using mixed precision. Though we initially implemented a step learning rate scheduler, early stopping prevented any learning rate changes from occurring.

\textbf{Data Augmentation:} Data augmentation was carried out using PyTorchVideo's \texttt{RandomResizedCrop} and \texttt{RandAugment} transforms\cite{SurgOp_Cubuk2020randaugment}. All transformations, including input conversion from uint8 to float, input standardization, and random horizontal flipping, were handled in the \texttt{create\_video\_transform} function. During preliminary tests, the minimum size for random cropping was set to 0.25, and RandAugment was set to a mild augmentation with $m = 6$ and $n = 2$. These values were later adjusted to 0.08 and $m = 10$, $n = 2$ for final testing.

\subsubsection{Model Training}

\textbf{Pre-training:} We used the weights from a pre-trained model on the Kinetics-400\cite{SurgOp_Kay2017kinetics} dataset. This dataset includes 306,245 video clips of approximately 10 seconds each, annotated with 400 action classes. Although it does not contain surgical videos and typically features shorter videos at higher frame rates, it is one of the few video datasets offering pre-trained backbones for action recognition.

\textbf{Class Balancing and Sampling:} A significant challenge in the dataset was the class imbalance, with the "other" class representing 75\% of the dataset but being the least relevant in practice. The label noise in this class often reduced generalization performance on other classes. To address this, we implemented a sampler in the data loader that selected samples with probability $\frac{1}{p_c}$, where $p_c$ is the proportion of the class in the dataset. This ensured uniform class balancing during training and improved performance across all classes. Under this setup, the notion of epochs became less meaningful. We evaluated the network every 1000 training steps and applied early stopping to prevent prolonged overfitting. Interestingly, the submitted networks were trained on only 1\% of the total dataset (30k samples out of over 3 million total samples), yet the large scale was necessary for proper class balancing.

\begin{figure}[tbh]
\begin{center}
\resizebox{5in}{!}{\includegraphics{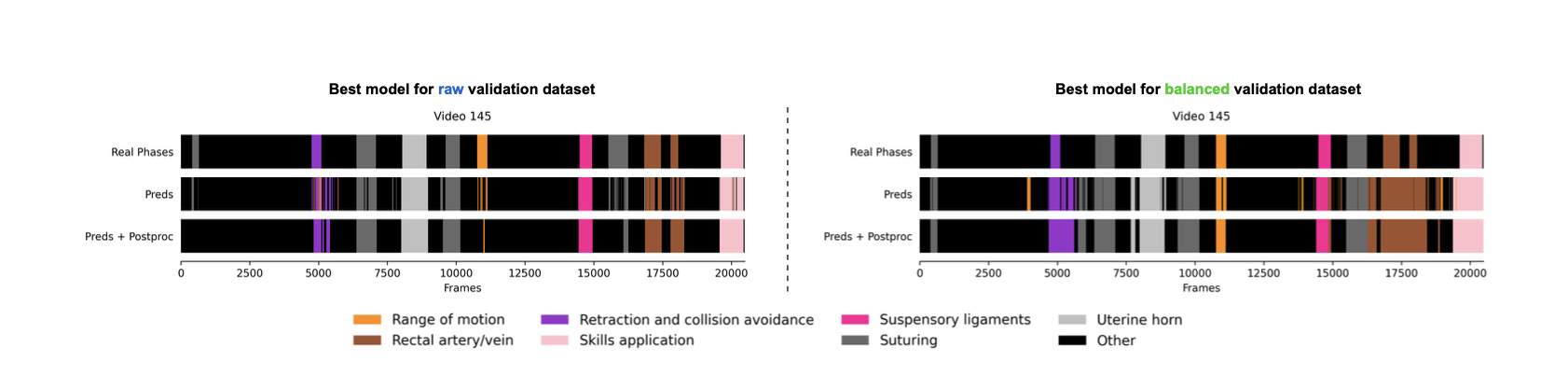}}
\caption{Comparison of two models: The left shows the model trained with partially balanced resampling, optimized for the original class distribution of complete videos (preliminary test F1 score: 0.7087). The right shows the model trained with fully balanced resampling, optimized for a balanced validation set (preliminary test F1 score: 0.7752; final test F1 score: 0.8521).}
\label{fig:C2_SurgOp_comparison}
\end{center}
\end{figure}

\textbf{Post Processing:} We did not want to rely on complex post-processing as the extra complexity would have made tuning the training parameters more tedious. We relied on a majority vote smoothing with a centered sliding window of size 125 seconds. This method helps by reducing phase prediction outliers, particularly fast and erratic phase changes, leading to more stable and consistent predictions. For a qualitative example of the effect of post-processing, see Figure~\ref{fig:C2_SurgOp_comparison}, which compares predictions before and after applying majority vote smoothing.

\textbf{Model Selection and Evaluation:} To evaluate the model, we reserved the last 15 videos (cases 139-154) as a validation set. This set, representing complete surgical procedures, was used to assess model performance using metrics such as accuracy, precision, recall, F1 score, and Jaccard index, both globally and per phase. Models were evaluated with and without post-processing. To monitor for overfitting without slowing down training, we sampled 500 batches from the validation set. A full evaluation, taking around 2 hours, was performed only at the end of training.

We observed that F1 scores are influenced by the distribution of the evaluation dataset. Specifically, the precision of minority classes decreases when the "other" class dominates, increasing false positives in these classes and lowering their overall F1 scores. Due to the unknown distribution of the challenge test set, we trained two models targeting different objectives: (1) maximizing the F1 score on the raw validation set of complete videos and (2) on a balanced validation set, respectively. The model optimized for the raw set, using partially balanced resampling, performed best on the "other" class, while the model trained with fully balanced resampling excelled in recognizing minority classes. This approach prepared us for various potential test set distributions. Figure~\ref{fig:C2_SurgOp_comparison} illustrates this strategy, comparing the model optimized for the original class distribution with complete videos (left) with the one optimized for a balanced validation set (right).

\subsubsection{Preliminary Performance}

We submitted multiple versions of the model with different sampling strategies. On the preliminary test set, our final three MViT submissions achieved F1 scores of 0.7087, 0.7519, and 0.7752. The best-performing preliminary model, trained with uniform sampling across all classes, was retained for the final submission and achieved a final test F1 score of 0.8521.

The majority vote smoothing applied during post-processing effectively reduced erratic phase predictions, resulting in more consistent outputs. Additionally, using balanced resampling techniques during training significantly improved the model's ability to generalize across different classes, particularly when faced with the highly imbalanced "other" class. The use of MViT, combined with pre-trained weights, enabled us to bypass the scarcity of annotated surgical datasets while achieving competitive performance.

Despite improvements in handling class imbalance, the dominance of the "other" class continued to introduce noise, particularly in cases where misclassifications in this class impacted the recall for other classes. We also observed discrepancies between our internal evaluation and the preliminary test set results, highlighting the difficulty in managing class distributions during both training and evaluation.

\FloatBarrier
\subsection{BCU}

We participated in Category 2 of the SurgVU challenge, focusing on surgical task recognition from long minimally invasive surgical videos. Our approach addresses the significant challenges posed by extensive video data through efficient data sampling, lightweight model ensembling, and targeted post-processing techniques. We achieved third place in the MICCAI 2024 SurgVU challenge with a mean weighted F1 score of 0.8216, demonstrating that simpler models can achieve performance comparable to state-of-the-art methods when integrated into intelligently designed machine learning pipelines.

\subsubsection{Method Description}

\begin{figure}[tbh]
\begin{center}
\resizebox{5in}{!}{\includegraphics{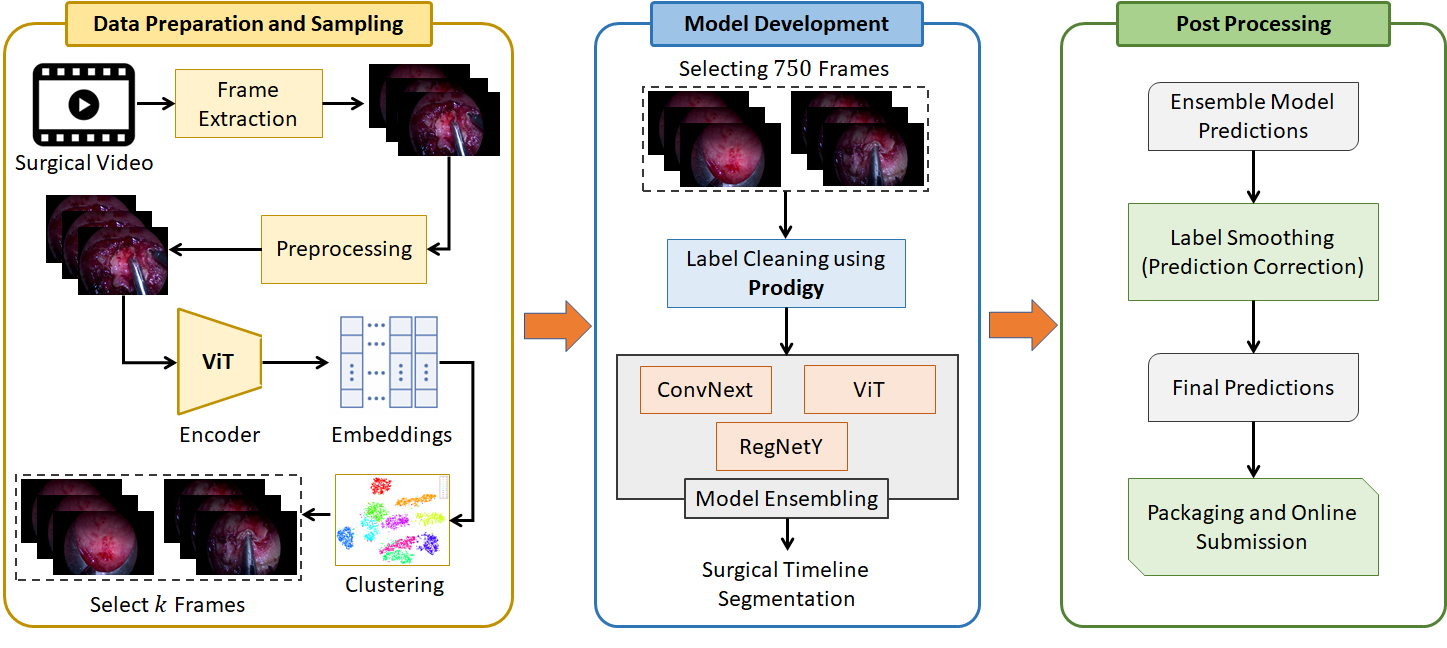}}
\caption{Our proposed solution architecture encompasses three major steps: (1) data preparation and sampling; (2) ensemble model development; and (3) post-processing for error correction.}
\label{fig:C2_BCU_architecture}
\end{center}
\end{figure}

Our methodology consists of three major components: data preparation and sampling, ensemble model development, and post-processing for error correction, as illustrated in Figure~\ref{fig:C2_BCU_architecture}.

\textbf{Data Preparation and Efficient Data Sampling:} One of the significant challenges we faced was managing the extensive size of the dataset, which included approximately 180 million frames from 280 video recordings captured at 60 fps. We developed a self-supervised learning-based data sampling strategy to curate a smaller yet representative subset of data. Our efficient data sampling methodology involves the following key steps:

\begin{enumerate}
\item \textbf{Frame Extraction:} We extracted individual frames from minimally invasive surgical videos at a rate of 1 FPS using FFmpeg, yielding approximately 3 million frames.
\item \textbf{Feature Encoding:} Each extracted frame was encoded into feature embeddings using a Vision Transformer (ViT) encoder, facilitating the capture of complex visual patterns and effectively differentiating between various surgical activities.
\item \textbf{Clustering:} The generated feature embeddings were clustered using the K-Means algorithm, enabling the identification of visually similar frames and effectively grouping the dataset based on inherent visual similarities.
\item \textbf{Sample Selection:} We selected a minimal set of representative frames from each cluster, reducing the dataset size from approximately 3 million frames to around 100,000 frames while retaining diverse and representative subsets.
\end{enumerate}

Through this efficient data sampling strategy, we achieved a substantial 20-fold reduction in data volume, making it feasible to train sophisticated machine learning models without compromising the diversity and representativeness of the surgical tasks captured within the dataset.

\textbf{Baseline Models:} Our first step in model development involved evaluating state-of-the-art architectures as baseline models. We focused on models capable of capturing long-range spatio-temporal patterns, a critical requirement for accurately recognizing surgical tasks. We trained various architectures, including ConvNeXt~\cite{BCU_Liu2022}, RegNetY~\cite{BCU_Radosavovic2020}, and Vision Transformer (ViT)~\cite{BCU_dosovitskiy2020image}. Each model was trained using the SurgVU dataset to assess its potential in learning the complex temporal relationships within surgical procedures.

\textbf{Ensemble Model:} To further enhance predictive performance, we adopted ensemble learning as our key strategy. The motivation behind combining multiple models was to leverage their complementary strengths, thereby producing more reliable and accurate predictions. We utilized three distinct models: ConvNeXt, RegNetY, and ViT. Each model excelled in different aspects of data processing, with ConvNeXt handling intricate spatial features, RegNetY offering robust scalability, and ViT capturing long-range temporal dependencies. We assigned weights to each model's output based on their performance on the training data, ensuring that the final ensemble output reflected the most accurate predictions. Mathematically, our generalized ensemble model can be represented as:
\[
\text{Ensemble} = W_1 \cdot \text{ConvNeXt} + W_2 \cdot \text{RegNetY} + W_3 \cdot \text{ViT}
\]

\subsubsection{Model Training}

We utilized the University of Birmingham's BlueBEAR High-Performance Computing service, running computations on NVIDIA A100 GPUs with 80 GB of memory. All models were implemented using the fastai library, and we employed the timm library to access deep learning models for transfer learning.

We created a custom dataloader that reads images from specified folder paths and maps each frame to its corresponding label using a dataframe. The dataloader splits the data into training and validation sets using an 80:20 ratio, based on our intelligent data splitting strategy. We developed a nested stratification approach based on both unique task labels and surgical time to create a balanced training and validation split that accurately reflects the dataset's diversity.

Our training methodology involved two phases: first, pretraining the model's head for 50 epochs with a learning rate of $1 \times 10^{-3}$; second, fine-tuning the entire model (after unfreezing all layers) for 30 epochs with a learning rate range from $1 \times 10^{-3}/400$ to $1 \times 10^{-3}/4$. We employed the AdamW optimizer with the ReLU activation function in our experiments.

\textbf{Post-Processing:} Although ensemble learning improved our models' performance by approximately 5\% in mean F1 score on the preliminary leaderboard, visual inspection of the predictions highlighted residual issues related to temporal inconsistencies. We observed instances of "phase shaking," where the model generated erroneous predictions at the transitions between surgical phases. To address these issues and enhance the temporal coherence of the predictions, we incorporated a post-processing step using neighbour smoothing with a predefined window size. The smoothing algorithm is formally defined in Algorithm~\ref{alg:smoothing}. Given a list of predicted class labels ${c_i}_{i=1}^N$ and a window size $w$, the algorithm processes each frame as follows:
\begin{algorithm}[H]
\caption{Neighbour Smoothing Algorithm}
\label{alg:smoothing}
\KwIn{Predicted classes $\{c_i\}_{i=1}^N$, window size $w$}
\KwOut{Smoothed predictions $\{\hat{c}_i\}_{i=1}^N$}
\For{$i \leftarrow 1$ \KwTo $N$}{
    $\texttt{start\_idx} \gets \max(1,\, i - w)$\;
    $\texttt{end\_idx} \gets \min(N,\, i + w)$\;
    $\texttt{neighbourhood} \gets \{c_j\ |\ j \in [\texttt{start\_idx},\, \texttt{end\_idx}]\}$\;
    $\texttt{counts} \gets$ frequency count of classes in $\texttt{neighbourhood}$\;
    $\texttt{max\_count} \gets \max(\texttt{counts.values()})$\;
    $\texttt{common\_classes} \gets \{c\ |\ \texttt{counts}[c] = \texttt{max\_count}\}$\;
    \eIf{$\text{len}(\texttt{common\_classes}) = 1$}{
        $\hat{c}_i \gets$ the single class in $\texttt{common\_classes}$\;
    }{
        $\hat{c}_i \gets c_i$\; \tcp{Retain original class in case of a tie}
    }
}
\end{algorithm}
This method scans the predicted sequence of classes and replaces each prediction with the most frequent class within a moving window centered on that frame. We found that a window size of 8 frames yielded the highest average F1 score. Employing this neighbour smoothing technique, we significantly reduced the occurrence of phase-shaking errors and improved the temporal consistency of the model's outputs.

\subsubsection{Preliminary Performance}

Our methodology led to progressive improvements, starting from individual models, enhanced through ensembling, and further refined via post-processing. Table~\ref{tab:C2_BCU_results} summarizes the performance metrics of the baseline models, the ensemble method, and the post-processing technique on our validation set.

\begin{table}[htbp]
\centering
\caption{Comparative analyses of our proposed models, ensemble method, and post-processing technique for surgical task recognition.}
\label{tab:C2_BCU_results}
\begin{tabular}{lcccc}
\hline
\textbf{Model} & \textbf{Accuracy} & \textbf{Precision} & \textbf{Recall} & \textbf{F1 Score} \\
\hline
RegNetY-008 & 0.5841 & 0.6753 & 0.5120 & 0.5406 \\
ConvNeXtV2-Tiny & 0.8182 & 0.8286 & 0.7045 & 0.7469 \\
ViT-Tiny & 0.8116 & 0.7860 & 0.7560 & 0.7681 \\
Ensemble & 0.8127 & 0.8327 & 0.7156 & 0.7557 \\
Post-Processing & \textbf{0.8347} & \textbf{0.8798} & 0.7243 & \textbf{0.7680} \\
\hline
\end{tabular}
\end{table}

Among the individual encoders, ConvNeXtV2-Tiny achieved the highest accuracy of 81.82\%, closely followed by ViT-Tiny with 81.16\%. The ensemble model achieved an accuracy of 81.27\% and the highest precision of 83.27\%, surpassing each baseline model in precision while maintaining competitive accuracy. Our post-processing technique further refined the predictions, leading to the highest accuracy of 83.47\%, the highest precision of 87.98\%, and an improved F1 score of 76.80\%.

Figure~\ref{fig:C2_BCU_visual_comparison} provides a visual comparison of predictions from the baseline models, the ensemble, and the post-processed outputs. The enhancement in the quality of surgical task recognition is evident, with the post-processing step effectively eliminating inconsistencies and improving overall prediction coherence.

\begin{figure}[tbh]
\begin{center}
\resizebox{5in}{!}{\includegraphics{TeamDocs2024/C2_BCU/c2_fig4.png}}
\caption{Colour-coded ribbon comparison of model predictions. Each colour represents a different phase label, illustrating the progression from baseline models to ensemble and post-processed outputs.}
\label{fig:C2_BCU_visual_comparison}
\end{center}
\end{figure}

In the final testing phase of the SurgVU MICCAI 2024 Grand Challenge, we achieved third place with a mean weighted F1 score of 0.8216 across all classes. Our model's performance demonstrates its competitiveness among state-of-the-art methods. Notably, we outperformed teams employing advanced algorithms such as Surgformer, underscoring the strength of our methodology despite its relative simplicity. Our approach relies on a straightforward baseline model without incorporating complex architectural enhancements or extensive computational resources, making it particularly suitable for practical applications where computational limitations are a concern.

\FloatBarrier
\subsection{SmartLab HKUST}

We participated in Category 2 of the SurgVU challenge, focusing on surgical step recognition from untrimmed surgical videos. Our approach reformulates frame-level step recognition as clip-level step classification over clips with fixed temporal resolution, enabling end-to-end surgical workflow understanding. We adopt the architecture and training strategy of Surgformer~\cite{HKUST_Yang2024surgformer} for fine-grained step recognition, thereby simplifying the conventional two-stage pipeline for untrimmed surgical video analysis.

\subsubsection{Method Description}

For Category 2, we address surgical step recognition in untrimmed surgical videos, which requires accurate frame-level understanding under complex temporal transitions and substantial visual redundancy. Conventional surgical scene understanding methods usually follow a two-stage paradigm, where the target frame and its preceding frames are first converted into spatial features, and temporal dependencies are then modeled by dedicated temporal modules to obtain spatial-temporal representations~\cite{HKUST_Czempiel2020tecno,HKUST_Liu2023skit}. Although effective for handling long video sequences and aggregating global temporal context, such stage-wise designs may limit joint spatial-temporal optimization and overlook the fine-grained spatial cues required for precise step recognition. More recently, end-to-end methods have shown promising performance in surgical scene understanding~\cite{HKUST_Yang2025surgpetl,HKUST_Yang2026large}, as they enable unified optimization of spatial and temporal modeling within a single framework.

Motivated by this trend, we reformulate frame-level step recognition as clip-level classification over short clips with fixed temporal resolution, thereby enabling end-to-end surgical workflow understanding while preserving the local spatial details and temporal dynamics required for fine-grained recognition. Our implementation is based on Surgformer~\cite{HKUST_Yang2024surgformer}, an end-to-end Transformer framework for surgical step recognition. Surgformer employs divided spatial-temporal attention~\cite{HKUST_Bertasius2021timesformer} to process a sparse set of sampled frames efficiently, reducing redundancy while maintaining discriminative spatial-temporal representations. In addition, it introduces Hierarchical Temporal Attention to capture both local and global temporal dependencies from a target frame-centric perspective. We adopt its architecture and training strategy for the SurgVU challenge.

Given an untrimmed surgical video, we sample $T$ frames from the target frame with a fixed stride $R$ to construct a clip volume $V$, and predict the corresponding surgical step for the target frame. The sampled clip is divided into non-overlapping patches in spatial-temporal order, which are then projected into spatial-temporal tokens. An additional class token is appended to aggregate global video-level information. After adding dynamic learnable 3D positional embeddings, the resulting token sequence is fed into a stack of Transformer blocks for end-to-end step recognition. Within each Transformer block, the tokens are first processed by Hierarchical Temporal Attention (HTA), which aggregates temporal information at each spatial location, and then by Aggregated Spatial Attention, which propagates the learned temporal context across spatial positions. Specifically, HTA models temporal relations among tokens at each spatial location from a target frame-centric perspective. By grouping temporal tokens into segments with different temporal resolutions, HTA captures both short-range and long-range temporal dependencies and refines the temporal representation for fine-grained surgical step recognition.

\begin{figure}[tbh]
\begin{center}
\resizebox{5in}{!}{\includegraphics{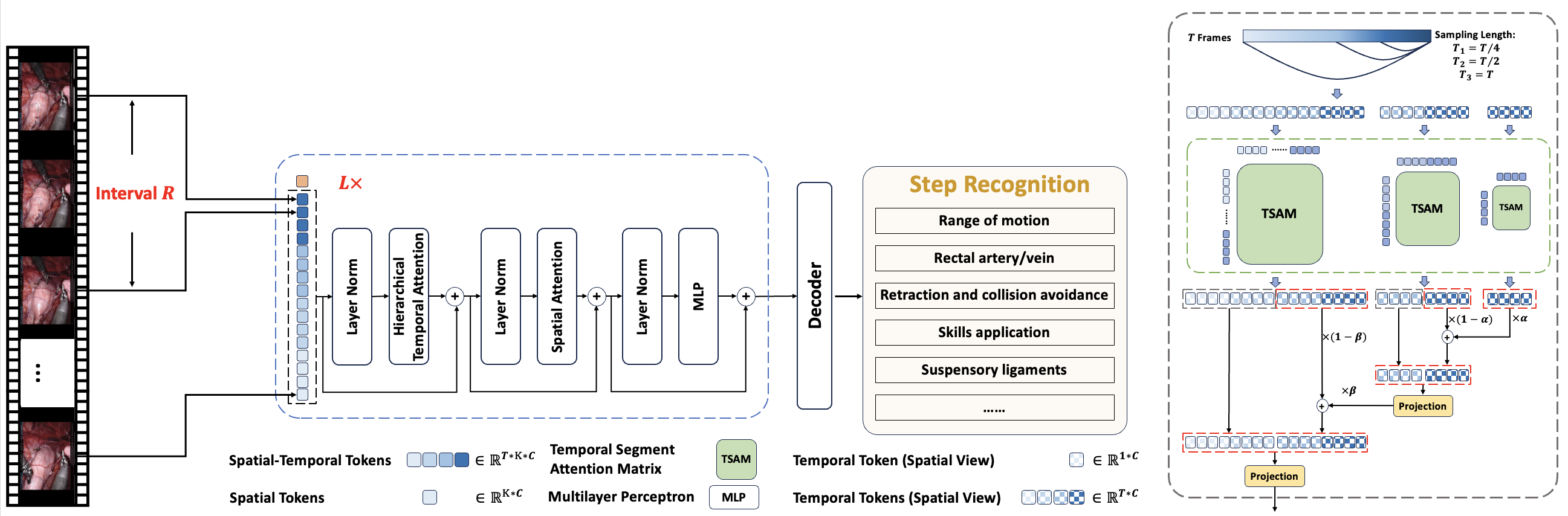}}
\caption{Overview of our Surgformer-based approach for surgical step recognition.}
\label{fig:C2_HKUST_architecture}
\end{center}
\end{figure}

\subsubsection{Model Training}

We uniformly sample videos at 1 FPS to reduce temporal redundancy. To further reduce storage requirements and alleviate I/O overhead during training, we resize the shorter side of each frame to 360 pixels. The dataset is split into 135 videos for training, corresponding to case\_000 to case\_134, and 20 videos for validation, corresponding to case\_135 to case\_154. No additional pre-processing is applied before training.

During training, random scaling and cropping are applied to generate input frames of size $224 \times 224$. We further adopt strong data augmentation strategies, including RandAugment~\cite{HKUST_Cubuk2020randaugment} and random erasing~\cite{HKUST_Zhong2020randomerasing}. Model weights are initialized from TimeSFormer~\cite{HKUST_Bertasius2021timesformer} pre-trained on Kinetics-400~\cite{HKUST_Kay2017kinetics}, while the remaining layers are randomly initialized. For model selection, we monitor image-level accuracy on the validation set during training and select the checkpoint with the best validation performance. For evaluation within the Docker container, we use the mean weighted F1-score to determine inference hyperparameters, including the sequence length and frame sampling interval.

\subsubsection{Preliminary Performance}

Based on the results in the final testing phase, recognizing surgical steps from clips with limited temporal resolution still achieves strong performance, demonstrating the effectiveness of our framework for surgical scene understanding. Moreover, increasing the temporal resolution at inference time does not lead to further performance gains, which suggests that accurate surgical step recognition can be achieved without relying heavily on long-term temporal context.

We also observe that the test data distributions in the preliminary and final testing phases are not fully balanced. This observation highlights the importance of considering data distribution when constructing the training and validation splits, rather than relying solely on random partitioning. A more distribution-aware split can help reduce potential bias and improve the robustness of the model across diverse surgical videos.
\FloatBarrier
\subsection{MIDAS}

We present a two-stage method for fusing multi-stage temporal convolutional networks (MS-TCN) and Kolmogorov–Arnold Networks (KAN)\cite{MIDAS_Liu2024KAN}. The goal of the surgical step recognition challenge is to perform classification on each frame of the video. In the challenge, instead of classifying every frame in the video, the task is to classify 1 frame per second of the whole video, which is approximately 40 to 60 minutes long. To perform the classification of this long video, which can contain multiple surgical steps in a single video, we needed to consider the long-term temporal correlation of the video frames. To understand this long-term temporal correlation and spatial representation, we propose fusing the MS-TCN and KAN.

\subsubsection{Method Description}

We utilize a hybrid design that comprises KAN and MS-TCN to jointly learn useful information from endoscopic videos. Our method constitutes a surgical step recognition task pipeline consisting of the following steps. In stage 1, we train the feature extractor with fc layer as the classification head. In stage 2, we train MS-TCN, KAN, and a fc layer with features extracted from stage 1.

\begin{figure}[tbh]
\begin{center}
\resizebox{5in}{!}{\includegraphics{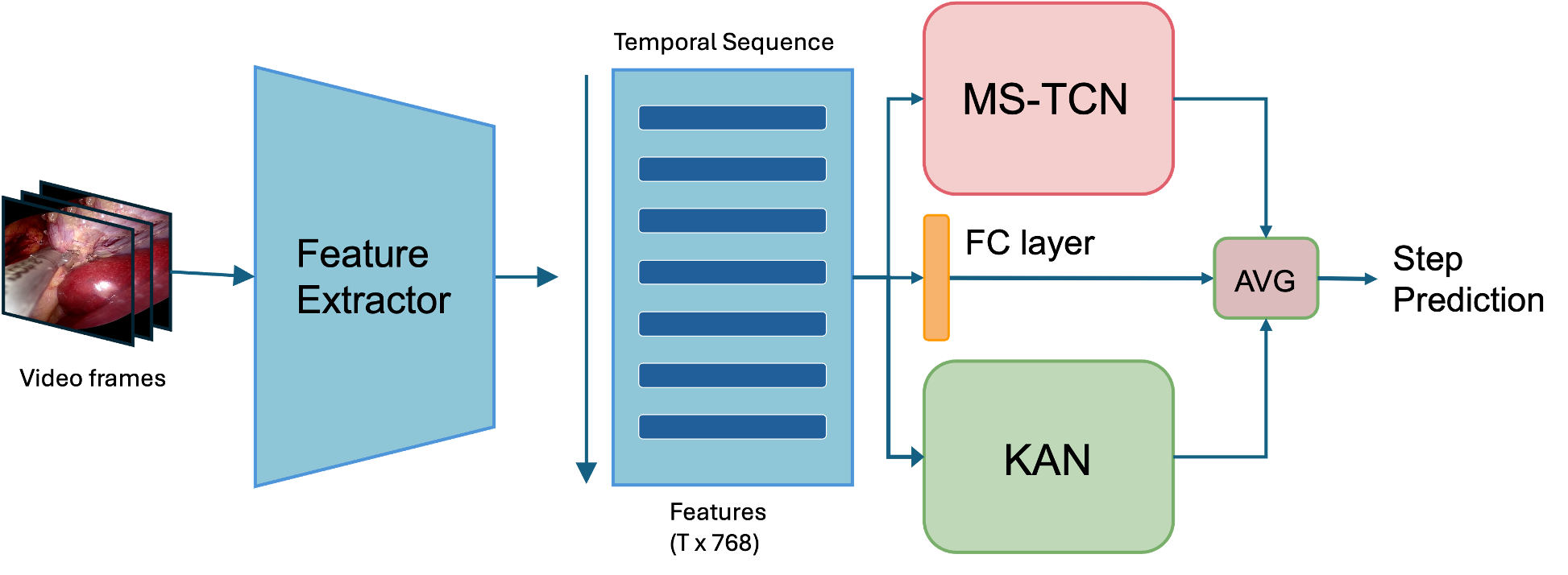}}
\caption{Workflow of our method for surgical step recognition}
\label{fig:C2_MIDAS_workflow}
\end{center}
\end{figure}

In stage 1, the feature extractor is trained frame-wise without temporal context where the video frames are considered as a single image for step recognition. Generally, in most step recognition datasets, the data are highly imbalanced and prone to overfitting. To tackle this problem, we implement focal loss for this task. Focal loss is effective in class imbalance tasks by adjusting the cross-entropy by applying a modulating factor to the loss of each sample based on its predicted probability\cite{MIDAS_Lin2017Focal}. This reduces the relative loss contribution from well-classified examples and focuses more on the hard-to-classify or misclassified examples.

\begin{figure}[tbh]
\begin{center}
\resizebox{5in}{!}{\includegraphics{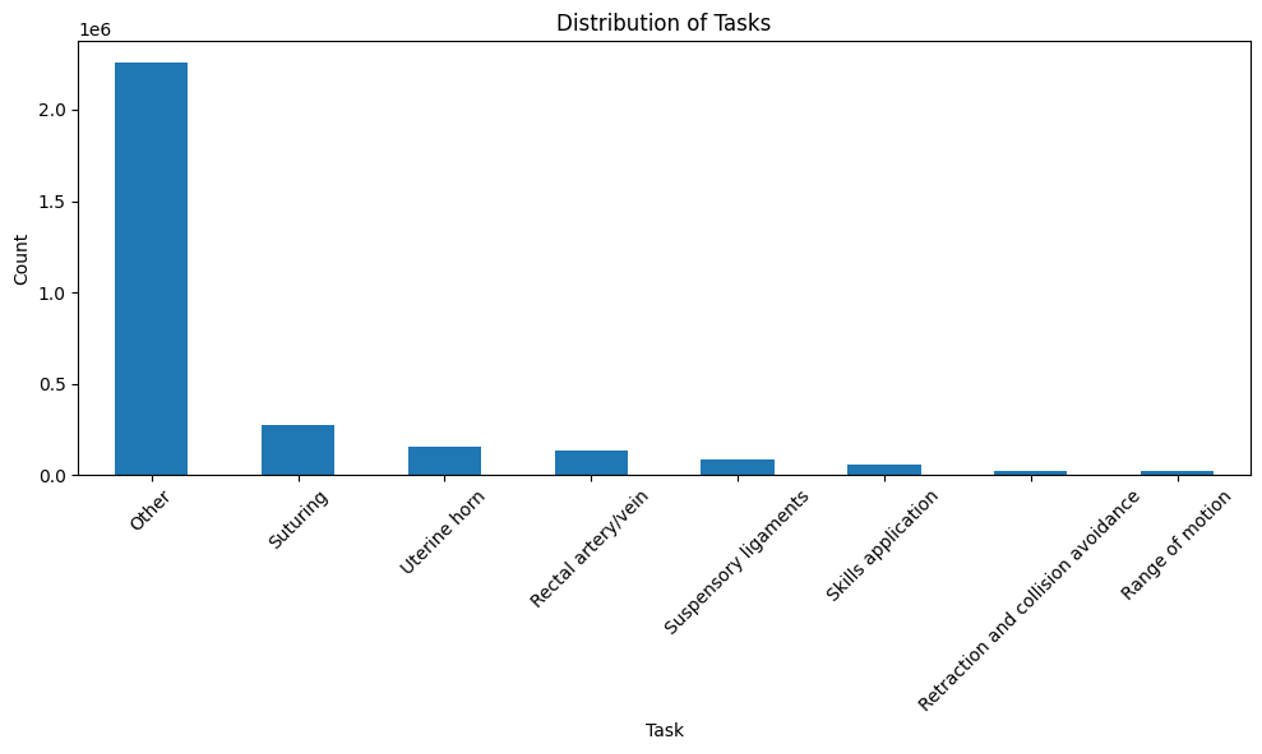}}
\caption{Data distribution of the steps in the dataset}
\label{fig:C2_MIDAS_data_distribution}
\end{center}
\end{figure}

As the sequential nature of the surgeries, the sequence of the steps in the video can be crucial in prediction. Temporal modeling captures the sequential structure\cite{MIDAS_Liu2023LoViT}. In stage 2, we incorporate MS-TCN from TeCNO\cite{MIDAS_Czempiel2020TeCNO} to capture the long-range temporal dependencies and patterns across the frames in the video. Moreover, we implemented the KAN, which tends to be more generalizable than MLP. We assume that learnable activation functions can lead to learning different patterns in the data that can lead to better generalization. KAN can capture complex patterns, fc layers capture high-level abstract representations, and MS-TCN captures temporal dependencies in sequential data. To combine the strengths of each method, we fuse them by averaging the prediction.

\subsubsection{Model Training}

In stage 1, we train the feature extractor with a fully connected layer as the classification head using focal loss to address class imbalance. In stage 2, we train MS-TCN, KAN, and a fully connected layer using features extracted from stage 1.

\subsubsection{Preliminary Performance}

In the final testing phase our method performed 69.16\% in F1 metric system. Although it has not shown its effectiveness in evaluation, we believe that this dual pipeline in the second stage has room for improvement and is key in refining the model's prediction from each pipeline. However, one of the most important sections of this task is building a generalizable feature extractor. Without a feature extractor that can extract a generalizable representation, the performance improvement through MS-TCN and KAN is limited. In the future, we will focus on building a generalizable feature extractor using foundation models.

\clearpage

\subsection{Results}

Challenge evaluations were conducted on a private, hidden test set comprising surgical video files at 1~FPS. Results were generated through the Grand Challenge automated algorithm submission and evaluation system. Participants did not have access to the test data or labels.

\subsubsection{Category 1: Surgical tool detection}

For Category~1, the standard COCO bounding box detection metric (mAP@[0.5:0.05:0.95]) \cite{lin2014microsoft} was used for evaluation and ranking. Table~\ref{tab:c1_results_2024} presents the final results for teams with complete submissions. The top two teams, PDMYR and InspireLab, achieved mAP scores above 0.42, comparable to the best results from prior years. MULTIS placed third with a mAP of 0.3478. Performance dropped sharply for the remaining teams.

\begin{table}[h!]
\centering
\caption{Category 1 Results -- 2024 (complete submissions only)}
\begin{tabular}{clr}
\toprule
\textbf{Rank} & \textbf{Team} & \textbf{mAP} \\
\midrule
1st  & PDMYR      & 0.4244 \\
2nd  & InspireLab & 0.4217 \\
3rd  & MULTIS     & 0.3478 \\
4th  & SJTUB      & 0.2754 \\
5th  & SKJP       & 0.031  \\
6th  & SEU-MIA    & 0.001  \\
\bottomrule
\end{tabular}
\label{tab:c1_results_2024}
\end{table}

\subsubsection{Category 2: Surgical task recognition}

For Category~2, the mean weighted F1-score across all surgical task classes was used for evaluation and ranking. Table~\ref{tab:c2_results_2024} presents the final results. All six teams with complete submissions achieved F1-scores above 0.69, indicating that the fully supervised task recognition problem was considerably more accessible than the weakly supervised detection task. PDMYR achieved the highest score of 0.8977. The top five teams all exceeded an F1-score of 0.81, demonstrating that modern video understanding architectures can effectively classify surgical steps when adequate supervision is provided.

\begin{table}[h!]
\centering
\caption{Category 2 Results -- 2024 (complete submissions only)}
\begin{tabular}{clr}
\toprule
\textbf{Rank} & \textbf{Team} & \textbf{Weighted F1} \\
\midrule
1st  & PDMYR          & 0.8977 \\
2nd  & SurgOp         & 0.8521 \\
3rd  & TeamBCU        & 0.8216 \\
4th  & SmartLab\_HKUST & 0.821  \\
5th  & InspireLab     & 0.8103 \\
6th  & MIDAS          & 0.6916 \\
\bottomrule
\end{tabular}
\label{tab:c2_results_2024}
\end{table}

\subsection{Discussion}

The 2024 challenge introduced surgical task recognition as a second category alongside the established tool detection task. This dual-category format enabled a comparison between weakly supervised detection (Category~1) and fully supervised recognition (Category~2) within the same competition.

For Category~1, the top-performing teams continued to rely on manually curated annotations generated through various human-in-the-loop strategies. PDMYR employed active learning with OCR-based instrument identification and tracking-assisted annotation using the DarkLabel tool, labeling over 220,000 frames. InspireLab adopted a semi-supervised approach with YOLOv10, iteratively expanding their labeled dataset to 19,016 images through manual correction of model predictions. MULTIS leveraged the EndoVis2017 dataset for pre-training and applied BoT-SORT tracking for sample mining. These results reinforce the observation from prior years that generating high-quality pseudo-labels or manually curated annotations remains the most effective strategy for this weakly supervised detection problem.

Notably, SJTUB introduced Segment Anything Model 2 (SAM2) for semi-automatic annotation, generating over 300,000 labeled frames. Despite this large annotated dataset, their final mAP of 0.2754 fell below the top teams, suggesting that annotation quality may matter more than quantity. SKJP's approach of training on synthesized images from EndoVis datasets yielded a low mAP of 0.031, highlighting a potential the domain gap between synthesized and real surgical scenes. SEU-MIA's vision-language approach using CLIP with DualCoOp and class activation maps achieved a similarly low mAP, suggesting that CAM-based localization remains insufficient for precise bounding box detection.

For Category~2, PDMYR's success stemmed from a robust regularization system that combined residual-based invalid frame removal, temporal resampling centered on positive samples, 3D-CutMix augmentation, GroupKFold cross-validation, and mode filtering for temporal smoothing. SurgOp demonstrated that a relatively simple single-stage MViT model with balanced resampling and majority vote smoothing could achieve competitive results. TeamBCU showed that an ensemble of lightweight image classifiers (ConvNeXt, RegNetY, ViT) with neighbour smoothing post-processing was effective even without explicit temporal modeling. These findings suggest that careful data balancing and temporal post-processing are at least as important as architectural sophistication for surgical step recognition.

Comparing the Category~1 results with prior years, the top mAP of 0.4244 (PDMYR) is slightly below the 2023 first-place result of 0.4669 (PUMCH). However, these comparisons must be interpreted cautiously since each year's test set differs. More importantly, the overall landscape of approaches has shifted: whereas earlier years saw more diverse architectural explorations (e.g., class activation maps, attention-based methods), the 2024 teams converged on YOLO-family detectors paired with increasingly sophisticated data curation pipelines. 

As in prior years, we note that while the teams are employing state-of-the-art classifier algorithms that perform well on natural and conventional image data sets, they failed to exhibit similar performance on these surgical videos. For these and many similar reasons, we are committed to encourage the ML community in future challenges to tackle the difficult problems encountered in surgical data science.
 
\newpage
\section{Results and methods from the MICCAI 2025 SurgVU challenge}

\subsection{SurgVU 2025 Challenge Description}

\subsubsection{Overview}
\begin{figure}[tbh]
 \begin{center}
\resizebox{5in}{!}{\includegraphics{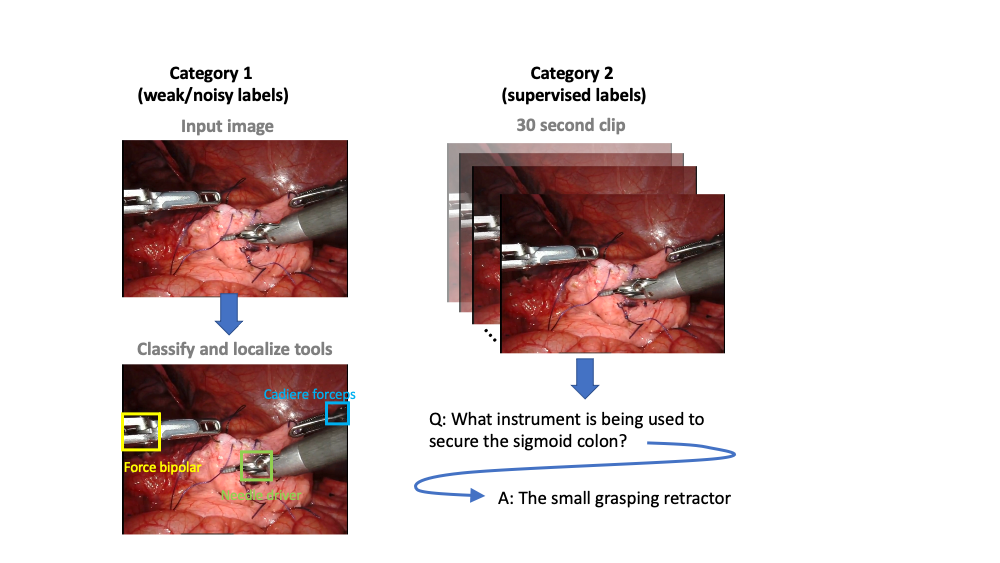}}
 \caption{Overview of challenge categories 1 and 2 (2025)}
  \label{fig:overview_categories_2025}
\end{center}
\end{figure}

The SurgVU 2025 sub-challenge, arranged as part of the Endoscopic Vision Challenge at MICCAI 2025, continues the multi-year challenge of tool classification and localization, but extends the tasks to include a new multimodal reasoning component (Figure~\ref{fig:overview_categories_2025}). 
Category 1 remains consistent with the 2022, 2023 and 2024 challenges, focusing on \textbf{surgical tool classification and localization} under a weakly supervised setting. However, this year with the addition of limited bounding box labels provided in a small validation set. 
Category 2 introduces a new task of \textbf{surgical visual question answering (VQA)}, requiring models to generate answers to open ended questions based on 30-second video clips. Training labels are tool presence labels, surgical steps and a description of the surgical step categories.

\subsubsection{Training Data}

The 2025 challenge used the same expanded dataset as the 2024 challenge: 280 long videos from 155 training sessions, captured at 60~fps with 720p resolution, totaling over 840 hours and 18 million frames (see Section~\ref{chapter_dataoverview}). Tool presence labels and surgical task labels were provided in the same format as in 2024.

For Category~1, training labels consisted of noisy tool presence labels and, as a new addition, a small validation set with ground-truth bounding box annotations. This validation set allowed teams to develop semi-supervised approaches leveraging a limited amount of strong supervision alongside the large-scale weakly labeled data.

For Category~2, the dataset labels were expanded to include detailed natural-language descriptions of each surgical task (found in the ``matched\_description'' column of each CSV file). These descriptions specified the anatomical elements involved, the surgeon's actions, and the tools in use, enabling teams to train vision-language models for the question answering task. A set of 10 sample videos with question-answer pairs was provided in the same format used for evaluation.

\subsubsection{Testing Data}

The testing dataset consisted of surgical training videos captured under the same conditions as the training set, sub-sampled to 1~FPS for inference. For Category~1, the test set was annotated with bounding boxes following the same protocol as prior years. For Category~2, each 30-second video clip in the test set was paired with open-ended questions, and five different ground-truth answers were provided per question for BLEU-based evaluation. As in prior years, the UI region was blurred in all test frames.

\subsubsection{Submission Process}

The submission process followed the same Type~2 challenge format on the Grand Challenge platform as in prior years; see Section~3.1 for details. For Category~1, the output format was identical to 2024 (bounding box predictions in JSON). For Category~2, the algorithm was required to generate a single predicted answer per question in JSON format. All teams were required to submit a final report alongside their algorithm to be considered a complete submission.

\subsection{Team Submissions}\label{team_submissions_25}

For the 2025 challenge, a total of 87 teams registered. In the preliminary phase, 96 submissions were received for Category~1 and 104 submissions for Category~2. In the final testing phase, 31 submissions were received for Category~1 and 36 submissions for Category~2. After filtering for teams that submitted both final results and a report, 6 complete submissions remained for Category~1 and 8 for Category~2. Three teams (Medibot, SKJP, and Algoritmi) participated in both categories. Table~\ref{table:TeamAffils2025} shows the participating teams with complete submissions, while Table~\ref{table:TeamMethodsSummary2025} summarizes the methodologies employed by each team. Team methodological details follow below. Note that these sub-sections were written by the participating teams.

\begin{table}[h!]
\centering
\small
\caption{Team affiliations and challenge categories -- 2025}
\begin{tabular}{ccp{5cm}ccc}
\toprule
\textbf{Team \#} & \textbf{Team name} & \textbf{Institution} & \textbf{Country} & \textbf{Category} & \textbf{Report} \\ \midrule
1 & PUH-SVU & Peking Union Medical, Hikvision, Beijing United Family Hospital & China & 1 only & Y \\
2 & Medibot & Shanghai Microport MedBot & China & 1 \& 2 & Y \\
3 & PDMYR & Southern Medical U. & China & 1 only & Y \\
4 & SK & Muroran Inst. of Tech. & Japan & 1 \& 2 & Y \\
5 & Algoritmi & Universidade do Minho & Portugal & 1 \& 2 & Y \\
6 & SurgTroopers & National Tsing Hua U. & Taiwan & 1 only & Y \\
7 & Capybara & Aillis, Inc. & Japan & 2 only & Y \\
8 & UoM-SurgicalAI & U. of Manchester, UCL & UK & 2 only & Y \\
9 & AMI & Kyung Hee U. & South Korea & 2 only & Y \\
10 & UT & U. of Tokyo & Japan & 2 only & Y \\
11 & gardenia & Hefei U. of Technology & China & 2 only & Y \\
\bottomrule
\end{tabular}
\label{table:TeamAffils2025}
\end{table}

\begin{sidewaystable*}
\caption{Summary of methodologies -- 2025}
\begin{adjustbox}{scale=0.55,center}
{\begin{tabular}{ p{9em} p{12em} p{8em} p{7em} p{7em} p{6em} p{8em} p{8em} p{8em} }
Team Name & Architecture & Backbone & Data Preprocessing & Pretrain & Image Augmentation & Use Additional Data & Loss & Output \\
\hline
PUH-SVU & RTMDet ensemble & CSPNeXt & MAE self-supervised pre-training, pseudo-labels via Co-DETR & MAE, Co-DETR (ViT-L, Swin-L) & N/A & OSTrack pseudo-labels, dense sampling for tail classes & Standard detection loss & Tool bounding box \\
Medibot (C1) & YOLOv8l + ResNet-50 (two-stage) & CSPDarknet, ResNet-50 & Pseudo-label generation & COCO, ImageNet & ThreeAug & EndoVis2017, validation set & Standard YOLO + cross-entropy & Tool bounding box \\
Medibot (C2) & Qwen2.5-VL-3B & Qwen2.5-VL & Video segmentation, text formatting & Qwen2.5-VL & N/A & Qwen3-8B-generated enriched descriptions & Cross-entropy (LoRA) & VQA answer \\
PDMYR & YOLOv5 ensemble (WBF) & CSPDarknet & HITL active learning, OCR-based sampling & ImageNet, CLIP & Mosaic, Mixup, MotionBlur & 220k manually labeled frames & Standard YOLO loss & Tool bounding box \\
SK (C1) & YOLOv10 & CSPDarknet & Pseudo-labels from few annotations & N/A & N/A & Few manually annotated frames & Standard YOLO loss & Tool bounding box \\
SK (C2) & InternVL3.5-1B & InternViT + Qwen & LLM-generated QA pairs & InternVL3.5 & N/A & LLM-generated QA pairs & Cross-entropy (adapter) & VQA answer \\
Algoritmi (C1) & YOLO ensemble (YOLOv8, YOLOe) & CSPDarknet & Validation set + few-shot fine-tuning & COCO & Flip, rotate, jitter & Small manually annotated set & Standard YOLO loss & Tool bounding box \\
Algoritmi (C2) & Flan-T5-base (RAG) & Flan-T5-base & Context retrieval from tool detections & Flan-T5 & N/A & Tool context documents & Cross-entropy & VQA answer \\
SurgTroopers & YOLOv8s & CSPDarknet & Border masking, balanced frame selection & COCO & Flip, shear, scale, mosaic & 4.8k manually labeled frames & Standard YOLO loss & Tool bounding box \\
Capybara & LLaVA-OneVision-7B + EfficientNetV2 classifiers & EfficientNetV2-S, LLaVA & Tool/organ classification, margin removal & SurgToolLoc 2022 (tool classifier) & N/A & SurgToolLoc 2022 dataset & N/A (zero-shot VLM) & VQA answer \\
UoM-SurgicalAI & InternVL3 / InternVL3.5 & InternViT + Qwen & Template QA generation, clip sampling & InternVL3 / 3.5 & Gaussian blur on UI & 8.8k template QA pairs & Zero-shot (best); DoRA, DCT-GaLore (explored) & VQA answer \\
AMI & EndoViT + Vicuna-7B & EndoViT, Vicuna & Tool-anchored task refinement, balanced QA & EndoViT & N/A & GPT-5 generated QA variants, 1.2M examples & Cross-entropy (LoRA) & VQA answer \\
UT & InternVL3-2B & InternViT + Qwen & Rule-based QA generation & InternVL3-2B & N/A & 70k synthetic QA pairs & Cross-entropy (LoRA) & VQA answer \\
gardenia & InternVL2.5-MoP-4B + MedViT & MedViT, Qwen & 30s clip segmentation, quality filtering & SurgeNetSmall, InternVL2.5 & N/A & 5.9k manually verified QA pairs & Cross-entropy & VQA answer \\
\hline
\end{tabular}}
\end{adjustbox}
\label{table:TeamMethodsSummary2025}
\end{sidewaystable*}

\clearpage

\subsection{PUH-SVU}
Based on past clinical and surgical video-processing experience, we noticed that the frequency and distribution of different surgical tools vary largely in a certain dataset. In the current challenge, we adopt a strategy to train multiple models for different data distributions, and then merge the advantageous categories of each model to form an output. Specifically, we use uniform sampling on the training video data to obtain Dataset 1. Then, we conducted dense sampling on the low-frequency, long-tail categories based on official existence labels, and obtained Dataset 2. We trained Model 1 and Model 2 separately using Dataset 1 and Dataset 2, and merged their respective advantageous categories for output. Results in this competition have shown that adopting this strategy can effectively improve the performance of the model. The strategy holds similar properties as previous works including BBN and Survey~\cite{PUHSVU_Zhou2020BBN,PUHSVU_Zhang2023Survey}.

\subsubsection{Method Description}
\begin{figure}[tbh]
\begin{center}
\resizebox{5in}{!}{\includegraphics{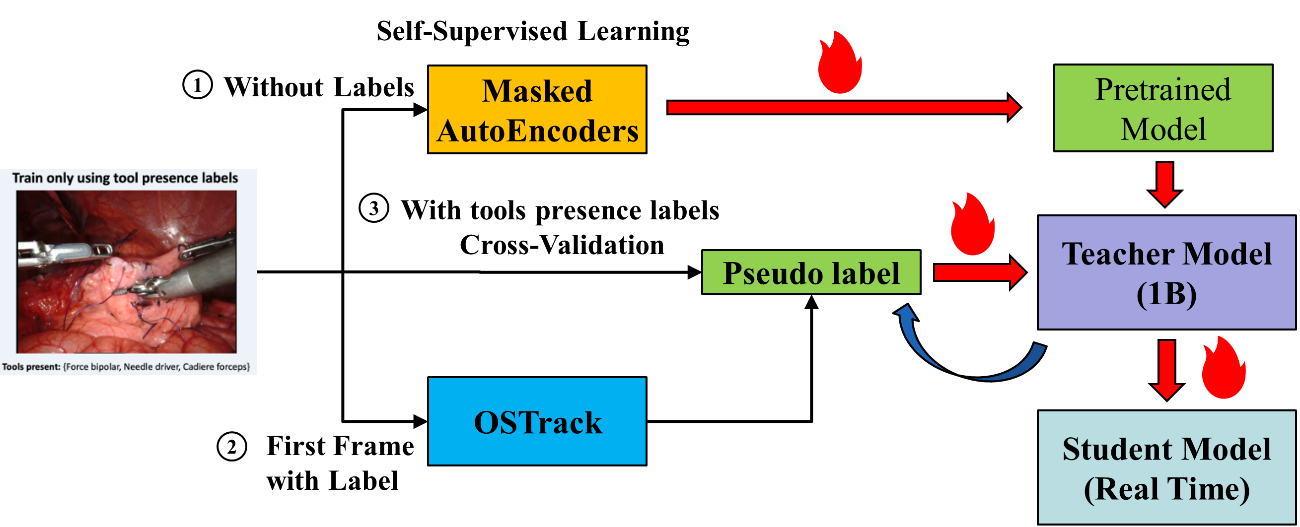}}
\caption{Schematic diagram of the pseudo-label strategy}
\label{fig:C1_PUH-SVU_pseudolabel}
\end{center}
\end{figure}

For Dataset 1, we adopted a pseudo-label strategy as is shown in Figure~\ref{fig:C1_PUH-SVU_pseudolabel}.

Our strategy is composed of 3 steps:

\textbf{Step 1:} Based on uniformly sampled unlabeled data (200W+), a ViT-L pretrained model is trained using the MAE scheme~\cite{PUHSVU_He2021MAE,PUHSVU_Dosovitskiy2020ViT}.

\textbf{Step 2:} A small number of manually annotated labels (5000+) are applied as OSTrack tracking anchors to generate pseudo-labels. Tool existence labels and expert rules are used to verify and filter the pseudo-labels. A training set of pseudo-labels (10W) is generated~\cite{PUHSVU_Ye2022OSTrack}.

\textbf{Step 3:} The pseudo-labeled training set is used to train a teacher model (Co-DETR)~\cite{PUHSVU_Zong2023CoDETR}. The prediction results of the teacher model on the training set are filtered based on expert rules, iteratively obtaining a pseudo-labeled training set with good consistency.

The pseudo-labeled training set generated is used to train student models for online evaluation. For both uniform and dense sampling training sets, the process of data pseudo-labeling is the same.

The initial performances of the models are shown in Table~\ref{tab:C1_PUH-SVU_offline} and Table~\ref{tab:C1_PUH-SVU_official}.

\begin{table}[htbp]
\centering
\caption{Performances of models on offline validation sets.}
\label{tab:C1_PUH-SVU_offline}
\begin{tabular}{llccc}
\hline
Model & BackBone & Param & mAP (IoU=0.5:0.95) & mAP (IoU=0.5) \\
\hline
RTMDet & CSPNeXt & 95M & 0.703 & 0.902 \\
Co-DETR & Swin-L & 0.3B & 0.718 & 0.928 \\
Co-DETR & ViT-L & 1B & 0.723 & 0.92 \\
\hline
\end{tabular}
\end{table}

\begin{table}[htbp]
\centering
\caption{Performances of models on official validation sets.}
\label{tab:C1_PUH-SVU_official}
\begin{tabular}{llccc}
\hline
Model & BackBone & Param & mAP (IoU=0.5:0.95) & mAP (IoU=0.5) \\
\hline
RTMDet & CSPNeXt & 95M & 0.564 & 0.918 \\
Co-DETR & Swin-L & 0.3B & 0.573 & 0.935 \\
Co-DETR & ViT-L & 1B & 0.581 & 0.94 \\
\hline
\end{tabular}
\end{table}

\begin{figure}[tbh]
\begin{center}
\resizebox{5in}{!}{\includegraphics{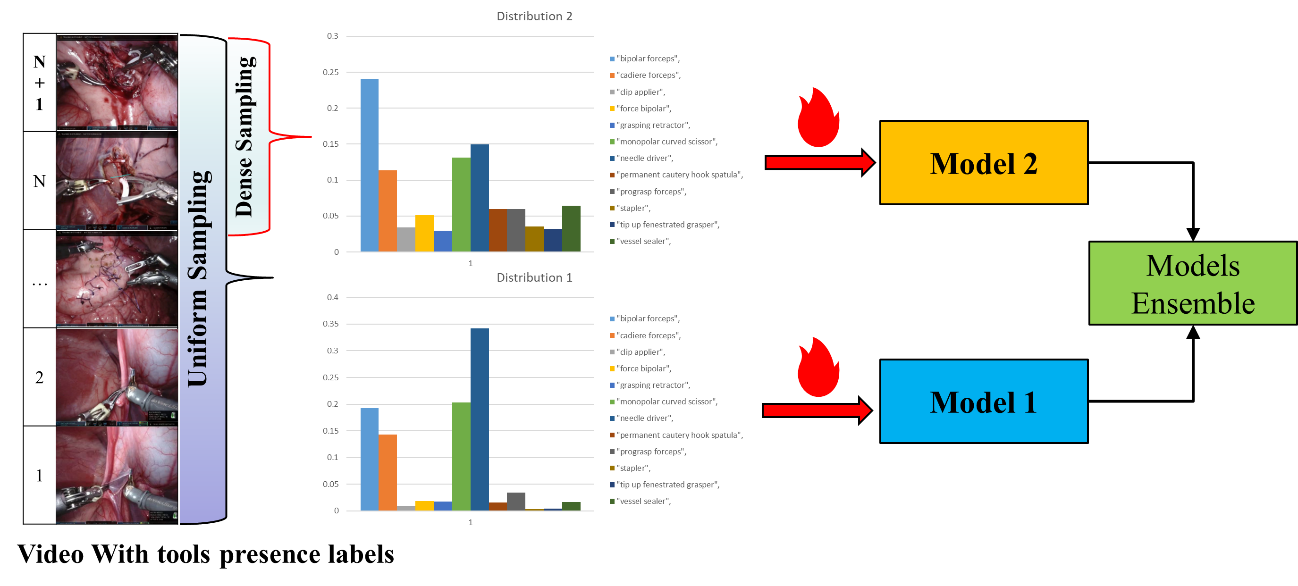}}
\caption{Dense sampling and ensemble of models}
\label{fig:C1_PUH-SVU_ensemble}
\end{center}
\end{figure}

We noticed that using a uniform sampling scheme resulted in a significant long tail phenomenon in the distribution of several kinds of targets. The optimization gradient of the model was dominated by several head categories, leading to poor performance of the long tail categories. We referred to the methods mentioned in BBN and Survey to sample and train models for the videos separately, as shown in Figure~\ref{fig:C1_PUH-SVU_ensemble}~\cite{PUHSVU_Zhou2020BBN,PUHSVU_Zhang2023Survey}.

We first performed uniform sampling on the officially released data to obtain a uniform sampling training Dataset 1 (10W+). In addition, we conducted targeted dense sampling on the tail categories based on the official device existence labels, and obtained a dense sampling training Dataset 2 (10W+) with a similar size compared to the uniform sampling training set, with a smaller difference in the number of targets between the head and tail. We trained RTMDet separately on training Dataset 1 and Dataset 2 to obtain Model 1 and Model 2~\cite{PUHSVU_Lyu2022RTMDet}. Finally, we analyzed the performance differences between each category of Model 1 and Model 2 in our own offline validation set 1 and the officially published offline validation set 2, and adopted the principle of optimal performance, using the advantageous categories of each model as the output categories in the final integrated model. Results of the preliminary round for evaluation are shown in Table~\ref{tab:C1_PUH-SVU_prelim}.

\begin{table}[htbp]
\centering
\caption{Performances of models on the preliminary round of evaluation.}
\label{tab:C1_PUH-SVU_prelim}
\begin{tabular}{clllll}
\hline
Index & Model & Training Set & Annotations & Training Strategies & mAP (IoU=0.5:0.95) \\
\hline
1 & RTMDet & Surgvu25& OSTrack Pseudo Label & Single Model & 0.5013 \\
2 & RTMDet & Surgvu25& Co-DETR Pseudo Label & Single Model & 0.5128 \\
3 & RTMDet & Surgvu25& Co-DETR Pseudo Label & Models Ensemble & 0.5223 \\
\hline
\end{tabular}
\end{table}

\subsubsection{Model Training}
After the preliminary test video was made available for download, we downloaded it and pseudo-labeled it using model index number 3 in Table~\ref{tab:C1_PUH-SVU_prelim}. We filtered the pseudo-labels using a confidence threshold of 0.5 to obtain preliminary test set 1. We included the preliminary test set 1 and the official validation set in the training set to perform a data-expanding trial. The performances of the trained models on the preliminary test set are shown in Table~\ref{tab:C1_PUH-SVU_expansion}.

\begin{table}[htbp]
\centering
\caption{Performances of models on the data-expanding trial.}
\label{tab:C1_PUH-SVU_expansion}
\begin{tabular}{clp{3.5cm}p{3cm}p{2.5cm}c}
\hline
Index & Model & Training Set & Annotations & Training Strategies & mAP (IoU=0.5:0.95) \\
\hline
1 & RTMDet & Surgvu25 & Co-DETR Pseudo Label & Models Ensemble & 0.5223 \\
2 & RTMDet & Surgvu25 + Prelim Phase & Co-DETR Pseudo Label & Models Ensemble & 0.5347 \\
3 & RTMDet & Surgvu25 + Prelim Phase + Official Validation & Co-DETR Pseudo Label & Models Ensemble & 0.5407 \\
\hline
\end{tabular}
\end{table}

We observed that expansion of training data was beneficial for the performance of the model.

\subsubsection{Final Performance}
Index number 3 in Table~\ref{tab:C1_PUH-SVU_expansion} was submitted to the final phase and a mAP of 0.5188 was achieved (Table~\ref{tab:C1_PUH-SVU_final}).

\begin{table}[htbp]
\centering
\caption{Performances in the final phase.}
\label{tab:C1_PUH-SVU_final}
\begin{tabular}{clp{3.5cm}p{3cm}p{2.5cm}c}
\hline
Index & Model & Training Set & Annotations & Training Strategies & mAP (IoU=0.5:0.95) \\
\hline
1 & RTMDet & Surgvu25 & Co-DETR Pseudo Label & Models Ensemble & 0.4865 \\
2 & RTMDet & Surgvu25 + Prelim Phase + Official Validation & Co-DETR Pseudo Label & Models Ensemble & 0.5188 \\
\hline
\end{tabular}
\end{table}

In this challenge, we validated many proposals, but most of them were ineffective. Due to space limitations, we are unable to list them all. The content of the above chapters is the validated effective solutions during our competition process. In summary, the performance improvement benefits from: 1. Pre-labeling of teacher model; 2. Multiple distribution sampling scheme; 3. Ensembled output of multiple models; 4. Data-expansion joining the official validation set and preliminary test set.

Still, the performance of our model can be further optimized. We humbly suggest that potential noises in the official test set (possibly due to multiple labelers and inconsistency within different categories) may also be improved in future challenges.

\FloatBarrier

\subsection{Medibot}

Our approach is motivated by the need for a solution that is accurate, real-time, and generalizable across diverse surgical environments, such as in vitro, in vivo and in animals. We participated in both Category 1 (surgical tool classification and localization) and Category 2 (surgical visual question answering).

For Category 1, we propose a two-stage surgical tool localization and classification method combined with an efficient and accurate semi-automatic data annotation method. For Category 2, we leverage large-scale multimodal models fine-tuned on surgical videos, adapting them to the domain's unique characteristics. Our approach reduces annotation overhead while achieving accurate and robust surgical visual understanding.

\subsubsection{Category 1: Method Description}

According to our experience, we proposed a two-stage method to do the tool location and classification respectively. We use the YOLOv8l model~\cite{Medibot_Reis2023} for the tool location because of its real-time performance and maturity. The location model only distinguishes whether it is a tool, but does not determine the category. This can improve the recall of the model. Then, the ResNet50 model~\cite{Medibot_He2015} is used to classify the location boxes into 12 categories.

For the surgical tools location and classification task, the accurate labeling of surgical tools in every frame is very time-consuming but important for the model training. To generate the high quality labels, we trained a location and classification model with YOLOv8l using the Category 1 validation set. Then we clipped out the segment from the videos according to the stop and end time in the task CSV file. We got the first frame of these clips only to create a dataset. We did the prediction on the dataset to obtain the pseudo labels. We manually modified the incorrect categories of these pseudo labels to the correct ones. We tried not to modify the box unless there was an obvious error. After counting the number of boxes in each category of all data, we found that some categories had too few boxes. We supplemented some data on tools with fewer categories in the tool CSV file and different backgrounds with the same way. Then, the training and validation dataset was formed by splitting the data by 8:2 portion. In order to increase the amount of data in the training set, the open source dataset EndoVis17 and the Category 1 validation set were all added in the training set. Finally, we obtained 10,997 images in the training set and 979 images in the validation set. The box count of every category is listed in Table~\ref{tab:C1_C2_Medibot_boxcount}.

\begin{table}[htbp]
\centering
\caption{The box count of every category in training set and validation set}
\label{tab:C1_C2_Medibot_boxcount}
\begin{tabular}{lcc}
\hline
Category & Trainset Boxes count & Valset Boxes count \\
\hline
1 & 6696 & 1110 \\
2 & 4074 & 258 \\
3 & 949 & 25 \\
4 & 674 & 94 \\
5 & 3085 & 258 \\
6 & 4486 & 266 \\
7 & 563 & 36 \\
8 & 502 & 15 \\
9 & 1024 & 68 \\
10 & 699 & 30 \\
11 & 568 & 19 \\
12 & 294 & 1 \\
\hline
\end{tabular}
\end{table}

\subsubsection{Category 1: Model Training}

We found that compared with the YOLOv8l 12-class detection model, the YOLOv8l single-class detection model performed better. So we selected the YOLOv8l single-class model as the detection model. We evaluated the detection performance on the validation set. We trained the model initialized with pretrained YOLOv8l weights for 1000 epochs with early stopping strategy. The image size was 640. The batch size was 300 with 6 A100 GPUs. The comparison between one-class and twelve-class YOLOv8l results on the validation set is shown in Table~\ref{tab:C1_C2_Medibot_yolo_comparison}.

\begin{table}[htbp]
\centering
\caption{The one-class vs twelve-class YOLOv8l result on validation set}
\label{tab:C1_C2_Medibot_yolo_comparison}
\begin{tabular}{lcc}
\hline
 & One-class & Twelve-class \\
\hline
AP@[IoU=0.50:0.95] & 0.490 & 0.435 \\
AP@[IoU=0.50] & 0.776 & 0.708 \\
AR@[IoU=0.50:0.95] & 0.558 & 0.497 \\
\hline
\end{tabular}
\end{table}

For all categories, we cropped regions using bounding boxes scaled by a factor of two relative to the original size. We initialized the model with pretrained ResNet-50 weights and then trained it on our dataset. We trained the model with an input size of 224 and a batch size of 600 for 400 epochs. In contrast to standard classification model training, we resized each cropped region directly to 224 × 224, without applying random scale cropping. We enabled model exponential moving average (EMA), mixed-precision training (AMP) and evaluation with EMA weights. In addition, the training process adopted ThreeAug data augmentation~\cite{Medibot_Touvron2022}. The classification model result is shown in Table~\ref{tab:C1_C2_Medibot_resnet_classification}.

\begin{table}[htbp]
\centering
\caption{The 12-class ResNet50 result on validation set}
\label{tab:C1_C2_Medibot_resnet_classification}
\begin{tabular}{lc}
\hline
Max accuracy 1 & EMA accuracy 1 \\
\hline
95.88\% & 93.27\% \\
\hline
\end{tabular}
\end{table}

\subsubsection{Category 1: Preliminary Performance}

\begin{figure}[tbh]
\begin{center}
\resizebox{5in}{!}{\includegraphics{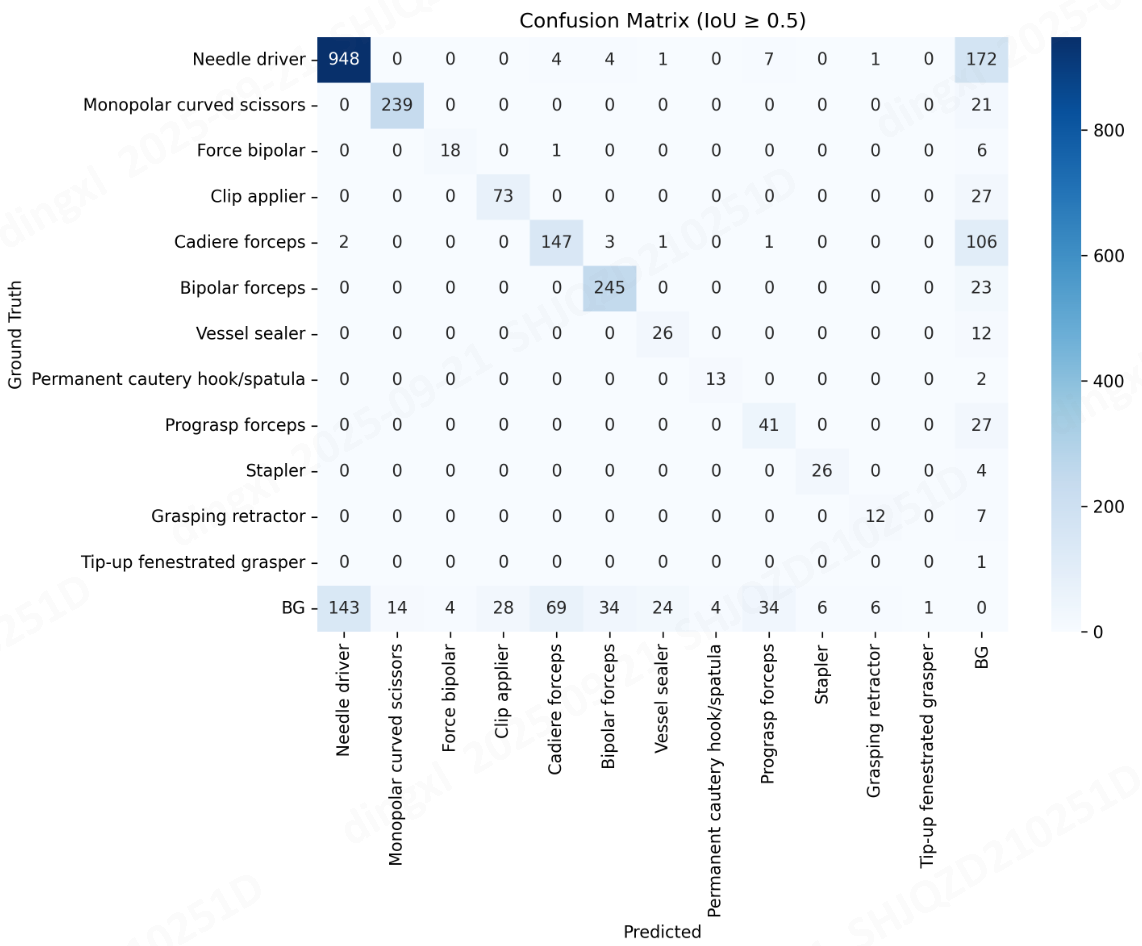}}
\caption{The confusion matrix of the classification result}
\label{fig:C1_C2_Medibot_confusion}
\end{center}
\end{figure}

Finally, we combined the results of the single-class detection model and the 12-class classification model for evaluation. The final evaluation results are listed in Table~\ref{tab:C1_C2_Medibot_final_results}. The confusion matrix of the classification result is shown in Figure~\ref{fig:C1_C2_Medibot_confusion}.

\begin{table}[htbp]
\centering
\caption{The 12-class location and classification mean mAP on validation set}
\label{tab:C1_C2_Medibot_final_results}
\begin{tabular}{lcc}
\hline
Category & AP@[IoU=0.50:0.95] & AP@[IoU=0.50] \\
\hline
1: Needle driver & 0.491 & 0.796 \\
2: Monopolar curved scissors & 0.665 & 0.905 \\
3: Force bipolar & 0.315 & 0.697 \\
4: Clip applier & 0.297 & 0.644 \\
5: Cadiere forceps & 0.248 & 0.489 \\
6: Bipolar forceps & 0.680 & 0.893 \\
7: Vessel sealer & 0.220 & 0.546 \\
8: Permanent cautery hook/spatula & 0.432 & 0.765 \\
9: Prograsp forceps & 0.167 & 0.460 \\
10: Stapler & 0.646 & 0.855 \\
11: Grasping retractor & 0.324 & 0.528 \\
12: Tip-up fenestrated grasper & 0.000 & - \\
\hline
Average & 0.374 & 0.632 \\
\hline
\end{tabular}
\end{table}

As the number of training data increased from 3,543 to 10,016, the indicators of the preliminary stage also continued to rise. The result of the submissions are shown in Table~\ref{tab:C1_C2_Medibot_submission_results}.

\begin{table}[htbp]
\centering
\caption{Submission result of different training data in different phases}
\label{tab:C1_C2_Medibot_submission_results}
\begin{tabular}{lcccc}
\hline
 & \multicolumn{3}{c}{Preliminary phase} & Final phase \\
\hline
Image Number & 2628 & 4845 & 10016 & 10976 \\
Mean mAP & 0.4587 & 0.6073 & 0.7124 & 0.5018 \\
\hline
\end{tabular}
\end{table}

The two-stage framework demonstrates strong performance in both detection and classification. Compared with the multi-class detector, the single-class detection model achieves superior accuracy. Moreover, a limited set of high-quality annotations proves to be more valuable for model training than a large quantity of noisy pseudo-labeled data. Within a certain range, increasing the amount of accurately labeled data consistently leads to improved model performance.

\subsubsection{Category 2: Method Description}

We used only the official challenge dataset for model training and evaluation. No private datasets or external annotations were incorporated. However, the official dataset is inherently multi-dimensional. To enable the model to effectively learn diverse visual-language information, we applied the following preprocessing steps:

\textbf{Video segmentation:} We first extracted only the portions of the original videos corresponding to the annotated task intervals and further divided them into 30-second clips.

\textbf{Text formatting:} The accompanying textual information was standardized and organized into a structured JSON format, with different dimensions of information explicitly represented. In addition to the fixed fields of tool names, task name, task description, and matched description, we further enriched the dataset with two additional dimensions: tool functions and operated organs and tissues. These fields were automatically generated from the matched description using the Qwen3-8B model~\cite{Medibot_Yang2025}.

\textbf{Video-text pairing:} Each video clip was paired with its corresponding JSON entry, along with a prompt.

In total, the preprocessing resulted in 162,000 video-question-answer pairs, of which 158,892 were used for training. The remaining 1,554 and 1,554 samples were allocated for validation and testing, respectively.

Our objective was to design a multimodal model for surgical visual question answering. Given the hardware constraint of a 16 GB GPU memory, we selected Qwen2.5-VL-3B~\cite{Medibot_Bai2025} as the base multimodal model, owing to its strong performance on a wide range of vision-language benchmarks. To adapt it to the surgical domain, we finetuned the model on our constructed dataset using two strategies: full-parameter fine-tuning and LoRA~\cite{Medibot_Hu2021} (Low-Rank Adaptation).

To further improve both the quality and consistency of answers, we adopted a two-stage answering strategy:

\textbf{Stage 1 (Scene Description):} The model is first prompted to generate a structured description of the overall surgical scene, including all information in the JSON. Crucially, instrument information is not inferred solely by the vision-language model. Due to significant noise in the training dataset, particularly in tool annotations, the model is prone to hallucinations when identifying instruments. To mitigate this, we incorporated a dedicated instrument detection model, developed as part of our team's submission to Category 1: Surgical Tool Classification and Localization, which serves as the authoritative source for instrument presence. The resulting merged description thus combines the contextual understanding from the vision-language model with accurate and reliable tool identification from the detection model.

\textbf{Stage 2 (Question-Specific Answering):} The scene description generated in Stage 1, together with the user's question, is passed back into the model. At this stage, the model produces a concise and question-focused response. The style and format of the answer are further controlled through carefully designed prompt templates.

This two-stage pipeline enables the model to first establish a holistic representation of the surgical scene, and then refine its reasoning to deliver accurate, coherent, and stylistically consistent answers.

\subsubsection{Category 2: Model Training}

We compared LoRA and full-parameter fine-tuning using three evaluation metrics: task recognition accuracy, as well as precision and recall for tools recognition. Since the test set is large and contained considerable noise, we manually curated a subset of 25 samples to ensure data quality. The results are shown in Table~\ref{tab:C1_C2_Medibot_c2_lora_comparison}.

\begin{table}[htbp]
\centering
\caption{Comparison of LoRA and full-parameter fine-tuning on Category 2}
\label{tab:C1_C2_Medibot_c2_lora_comparison}
\begin{tabular}{lcc}
\hline
Metric & LoRA & Full para \\
\hline
Task recognition accuracy & 0.96 & 0.72 \\
Tools detect precision & 0.69 & 0.63 \\
Tools detect recall & 0.91 & 0.77 \\
\hline
\end{tabular}
\end{table}

\subsubsection{Category 2: Preliminary Performance}

Table~\ref{tab:C1_C2_Medibot_c2_detection_impact} indicates whether the detection model was incorporated, allowing us to assess its impact on overall model performance.

\begin{table}[htbp]
\centering
\caption{Impact of detection model on Category 2 performance}
\label{tab:C1_C2_Medibot_c2_detection_impact}
\begin{tabular}{lcc}
\hline
Metric & Qwen2.5 VL & Qwen2.5 VL + detection \\
\hline
Task recognition accuracy & 0.96 & 0.96 \\
Tools detect precision & 0.69 & 0.95 \\
Tools detect recall & 0.91 & 0.87 \\
\hline
\end{tabular}
\end{table}

Additionally, based on the 11 official test samples, we created 34 additional similar examples to evaluate Stage 2. By carefully adjusting the prompts, our local evaluation on a total of 45 test samples achieved a BLEU score of 0.73.

Our submission explored fine-tuning strategies and multimodal model augmentation for surgical video question answering. Key takeaways include: (1) LoRA adaptation yielded better accuracy and stability, suggesting that smaller, efficient adaptation methods are preferable when annotated data is limited; (2) Adding a dedicated detection module improved tool recognition precision, while vision-language models are versatile, classical detection networks still excel in specialized recognition tasks; (3) Current training data contain considerable noise, such as mismatches between instrument names and video content. Future work could focus on more thorough data cleaning and leverage larger models to further improve performance; (4) The BLEU metric alone may not fully reflect model performance. Future evaluations should explore alternative assessment approaches, such as multiple-choice or open-ended question formats, to better capture the quality and correctness of model responses.

\FloatBarrier

\subsection{PDMYR}

Our team participated in the Category 1 track, which involves a typical object detection task that requires extensive manual labeling. To minimize manual effort and enhance model performance, we adopted a Human-in-the-Loop (HITL) strategy. Initially, we actively selected a small batch of representative video frames for manual labeling and trained a first-stage model on this data. We then generated predictions on another batch of unlabeled data, reviewed the results, and manually labeled a minimal portion where the model performed poorly. This iterative process continued until we were satisfied with the model's predictive performance.

\subsubsection{Method Description}

The competition provided only information about the installation of surgical tools on the da Vinci robot, without object detection box labels. While weakly supervised object detection methods can predict object locations using image-level labels, they are not yet practical for real-world applications due to significant precision gaps compared to fully supervised methods. Therefore, we needed to manually label the objects using the tool installation information as guidance.

Lacking a professional data labeling team, we aimed to achieve high model performance with minimal labeling effort. We designed an active learning-based sample mining and interactive labeling strategy within the HITL framework, conducting multiple rounds of cyclic labeling on the dataset.

\begin{figure}[tbh]
\begin{center}
\resizebox{5in}{!}{\includegraphics{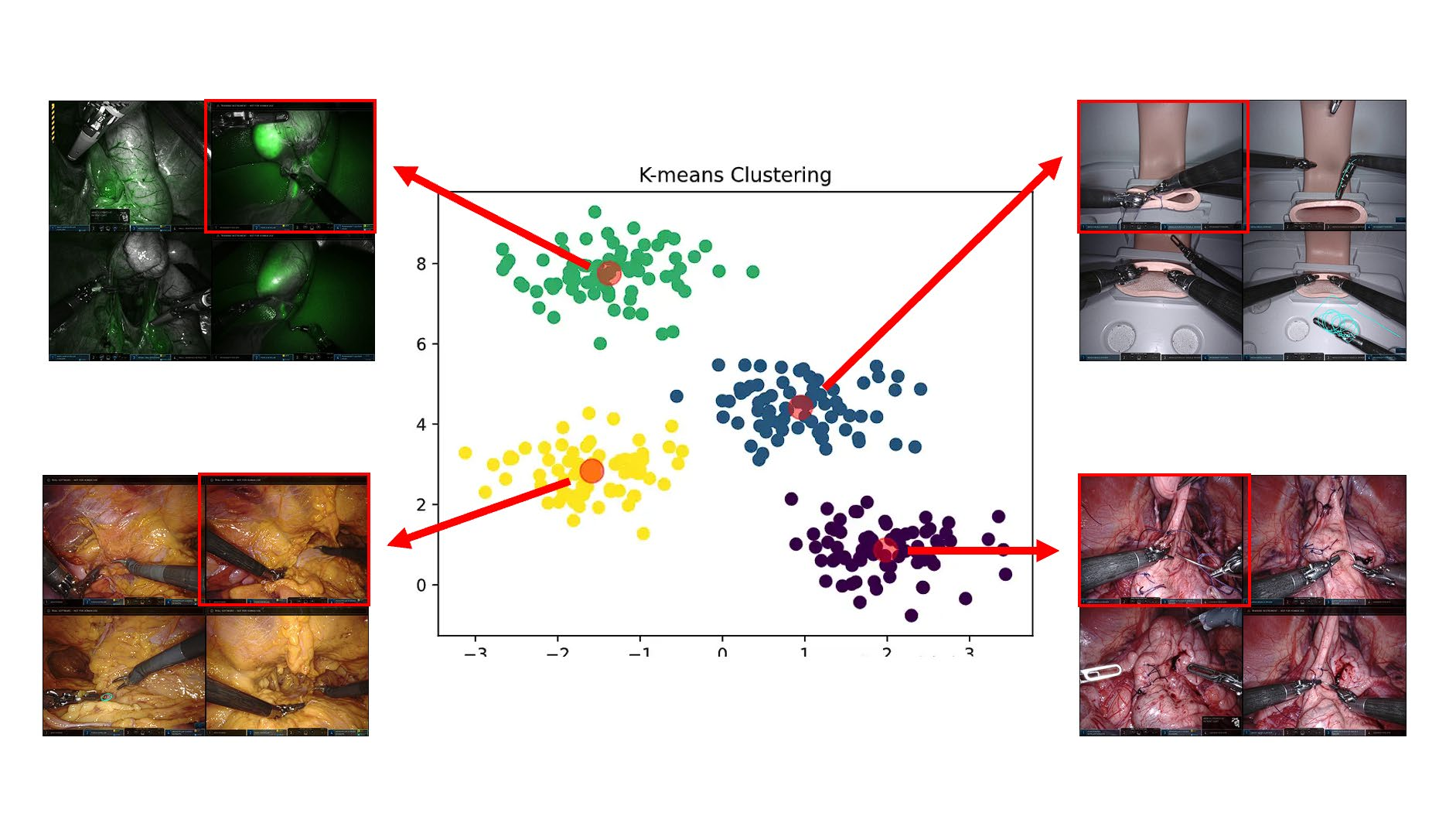}}
\caption{Feature-based clustering for initial sample selection. Images are clustered using pre-trained features, and representative samples closest to cluster centers are selected for manual labeling.}
\label{fig:C1_PDMYR_clustering}
\end{center}
\end{figure}

\paragraph{Feature-Based First-Round Sample Selection}
To obtain initial training data, we sampled the entire dataset at 1 frame per second (fps) and extracted features from all images using pre-trained backbone networks like ImageNet classifiers or CLIP. We performed K-Means clustering into 1,000 clusters and selected the images closest to each cluster center as representative samples for labeling (Figure~\ref{fig:C1_PDMYR_clustering}).

\begin{figure}[tbh]
\begin{center}
\resizebox{5in}{!}{\includegraphics{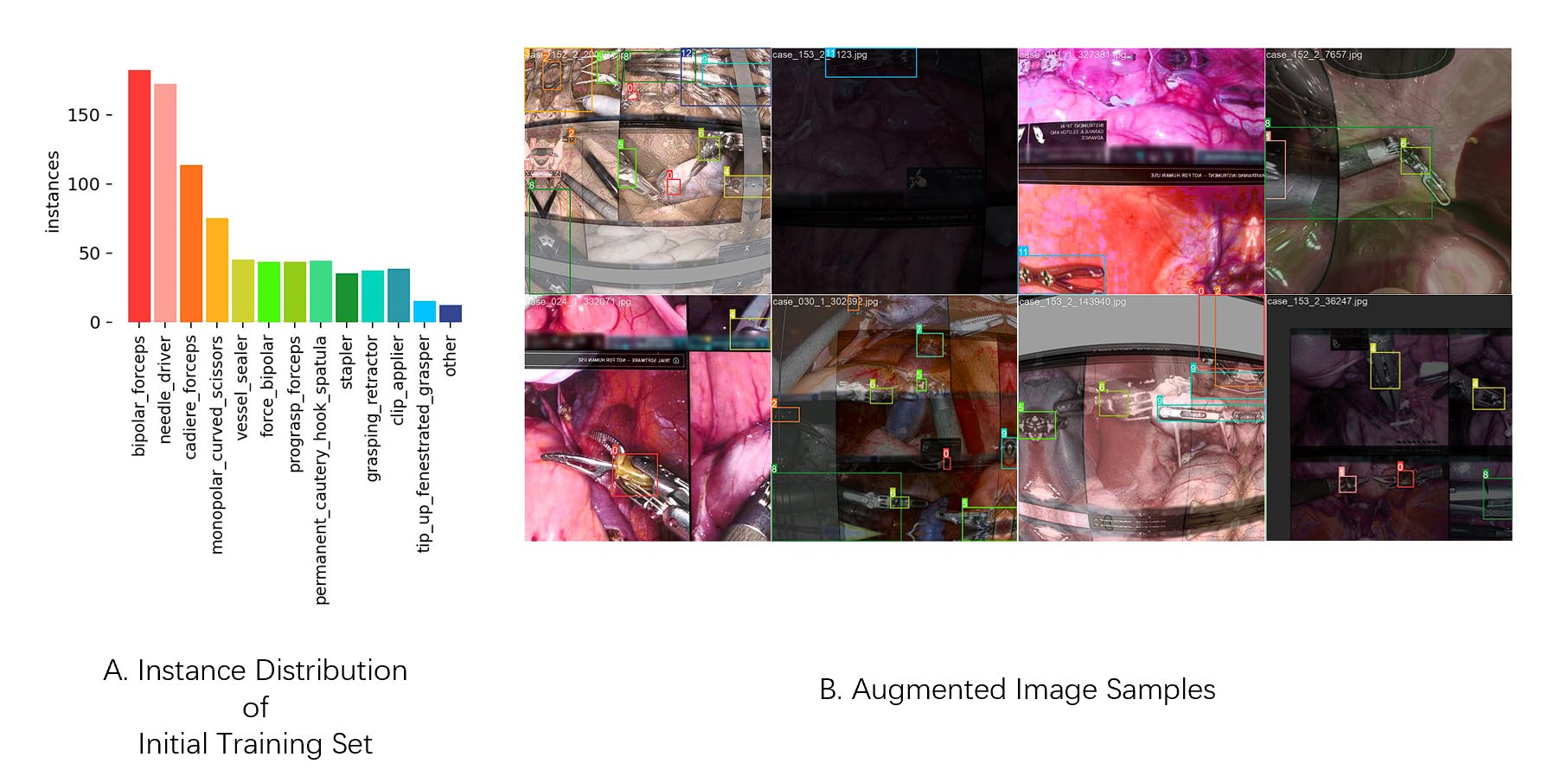}}
\caption{Data distribution analysis and augmentation strategies. (A) Distribution of high-frequency and low-frequency instruments in the dataset. (B) Data augmentation techniques including MotionBlur, OpticalDistortion, GridDistortion, Mosaic, and Mixup.}
\label{fig:C1_PDMYR_distribution}
\end{center}
\end{figure}

\paragraph{Active Mining of Difficult Samples}
After labeling these 1,000 images, we analyzed the data distribution and identified two major issues: (1) a significant imbalance between high-frequency and low-frequency instruments (Figure~\ref{fig:C1_PDMYR_distribution}A), and (2) numerous challenging scenarios where instruments were difficult to distinguish due to factors like fog, bleeding, dirty lenses, and exposure errors.

\begin{figure}[tbh]
\begin{center}
\resizebox{5in}{!}{\includegraphics{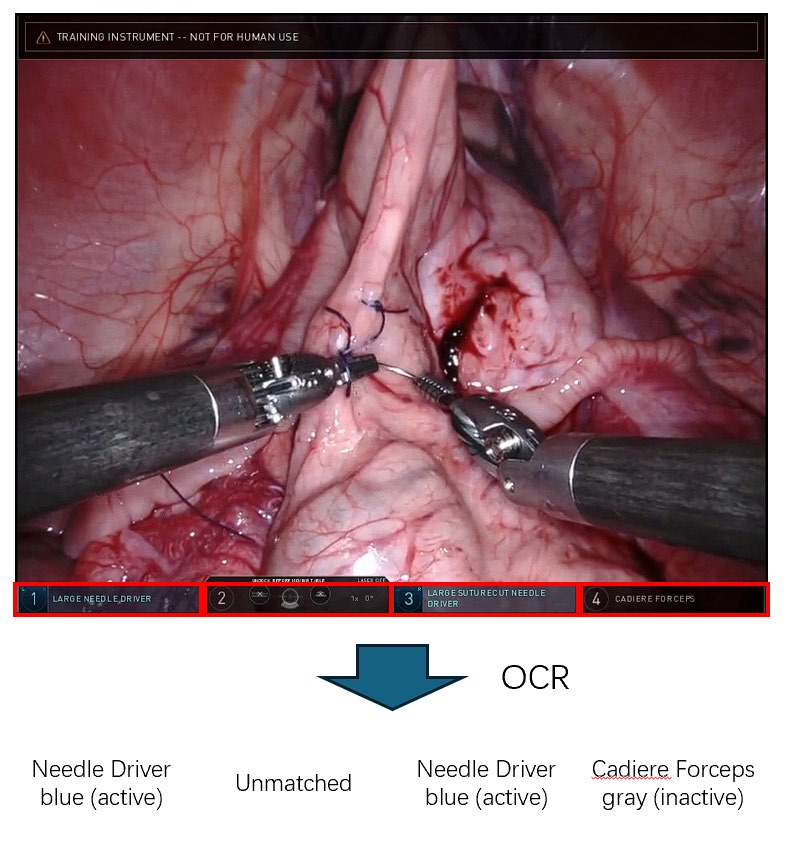}}
\caption{OCR-based instrument identification from da Vinci system GUI. The EasyOCR package is used to extract instrument information directly from the graphical user interface, enabling accurate identification of frames containing low-frequency instruments.}
\label{fig:C1_PDMYR_ocr}
\end{center}
\end{figure}

To address the low-frequency instrument issue, we focused on video segments where these instruments appeared. Meanwhile, high-frequency instruments also appeared in these segments, this approach increased the diversity of our labeled data. We attempted to use the official labels indicating instrument presence but found them unreliable due to duplicates and errors. Instead, we extracted instrument information from the graphical user interface (GUI) of the da Vinci system using optical character recognition (OCR) with the EasyOCR package, accurately identifying frames with low-frequency instruments (Figure~\ref{fig:C1_PDMYR_ocr}).

\begin{figure}[tbh]
\begin{center}
\resizebox{5in}{!}{\includegraphics{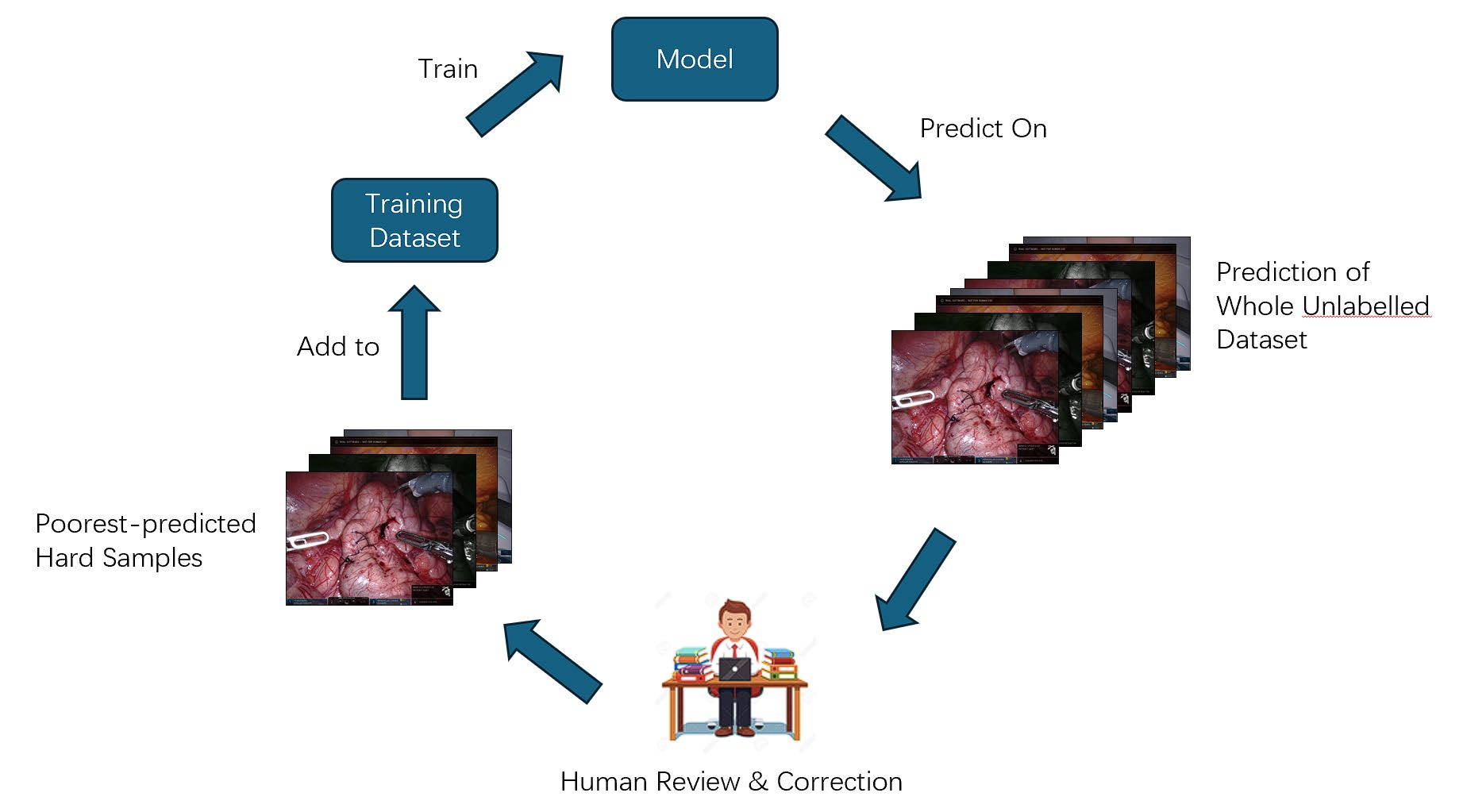}}
\caption{Active mining of difficult samples. Sequences with poor model predictions are identified and labeled to improve model robustness in challenging scenarios such as fog, bleeding, and lens defects.}
\label{fig:C1_PDMYR_difficult}
\end{center}
\end{figure}

For challenging scenes, we implemented two strategies: (1) Within the HITL process, we predicted the entire unlabeled dataset and selected sequences with the poorest predictions for labeling. These sequences often contained defects, and labeling them enhanced the model's robustness (Figure~\ref{fig:C1_PDMYR_difficult}). (2) We designed data augmentation techniques reflecting the defects in the data, such as MotionBlur, OpticalDistortion, GridDistortion, along with built-in augmentations like Mosaic and Mixup in YOLOv5 (Figure~\ref{fig:C1_PDMYR_distribution}B).

\paragraph{Tracking-Assisted Video Annotation}
Previous competition winners utilized target tracking to assist in instrument identification across consecutive frames. However, existing pre-trained single-object tracking (SOT) and multi-object tracking (MOT) models were ineffective at 1 fps due to significant variability between frames. Effective tracking requires high frame rates and manual corrections to ensure accuracy.

\begin{figure}[tbh]
\begin{center}
\resizebox{5in}{!}{\includegraphics{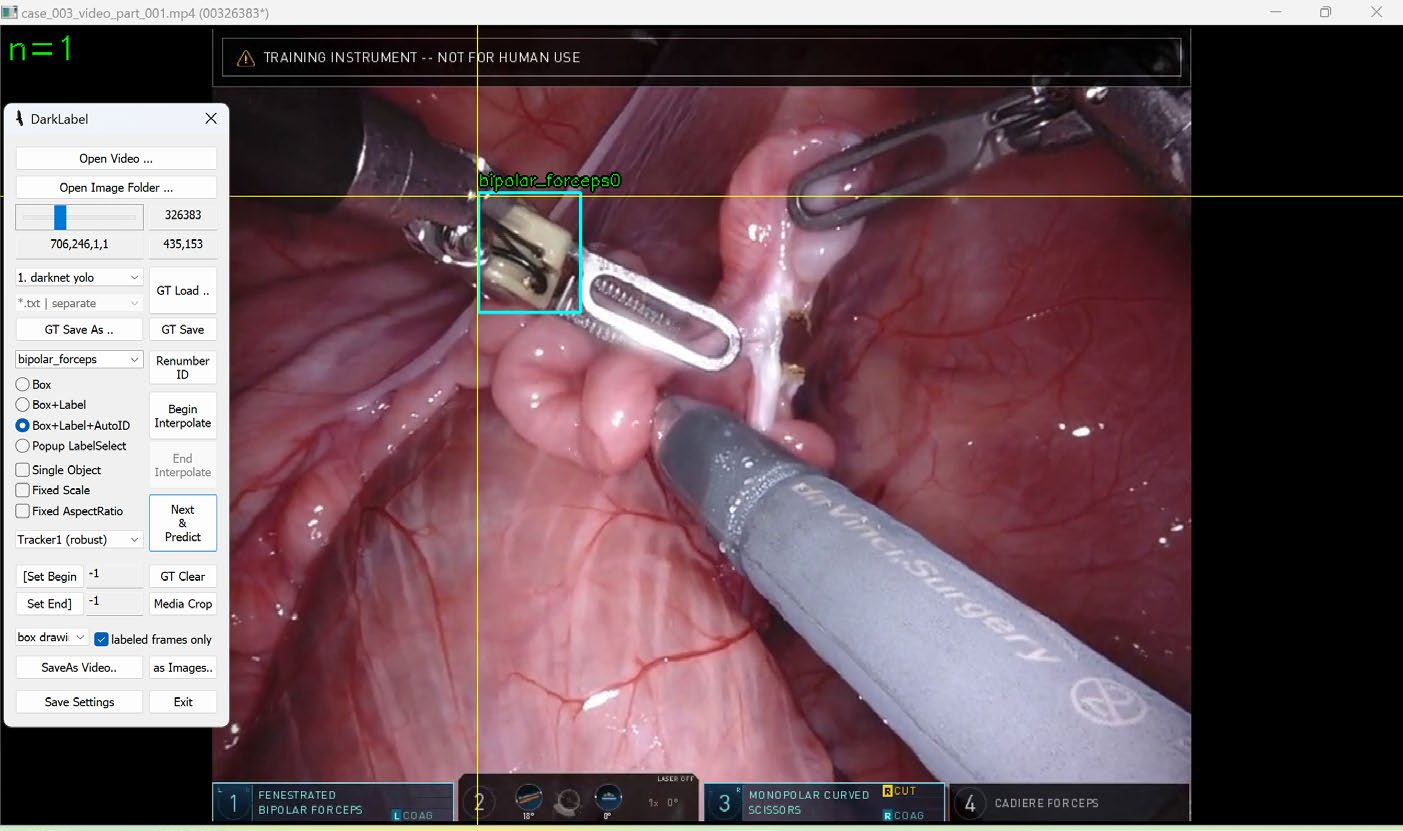}}
\caption{DarkLabel tool interface for tracking-assisted video annotation. The tool incorporates traditional tracking algorithms with an intuitive interface for immediate manual correction of tracking errors.}
\label{fig:C1_PDMYR_tracking}
\end{center}
\end{figure}

We used the DarkLabel tool, which incorporates a traditional tracking algorithm similar to Kernelized Correlation Filters (KCF). Although less advanced than deep learning-based trackers, DarkLabel offers an excellent user interface that facilitates immediate manual correction of tracking errors (Figure~\ref{fig:C1_PDMYR_tracking}). This approach enabled us to annotate 220,000 frames in just three days, significantly reducing annotation costs.

\subsubsection{Model Training}

In this task, the quantity and quality of annotations are crucial, so the model architecture contributes less to the final performance. We chose YOLOv5 for its speed and ease of iteration. As a single-stage detector, YOLOv5 effectively relies on global information for instrument classification without requiring additional Region of Interest (ROI) processing.

To prevent label leakage, we segregated different cases in the training and validation sets based on the provided case information. We conducted extensive tuning and parameter searches within YOLOv5, including 300 rounds of hyperparameter evolution to identify the optimal training parameters (Table~\ref{tab:C1_PDMYR_hyperparams}). We used a batch size of 12 across four NVIDIA RTX 4090 GPUs and maximized the input image resolution.

\begin{table}[htbp]
\centering
\caption{Hyperparameter evolution results for YOLOv5 training configurations.}
\label{tab:C1_PDMYR_hyperparams}
\begin{tabular}{lc}
\hline
Config & mAP \\
\hline
hyp.scratch-low & 0.4515 \\
hyp.scratch-high & 0.5022 \\
custom evolution & 0.5548 \\
\hline
\end{tabular}
\end{table}

\subsubsection{Preliminary Performance}

Our best individual model was YOLOv5x with a resolution of 1280. We combined predictions from models of various sizes and resolutions using Weighted Box Fusion with weights [0.75, 1.0, 1.0, 0.5, 0.5] to produce our final submission (Table~\ref{tab:C1_PDMYR_results}).

\begin{table}[htbp]
\centering
\caption{Performance comparison of different YOLOv5 model configurations and ensemble results.}
\label{tab:C1_PDMYR_results}
\begin{tabular}{lccc}
\hline
Model Size & Input Resolution & Training Image Num. & mAP \\
\hline
YOLOv5-X & 1280 & 220k & 0.5548 \\
YOLOv5-X & 1280 & 170k & 0.5462 \\
YOLOv5-X & 640 & 170k & 0.5470 \\
YOLOv5-L & 1600 & 220k & 0.5453 \\
YOLOv5-M & 1920 & 220k & 0.5336 \\
WBF-ALL & - & - & 0.5661 \\
\hline
\end{tabular}
\end{table}

Despite employing various active learning and assistive techniques to enhance labeling efficiency, our effectively labeled data represented less than 3\% of the entire 840-hour video dataset. While pseudo-labeling and semi-supervised training are effective for object detection data, time constraints prevented us from adopting these methods before the competition deadline. We anticipate that incorporating these approaches in future competitions will further improve performance.

\newpage
\FloatBarrier
\subsection{SKJP}
Our team participated in both Category 1 (surgical tool classification and localization) and Category 2 (surgical visual question answering) of the SurgVU 2025 challenge. For Category 1, we proposed to utilize pseudo labels to train an object detection model to tackle the challenging task of training with weakly annotated datasets. For Category 2, we developed a surgical visual question answering method using LLM-generated QA pairs to train a vision-language model.

\subsubsection{Method Description}

\textbf{Category 1: Surgical Tool Detection.} Training object detection models by using weak annotated datasets is a challenging task. To tackle this problem in surgical tool detection, we propose to utilize pseudo labels for object detection tasks constructed from few human-annotated labels.

First, we annotate bounding boxes for few videos. The target part of the surgical tools is clevis, but the annotation is conducted for tip, clevis and shaft because we use the bounding box information of tip and shaft in the inference stage. Next, we train the first object detection model by using the annotated images. We use YOLOv10 \cite{SKJP_Wang2024} as the object detector.

Then, we extract bounding boxes for other videos by using the first object detector. And the second object detector is trained by using the extracted bounding boxes as pseudo labels.

\begin{figure}[tbh]
\begin{center}
\resizebox{5in}{!}{\includegraphics{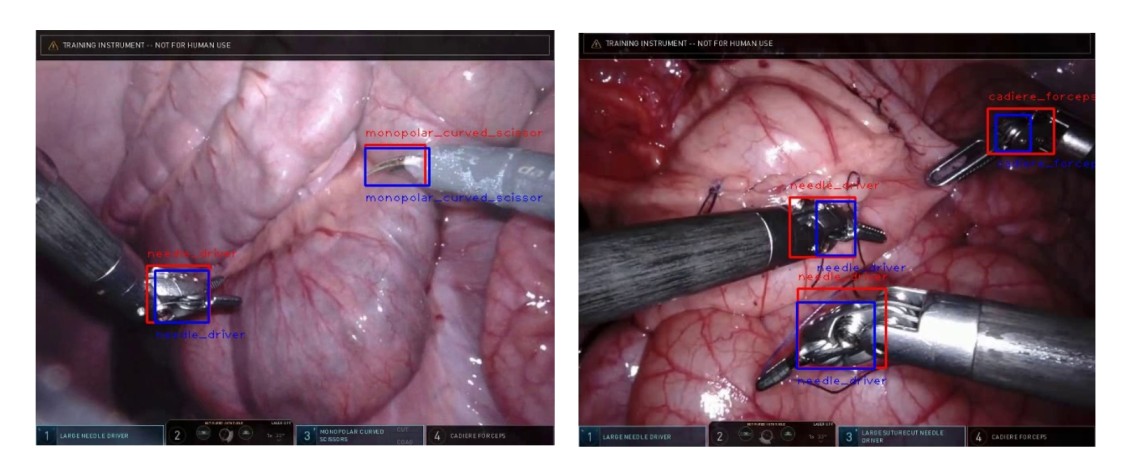}}
\caption{Examples of input image and surgical tool detection results. The red boxes show the prediction results and the blue boxes show the ground truth.}
\label{fig:C1_C2_SKJP_detection_results}
\end{center}
\end{figure}

\textbf{Category 2: Surgical Visual Question Answering.} Visual Question Answering (VQA) for endoscopic surgery videos is a technology that can automatically respond to questions regarding intraoperative situations, surgical instruments, and anatomical structures, making it useful for surgical education, skill assessment, and intraoperative support.

Descriptions of videos are provided as labels, indicating which surgical instruments are used during which time periods and what tasks are performed during which time periods. Since only these descriptions are given without QA pairs and the texts are selected from about ten fixed templates, we generate QA pairs using LLM.

In the prompt for generating QA pairs, we instruct LLM to generate three questions for each video and to generate five different answers for each question. In addition, surgical tool names detected by an object detector trained for Category 1 are included in the prompt, along with sample QA pairs provided as few-shot examples. Since the designated video segments are at least 30 seconds long, we extract the central 30 seconds.

Using the videos and the QA pairs generated in this way, we train VLM. We use InternVL3.5-1B (hereafter referred to simply as InternVL) for VLM \cite{SKJP_Wang2025,SKJP_InternVL}. An adapter layer is inserted after the vision-language connector of InternVL. The original InternVL weights keep frozen, and only the adapter is trained. Eight frames are evenly sampled from each video as input. Regarding the QA pairs, multiple pairs (15 per video) are available, and these are randomly selected for each epoch.

\subsubsection{Model Training}

\textbf{Category 1.} The conditions for training YOLOv10 model are as follows. The batch size is 90, the image size is 640×640 and the number of epochs is 500. The other training conditions are the same as the default setting.

\textbf{Category 2.} Training was conducted for 100 epochs with a batch size of 1. The optimizer was AdamW, with a warmup of 2000 steps, followed by a linear learning rate schedule.

\subsubsection{Preliminary Performance}

\textbf{Category 1.} Performance evaluation at the final testing phase was 0.226 mAP. Some examples of tool detection results for public test dataset are shown in Fig. \ref{fig:C1_C2_SKJP_detection_results}.

\textbf{Category 2.} Performance evaluation at the final testing phase was 0.202 in BLEU score.

\FloatBarrier
\subsection{Algoritmi}

Automatic detection of surgical tools in endoscopic videos is an essential step toward improving surgical workflow analysis. In this challenge, we addressed two complementary tasks: detecting tools from video frames and answering natural language questions about the surgical video provided.

For tool detection, we employed an ensemble of YOLO-based models, which offers robustness by using complementary predictions across different detectors. For question answering, we used a lightweight language model enhanced with contextual information from the detected tools, allowing the system to generate interpretable responses without requiring large-scale video-language pretraining. By combining ensemble-based detection with a context-aware language model, we balance accuracy with computational feasibility, while demonstrating a practical approach toward multimodal surgical AI.

\subsubsection{Method Description}

\begin{figure}[tbh]
\begin{center}
\resizebox{3.5in}{!}{\includegraphics{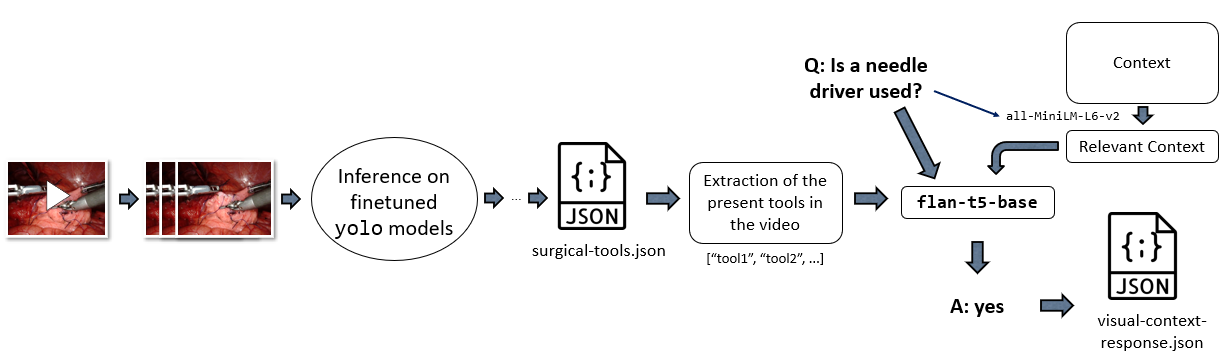}}
\caption{Workflow of our proposed solution for both categories.}
\label{fig:C1_C2_Algoritmi_workflow}
\end{center}
\end{figure}

Our approach consisted of two main components: (1) surgical tool detection from endoscopic videos, and (2) question answering using tool-context information. The overall workflow is illustrated in Figure~\ref{fig:C1_C2_Algoritmi_workflow}.

\textbf{Data preprocessing and augmentation.} We used the validation dataset provided by the challenge organizers for both model development and evaluation. Each video was split into frames. To improve robustness, we applied data augmentation using the Albumentations library~\cite{Algoritmi_Buslaev2020} with the following transformations: random horizontal flip, random vertical flip, random rotation between $-20^\circ$ and $+20^\circ$, random shear and random color jitter. This augmentation strategy helped increase the amount of training data and the models generalize to the different orientations and lighting conditions.

\textbf{Tool detection.} We trained multiple object detection models based on the You Only Look Once (YOLO)~\cite{Algoritmi_Redmon2016} family of architectures, including YOLOv8~\cite{Algoritmi_Yaseen2024} and YOLOe~\cite{Algoritmi_Wang2025}. The ensemble was designed to combine complementary strengths across models. Since the validation dataset did not contain all surgical tool classes, we further fine-tuned one of the best-performing YOLOe models using a small manually annotated dataset (few-shot learning). This step allowed the detector to recognize underrepresented classes.

\textbf{Post-processing.} To refine the raw detections, we applied post-processing steps designed to enforce temporal consistency: (1) For frames missing detections that appeared in adjacent frames, a secondary inference was performed with a lower confidence threshold. (2) Detections that appeared in only a single frame across the video were removed to reduce false positives. (3) In cases where multiple bounding boxes overlapped for the same tool, only the highest-confidence box was retained. These refinements helped reduce wrong detections and improved the stability of tool tracking across frames.

\textbf{Question answering.} Due to hardware constraints that limited the use of large video-language models, we implemented a lightweight approach. First, relevant context about surgical tools and tasks was collected, such as the function of each tool, procedural steps, and example question-answer pairs. These document chunks were encoded using the all-MiniLM-L6-v2~\cite{Algoritmi_Wang2020} embedding model to allow semantic retrieval of relevant context for each question. The retrieved context was then fed into a Flan-T5-base~\cite{Algoritmi_Chung2022} language model to generate answers. This approach allowed us to answer natural language questions while keeping computational requirements low.

\subsubsection{Model Training}

We evaluated our pipeline using the test split of the augmented validation set. For tool detection, the ensemble of YOLO models, combined with the post-processing steps, significantly improved performance compared to using a single model alone. The ensemble increased robustness across tool classes, while post-processing reduced missing detections, removed one-frame detections, and resolved overlapping bounding boxes.

\subsubsection{Preliminary Performance}

For tool detection, the ensemble of YOLO models, combined with the post-processing steps, significantly improved performance compared to using a single model alone. The ensemble increased robustness across tool classes, while post-processing reduced missing detections, removed one-frame detections, and resolved overlapping bounding boxes.

When it comes to question answering, the retrieval-augmented Flan-T5-base language model was able to handle simple questions effectively, providing concise and accurate answers when context was clear. While the approach worked well for straightforward queries, more complex or multi-step reasoning questions remain a challenge.

Overall, the detection component worked well, with the ensemble and post-processing steps substantially improving stability and recall compared to a single detector. However, some tools were occasionally misclassified, and several underrepresented classes were not learned effectively due to limited training examples. Regarding the second task, our approach was able to provide correct responses to simple queries but struggled with more complex questions. Performance was also closely tied to the accuracy of the tool detections, meaning that errors in detection propagated into the question answering stage.

\FloatBarrier

\subsection{SurgTroopers}

We first explored a weakly supervised pipeline using a Grad-CAM–based CNN with tool presence labels, which provided useful insights but lacked detection accuracy. To address this, we curated and annotated representative frames with bounding boxes, creating a more balanced dataset. Using these annotations, we trained a YOLOv8-based detector with tailored preprocessing and data augmentation. This fully supervised approach produced stronger generalization and more reliable detection results, highlighting the limitations of weak supervision and the value of curated annotations.

\subsubsection{Method Description}

\textbf{Dataset.} The dataset provided consists of 155 training sessions captured at 60 fps with a resolution of 720p from one endoscopic channel. In total, the dataset spans over 840 hours of video, where each case is accompanied by tool presence labels, indicating which instruments are in use at a given time, but without bounding box annotations.

\begin{figure}[tbh]
\begin{center}
\resizebox{3in}{!}{\includegraphics{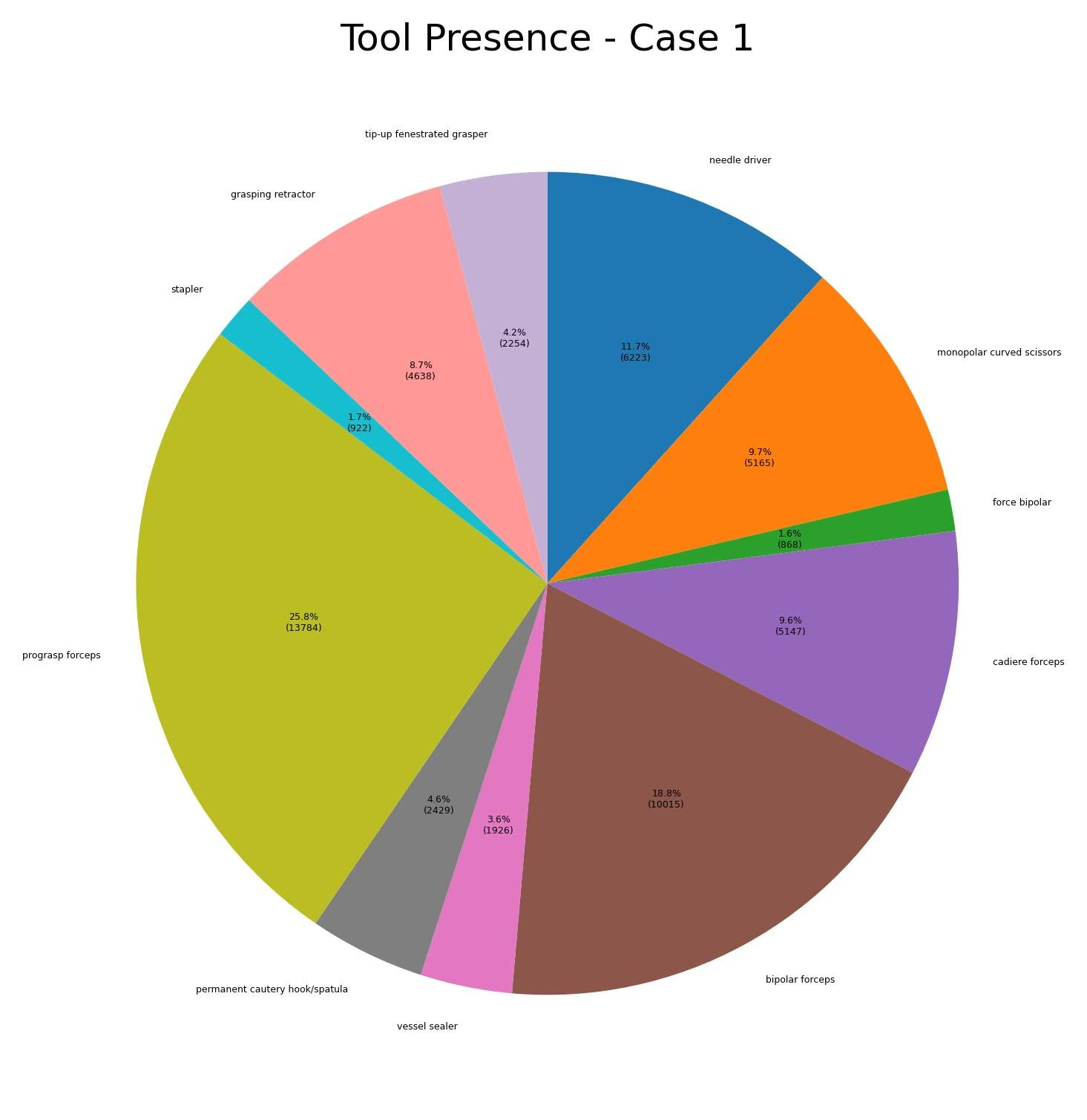}}
\caption{A pie chart showing the tool presence statistics for case 001.}
\label{fig:C1_SurgTroopers_tool_stats}
\end{center}
\end{figure}

To make the dataset computationally feasible, we downsampled videos to 2 fps, resulting in approximately 6 million frames. From this pool, we generated tool presence statistics for each case, as shown in Figure~\ref{fig:C1_SurgTroopers_tool_stats}, using the given labels, which revealed strong class imbalance. For example, commonly used tools such as needle drivers appeared frequently, while others such as clip appliers were underrepresented. Based on these statistics, we manually selected 60–70 representative frames per class, prioritizing frames that contained underrepresented classes. The selected frames were then annotated with bounding boxes and tool categories using Label Studio and exported in YOLO format, creating a dataset with balanced coverage across instruments.

\textbf{Preprocessing.} The data was split 80/20 for training and validation per Ultralytics convention. Border masking removed black margins to focus on the surgical field, and images were auto-resized to 512×512 by YOLOv8's dataloader during training.

\textbf{Data Augmentation.} To enhance generalization and mitigate limited data, we applied several augmentations during training. Horizontal flipping (50\% probability) exposed the model to mirrored orientations, while geometric transforms such as shearing (10°), scaling ($\sim$20\%), and translations ($\leq$5\%), introduced viewpoint and positioning variations.

We also used mosaic augmentation (20\% probability), combining four images into one to increase object diversity; this was disabled after 10 epochs to stabilize training. Augmentation is applied to only training data, to reflect the real-world unseen data as much as possible.

\textbf{Model.} For the challenge, we previously explored opportunities in employing the ResNet50-based model with Grad-CAM, but decided to abandon the approach due to poor results and other constraints. An experiment was then performed on multiple Ultralytics YOLO models including YOLOv8s, YOLOv8m, YOLOv11s, and YOLOv11m. Ultimately, we employed Ultralytics' YOLOv8s model which has about 11 million parameters~\cite{SurgTroopers_Ultralytics2024} for training on approximately 4,800 labeled images. Furthermore, we would like to experiment using YOLO which has a CNN-based backbone that supports both classification and bounding boxes to rapidly build and train an object detection model.

\subsubsection{Model Training}

\begin{figure}[tbh]
\begin{center}
\resizebox{3in}{!}{\includegraphics{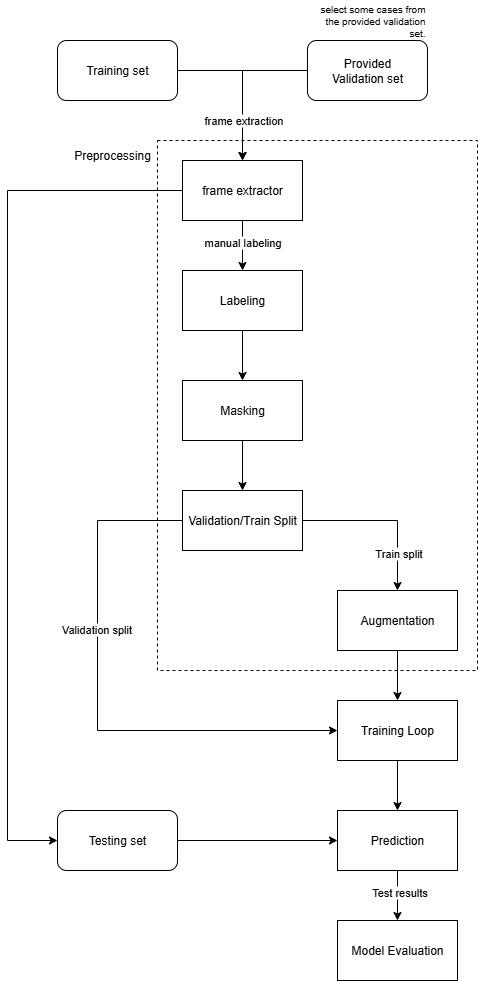}}
\caption{Overall process. The process begins with the Training set and a Provided Validation set. Some cases are selected from the validation set, and both datasets go through frame extraction, then applied the preprocessing. The training split is then passed through the training loop, producing a model that generates predictions on both the validation and testing set. The results are then used for model evaluation.}
\label{fig:C1_SurgTroopers_process}
\end{center}
\end{figure}

As shown in Figure~\ref{fig:C1_SurgTroopers_process}, the process begins with the Training set and a Provided Validation set. Some cases are selected from the validation set, and both datasets go through frame extraction, then applied the preprocessing. The training split is then passed through the training loop, producing a model that generates predictions on both the validation and testing set. The results are then used for model evaluation.

The model was trained on RTX 5060 Ti and was experimented with multiple parameter values. From the experiment, we found that training the model with an epoch of 50 and patience set to 10 is balanced. The number of workers is set to '4'. In addition, we set the batch size to '-1' which automatically utilizes only 60\% of GPU resources. Similarly, the optimizer is set to be 'auto' which will automatically select the suitable optimizer. The learning rate is set to 5e-4.

\subsubsection{Preliminary Performance}

\begin{figure}[tbh]
\begin{center}
\resizebox{3in}{!}{\includegraphics{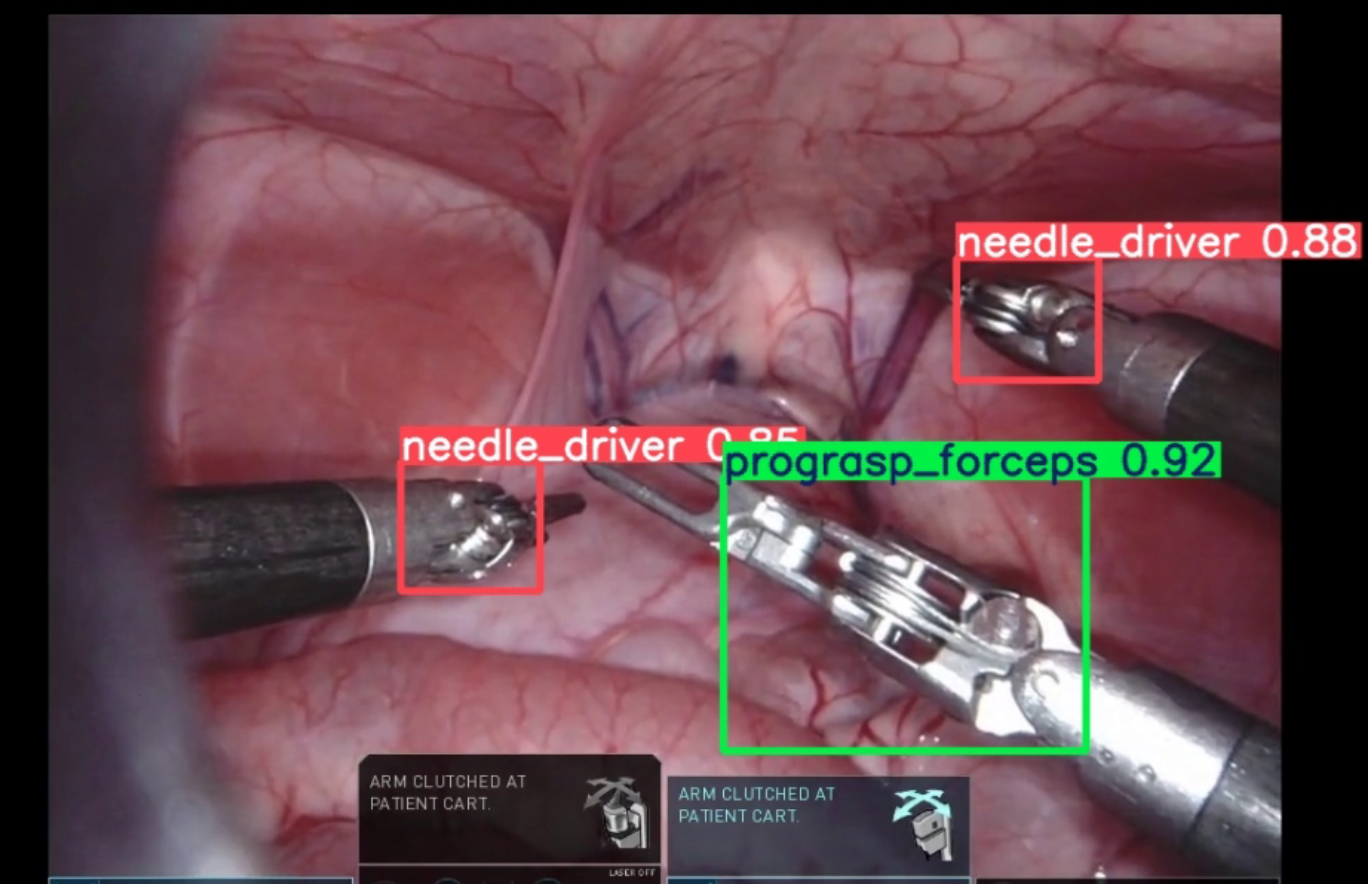}}
\caption{Bounding box result example.}
\label{fig:C1_SurgTroopers_bbox_result}
\end{center}
\end{figure}

\begin{figure}[tbh]
\begin{center}
\resizebox{3in}{!}{\includegraphics{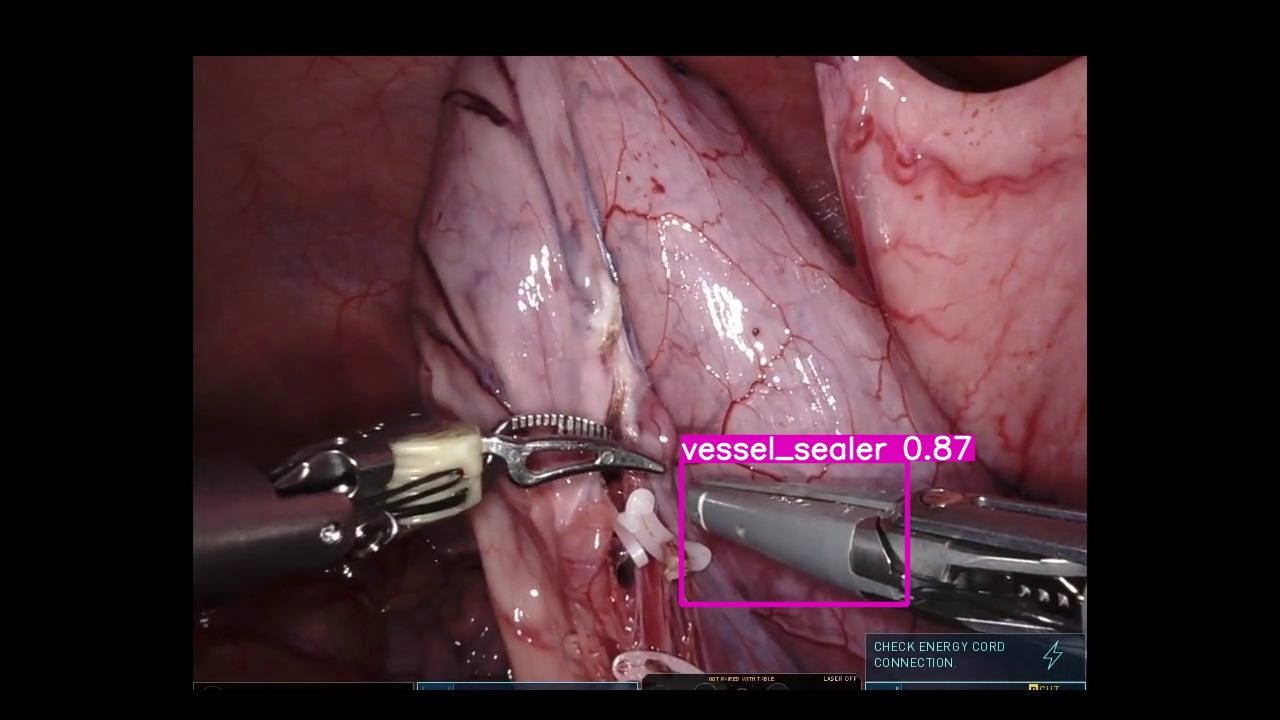}}
\caption{Undetected tool example.}
\label{fig:C1_SurgTroopers_undetected}
\end{center}
\end{figure}

The method resulted in a mean average precision of 0.2087 on the final test set. While this number is modest, our observations suggest the model has started to learn meaningful patterns. In particular, it performs quite well on tools that appear frequently in the dataset, showing consistent detections with good bounding box alignment, as shown in Figure~\ref{fig:C1_SurgTroopers_bbox_result}. On the other hand, less common tools (long-tail classes) and confusion between the background and the tools remain challenging for the model. Detections often fail when instruments are seen from unusual angles or partly occluded, as shown in Figure~\ref{fig:C1_SurgTroopers_undetected}.

\begin{figure}[tbh]
\begin{center}
\resizebox{4in}{!}{\includegraphics{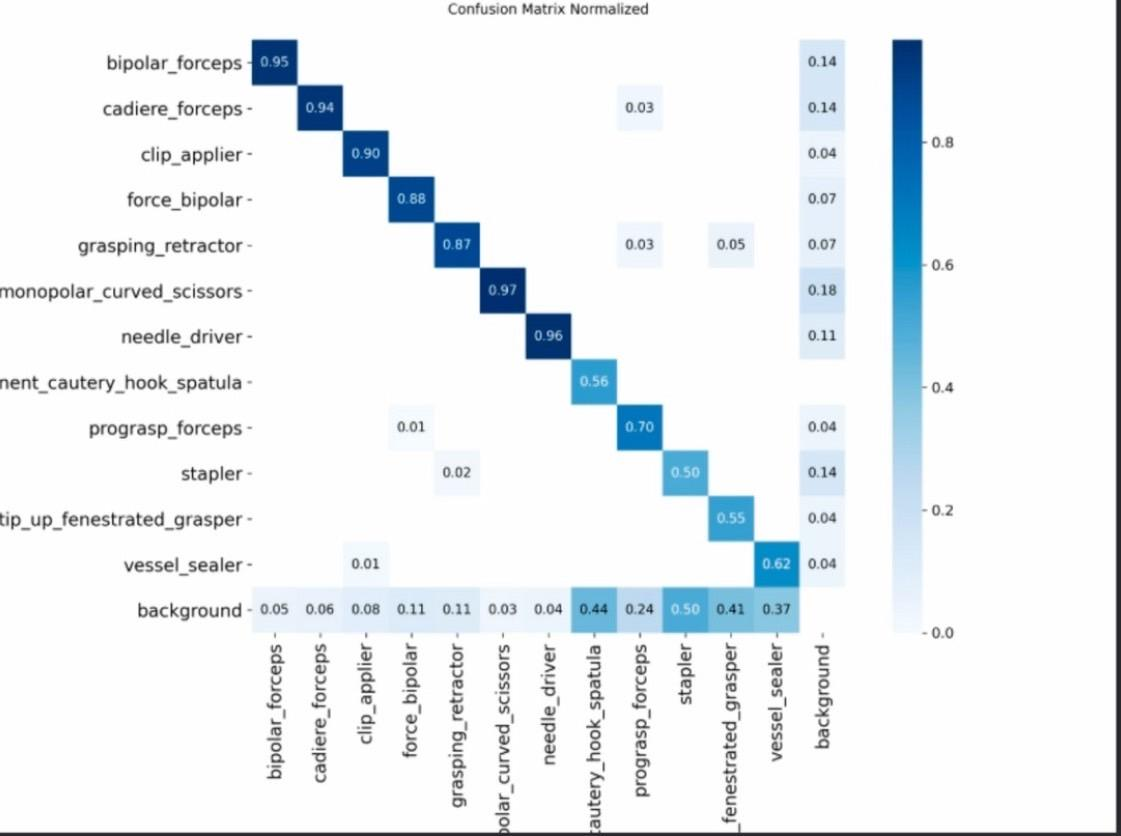}}
\caption{Confusion Matrix (Normalized) of Model Performance.}
\label{fig:C1_SurgTroopers_confusion}
\end{center}
\end{figure}

In terms of localization, bounding box accuracy is generally reliable once the model detects a tool, the box usually covers it properly. The main drawback we noticed is the tendency to generate duplicate boxes around the same tool with different confidence scores, which indicates some prediction uncertainty and effect on mean average precision score.

For our final phase, we tested YOLOv8m as an alternative to YOLOv8s using the same training settings. However, YOLOv8m slightly underperformed (mAP 0.1502), showing that increased model size alone doesn't guarantee better results and further tuning and dataset-specific optimization may be needed.

In this study, we explored both weakly and fully supervised approaches for surgical tool detection in RAS. While the Grad-CAM–based ResNet-50 showed some promise, it lacked the precision required for reliable detection. Due to time constraints, we shifted to a fully supervised YOLOv8 detector trained on manually annotated data. With tailored preprocessing and augmentation, the model showed improved performance, particularly for common instruments. Our findings highlight the importance of high-quality annotations, balanced datasets, and appropriate model selection in advancing surgical data science.

Our approach faced class imbalance, with rare tools like clip appliers and staplers underrepresented in the dataset. Tool similarity and limited data volume reduced generalization, especially for unusual angles or occlusions. Duplicate bounding boxes also signaled prediction uncertainty. While switching from YOLOv8s to YOLOv8m didn't help, future improvements could include hyperparameter tuning, active learning, temporal modeling, or integrating a secondary CNN for refinement.

\FloatBarrier

\subsection{Capybara}

Our team from Aillis, Inc. (Tokyo, Japan) participated in Category 2: Surgical Visual Question Answering (VQA). 
While Vision–Language Models (VLMs) have demonstrated strong capability in generating general descriptions of surgical videos, they often exhibit limited accuracy in identifying domain-specific entities such as surgical instruments and anatomical structures. 
A straightforward approach to this task is to fine-tune VLMs using the provided video–caption pairs. 
However, the dataset presents two major challenges: (1) the annotations are noisy, and (2) the diversity of textual supervision is limited, with only 21 unique captions (based on the “match description” field). 
Consequently, preliminary fine-tuning experiments with the provided descriptions yielded unsatisfactory results.

Analysis of the public sample set (11 videos) indicates that most questions are primarily about surgical instruments, anatomical structures, and procedural actions (e.g., tissue cutting, manipulation with forceps). 
This observation suggests that accurate recognition of tools and organs may provide sufficient cues for inferring surgical actions, even in the absence of reliable ground-truth descriptions. 
For example, the presence of scissors is strongly associated with cutting actions. 
Motivated by this insight, we propose a framework that explicitly detects surgical instruments and anatomical structures from video frames and integrates these detections with general video descriptions for surgical VQA. 

\subsubsection{Method Description}

\begin{figure}[tbh]
\begin{center}
\resizebox{5in}{!}{\includegraphics{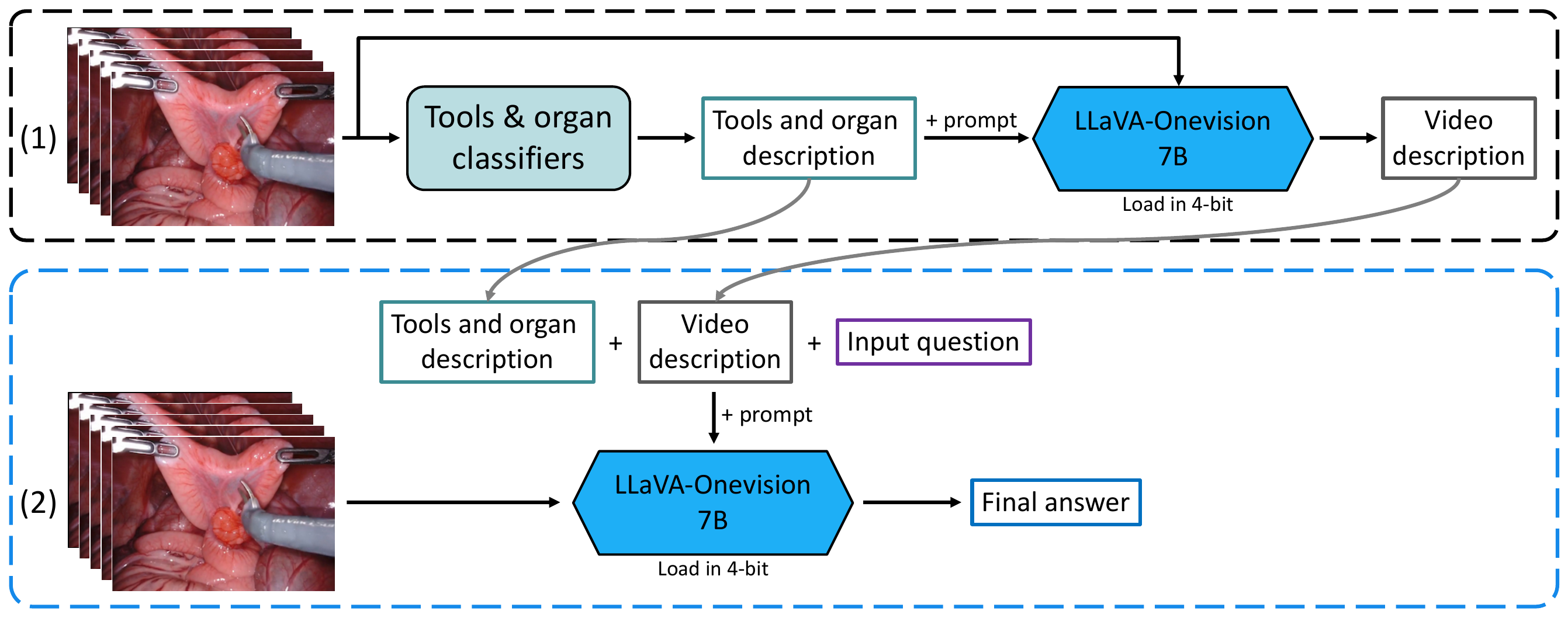}}
\caption{Overview of the proposed method showing the two-step process: (1) creating a video description using surgical tool and organ classifiers, and (2) generating the final answer using the VLM with the video description, tools/organ information, and video frames.}
\label{fig:C2_Capybara_overview}
\end{center}
\end{figure}

Figure~\ref{fig:C2_Capybara_overview} illustrates the overview of our proposed method. 
The method comprises two stages: (1) video description generation and (2) answer prediction. 
In the first stage, surgical tools and organ classifiers are applied to individual video frames to identify the presence of relevant tools and organs, producing a structured tool–organ description. 
This description is then combined with a predefined prompt to generate a comprehensive video description. 

In the second stage, the tool–organ description, the generated video description, the input question, and the video frames are jointly provided to the Vision–Language Model (VLM) to produce the final answer. 
Notably, the same VLM and input video frames are utilized in both stages. 
Based on preliminary experiments, we adopt the LLaVA-OneVision-7B model \cite{Capybara_Li2024} without additional fine-tuning. 

For preprocessing, black margins on the left and right sides of each frame were removed, and the tool list region at the bottom of the frame was obscured using Gaussian blurring before inference.

\textbf{The Surgical Tool Classifier:} The tool presence labels in the provided dataset are noisy, and the clip durations vary substantially (0 – 5000 seconds), making it challenging to construct a reliable training set for surgical tool classification.  
To address this, we leveraged a public dataset from the SurgToolLoc Challenge (MICCAI 2022) \cite{Capybara_Zia2023}, which contains ~24,600 clips of 30 seconds each with tool presence annotations. 
Although this dataset also contains some noise, it significantly reduces the effort required for data curation. 

We excluded clips that were too small or large, and built a multi-label dataset with 22k training clips and 2.4k validation clips (30 frames each). 
All video frames were preprocessed as described previously.
We trained EfficientNetV2-Small \cite{Capybara_Tan2021} with 512$\times$512 input, applying label smoothing to handle noisy labels. 
The model achieved a macro F1-score of 97\% on the validation set. 
At test time, surgical tools are detected frame by frame.

\textbf{The Organ Classifier:} There are seven types of organs from the challenge dataset description (“match description” in CSV files). 
With the associate videos, we extracted 2700 clips for training the organ classifier (2300 clips for training, 400 clips for validation; 30 frames/clip). 
Assuming each clip has one organ, this is the single-class classification. 
The model, input size, and pre-processing method are the same as in the surgical tool classifier. 
The trained model achieved a macro F1-score of 98\% on the validation set. 
At test time, the organ prediction is based on voting (e.g., if rectum is predicted in most frames, the final prediction is rectum). 

\begin{figure}[tbh]
\begin{center}
\resizebox{5in}{!}{\includegraphics{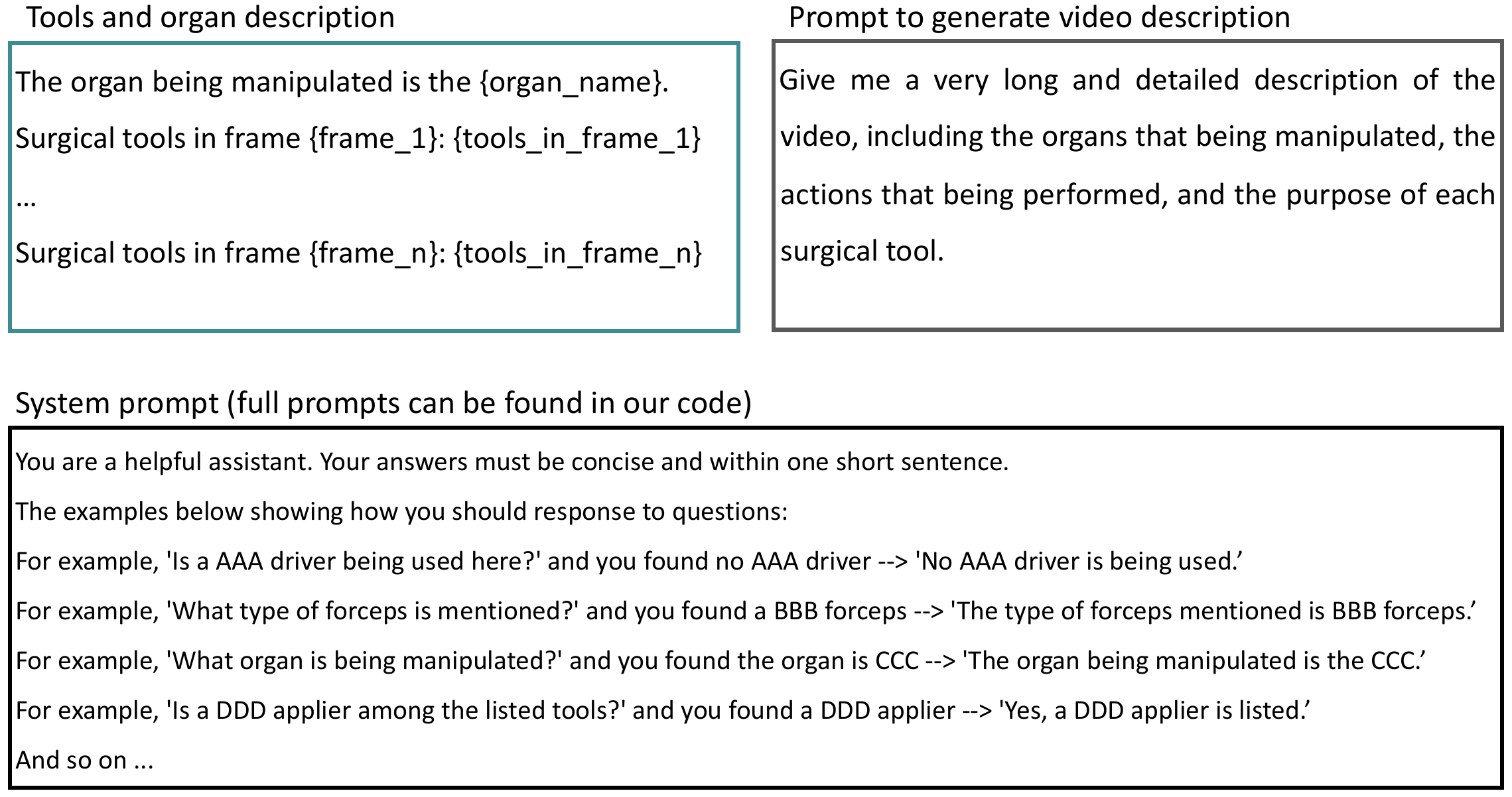}}
\caption{Template of the tools and organ description, and the prompts for generating video description and final answer.}
\label{fig:C2_Capybara_prompts}
\end{center}
\end{figure}

\textbf{Generating the Video Description and Final Answer:} 
Figure~\ref{fig:C2_Capybara_prompts} presents the template used for constructing tool–organ descriptions, as well as the prompts for video description generation and answer prediction. 
Given that the ground-truth answers in the sample set are expressed as short sentences, we design the system prompt to encourage concise responses. 
The prompt was selected through iterative experimentation, where multiple candidates were evaluated, and the one achieving the highest BLEU score on the public sample set (11 videos) was chosen.

\subsubsection{Model Training}

\textbf{Surgical Tool Classifier:}
\begin{itemize}
\item Dataset: 22k training clips and 2.4k validation clips from SurgToolLoc Challenge (MICCAI 2022), 30 frames per clip
\item Model: EfficientNetV2-Small with 512×512 input
\item Training technique: Label smoothing to handle noisy labels
\item Performance: 97\% macro F1-score
\end{itemize}

\textbf{Organ Classifier:}
\begin{itemize}
\item Dataset: 2700 clips extracted from challenge dataset (2300 training, 400 validation), 30 frames per clip
\item Task: Single-class classification (7 organ types)
\item Model: EfficientNetV2-Small with 512×512 input
\item Performance: 98\% macro F1-score
\end{itemize}

\textbf{VLM:} LLaVA-OneVision 7B model (original model without additional fine-tuning)

\subsubsection{Preliminary Performance}

Our best final submission samples 5 frames/video as input. 
We achieved 0.4237 BLEU in the Prelim phase, and 0.4215 BLEU in the Final phase (rank 1st). 
Among other final submissions, the variant with 21 frames/video input got 0.4022 BLEU (rank 2nd), while replacing LLaVA-OneVision with the reasoning model R-4B \cite{Capybara_Jiang2025} yielded lower result with 0.3448 BLEU (rank 4th).

Throughout the challenge, we experimented with several VLMs and observed that models using the Qwen text encoder (e.g., Qwen2, Qwen3) were easier to control the output compared to those based on LLaMA or BERT. 
We also found that VLMs with reasoning capabilities better captured video semantics, but were more difficult to control. 
Nevertheless, with appropriate configuration, such models could be a promising approach.

Additionally, we observed that VLMs quantized to 4-bit precision showed inconsistent behavior across GPU architectures (e.g., Turing, Ampere, Hopper), producing different outputs for the same input. 
Since the evaluation platform uses Tesla T4 (Turing), we recommend testing methods on the same architecture for reproducibility. 
As future work, enriching video descriptions through robust object detection or collecting higher-quality captions could further improve performance. 

\FloatBarrier
\subsection{UoM-SurgicalAI}

We participated in Category 2 (Surgical Visual Question Answering) of the SurgVU 2025 challenge. Our work explores whether well-crafted zero-shot prompting can outperform parameter-efficient fine-tuning (PEFT) methods in weakly supervised surgical VQA. Surgical visual question answering~\cite{UoMSurgicalAI_Seenivasan2022} supports robotic procedures by addressing clinically relevant queries on instruments, anatomy, and tasks. Yet, limited and weakly supervised datasets make model training challenging. While PEFT methods~\cite{UoMSurgicalAI_Farina2025} are commonly used, they require additional effort and may overfit scarce labels. Using the InternVL3~\cite{UoMSurgicalAI_Zhu2025} family of Vision Language Models (VLMs), we design structured system messages with explicit roles, task and tool scope, and concise answering rules. Our results show that under limited supervision, zero-shot prompting achieves stronger generalization than fine-tuning.

\subsubsection{Method Description}

\begin{figure}[tbh]
\begin{center}
\resizebox{3.5in}{!}{\includegraphics{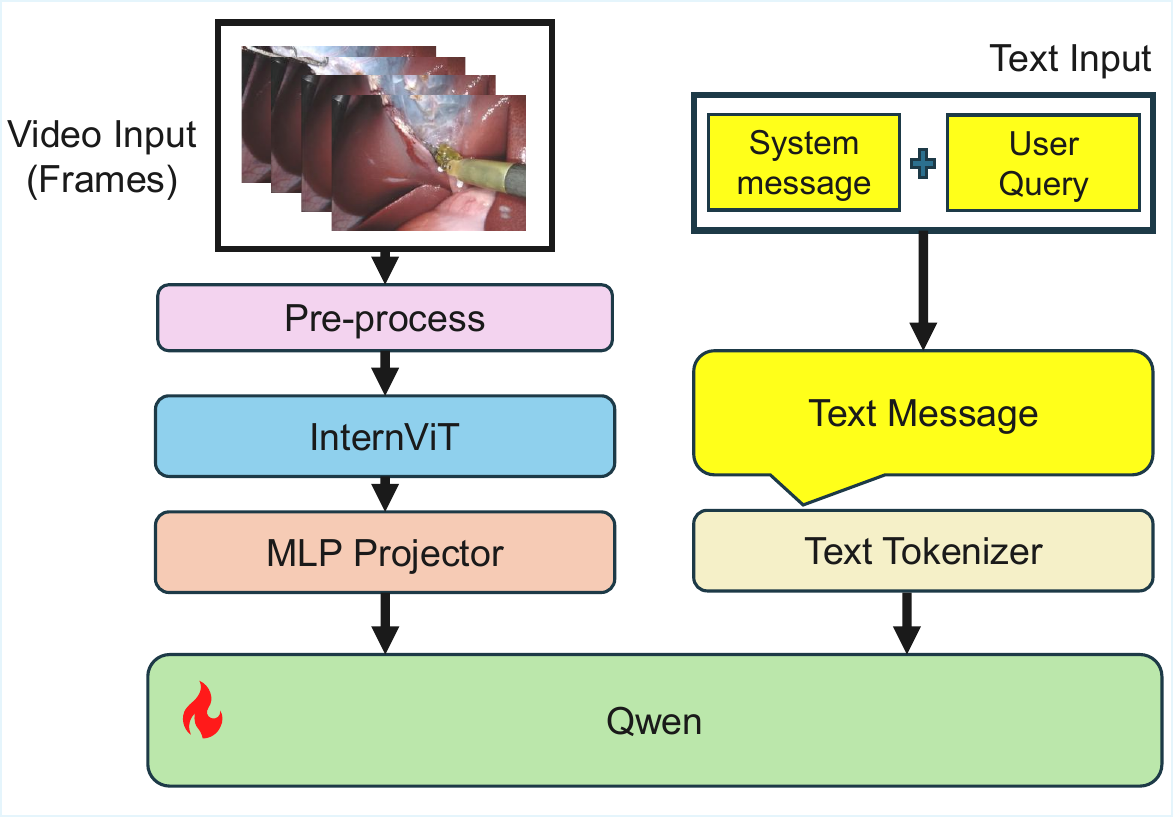}}
\caption{The InternVL network couples an InternViT vision encoder and a Qwen language backbone. Different versions of the InternVL model are distinguished by different versions of the vision encoder and the language backbone.}
\label{fig:C2_UoM-SurgicalAI_architecture}
\end{center}
\end{figure}

We employed the InternVL3 family of Vision Language Models for our approach. The VLMs we used are InternVL3-2B-Instruct and InternVL3.5-4B-Instruct. The InternVL3-2B-Instruct model couples an InternViT-300M vision encoder (24 Transformer blocks; hidden size 1024; 304M parameters) with a Qwen2.5-1.5B language backbone (28 layers; hidden size 1536; 12 heads; context length up to 32K). The InternVL3.5-4B-Instruct model has the same vision encoder as the InternVL3-2B-Instruct model, but its language backbone is changed to Qwen3-4B (36 layers; hidden size 2560; 32 heads; context length up to 41K). Unlike the strategy used during model pre-training, where the vision encoder, MLP projection layer, and language backbone are all trained together, our study only fine-tunes the language backbone as shown in Figure~\ref{fig:C2_UoM-SurgicalAI_architecture}.

We explore the optimal zero-shot performance of the model through systematic design of the system message. Several alternative formulations were tested. The system message that achieved the best performance was designed with explicit role definition, constrained task and tool scope, strict answering rules, and illustrative examples to guide the model toward producing concise, clinically relevant, and structurally consistent responses in surgical contexts.

\subsubsection{Dataset}
The training and validation datasets were constructed through a three-step process:  

\paragraph{Step 1: Clip Sampling and pre-processing.} For each surgical case, two video clips of 30 seconds each were randomly sampled. Clips that did not contain the Da Vinci interface or exhibited unnatural occlusions were excluded following manual inspection. Subsequently, Gaussian blurring was applied to the interface regions of the selected clips in order to obscure textual information. This step was performed to prevent the model from inadvertently learning textual cues embedded in the interface.

\paragraph{Step 2: Category Balancing.} The frequency distribution of each unique category was analyzed. Additional clips were manually extracted to ensure that underrepresented categories were adequately included. Furthermore, files were reallocated between the training and validation sets to guarantee that every unique category appeared at least once in both subsets.  

\paragraph{Step 3: Question-Answer Generation.} A set of template-based question-answer (QA) pairs was created and used to annotate each video clip. These annotations were applied consistently across the training and validation sets to facilitate downstream model training and evaluation. There are 12 unique tools and 7 unique tasks, so we ensured they appear once in both train and validation set. For each clip we created a CSV file that tracks the tasks and instruments that occur in each clip, we manually did this for the validation set that was provided; however, given that task labels were missing, there can be inaccuracies. We then created template questions inspired by the provided validation set. The first couple of questions will be repeated for all 12 tools, \textit{"What type of forceps are used?"} is only asked if there are forceps being used and it shall show which forceps are being used that can be one or more than one. The last question \textit{"What task is being performed in this clip?"} is the same for all the clips. We created these QA (Question-Answer) text files for each clip with these questions. On average there are $38$ or $39$ QA pairs for each clip. Table~\ref{tab:C2_UoM-SurgicalAI_dataset} summarizes the frequency of tools and tasks in train and validation set. In summary, the training set contains 117 cases with a total of 8,845 QA pairs, while the validation set contains 16 cases with a total of 797 QA pairs.

\paragraph{Example QA templates.}
\begin{itemize}
    \item ``Is a \$tool\_name\$ among the listed tools?'' $\mid$ \emph{No, a \$tool\_name\$ is not listed.} / \emph{Yes, a \$tool\_name\$ is listed.}
    \item ``Was a \$tool\_name\$ used in this clip?'' $\mid$ \emph{No, a \$tool\_name\$ was not utilized.} / \emph{Yes, a \$tool\_name\$ was utilized.}
    \item ``Are there forceps being used here?'' $\mid$ \emph{No, forceps are not mentioned.} / \emph{Yes, forceps are mentioned.} 
    \item ``What type of forceps are used?'' $\mid$ \emph{[forceps type]].}
    \item ``What task is being performed in this clip?'' $\mid$ \emph{[task name]}
\end{itemize}

\begin{table}[htbp]
\centering
\caption{Summary of tools and tasks in train/validation set.}
\label{tab:C2_UoM-SurgicalAI_dataset}
\begin{tabular}{lcc|lcc}
\hline
\multicolumn{3}{c|}{Tool} & \multicolumn{3}{c}{Task} \\
Name & Train & Val & Name & Train & Val \\
\hline
Large needle driver & 115 & 6 & Suturing & 86 & 5 \\
Monopolar curved scissors & 82 & 7 & Uterine horn & 35 & 1 \\
Force bipolar & 8 & 1 & Suspensory ligaments & 28 & 5 \\
Clip applier & 5 & 1 & Rectal artery/vein & 26 & 3 \\
Cadiere forceps & 98 & 5 & Skills application & 10 & 3 \\
Bipolar forceps & 94 & 14 & Range of motion & 6 & 1 \\
Vessel sealer & 9 & 1 & Retraction/collision avoidance & 31 & 2 \\
Permanent cautery hook/spatula & 4 & 2 & & & \\
Prograsp forceps & 39 & 1 & & & \\
Stapler & 2 & 1 & & & \\
Grasping retractor & 3 & 3 & & & \\
Tip-up fenestrated grasper & 1 & 1 & & & \\
\hline
\end{tabular}
\end{table}

After building the dataset, we first explored the optimal zero-shot performance of the InternVL3 and InternVL3.5 models through different system message prompts, and subsequently fine-tuned the InternVL3 models using two PEFT methods, DoRA~\cite{UoMSurgicalAI_Liu2024} and DCT-GaLore~\cite{UoMSurgicalAI_Modoranu2025}. Finally, we evaluated the performance of both zero-shot models and fine-tuned models on our self-constructed local validation set. For evaluation, we used Bilingual Evaluation Understudy (BLEU)~\cite{UoMSurgicalAI_Papineni2002}, Recall-Oriented Understudy for Gisting Evaluation-Longest Common Subsequence (ROUGE-L)~\cite{UoMSurgicalAI_Lin2004}, and Metric for Evaluation of Translation with Explicit ORdering (METEOR)~\cite{UoMSurgicalAI_Banerjee2005} as the metrics.

\subsubsection{Model Training}

In all fine-tuning experiments, we set the batch size to 6 and fine-tuned for 20 epochs. A dynamic learning rate was employed with a decay rate of 0.8 and a patience of 3. DCT-GaLore and DoRA are applied to the Q, K, V, O layers of the Transformer blocks of the language modules. For DoRA fine-tuning, we set the initial learning rate to $2 \times 10^{-5}$, the rank to 8, the scaling factor to 16, and the dropout rate to 0.1. For DCT-GaLore fine-tuning, we set the initial learning rate to $3 \times 10^{-6}$, the rank to 128, the update projection frequency to 50, the GaLore scale to 1, and the projection type to 1.0. All fine-tuning experiments were conducted within the PyTorch framework on an NVIDIA A100\_80 GPU. The inference experiments were all conducted within the PyTorch framework on an NVIDIA Tesla P100 GPU.

\subsubsection{Preliminary Performance}

Table~\ref{tab:C2_UoM-SurgicalAI_results} summarizes the performance of both fine-tuned and zero-shot models across local validation, prelim, and final phases. On the local validation set, PEFT (DoRA and DCT) achieved the highest scores in ROUGE-L, METEOR, and BLEU, indicating that adaptation can better fit the limited training data. However, in the official evaluation phases, the zero-shot models demonstrated stronger generalization: the zero-shot InternVL3 achieved a higher BLEU score (0.2776) than both fine-tuned counterparts in the final phase, while the zero-shot InternVL3.5 further improved to 0.3656. These results highlight that under weakly supervised conditions and limited information, carefully designed zero-shot prompting can outperform PEFT-based fine-tuning.

\begin{table}[htbp]
\centering
\caption{The performance of best fine-tuned model and zero-shot model on our local self-built validation set and validation set in prelim phase and final phase.}
\label{tab:C2_UoM-SurgicalAI_results}
\begin{tabular}{lcccccc}
\hline
& \multicolumn{3}{c}{Local Validation} & Prelim & Final \\
Model & ROUGE-L & METEOR & BLEU & BLEU & BLEU \\
\hline
InternVL3 (DoRA) & 0.8969 & 0.9145 & 0.8528 & 0.6125 & 0.2367 \\
InternVL3 (DCT-GaLore) & 0.8468 & 0.8834 & 0.7747 & 0.5678 & 0.2134 \\
InternVL3 (Zero-shot) & 0.7387 & 0.8277 & 0.5701 & 0.4595 & 0.2776 \\
InternVL3.5 (Zero-shot) & 0.7290 & 0.8226 & 0.5335 & --- & 0.3656 \\
\hline
\end{tabular}
\end{table}

Our study shows that carefully designed zero-shot prompting can generalize better than parameter-efficient fine-tuning (PEFT) in weakly supervised surgical VQA, despite PEFT achieving higher local validation scores. Structured system messages with explicit roles and constrained scope guided models toward concise and clinically relevant answers, while fine-tuned models risked overfitting. These results support the notion that fine-tuning on a limited dataset may lead to low generalization.

\FloatBarrier

\subsection{AMI}

Our team from Kyung Hee University, Republic of Korea, participated in Category 2 (Surgical Visual Question Answering). Because our available annotations paired tools with video frames but defined tasks only at a very coarse, sequence-level granularity, a naïve QA construction would fail to capture fine-grained steps. To overcome this, we anchored QA generation on tool occurrences and, whenever a specific tool appeared, we refined the task into concrete steps/actions likely underway. This yielded a supervision signal that is temporally local and operationally meaningful, enabling the model to learn actionable visual–text correspondences rather than vague procedure labels.

\subsubsection{Method Description}

Our approach introduces three key innovations for surgical video question answering:

\textbf{Tool-anchored task refinement:} Instead of treating tasks as monolithic labels, we use the presence of particular tools as priors to disambiguate which step/action is being performed.

\textbf{Presence/absence awareness:} For each question type, we balance positive (Yes) and negative (No) QA pairs so the model learns to detect presence/absence of tools, actions, and tasks, not just memorize affirmative patterns.

\textbf{Two-stage training with a visual adapter:}
\begin{itemize}
\item \textbf{Stage 1:} Train a visual adapter that maps EndoViT\cite{AMI_Batic2024} feature space to Vicuna's text space, aligning vision features with the LLM's embedding geometry for downstream QA.
\item \textbf{Stage 2:} LoRA fine-tuning of Vicuna on our curated QA set (augmented with GPT-5–generated variants) to improve domain specificity while remaining parameter- and data-efficient.
\end{itemize}

\begin{figure}[tbh]
\begin{center}
\resizebox{5in}{!}{\includegraphics{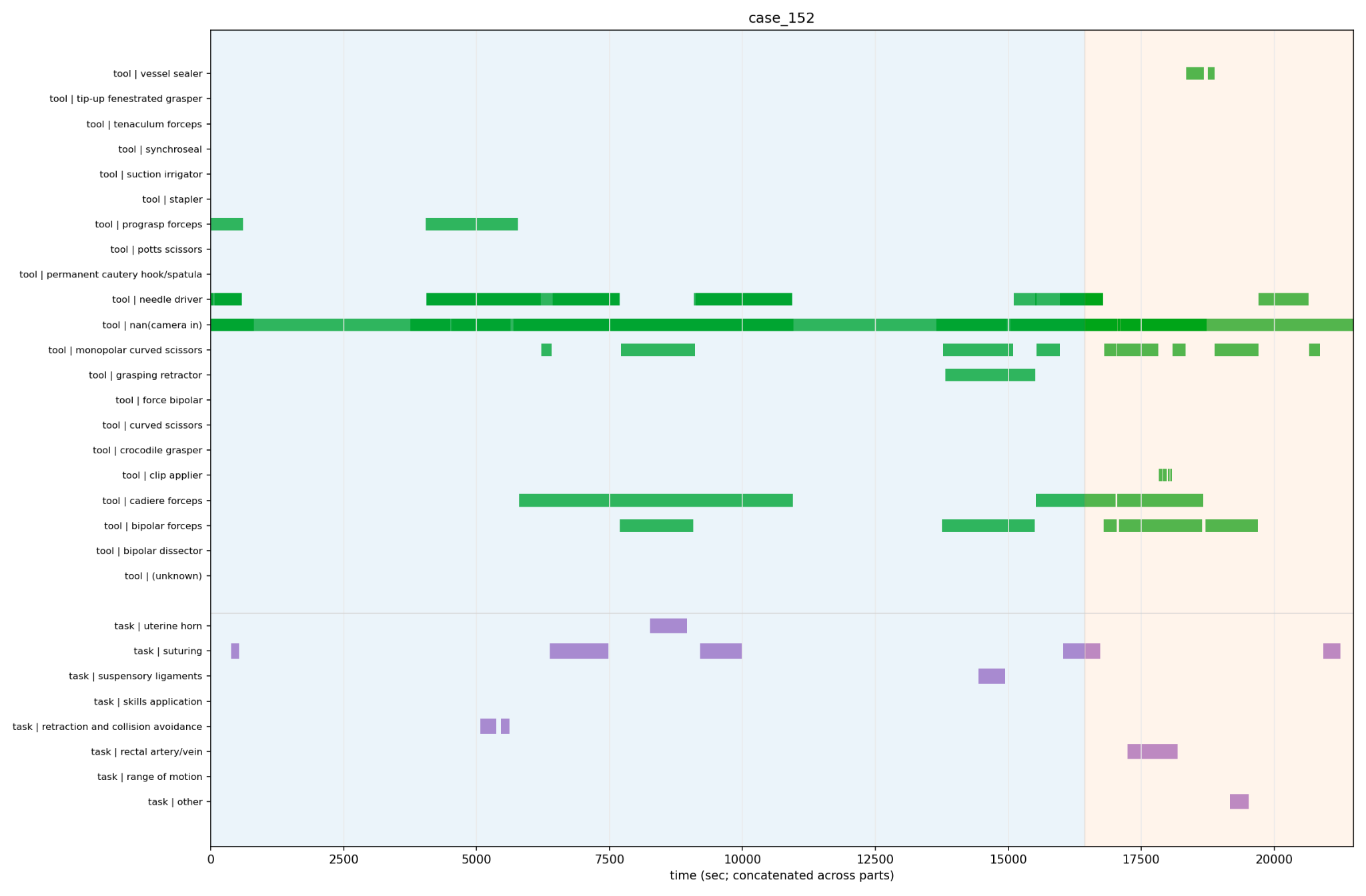}}
\caption{Tool and task labels merged into a single timeline, showing how tool occurrences are used to refine coarse task labels into fine-grained steps/actions.}
\label{fig:C2_AMI_timeline}
\end{center}
\end{figure}

\textbf{QA Set Construction:} We first merged the tool and task annotations into a single timeline (Figure~\ref{fig:C2_AMI_timeline}). Tool annotations were used as-is, while coarse task labels were decomposed into tool-specific fine-grained steps/actions. We then used GPT-5 to generate paraphrased question variants and to balance Yes/No pairs without altering the underlying labels.

\begin{figure}[tbh]
\begin{center}
\resizebox{5in}{!}{\includegraphics{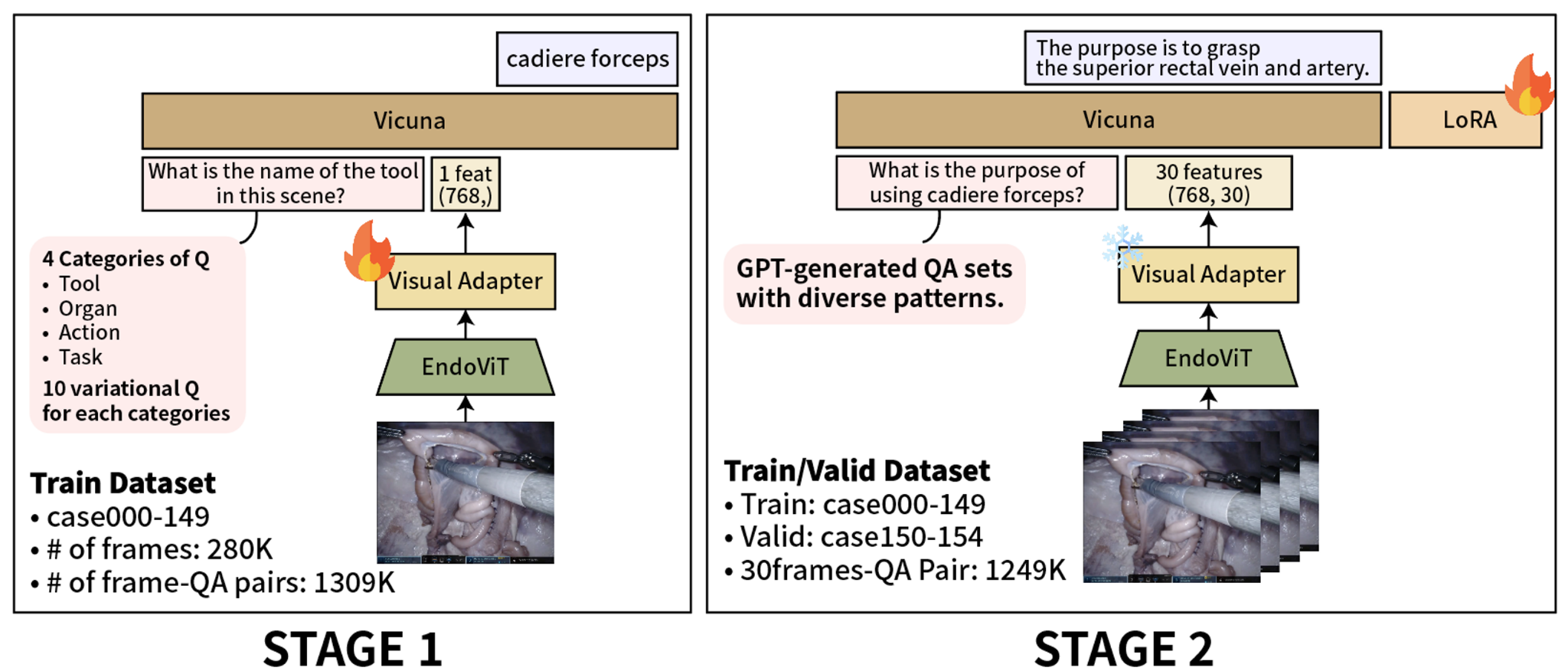}}
\caption{Overall training process showing the two-stage approach: Stage 1 trains a visual adapter to map EndoViT features to Vicuna's text space, and Stage 2 applies LoRA fine-tuning on the curated QA set.}
\label{fig:C2_AMI_training}
\end{center}
\end{figure}

\subsubsection{Model Training}

\textbf{Data \& Pre-processing:}
\begin{itemize}
\item \textbf{Split:} 155 videos total — cases 000–149 for training; 150–154 for internal validation.
\item \textbf{Frame extraction:} 1 fps.
\item \textbf{Features:} EndoViT used for feature extraction.
\end{itemize}

\textbf{Methodology:}
\begin{itemize}
\item \textbf{Backbone LLM:} Vicuna-7B-v1.5\cite{AMI_Vicuna} run with 4-bit quantization.
\item \textbf{Two-stage learning} (inspired by VTimeLLM\cite{AMI_Huang2024}):
\begin{itemize}
\item \textbf{Stage 1 — Visual adapter training:}
\begin{itemize}
\item Adapter: MLP (768→4096) mapping EndoViT features to the Vicuna text space.
\item Loss: token-level cross-entropy.
\item Learning rate: 1e-3.
\item Data: ~280K images; 1 epoch.
\end{itemize}
\item \textbf{Stage 2 — LoRA fine-tuning:}
\begin{itemize}
\item LoRA (Low-Rank Adaptation) rank 64
\item Loss: token-level cross-entropy.
\item Learning rate: 1e-4.
\item Data: QA set ~1.249M examples; 1 epoch.
\end{itemize}
\end{itemize}
\end{itemize}

\subsubsection{Preliminary Performance}

\textbf{What Worked Well:}
\begin{itemize}
\item \textbf{Stage-1 Visual Adapter:} Despite its simplicity, the MLP adapter aligned non-VLM features from EndoViT to Vicuna's language space surprisingly well. In internal checks, the model became reliably good at distinguishing whether specific objects/tools were present.
\item \textbf{Balanced, Contrasting Answers for the Same Question Formulation:} Exposing the model to scenarios where identical question wording required different answers (e.g., Yes/No depending on the frame) improved its ability to truly disambiguate the question, rather than memorize an answer pattern.
\end{itemize}

\textbf{What Did Not Work Well:}
\begin{itemize}
\item \textbf{Single-Answer Bias Per Question Type:} When a given question type was consistently paired with the same answer during training, the model tended to memorize that mapping and ignored the visual evidence from the video. This led to clear overfitting to answer priors.
\end{itemize}

\textbf{Future Directions:}
\begin{itemize}
\item \textbf{Frame Selection for Token-Heavy Inputs:} We did not explore frame selection strategies in this submission. A promising direction is to select a small set of frames that are most likely to support answering the question (e.g., via relevance scoring or lightweight retrieval) and feed only those into the model.
\end{itemize}

\FloatBarrier
\subsection{UT}

We participated in Category 2 (Surgical Visual Question Answering) of the SurgVU 2025 challenge. Our work addresses the challenge of surgical video question answering (VQA) by leveraging a large vision-language model (VLM), InternVL3-2B. The challenge provided annotations specifying whether surgical instruments are present in each frame and the surgical phase (task) being performed. We aimed to exploit these annotations to generate a large-scale synthetic QA dataset for model training.

The motivation behind our approach was twofold: (1) maximize the use of the limited annotations provided, and (2) train a general-purpose VLM for surgical VQA without introducing external data. Our method is unique in that the training data was not directly annotated manually but instead derived from rule-based generation leveraging provided metadata. While rule-based QA generation has limitations in diversity, it offered a straightforward way to bootstrap training data and test whether InternVL3-2B can adapt effectively to the surgical VQA domain.

\subsubsection{Method Description}

From the challenge dataset, we generated four categories of QA pairs:
\begin{itemize}
\item Instrument presence: Does a particular instrument appear in the frame?
\item Organ presence: Is a specific organ present in the frame?
\item Phase classification: What surgical task is being performed?
\item Instrument purpose: What is the purpose of the instrument being used? (rule-based mapping)
\end{itemize}

Approximately 70,000 QA pairs were generated in this way. Importantly, no external data or private annotations were used.

We fine-tuned InternVL3-2B, a multimodal large language model, using LoRA (Low-Rank Adaptation) to efficiently adapt the model to the task while keeping memory and compute costs manageable. The training data consisted of approximately 70k rule-based QA pairs derived from the challenge annotations.

\subsubsection{Model Training}

The following configuration was used:
\begin{itemize}
\item Framework: ms-swift (CLI: swift sft)
\item Model: OpenGVLab/InternVL3-2B
\item LoRA settings: rank = 64, $\alpha$ = 16, dropout = 0.05
\item Precision: bfloat16 enabled, fp16 disabled
\item Optimizer: AdamW (torch implementation)
\item Learning rate: $5 \times 10^{-5}$ with warmup ratio 0.03
\item Batch size: batch size = 1, gradient accumulation steps = 8
\item Max pixels: 1,003,520
\item Steps: trained up to approximately 28,000 steps ($\approx$ 3 epochs)
\end{itemize}

\subsubsection{Preliminary Performance}

Performance improved consistently up to around 3,000 steps, with clear gains observed in BLEU scores and qualitative QA accuracy. However, performance degraded significantly after approximately 28,000 steps (3 epochs), where BLEU scores dropped sharply. This suggests overfitting to rule-based QA patterns, limiting generalization to unseen QA forms. The results indicate that while synthetic rule-based data can effectively bootstrap VQA training, it lacks the linguistic and conceptual diversity needed to sustain performance at larger training scales.

Our submission demonstrates both the potential and limitations of rule-based QA generation for surgical video understanding. InternVL3-2B adapted quickly to the surgical QA task with only approximately 70k synthetic QA pairs, and LoRA fine-tuning allowed efficient training without excessive computational resources. Early training phases yielded encouraging improvements in BLEU and QA accuracy. However, overfitting occurred with extended training, leading to degraded performance, and rule-based QA generation lacked diversity, making the model brittle to novel QA phrasing or unseen contexts.

\FloatBarrier
\subsection{gardenia}

In recent years, multimodal large language models (MLLMs) have demonstrated remarkable understanding and generation capabilities across general tasks. The careful adaptation of such models to medical domains has the potential to significantly enhance the efficiency and consistency of medical data annotation, while interactive question-answering-based explanations could lower the barriers to medical education and training. This would allow regions with limited access to medical resources to benefit from high-quality knowledge services and educational content~\cite{gardenia_Li2024,wang2025endochat}. However, the high demands for rigor and privacy compliance in the medical field make high-quality annotated data exceedingly scarce, limiting the application and generalization of general-purpose MLLMs in real-world medical processes.

In the field of Surgical Visual Question Answering (Surgical VQA)\cite{chen2024survey,lin2023medical,hu2024interpretable}, existing public resources are limited and primarily consist of image-based question-answer pairs\cite{seenivasan2022surgical,bai2023surgical,seenivasan2023surgicalgpt}. There is a lack of video-based question-answer pairs that capture the temporal evolution of semantics and the procedural context throughout surgery. Motivated by these challenges, we constructed a small-scale, balanced, and diverse surgical video question-answer dataset for the SurgVU challenge. Our dataset aims to fill the gap in temporal annotations and improve the relevance of evaluation tasks. Based on this dataset, we propose SurgeMedVL, a model that replaces the visual backbone of InternVL2.5-MoP\_4B\cite{chen2024internvl} with MedViT\cite{manzari2023medvit} to inject medical visual priors, and makes lightweight adaptations to the cross-modal alignment and decoding strategies. We fine-tuned the model end-to-end on our video question-answer data.
\begin{figure}[tbh]
\centering
\includegraphics[width=\linewidth]{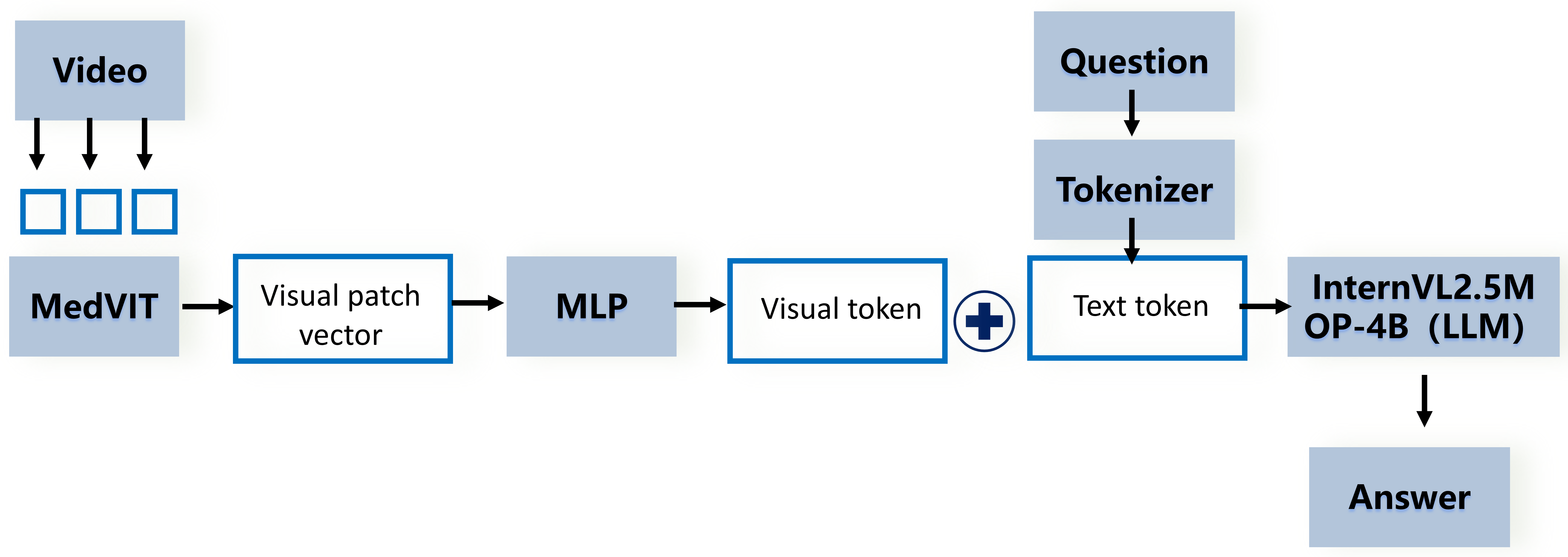}
\caption{Overview of the proposed SurgeMedVL framework. Visual features extracted from surgical video frames are projected into visual tokens and fused with text tokens from the question for answer generation.}
\label{fig:C2_gardenia_framework}
\end{figure}

\subsubsection{Method Description}

To support the temporally sensitive task of Surgical VQA, we constructed a dynamic question-answer dataset based on the training videos provided by the SurgVU challenge. Unlike existing medical VQA resources that primarily rely on static images, our dataset is grounded in the visual evidence of \textbf{Instrument-Verb-Target} relationships. By anchoring questions and answers directly to key events within specific video segments, our dataset provides a more realistic representation of the dynamic surgical field for both training and evaluation.

Our dataset comprises a total of 5,911 question-answer pairs, encompassing five primary task categories: Visual Perception (VP), Temporal and Procedural (TP), Relational Semantics (RS), Evaluative Question Types (EX), and Basic Judgment (JD). The detailed distribution of these question types is summarized in Table 1. Notably, the dataset exhibits a long-tailed distribution; fundamental visual tasks such as tool presence, anatomy identification, and tool counting form the majority of the queries, while more complex procedural and relational semantics constitute the challenging long tail.

\begin{table}[htbp]
\centering
\caption{Distribution of question types in the proposed SurgeMedVL dataset.}
\label{tab:dataset_statistics}
\begin{tabular}{lllrr}
\toprule
\textbf{Main Category}  & \textbf{Sub-category} & \textbf{Count} & \textbf{Percentage} \\
\midrule
\multirow{3}{*}{Visual Perception (VP)} 
 & Counting Surgical Instruments & 1,237 & 20.9\% \\
 & Anatomy Identification & 1,411 & 23.9\% \\
 & Instrument Type and Attributes & 60 & 1.0\% \\
\midrule
\multirow{4}{*}{Temporal and Procedural (TP)} 
 & Procedure Identification & 4 & 0.1\% \\
 & State Change and Outcome & 17 & 0.3\% \\
 & Action Detection & 22 & 0.4\% \\
 & Current Surgical Task or Phase & 1,081 & 18.3\% \\
\midrule
\multirow{3}{*}{Evaluative Questions (EX)} 
 & Tool Presence and Use & 1,537 & 26.0\% \\
 & Need for Action & 15 & 0.3\% \\
 & Multiple-choice Recognition & 211 & 3.6\% \\
\midrule
Basic Judgment (JD) 
 & Surgical Approach Classification & 304 & 5.1\% \\
\midrule
\multirow{2}{*}{Relational Semantics (RS)} 
 & Instrument--Action--Target Relation & 7 & 0.1\% \\
 & Purpose of the Action or Tool & 5 & 0.1\% \\
\midrule
\textbf{Total} & & & \textbf{5,911} & \textbf{100.0\%} \\
\bottomrule
\end{tabular}
\end{table}

We strictly derived our dataset from the official training set provided by the challenge organizers. Based on this, we implemented a reproducible pipeline for generating video-level question-answer samples through fixed-duration segmentation and rigorous quality screening. Specifically, all original videos were first divided into 30-second clips, yielding approximately 36,000 candidate segments. To mitigate potential issues with truncated, damaged, or anomalous segments generated during the slicing process, we applied a two-stage quality control protocol:

\begin{itemize}
    \item \textbf{Temporal Integrity:} Clips with durations shorter than 25 seconds were discarded to guarantee sufficient temporal context.
    \item \textbf{Visual Quality:} Using an automated quality estimator, we calculated the proportion of blurry or black frames in each segment. Clips containing over 80\% of such frames were discarded to ensure adequate visual evidence.
\end{itemize}

Following this filtration process, approximately 22,000 usable video clips were retained as candidates for subsequent question-answer construction.

To ensure clarity and consistency in the supervision signals, each retained segment was assigned a single, unique question type. Subsequently, we selected representative segments from the candidate pool for manual annotation. Annotators provided standardized answers based on predefined question templates and a specialized terminology lexicon, recording optional timestamps to anchor visual evidence where applicable. Ultimately, we obtained a total of 5,911 manually verified video-text question-answer pairs, forming the benchmark set used for model training and evaluation. The remaining segments, assigned with question types but lacking human answers, are retained for future self-supervised exploration.

For our multimodal architecture, we employ InternVL-2.5-MoP-4B, replacing only the visual backbone without introducing any additional structures. Specifically, we substitute its native ViT with MedViT, which has been fine-tuned on the SurgeNetSmall~\cite{de2025scaling} dataset. Once the video clips are encoded by MedViT into visual representations, they are directly fed into the original MLP and the preconfigured cross-modal alignment pathway of InternVL, facilitating text-conditioned answer generation. To mitigate the distribution shift caused by the backbone replacement and to prevent ``catastrophic forgetting'' in the language model, we adopt a progressive, joint training schedule. This ensures stable adaptation while preserving the original module structures.

\subsubsection{Model Training}

The training process follows a progressive and joint schedule to ensure stable adaptation. We first fine-tune the replaced MedViT visual backbone on our surgical video dataset while keeping the cross-modal alignment and language components frozen. This allows the visual encoder to adapt to the surgical domain without disrupting the pre-trained multimodal capabilities. Subsequently, we perform end-to-end fine-tuning on our 5,911 video question-answer pairs, jointly optimizing the visual encoder, cross-modal alignment, and language decoder to enable coherent answer generation conditioned on both visual and textual inputs.

\subsubsection{Preliminary Performance}

Experimental results indicate that our method benefits from temporal information in surgical videos when answering questions about instrument use, anatomical structures, and procedural context. Performance is especially strong in the most frequent question categories, such as tool presence and use, anatomy identification, and current surgical task or phase recognition. The progressive training strategy also contributes to stable domain adaptation by reducing forgetting during the transfer from a general-purpose MLLM to the surgical setting.

\clearpage

\subsection{Results}

Challenge evaluations were conducted on a private, hidden test set comprising surgical video files at 1~FPS, using the Grand Challenge automated submission and evaluation system. Participants did not have access to the test data or labels.

\subsubsection{Category 1: Surgical tool detection}

For Category~1, the standard COCO bounding box detection metric (mAP@[0.5:0.05:0.95]) \cite{lin2014microsoft} was used for evaluation and ranking. Table~\ref{tab:c1_results_2025} presents the final results for teams with complete submissions. Notably, the top two teams---PUH-SVU and Medibot---both surpassed the 0.50 mAP threshold, exceeding the best results from all prior years of this challenge. PDMYR placed third with a mAP of 0.4354. Performance dropped considerably for the remaining teams.

\begin{table}[h!]
\centering
\caption{Category 1 Results -- 2025 (complete submissions only)}
\begin{tabular}{clr}
\toprule
\textbf{Rank} & \textbf{Team} & \textbf{mAP} \\
\midrule
1st  & PUH-SVU      & 0.5188 \\
2nd  & Medibot      & 0.5018 \\
3rd  & PDMYR        & 0.4354 \\
4th  & SK           & 0.2266 \\
5th  & Algoritmi    & 0.161  \\
6th  & SurgTroopers & 0.1577 \\
\bottomrule
\end{tabular}
\label{tab:c1_results_2025}
\end{table}

\subsubsection{Category 2: Surgical visual question answering}

For Category~2, the mean BLEU score \cite{UoMSurgicalAI_Papineni2002} was used for evaluation. For each question, the algorithm produced a single predicted answer, and the BLEU score was computed against each of five ground-truth reference answers; the maximum score among the five comparisons was taken as the score for that question. The final metric was the mean of these maximum BLEU scores across all questions. Table~\ref{tab:c2_results_2025} presents the final results. Capybara achieved the highest BLEU score of 0.4215, followed by UoM-SurgicalAI at 0.3656 and Medibot at 0.3404. Scores ranged from 0.1231 to 0.4215 across the eight teams with complete submissions, reflecting the difficulty of the open-ended VQA task.

\begin{table}[h!]
\centering
\caption{Category 2 Results -- 2025 (complete submissions only)}
\begin{tabular}{clr}
\toprule
\textbf{Rank} & \textbf{Team} & \textbf{BLEU} \\
\midrule
1st  & Capybara        & 0.4215 \\
2nd  & UoM-SurgicalAI  & 0.3656 \\
3rd  & Medibot         & 0.3404 \\
4th  & AMI             & 0.295  \\
5th  & UT              & 0.2652 \\
6th  & gardenia        & 0.2615 \\
7th  & SK              & 0.2021 \\
8th  & Algoritmi       & 0.1231 \\
\bottomrule
\end{tabular}
\label{tab:c2_results_2025}
\end{table}

\subsection{Discussion}


For Category~1, the top two teams achieved mAP scores above 0.50 for the first time in the four-year history of this challenge. PUH-SVU's winning approach combined self-supervised pre-training (MAE on 2 million+ unlabeled frames), pseudo-label generation via Co-DETR, and a multi-distribution sampling strategy to address the long-tail problem. Medibot employed a two-stage pipeline that separated detection (single-class YOLOv8l) from classification (ResNet-50), finding that decoupling these tasks improved overall recall. PDMYR continued their successful HITL active learning approach from the prior year. These results demonstrate continued year-over-year improvement: the 2025 first-place mAP of 0.5188 exceeds the 
the 2024 first-place of 0.4244 (PDMYR), the 2023 first-place result of 0.4669 (PUMCH) and the 2022 first-place of 0.4452 (HRI\_MV).
Importantly, however, the availability of a small validation set with ground-truth bounding box labels in 2025 likely contributed to this improvement. 

A clear trend across all four years is the dominance of pseudo-label and human-in-the-loop annotation strategies over purely weakly supervised approaches. The teams that achieved the highest performance consistently invested in generating high-quality training labels---whether through manual annotation, tracking-assisted labeling, or iterative pseudo-label refinement. Purely weakly supervised methods (e.g., class activation maps, vision-language models without bounding box supervision) have consistently underperformed. Moreover, this year's addition of the validation set and the commensurate increases in improvement drive home the importance of high-quality labels.

Category~2 introduced a fundamentally different challenge: generating natural-language answers to open-ended questions about surgical video clips. The winning team, Capybara, took a notably pragmatic approach: rather than fine-tuning a VLM, they used LLaVA-OneVision-7B in a zero-shot setting, augmented with dedicated surgical tool and organ classifiers trained on the SurgToolLoc 2022 dataset. This two-stage pipeline---first generating structured descriptions of instruments and anatomy, then prompting the VLM with this context---proved more effective than end-to-end fine-tuning approaches. UoM-SurgicalAI corroborated this finding, demonstrating that carefully designed zero-shot prompting with InternVL3.5 outperformed parameter-efficient fine-tuning (DoRA and DCT-GaLore) under the weakly supervised conditions of this challenge. 
This result is notable because it suggests that, for domain-specific VQA with limited supervision, leveraging the zero-shot capabilities of large pre-trained VLMs with structured prompting may be preferable to fine-tuning, which risks overfitting to limited training signals.

Among the fine-tuning approaches, AMI's two-stage training with a visual adapter mapping EndoViT features to Vicuna's text space, followed by LoRA fine-tuning on GPT-generated QA pairs, achieved a competitive BLEU score of 0.295. Medibot's approach of integrating a dedicated detection model from Category~1 into the VQA pipeline to improve tool recognition precision illustrates the potential for cross-task synergy within the challenge. On the other hand, teams that relied on rule-based QA generation (UT) or retrieval-augmented generation with smaller language models (Algoritmi) achieved lower scores, suggesting that the diversity and quality of training supervision are critical for this task.

The introduction of the VQA task represents a natural evolution of the challenge, moving from perception (tool detection) toward understanding (reasoning about surgical scenes). While BLEU scores remain modest overall, the variety of approaches explored---from zero-shot VLMs to fine-tuned multimodal models to retrieval-augmented pipelines---establishes a valuable baseline for future iterations of this task.

\newpage
\section{Conclusion}


Over four years of challenges (2022--2025), working from essentially a single dataset, this series has engaged dozens of teams and highlighted just as many diverse approaches to surgical video-based machine intelligence. In the core task of weakly supervised tool classification and localization, top performances have demonstrated year-over-year progress.
The challenges have also broadened in scope. In 2024, we introduced surgical task recognition as a fully supervised category. In 2025, we introduced surgical visual question answering, a multimodal reasoning task that demands integration of visual perception with natural-language generation. 
The on-going challenge has demonstrated both the broad applicability of our dataset, and the formidability of the machine learning problems that rely on it.



Despite the many skilled solutions employed, team performance on the main challenge has been modest. Weakly supervised tool classification and localization has remained at or below a mAP of 0.5. This is substantially behind fully supervised state-of-the-art methods on general computer vision benchmarks. It is noteworthy that year-over-year improvements have been driven primarily by increasingly sophisticated data curation strategies---human-in-the-loop annotation, pseudo-label generation via tracking and teacher models, and self-supervised pre-training. 
While the fully-supervised category detection challenge found greater success, the VQA task remains nascent, with BLEU scores that leave considerable room for improvement.
Overall this corroborates the inherent difficulty of visual scene understanding in the surgical data science arena and suggests there is room for architectural innovation. 


%
Recent years have produced remarkable advancements in computer vision, particularly with well-established public datasets. While the same performance cannot yet be claimed for surgical data science, the trajectory is encouraging. We will continue to host this competition and evolve its tasks, as we believe that classifying, detecting, and reasoning about surgical scenes in these highly unstructured settings is crucial for the future of surgical data science and computer vision in general. The publicly released dataset accompanying these challenges \cite{SurgVUpaper} will, we hope, serve as a lasting resource for the community.

\bibliographystyle{unsrt}
\bibliography{references}

\newpage


\end{document}